\pdfoutput=1
\documentclass{article} 
\usepackage{amsmath}
\usepackage[final]{colm2026_conference}
\usepackage[protrusion=true,expansion=true,tracking=true,kerning=true,spacing=true,activate=true,final,verbose=silent]{microtype}
\usepackage[hyperfootnotes=false]{hyperref}
\usepackage{url}
\SetProtrusion{encoding=T1,family=zi4}{}
\usepackage{xspace}
\usepackage{array}
\usepackage{booktabs}
\usepackage{graphicx}
\usepackage{enumitem}
\usepackage[T1]{fontenc}
\usepackage{titletoc}
\usepackage{xcolor}

\usepackage{amsthm}
\usepackage{placeins}
\usepackage{svg}

\usepackage[utf8]{inputenc}
\makeatletter
{\nfss@catcodes
\input{t1zi4.fd}
\input{t1ppl.fd}}
\makeatother
\DeclareFontShape{T1}{zi4}{b}{it}{<->ssub * zi4/b/n}{}
\DeclareFontShape{T1}{ppl}{m}{scit}{<->ssub * ppl/m/sc}{}

\usepackage{fvextra}
\DefineVerbatimEnvironment{MyVerbatim}{Verbatim}{
  breaklines=true,
  breakanywhere=true,
  fontsize=\scriptsize
}

\usepackage{etoc}
\newtheorem{theorem}{Theorem}
\usepackage{listings}
\lstdefinelanguage{smt2}{
  morekeywords={declare-sort,declare-fun,assert,forall,exists,define-fun,let,check-sat,declare-const,define-sort,define-fun-rec,define-funs-rec,declare-datatype,declare-datatypes,declare-rel,declare-var,rule,query,par,as,match,case,project,select,store,ite,true,false,Int,Bool,Real,Array,Set},
  sensitive=true,
  morecomment=[l]{;},
  morestring=[b]"}
\usepackage{multirow}
\usepackage{rotating}
\usepackage[textsize=t]{todonotes}
\usepackage{tikz}
\usetikzlibrary{arrows.meta,positioning,fit,calc}
\usepackage{relsize}

\newcommand{\TODO}[1]{{\color{blue} TODO: #1}}

\usepackage{subcaption}
\usepackage{amssymb}

\definecolor{bannerbg}{HTML}{F7F8FA}
\definecolor{bannerframe}{HTML}{D6DBE3}
\definecolor{bannertitle}{HTML}{1F3A8A}

\definecolor{kw}{HTML}{0F766E}      
\definecolor{op}{HTML}{7C3AED}      
\definecolor{fn}{HTML}{1D4ED8}      
\definecolor{sym}{HTML}{111827}     
\definecolor{var}{HTML}{B45309}     
\definecolor{cmt}{HTML}{6B7280}     
\definecolor{bad}{HTML}{B91C1C}     

\usepackage{lineno}

\definecolor{darkblue}{rgb}{0, 0, 0.5}
\hypersetup{colorlinks=true, citecolor=darkblue, linkcolor=darkblue, urlcolor=darkblue}

\title{
\begin{minipage}[c]{0.76\textwidth}
\raggedright
\LARGE\bfseries
\textsc{SatIR}:
Scalable High-Recall\\
Constraint-Satisfaction-Based\\
Information Retrieval for \\
Clinical Trials Matching
\end{minipage}
\hspace{-0.02\textwidth}
\begin{minipage}[c]{0.18\textwidth}
\centering
\href{https://satir.genie.stanford.edu}{%
    \includegraphics[width=0.72\linewidth]{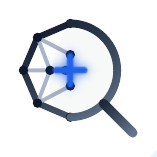}
}

\vspace{1mm}
\makebox[\linewidth][c]{%
    \small\href{https://satir.genie.stanford.edu}{satir.genie.stanford.edu}
}
\end{minipage}
}

\author{%
Zikai Zhou$^{1}$ \quad Yufei Jin$^{2,*}$ \quad Yilin Xu$^{3,*}$\\
\textbf{Yu-Chiang Wang$^{4}$ \quad Chieh-Ju Chao$^{4}$ \quad}
\textbf{Monica S. Lam}$^{1}$ \\
\\
$^{1}$Department of Computer Science, Stanford University, Stanford, CA, USA \\
$^{2}$Samueli Electrical and Computer Engineering, UCLA, Los Angeles, CA, USA \\
$^{3}$Department of Computer Science, Emory University, Atlanta, GA, USA \\
$^{4}$Mayo Clinic, Rochester, MN, USA \\
$^{*}$Equal contribution \\
\texttt{zikai@stanford.edu, lam@stanford.edu}
}

\usepackage{pgfplots}
\pgfplotsset{compat=1.18}
\usepackage{sansmath}

\newcommand{\system}{EMC\xspace}
\usepackage{setspace}
\usepackage{tabularx}
\usepackage[most]{tcolorbox}

\usepackage{longtable}
\usepackage{pdflscape}
\usepackage{caption}
\usepackage{makecell}
\usepackage{ragged2e}

\usepackage{eso-pic}
\newcolumntype{L}[1]{>{\RaggedRight\arraybackslash}p{#1}}
\newcolumntype{C}[1]{>{\centering\arraybackslash}p{#1}}

\begin{document}

\ifcolmsubmission
\linenumbers
\fi

\maketitle

\newcommand{\Text}{\mathrm{Text}}

\newcommand{\cyrus}[1]{{\color{blue}[CZ: #1]}}

\definecolor{accent}{HTML}{2E86DE} 

\newcommand{\name}{\textsc{SatIR}\xspace}
\newcommand{\representation}{\textsc{TrialRepr}\xspace}
\newcommand{\blname}{\textsc{TrialGPT}\xspace}
\newcommand{\fullnamerender}{%
  \textbf{\textcolor{accent}{FOrma}}l\ %
  \textbf{R}easoning\ and\ %
  \textbf{M}odeling\ %
  \textbf{A}gent%
}
\newcommand{\fullname}{Formal Reasoning and Modeling Agent}

\newcommand{\fillme}{{\bf XXX}\xspace}
\newcommand{\eg}{{\it e.g.,}\xspace}
\newcommand{\ie}{{\it i.e.,}\xspace}
\newcommand{\etal}{{\it et.~al}\xspace}
\newcommand{\bigO}{\mathrm{O}}

\newenvironment{packeditemize}{\begin{list}{$\bullet$}{\setlength{\itemsep}{0.5pt}\addtolength{\labelwidth}{-4pt}\setlength{\leftmargin}{2ex}\setlength{\listparindent}{\parindent}\setlength{\parsep}{1pt}\setlength{\topsep}{2pt}}}{\end{list}}

\newcommand{\ruleset}{constraint set}
\newcommand{\Ruleset}{Constraint set}
\newcommand{\rulesets}{constraint sets}
\newcommand{\Rulesets}{Constraint sets}
\newcommand{\rulesetdefinition}{\textbf{Definition:} A \emph{constraint set} is a named collection of deterministic predicates that must all evaluate to true on a record for that record to be deemed compliant.}

\newcommand{\recorddefinition}{\textbf{Definition:} A \emph{record} is one self-contained, immutable mapping from schema field names to typed, time-stamped values describing a single real-world entity or event.}

\newcommand{\recordconstraintpair}{record-contraint-set pair}
\newcommand{\recordconstraintpairs}{record-contraint-set pairs}

\newcommand{\uls}{unified terminology system}
\newcommand{\ulses}{unified terminology systems}

\newcommand{\applicationfield}{patient recruitment for clinical trials}

\newcommand{\SAT}{\mathrm{SAT}}

\newcommand{\Clin}{\mathsf{Clin}}          
\newcommand{\surf}{\mathrm{surf}}          
\newcommand{\can}{\mathrm{can}}            
\newcommand{\struct}{\mathrm{struct}}      
\newcommand{\sem}{\mathsf{sem}}            
\newcommand{\sys}{\mathrm{sys}}            

\newcommand{\impl}{\mathrm{impl}}

\newcommand{\SameConcept}{\mathsf{SameConcept}}

\newcommand{\I}{\mathbb{I}}
\newcommand{\TopN}{\mathrm{TopN}}
\newcommand{\Dedup}{\mathrm{Dedup}}
\newcommand{\Uniform}{\mathrm{Uniform}}
\newcommand{\TP}{\mathrm{TP}}

\newcommand{\TopK}{\operatorname{TopK}}

\newcommand{\FP}{\mathrm{FP}}
\newcommand{\TN}{\mathrm{TN}}
\newcommand{\FPR}{\mathrm{FPR}}
\newcommand{\FNR}{\mathrm{FNR}}
\newcommand{\BalAcc}{\mathrm{BalAcc}}

\newcommand{\modeccr}{treat-chief}
\newcommand{\modeall}{treat-any}
\newcommand{\modeallexplore}{relevant-to-any}


\definecolor{smtblue}{RGB}{33,113,181}
\definecolor{tgred}{RGB}{205,70,65}
\definecolor{tiegray}{RGB}{128,128,128}

\newcommand{\smtcolor}{smtblue}
\newcommand{\tgcolor}{tgred}
\newcommand{\tiecolor}{tiegray}

\newcommand{\smtimprove}[1]{{\scriptsize\textbf{\textcolor{smtblue}{(+#1\%)}}}}

\newcommand{\smt}[1]{\textcolor{smtblue}{\textbf{S:} #1}}
\newcommand{\tg}[1]{\textcolor{tgred}{\textbf{T:} #1}}

\newcommand{\winbarendpoints}[5]{%
\begin{tikzpicture}[x=1mm,y=1mm,baseline=(bar.base)]
    \def\L{8.0}
    \def\W{26}
    \def\R{8.0}

    \pgfmathsetmacro{\A}{\W*#1/100}
    \pgfmathsetmacro{\B}{\W*#2/100}
    \pgfmathsetmacro{\C}{\W*#3/100}

    \path[use as bounding box] (0,-0.3) rectangle ({\L+\W+\R},4.3);
    \node[inner sep=0pt, outer sep=0pt] (bar) at ({\L + \W/2},2) {};

    \ifdim #1pt > 0pt
        \node[font=\tiny\bfseries, text=\smtcolor, anchor=east]
            at ({\L-0.25},2) {(+#4)};
    \fi

    \ifdim #3pt > 0pt
        \node[font=\tiny\bfseries, text=\tgcolor, anchor=west]
            at ({\L+\W+0.25},2) {(+#5)};
    \fi

    \fill[\smtcolor] (\L,0) rectangle ({\L+\A},4);
    \fill[\tiecolor!35] ({\L+\A},0) rectangle ({\L+\A+\B},4);
    \fill[\tgcolor] ({\L+\A+\B},0) rectangle ({\L+\A+\B+\C},4);
    \draw[black!30] (\L,0) rectangle ({\L+\W},4);

    \ifdim \A pt > 6pt
        \node[font=\tiny\bfseries, text=white] at ({\L+\A/2},2) {#1};
    \fi
    \ifdim \B pt > 6pt
        \node[font=\tiny\bfseries, text=black] at ({\L+\A+\B/2},2) {#2};
    \fi
    \ifdim \C pt > 6pt
        \node[font=\tiny\bfseries, text=white] at ({\L+\A+\B+\C/2},2) {#3};
    \fi
\end{tikzpicture}%
}

\newcommand{\coveragefocusbar}[4]{%
\begin{tikzpicture}[x=1mm,y=1mm,baseline=(bar.base)]
    \def\W{26}
    \def\H{4}
    \pgfmathsetmacro{\L}{0.4*\W}
    \pgfmathsetmacro{\R}{0.2*\W}
    \pgfmathsetmacro{\M}{0.4*\W}

    \pgfmathsetmacro{\U}{#2 + #3}

    \ifdim \U pt > 0pt
        \pgfmathsetmacro{\B}{\M*(#2/\U)}
        \pgfmathsetmacro{\C}{\M*(#3/\U)}
    \else
        \pgfmathsetmacro{\B}{0}
        \pgfmathsetmacro{\C}{0}
    \fi

    \node[inner sep=0pt, outer sep=0pt] (bar) at (\W/2,2) {};

    \fill[tiegray!45] (0,0) rectangle (\L,\H);
    \fill[smtblue] (\L,0) rectangle ({\L+\B},\H);
    \fill[tgred] ({\L+\B},0) rectangle ({\L+\B+\C},\H);
    \fill[white] ({\W-\R},0) rectangle (\W,\H);

    \draw[black!35] (0,0) rectangle (\W,\H);

    \fill[white] ({\L-1.15},-0.15) rectangle ({\L+0.55},{\H+0.15});
    \fill[white] ({\W-\R-0.55},-0.15) rectangle ({\W-\R+1.15},{\H+0.15});

    \draw[black!35] (0,0) rectangle (\W,\H);

    \draw[black!85, line width=0.7pt] ({\L-0.95},0.15) -- ({\L-0.15},{\H-0.15});
    \draw[black!85, line width=0.7pt] ({\L-0.45},0.15) -- ({\L+0.35},{\H-0.15});

    \draw[black!85, line width=0.7pt] ({\W-\R-0.35},0.15) -- ({\W-\R+0.45},{\H-0.15});
    \draw[black!85, line width=0.7pt] ({\W-\R+0.15},0.15) -- ({\W-\R+0.95},{\H-0.15});

    \node[font=\tiny\bfseries, text=black] at (\L/2,2) {#1};

    \ifdim \B pt > 5pt
        \node[font=\tiny\bfseries, text=white] at ({\L+\B/2},2) {#2};
    \else
        \ifdim #2pt > 0pt
            \node[font=\tiny\bfseries, text=smtblue, anchor=south] at ({\L+max(\B/2,0.8)},4.5) {#2};
        \fi
    \fi

    \ifdim \C pt > 5pt
        \node[font=\tiny\bfseries, text=white] at ({\L+\B+\C/2},2) {#3};
    \else
        \ifdim #3pt > 0pt
            \node[font=\tiny\bfseries, text=tgred, anchor=south] at ({\L+\B+max(\C/2,0.8)},4.5) {#3};
        \fi
    \fi

\end{tikzpicture}%
}
\newcommand{\Ont}{\mathcal{M}}                 
\newcommand{\Concepts}{K}            
\newcommand{\Sub}{\sqsubseteq}                 
\newcommand{\Patients}{P}            
\newcommand{\Trials}{C}              
\newcommand{\Objectives}{O}          
\newcommand{\Facts}{F}            
\newcommand{\PatientAttr}{A_\textrm{p}}            
\newcommand{\Relations}{R}           
\newcommand{\Qualifiers}{Q}          
\newcommand{\Rel}{R}
\newcommand{\PC}{\textit{PC}}
\newcommand{\TC}{\textit{TC}}
\newcommand{\Rm}{R_\textrm{M}}
\newcommand{\Rp}{R_\textrm{P}}
\newcommand{\Rt}{R_\textrm{T}}
\newcommand{\CTM}{\textsc{CTM}}
\newcommand{\CTR}{\textsc{CTR}}
\newcommand{\CNF}{\textsc{CNF}}
\newcommand{\SQL}{\textsc{SQL}}

\newcommand{\cnf}{\textit{cnf}}
\newcommand{\PD}{\textit{PD}}
\newcommand{\CD}{\textit{CD}}
\newcommand{\PT}{\textit{PT}}
\newcommand{\TT}{\textit{TT}}
\newcommand{\OT}{\textit{OT}}
\newcommand{\ECNF}{\textsc{ECNF}}
\newcommand{\CNFD}{\textsc{CNFD}}
\newcommand{\DA}{\textsc{DA}}
\newcommand{\AB}{\textsc{AB}}
\newcommand{\AN}{\textsc{AN}}
\newcommand{\FactSet}[1]{\mathsf{Facts}(#1)}   
\newcommand{\TextP}[1]{\mathsf{T_\mathrm{p}}(#1)}   
\newcommand{\ReqSet}[1]{\mathsf{Req}(#1)}      
\newcommand{\Record}[1]{\mathsf{Rec}(#1)}      
\newcommand{\nonK}{K^\textrm{N}}
\newcommand{\Trial}{\mathsf{Trial}}            

\newcommand{\Intent}[1]{\mathsf{Intents}(#1)}              
\newcommand{\Attr}{\mathsf{Attr}}              
\newcommand{\ProjFormula}[1]{\Psi_{#1}}        
\newcommand{\TrialFormula}[1]{\Phi_{#1}}       
\newcommand{\Atoms}{\mathsf{Atoms}}            

\newcommand{\Policy}{\pi}                      
\newcommand{\PolicyMiss}{\pi_{\mathrm{miss}}}  
\newcommand{\PolicySpec}{\pi_{\mathrm{spec}}}  

\newcommand{\Interp}[2]{I_{#1,#2}}             
\newcommand{\SemLift}{\mathsf{Lift}}           

\newcommand{\Retriever}{\mathsf{Retrieve}}     
\newcommand{\Cand}{\mathsf{Cand}}              

\newcommand{\Match}{\textsc{Match}}            
\newcommand{\Unsat}{\mathsf{Unsat}}            

\newcommand{\Eligible}{\textsc{eligible}}
\newcommand{\Ineligible}{\textsc{ineligible}}
\newcommand{\Unknown}{\textsc{unknown}}

\newcommand{\True}{\textsc{true}}
\newcommand{\False}{\textsc{false}}

\newcommand{\AttrRel}{\mathcal{R}_{\mathrm{attr}}}

\newcommand{\Compat}{\mathsf{Compat}}

\begin{abstract}

Many real-world retrieval and matching problems require more than topical relevance: a candidate must satisfy the specific constraints of one profile among many, not just be relevant to it. Clinical trials are a high-stakes instance of this challenge: they are central to evidence-based medicine, yet many struggle to meet enrollment targets, despite the availability of over half a million trials listed on ClinicalTrials.gov, which attracts approximately two million users monthly. Existing retrieval techniques, largely based on keyword and embedding-similarity matching, treat eligibility constraints as soft signals rather than binding requirements, resulting in low recall, low precision, and limited interpretability.

We propose \name{}\footnote{\url{https://github.com/stanford-oval/clinical-trial-matching}.}, a scalable, efficient, high-precision, high-recall, interpretable clinical trial retrieval method based on {\em formal constraint satisfaction}. Leveraging established medical ontologies, we use Large Language Models (LLMs) to convert informal reasoning—regarding ambiguity, implicit clinical assumptions, and incomplete patient records—into explicit, precise, controllable, and interpretable formal Satisfiability Modulo Theories (SMT) constraints. For scalable and efficient retrieval, we project the SMT matching problem onto relational algebra, enabling an efficient database implementation that retains high recall while sacrificing little precision.

\name{} consistently improves eligibility-aware retrieval over similarity-based baselines on the SIGIR 2016 dataset and a benchmark derived from TREC 2022. 
Relative to TrialGPT-style retrieval, \name{} retrieves 32\%--72\% more relevant-and-eligible trials per patient on SIGIR 2016 and achieves 1.8--3.2$\times$ higher eligible-trial recall on the TREC benchmark. Retrieval is fast, requiring only 146 milliseconds per patient over 3,621 SIGIR trials.

\end{abstract}

\section{Introduction}
\label{sec:introduction}



\begin{figure}[htbp]
  \centering
  \includegraphics[width=0.9\columnwidth]{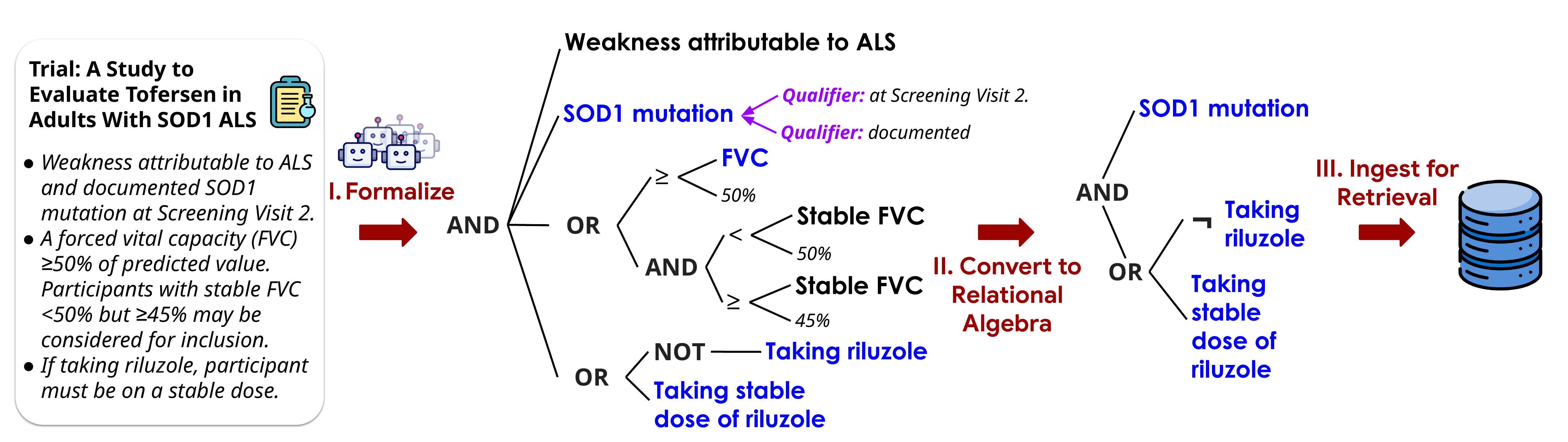}
  \caption{Example: A tofersen SOD1-ALS study showing how \name{} turns
  free-text eligibility criteria into formal, interpretable constraints that
  can be executed in a database for scalable retrieval.\protect\footnotemark{}
  Terms in \textbf{\textcolor{blue}{blue}} are ontology-grounded concepts.
  In Stage II, \name{} keeps only clauses that can be safely translated into
  database constraints without ruling out true matches; clauses that cannot be
  translated this way are dropped.\vspace{-0.1in}}
  \label{fig:framework_overview}
\end{figure}
\footnotetext{The example is simplified from the tofersen study NCT02623699.
See \url{https://nypost.com/2025/03/18/health/mother-with-rare-als-touts-miracle-drug-that-has-stopped-her-disease/}
for a motivating patient story.}

Many matching and retrieval tasks require {\em per-profile matching}: checking a candidate record against a large collection of profiles, where each profile defines its own distinct set of constraints. In job recommendation, a candidate must be matched to one posting's specific qualifications \citep{kan2026conflating}. In clinical trial matching, a patient must be matched to one trial's specific eligibility criteria. In medical diagnosis, a patient's symptoms must be matched against the diagnostic criteria for one specific condition among the many thousands catalogued in medicine \citep{nasem2015diagnosis}. In industrial process fault diagnosis, a reported equipment malfunction must be matched against the specific symptom signature of one candidate cause among many, drawn from a library of known fault conditions and their corresponding symptoms \citep{venkatasubramanian2003reviewII}.

\paragraph{Clinical Trial Matching}

Clinical trial matching is a high-stakes instance of this broader challenge. Clinical trials are central to evidence-based medicine, yet recruiting eligible participants remains a persistent bottleneck: despite recruitment investments estimated at approximately \$1.9 billion annually \citep{desai2020recruitment}, nearly 80\% of trials miss their enrollment targets or timelines \citep{brogger2020online}.

Depending on which side initiates the search, the problem splits into two subproblems. The first is {\em trial-to-patient} matching: a sponsor searching patients for their clinical trial—a conventional {\em constraint-based} query on a patient database. The second is {\em patient-to-trial} matching: a user, such as a cancer patient who has exhausted approved options, searches across all open trials for any they are eligible for. This is substantially harder: rather than one fixed constraint set applied to many patient records, there are as many distinct constraint sets as there are trials, each checked against the same single record. We refer to this direction—each profile's own constraint set checked against the same record—as {\em constraint-satisfaction-based} retrieval. Solving patient-to-trial matching well is both the harder problem and the one with the clearest case for investment on patients' behalf.

Most recent trial-matching systems, including \emph{TrialGPT}~\citep{jin2024matching}, represent patients and trials as keywords, text, or dense embeddings, retrieve candidates by lexical or semantic similarity, and use an LLM to rank or evaluate the resulting pairs. Such representations capture topical relatedness well, but eligibility depends on sharp constraints—negation, temporality, numeric thresholds, laterality—that must be satisfied jointly, not scored independently. Similarity-based retrieval treats these constraints as soft signals rather than requirements to be individually verified, so relevant-but-ineligible candidates crowd out eligible ones, with no way to inspect why a candidate was retrieved or withheld.

\paragraph{Problem Statement.}
This paper studies how to retrieve, at \textit{scale}, \textit{potentially} matching patient--trial pairs from a given set of each, subsuming both subproblems above. The objectives are: \textbf{high recall} (no suitable trial missed), \textbf{high precision} (minimizing matching burden with false leads, increasing screening workload, churn, and protocol risk \citep{gregory2017electronic,wendler2021inevitability,lee2025advancements}), \textbf{efficiency} (operating at scale), \textbf{controllability} (a uniform matching protocol across patients and trials), and \textbf{interpretability} (errors easy to detect and attribute). Once a retriever narrows the space of candidate pairs, more expensive matching algorithms can be deployed to adjudicate them \textit{exactly}.

\paragraph{Contributions.} We propose \name, a scalable information-retrieval technique based on \textbf{constraint satisfaction} that outperforms embedding-based retrieval on all five criteria. 
Trial and patient constraints are not only logically complex; they are typically written in informal, ambiguous natural language, presuppose implicit domain knowledge, and must be applied against records that are themselves incomplete.
We factor the task into two steps: \textit{understanding} constraints and \textit{retrieving} against them. For the former, we use LLMs to parse trials and patient records, with all their subtleties, into formal Satisfiability Modulo Theories (SMT)~\citep{barrett2018satisfiability} formulas. For the latter, we project the resulting SMT matching problem onto relational algebra, yielding an efficient database implementation that retains high recall while sacrificing little precision.

\begin{enumerate}[leftmargin=*, noitemsep, topsep=0pt]
\item{\textbf{Grounded, context-aware formalization.}}
We introduce \representation, an SMT-based representation for clinical trials and patient records that is
(i) \textbf{expressive}, capturing Boolean logic, numeric thresholds, temporal windows, and exceptions;
(ii) \textbf{medically grounded}, incorporating ontology-aware concept identity and entailment in a SNOMED-based model \citep{snomed2025model}, with standardized units and temporal conventions; and
(iii) \textbf{context- and intent-aware}, distinguishing hard and soft constraints while accounting for patient preferences and objectives.
We also present an interpretable approach to missing patient data, using LLMs to assess the \emph{salience} of missing information.

\item{\textbf{Accurate semantic parsing.}}
We propose an LLM-based semantic parser that accurately translates under-specified clinical trials and diverse patient records into \representation.

\item{\textbf{Scalable trial retrieval with high recall.}}
We propose a novel scalable constraint-based retriever by leveraging 
relational algebra. By storing a projection of \representation as first-class data in databases, our system retrieves a highly targeted set of candidate clinical trials for all patients, according to a given objective, efficiently with SQL queries.
\end{enumerate}

Experimentally, \name{} outperforms embedding-based baselines across all three objectives, retrieving 32--72\% more relevant-and-eligible trials per patient and achieving 92--94\% micro recall on SIGIR 2016. These gains generalize to TREC 2022, where recall improves by 1.8--3.2$\times$, while retrieval takes only 146 ms per patient and remains stronger even against TrialGPT at a budget of 200 candidate trials per patient.




\begin{figure}[t]
  \centering
  \includegraphics[width=0.85\columnwidth]{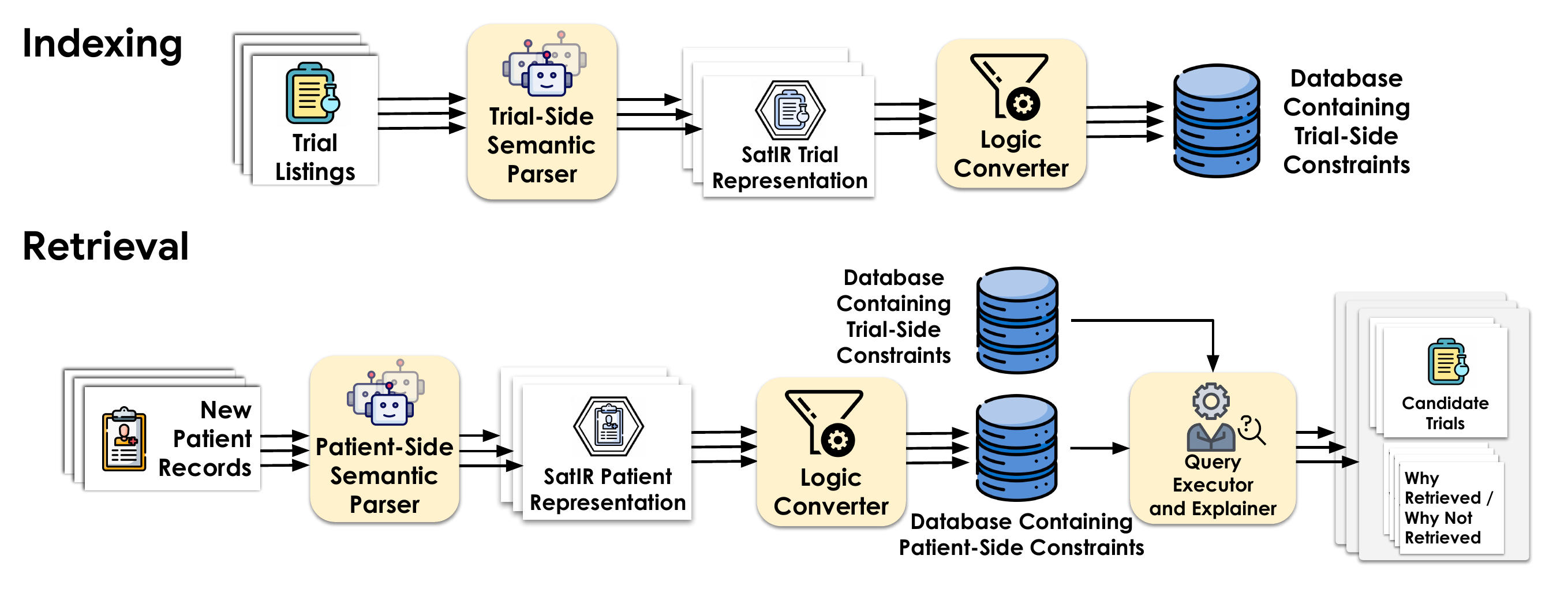}
    \caption{\name{} system overview. Trial listings are parsed and converted
    offline into formal trial-side constraints stored in a database. At retrieval
    time, each new patient record is independently parsed into patient-side
    constraints. A database query executor jointly evaluates the two
    representations to retrieve candidate trials satisfying the matching objective
    and explain why each trial was or was not retrieved.\vspace{-0.1in}}
  \label{fig:stacked}
\end{figure}
\section{Formalization of the Clinical Trial Retrieval Task}
\label{sec:formalization}

\textbf{Design Principles.} Our approach separates \textit{understanding} constraints from \textit{satisfying} them. We use LLMs to translate natural-language text into a formal representation, then use formal methods to perform the match. This separation makes the understanding phase interpretable and lets the matching phase carry formal guarantees.

Clinical trial matching cannot, and need not, be \emph{fully} captured in a formal language for retrieval: no finite symbolic language can express every clinical nuance. Rather than forcing all text into a rigid schema, we design a representation that is \textbf{conservative} and \textbf{recall-oriented}, avoiding the exclusion of potentially eligible trials, while remaining as \textbf{expressive} as possible to filter unsuitable ones and improve \textbf{accuracy}.

\representation represents clinical trials and patients as SMT (Satisfiability Modulo Theories) formulas, canonicalizing terminology with the SNOMED CT medical ontology, which contains over 350{,}000 concepts and about three million relationships~\citep{snomed_us_release_notes_2019}. Because SNOMED CT does not cover all concepts appearing in patient records and trial text, we extend it with {\em non-canonical variables} whose semantics are specified by free-text descriptions, preserving fidelity to the source text. See Section \ref{sec:rq2_faithful_parsing} for details.

\textit{Explicit knowledge representation for formal reasoning.}
All information used for patient--trial matching must be represented explicitly. We therefore leverage SNOMED's subsumption structure to support formal reasoning over canonicalized concepts. See Section \ref{sec:subsumption} for details.

\textit{Handling missing information with a salience assumption.}
Patient records are often incomplete. We address this by assuming that {\em salient facts about a patient would be documented}. When information is missing, we use LLMs to assess whether the record supports, refutes, or is inconclusive about the condition of interest. Making these assessments explicit keeps retrieval transparent and interpretable. We discuss this in Section \ref{sec:salience}.

\subsection{The Medical Ontology}\label{sec:medical_ontology}

A \textbf{medical ontology} for clinical trial matching is
$
O = (R, K, Q, \sqsubseteq),
$
where $R = \Rm \cup \Rp \cup \Rt$ (elaborated in Appendix \ref{app:naming_schema}), with $\Rm$ a set of medical \textbf{relations}
\{\textsc{hasfindingof}, \textsc{hasundergone}, ...\}, $\Rp$ a set of patient facts pertaining to
clinical trials including their intents in clinical trial matching and preferences \{chiefcomplaint, ...\}, and $\Rt$ a set of intents of trials
\{treat, prevent, diagnose, ...\}. Further, $K$ is a set of medical \textbf{concepts},
and $Q$ is a set of \textbf{qualifiers}. Finally, $
\sqsubseteq
=
\sqsubseteq_\textrm{R}
\cup
\sqsubseteq_\textrm{K}
\cup
\sqsubseteq_\textrm{causal}
\cup
\sqsubseteq_\textrm{Q}
$ (Appendix \ref{app:subsumption})
is the family of \textbf{subsumption} relations.
\begin{itemize}[leftmargin=*, noitemsep,topsep=0pt]
\item $\sqsubseteq_\textrm{R}\subseteq R\times R$ denotes subsumption over medical relations. We write $O_R=(R,\sqsubseteq_R)$, where $r_p \sqsubseteq_R r_c$ means $r_p$ is more specific than, and therefore supports, $r_c$; e.g., $\textsc{HasDiagnosisOf} \sqsubseteq_R \textsc{HasFindingOf}$.

\item $\sqsubseteq_\textrm{K}\subseteq K\times K$ denotes subsumption over concepts. A variable with concept $k_p$ supports one with concept $k_c$ whenever $k_p \sqsubseteq_K k_c$; e.g., $\textsc{AcuteAppendicitis} \sqsubseteq_K \textsc{Appendicitis}$.

\item $\sqsubseteq_\textrm{causal}\subseteq (R,K)\times(R,K)$ denotes subsumption relations based on causal or dependency structures across relations and concepts. 
We derive these from SNOMED graph relations and retain only transformations consistent with $\Rel$ (e.g., 
$
(\textsc{HasFindingOf}, \textsc{PostoperativeHemorrhage})
\sqsubseteq_{\mathrm{causal}}
(\textsc{HasUndergone}, \textsc{SurgicalProcedure})
$
).

\item $\sqsubseteq_\textrm{Q}\subseteq Q\times Q$ denotes subsumption over qualifiers, including cases where a more specific qualifier, or a qualified variable, supports a less specific one.
\end{itemize}








\subsection{The \representation Representation}

An \textbf{SMT formula} is an arbitrary composition of \textit{atomic constraints} using logical operators such as conjunction, disjunction, counting, and negation. An \textbf{atomic constraint}
$a=(v,\mathsf{cmp},t)\in A$
consists of a variable $v\in V$, a comparison operator $\mathsf{cmp}\in\{<,\le,=,\neq,\ge,>\}$, and a comparison target $t\in T$; for example, $((\mathsf{hasCondition},\mathsf{Type2Diabetes},\mathsf{within5y}),=,\mathsf{True})$ requires type 2 diabetes within the past five years, and $((\mathsf{hasMeasurement},\mathsf{HemoglobinA1c},\mathsf{current}),<,7.0)$ requires a current hemoglobin A1c below $7.0$.

A \textbf{variable} $v\in V:R \times K \times Q$ is either: $\text{(i) \textbf{canonical}: } v=(r,k,q), \quad r\in R,\; k\in K,\; q\in Q\cup\{\emptyset\},$
such as $(\mathsf{hasCondition},\mathsf{Type2Diabetes},\mathsf{within5y})$, or
$
\text{(ii) \textbf{non\mbox{-}canonical}: } v=\mathit{id},
$
such as $\mathsf{HasAdequateSocialSupportForFollowUp}$, where $\mathit{id}$ is a mnemonic identifier annotated with free text, used when the variable cannot be faithfully expressed using $R$, $K$, and $Q$.

\begin{figure}[!htbp]
\vspace{-0.2in}
  \centering
  \includegraphics[width=0.9\columnwidth]{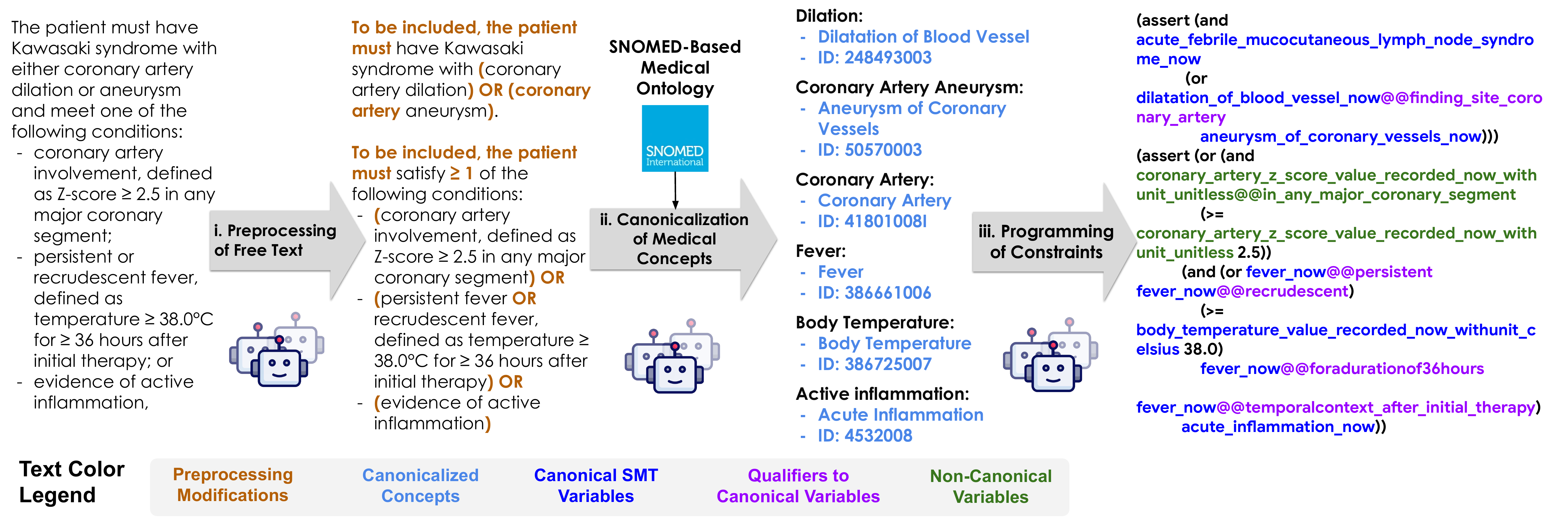}
  \caption{Architecture of Semantic Parsing Workflows in \name{}. $@@$ denotes qualifiers.\vspace{-0.15in}}
  \label{fig:semantic-parsing-architecture}
\end{figure}

Given an ontology $O=(R,K,Q,\sqsubseteq)$, a set of clinical trials $C$ with free-text descriptions $\TT$, a set of patient records $P$ with free-text descriptions $\PT$, and a free-text matching objective $\OT$, the \representation\ consists of:  
(i) for each trial $c\in C$, a \textbf{trial constraint} $\TC(c)$, an SMT formula over variables in $V:\Rm\cup\Rt \times K \times Q$, which may combine canonical constraints, such as ``type 2 diabetes within five years,'' with non-canonical constraints, such as        ``adequate social support for follow-up'';  
(ii) for each patient $p\in P$, a \textbf{patient constraint} $\PC(p)$, an SMT formula over canonical variables in $V: \Rm\cup\Rp \times K\times Q$, such as assertions that the patient was diagnosed with type 2 diabetes three years ago and has a current hemoglobin A1c of $6.8$; and  
(iii) a formalized matching \textbf{objective} $\theta$, such as ``treat chief complaints.''

\subsection{Semantic Parser from Clinical Text to \representation}
\label{sec:rq2_faithful_parsing}
Accurate retrieval depends on accurately translating natural-language constraints into formal SMT statements grounded in a comprehensive medical ontology. In practice, we found single-shot in-context LLM parsing error-prone and unreliable. We therefore use a three-stage pipeline (Figure~\ref{fig:semantic-parsing-architecture}): (1) restructure the text to facilitate parsing, (2) link entities to SNOMED CT, and (3) generate the SMT program.

\textbf{Stage 1: Structured Decomposition of Requirements.}
\label{sec:preprocessing_requirements}

Clinical trials often specify distinct \textbf{cohorts}, such as control and experimental groups, each with their own eligibility criteria. We create separate requirement lists for each cohort and apply the following steps on them. 
Further details appear in Appendix~\ref{app:preprocessing}.

\begin{enumerate}[leftmargin=*, noitemsep, topsep=0pt]
\item \textbf{Polarity.} Trials often list inclusion and exclusion criteria in separate sections. We generate a unified, context-independent list by reversing the polarity of exclusion criteria. 
\item \textbf{Span Contiguity.} We rewrite clinically meaningful spans so that relevant entities appear as contiguous phrases, making concept mentions easier to canonicalize.
\item \textbf{Logical Structure.} We rewrite the logical statements to make conjunction, disjunction, and precedence explicit, so as to clarify the logical relationships within constraints.
\item \textbf{Decomposition.} We break down constraint descriptions into separate, self-contained sentences, creating modular units that can later be compiled and validated independently.
\end{enumerate}

\textbf{Stage 2: Entity Canonicalization.}
\label{sec:concept_canonicalization}
Canonicalizing medical terms simplifies matching between patients and trials. We link entities to our ontology \citep{dredze2010entity,zheng2010learning}; medical facts that cannot be represented faithfully are retained as non-canonical variables. See Appendix~\ref{app:canonicalization} for details.

\textbf{Stage 3: Incremental SMT Programming.}
\label{sec:programming_smt_assertions}
Translating a trial into an SMT program—typically around 360 lines—is non-trivial. We therefore process one self-contained free-text constraint at a time, introducing canonical and non-canonical variables, extracting qualifiers, and generating SMT declarations and \texttt{assert} statements as needed. Before commitment, each fragment is checked for syntactic and sort correctness, consistency with the current program, and semantic faithfulness, including polarity, qualifiers, and variable use. Only validated fragments are retained. See Appendices~\ref{app:smt_programming} and \ref{app:smt_example} for details and an example.

Patient parsing follows a simplified version of Stages 1--3, with additional treatment of time-window qualifiers (Appendices~\ref{app:patient_coding_pipeline} and \ref{app:patient_intervalization}).

\subsection{Augmenting \representation for Formal Reasoning about Clinical Trials}\label{sec:subsumption}

As discussed in Section \ref{sec:formalization}, reasoning about constraints must account for the family of subsumption relations $\sqsubseteq = \sqsubseteq_\textrm{R} \cup \sqsubseteq_\textrm{K} \cup \sqsubseteq_\textrm{causal} \cup \sqsubseteq_\textrm{Q}$. We compile these subsumption relations into the SMT constraints themselves. 
Concretely, on the patient side, we encode
$\sqsubseteq_{\mathrm{K}}, \sqsubseteq_{\mathrm{R}}, \sqsubseteq_{\mathrm{Q}}$, and
$\sqsubseteq_{\mathrm{causal}}$
as additional constraints and inject them into the patient constraint set for each patient $p \in P$. On the trial side, we encode $\sqsubseteq_{\mathrm{Q}}$
as additional constraints and inject them into the trial constraint set for each trial $c \in C$. Thus, constraint solving is ultimately performed over augmented patient and trial formulas rather than over raw extracted variables alone. We discuss how these subsumption relations are implemented in Appendix \ref{app:subsumption} and \ref{app:qualifier_subsumption}.

\subsection{Augmenting \representation to Handle Patient Data Missingness}
%


Patient records are often incomplete or under-specified. In this section, we describe how we augment \representation to handle such incompleteness, primarily through \textit{salience}-based reasoning over missing or underspecified evidence, and, in some cases, by inferring diagnoses from patient notes.

\textbf{Salience as a principle for handling missingness.} \label{sec:salience}
To improve retrieval accuracy, we make a fundamental assumption: {\em any salient information about a patient's condition will be documented in their medical record}. This allows us to address missing data by focusing on whether the absent information is truly salient. Because salience is not formally defined in the medical literature, we use targeted LLM queries to assess concept importance. We record these judgments explicitly, keeping matching decisions transparent, interpretable, and open to expert review. In contrast, end-to-end LLM matching is much harder to inspect.

\begin{figure}[htbp]
\vspace{-0.2in}
    \centering
\begin{subfigure}[htbp]{0.47\textwidth}
    \centering
    \includegraphics[width=\linewidth]{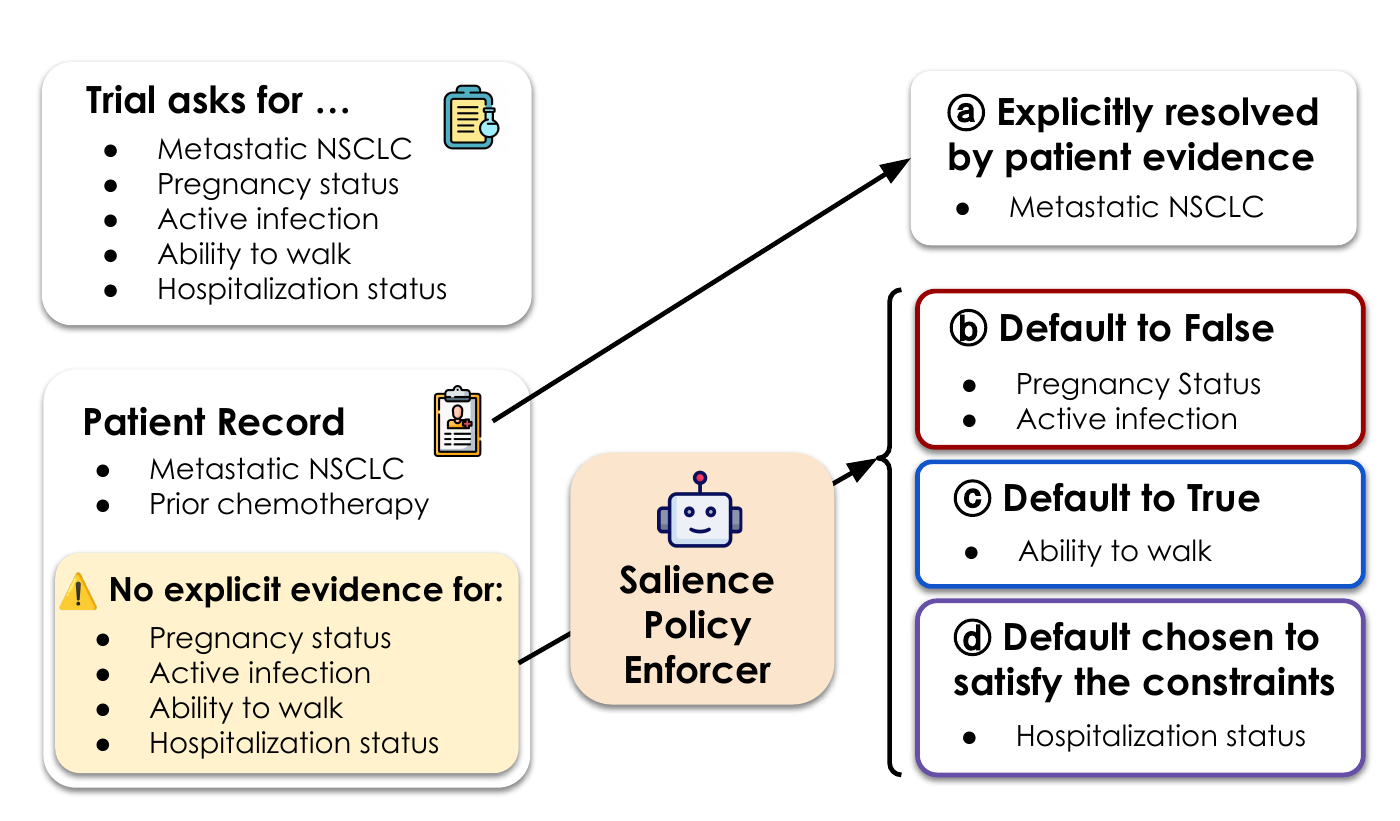}
    \caption{Whole-fact missingness.}
    \label{fig:whole-fact-salience}
\end{subfigure}
\hfill
\begin{subfigure}[htbp]{0.47\textwidth}
    \centering
    \includegraphics[width=\linewidth]{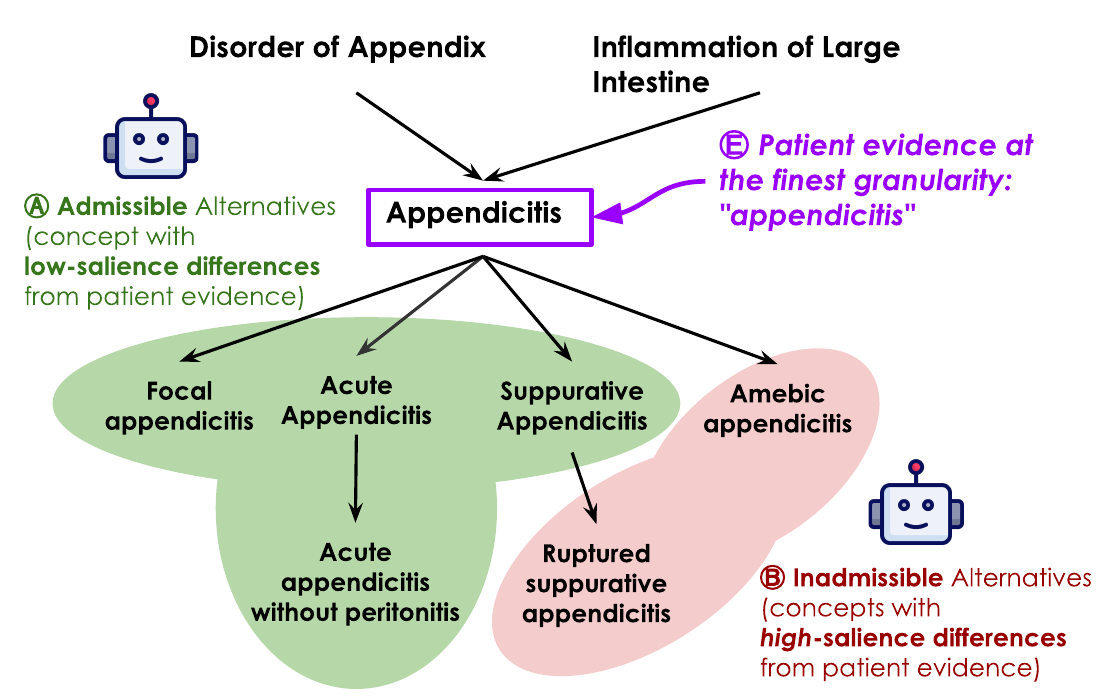}
    \caption{Specificity mismatch.}
    \label{fig:speicificity-salience}
\end{subfigure}
\caption{Salience-based reasoning under incomplete patient evidence.
\textbf{(a)} When evidence for a trial variable is absent from the patient record, salience determines whether the variable is false, true, or unconstrained.
\textbf{(b)} When patient evidence is less specific than a trial variable, salience determines which ontology descendants remain admissible: plausibly
omitted distinctions are retained, while highly salient conditions require
explicit evidence.\vspace{-0.1in}}
    \label{fig:side_by_side}
\end{figure}

\textit{Missingness in records}. Patient records often lack information to support or refute specific constraints. For each clinical trial condition, we use the LLM to determine whether potentially missing information is salient, and correspondingly whether it should be interpreted as supporting the condition, refuting it, or remaining inconclusive (Figure \ref{fig:whole-fact-salience}). 

\textit{Under-specificity in records}. As discussed in Section~\ref{sec:subsumption}, a patient is logically eligible for trials \textit{targeting conditions that subsume the patient's diagnosis}. However, since medical records can be under-specified, it may also be reasonable to match patients to trials when \textit{their diagnosis subsumes the trial's targeted condition}. Whether we should do so depends on the salience of the targeted condition. As shown in Figure~\ref{fig:speicificity-salience}, a patient documented only with ``Appendicitis'' may still match a trial for ``Acute Appendicitis,'' since the record may omit that extra specificity. But the same patient should not match a trial for ``Ruptured Suppurative Appendicitis,'' because such a salient condition would likely be explicitly recorded.

For each trial condition, we use the LLM to assess its salience relative to its parent concept in the medical ontology. If it is not highly salient, patients diagnosed with the parent condition are also considered potential matches.


\textbf{Inferring diagnoses from patient notes.}\label{sec:infer_diagnosis}
Some patient notes describe symptoms or clinical findings without stating a diagnosis explicitly. In these cases, we augment \representation by inferring likely diagnoses using the LLM pipeline from Section~\ref{sec:rq2_faithful_parsing}. This component is included mainly to address a limitation of the evaluation dataset~\citep{koopman2016test}, since real clinical documentation usually states diagnoses explicitly. Further details are provided in Appendix~\ref{app:patient_inference}.

\section{Constraint-Satisfaction-Based Retrieval}
\label{sec:retrieval}
Our SMT-based \representation is designed to capture strict eligibility and relevance constraints \textit{for retrieval} with high recall and precision. SMT is used for its expressiveness, but checking each patient--trial pair with an SMT solver directly would be prohibitively expensive at scale.

\name{}'s key idea is to \textit{project} SMT constraints into relational constraints and find matching pairs with far greater efficiency using standard database techniques. This projection guarantees that every pair matching under the SMT constraints also matches under the projected relational constraints—so \name{} retains \textit{perfect recall with respect to the SMT formulas}.

\textbf{Theorem.}
\name{} returns all pairs satisfying patient, trial, objective constraints in SMT (\TC, \PC, $\theta$), achieving 100\% recall.  
\[
\TC \cup \PC \cup \theta \models \name(\TC \cup \PC \cup \theta).
\]

\textbf{Remarks.}
Because relational constraints are strictly less expressive than SMT, some pairs satisfying the relational constraints may fail the original SMT constraints; \name{} may therefore return non-matching pairs. Precision is thus governed by the expressiveness gap between SMT and relational algebra for the given constraints.

\textbf{Projection from SMT to quantifier-free Conjunctive Normal Form (\CNF).}
Since patient records contain only concrete facts rather than Boolean clauses, we can project the SMT formulas by applying quantifier elimination \textit{independently} to patient records and clinical trials. We apply quantifier elimination to remove non-canonical variables and their associated subformulas, yielding a quantifier-free formula while preserving entailment~\citep{cooper1972,demoura2008z3}.


We represent each quantifier-free constraint in CNF, so each entity $e \in C \cup P$ takes the form $\CNF(e)=\bigwedge_{i=1}^m d_i$, where $d_i=\bigvee_{j=1}^{n_i} a_{ij}$ and each $a_{ij}\in A$ is an atomic constraint over canonical variables.


\textbf{Representing CNF constraints as first-class database objects.}
\name{} stores CNF constraints in a patient database $\PD$ and a clinical
trial database $\CD$, each containing five tables:
\begin{itemize}[leftmargin=*, noitemsep, topsep=0pt]
    \item \textbf{Entity-to-CNF} $\ECNF(e,\cnf)$ links each entity $e$ to its
    CNF constraint.
    
    \item \textbf{CNF-to-clause} $\CNFD(\cnf,d)$ maps each CNF to its $m$
    disjunctive clauses.
    
    \item \textbf{Clause-to-atom} $\DA(d,a)$ links each disjunctive clause to
    its $n$ atomic constraints.
    
    \item \textbf{Boolean atom} $\AB(a,v,\mathsf{cmp},t)$ stores unique
    Boolean atoms.
    
    \item \textbf{Numerical atom} $\AN(a,v,\mathsf{cmp},t)$ stores unique
    numerical atoms, with $t$ represented by lower and upper bounds and
    endpoint inclusivity.
\end{itemize}

Once trials and patients are represented in this form, matches for a given objective~$\theta$ can be retrieved with a straightforward SQL query over the database tables, denoted~$\SQL(O,\CD,\PD,\theta)$; see Appendix~\ref{app:sql_prefilter}.

\section{Clinical Trial Matching}
A \textbf{Clinical Trial Matching} (\CTM) task takes a medical ontology $O$, clinical trials $C$, patient records $P$, their free-text descriptions $\TT$ and $\PT$, and a match objective $\OT$, and seeks all trial--patient pairs satisfying the objective and all trial and patient constraints. $\CTM(O,\TT,\PT,\OT)=\Match(c,p), \quad \forall (c,p)\in \name(\representation(O,\TT,\PT,\OT)).$

If the constraints in \representation are all {\em necessary}, then \name has perfect recall. However, 
the LLM-based semantic parser can introduce such errors, though the formal representation it produces remains highly interpretable and auditable. Since \representation may not capture all constraints, \Match{} is implemented by applying an LLM to match the original trial and patient free text (Appendix~\ref{app:nl_match}).

\section{Experimental Results}
\label{sec:results}

\subsection{Evaluation Methodology}
\label{sec:evaluation_methodology}

\textbf{Datasets.}
We evaluate on the SIGIR 2016~\citep{koopman2016test} and TREC 2022~\citep{roberts2022overview} clinical trial matching benchmarks. SIGIR 2016 includes 59 synthetic patients and 3,621 ClinicalTrials.gov trials~\citep{clinicaltrialsgov}; TREC 2022 includes 50 synthetic topics and 26,581 trials. For cost efficiency, we construct \textsc{TREC-2022-RetrievalSubset} with all 50 topics and 2,658 trials: all 670 judged trials and 1,988 randomly sampled trials.

\textbf{Systems.} \name{} uses GPT-4.1, except GPT-5 for semantic-parser repair (Appendix~\ref{app:ir_repair_pipeline}). We compare against TrialGPT~\citep{jin2024matching}, its clinician-keyword baseline, and variants replacing its dense retriever with PubMedBERT~\citep{gu2021domain}, SapBERT-PubMedBERT~\citep{deka2022improved}, BMRetriever~\citep{xu2024bmretriever}, or BGE-large-en-v1.5~\citep{xiao2024c}. TrialGPT-style baselines use GPT-4.1~\citep{openai_gpt41_2025} and GPT-5~\citep{singh2025openai}.

\textbf{Retrieval Objectives.}
We evaluate three objectives: (1) trials treating the patient's chief complaint (\modeccr), (2) any patient condition (\modeall), and (3) trials broadly relevant to any patient condition (\modeallexplore); see Appendix~\ref{app:retrieval_objectives}. \name{} implements them through constraint-based filtering. Because TrialGPT lacks objective-specific retrieval, we evaluate objective-aware and objective-agnostic keyword generation and report the strongest variant.

\textbf{Evaluation Protocols.}
We use two complementary evaluation protocols.

\emph{Qrel-based evaluation (TREC 2022 only).}
For TREC 2022, we use its official annotations (qrels): irrelevant (0), relevant but ineligible (1), and relevant-and-eligible (2). SIGIR 2016 qrels cover only a subset of pairs, do not encode our retrieval objectives, and contain inconsistencies; we report results with the original labels separately in Appendix~\ref{app:dataset_faithful_results}.

\emph{Objective-specific judge evaluation (both datasets).}
For both datasets, a GPT-5 judge determines whether each retrieved trial is relevant to the specified objective and eligible for the patient. This captures retrieval utility under our objectives, which existing labels do not fully represent. We validate the judge through clinician review below.

\textbf{Metrics.}
\name{} returns all satisfying trials, whereas TrialGPT-style methods return ranked lists. For comparable retrieval budgets, we truncate each baseline to the mean number of trials returned per patient by \name{}.

Under both datasets, we measure mean retrieved relevant-and-eligible trials per patient and recall. On SIGIR 2016, we additionally report two patient-level outcomes: whether a method retrieves more useful trials than the strongest baseline (Utility winner) and whether it retrieves at least one (Patients served). Patient-level comparisons use the strongest fixed TrialGPT baseline with GPT-5.

\textbf{Judge Validation.}
We validate the judge used in the objective-specific evaluation protocol through independent clinician review. Two clinicians independently reviewed 27 balanced judge decisions, accepting 25 and 24, respectively. Of four unique disagreements, three were clinically ambiguous and one was a clear error. Agreement was strong (Gwet's AC1=0.82)~\citep{gwet2008computing}; we use AC1 instead of Cohen's $\kappa$ because acceptance was highly prevalent~\citep{cohen1960coefficient}. See Appendices~\ref{app:clinician_annotation} and~\ref{app:clinician_review_table}.

Statistical tests, ablations, failure analysis, and experimental settings appear in Appendix~\ref{app:experiment_umbrella}.

\begin{table}[htbp]
\centering
\footnotesize
\setlength{\tabcolsep}{2pt}
\renewcommand{\arraystretch}{1.2}
\begin{tabular}{lcccccccc}
\toprule
\textbf{Objective} & \textbf{Retrieved/Patient} & \textbf{BMRet} & \textbf{ClinBest} & \textbf{S-PubMedB} & \textbf{BGE} & \textbf{PubMedB} & \textbf{TG} & \textbf{\name{}} \\
\midrule
\modeccr & 36 & 1.44 & 1.75 & 1.97 & 2.07 & 2.00 & 2.17 & \textbf{3.25} \\
\modeall & 64 & 2.00 & 2.54 & 2.75 & 2.85 & 2.68 & 2.98 & \textbf{5.12} \\
\modeallexplore & 121 & 6.39 & 7.24 & 8.54 & 8.68 & 8.92 & 8.93 & \textbf{11.76} \\
\bottomrule
\end{tabular}
\caption{SIGIR 2016 results: mean number of retrieved trials per patient that are both relevant and eligible. Patient-level paired bootstrap CIs in Appendix~\ref{app:ci_tables} show that \name{} significantly outperforms every baseline under all three objectives.}
{\raggedright\footnotesize
\textit{Abbreviations:} BMRet = BMRetriever; ClinBest = best clinician-keyword variant; S-PubMedB = SapBERT-PubMedBERT; BGE = BGE-large-en-v1.5; PubMedB = PubMedBERT; TG = TrialGPT.\par}
\label{tab:retrieval_utility_main}
\end{table}

\begin{table}[htbp]
\centering
\footnotesize
\setlength{\tabcolsep}{4pt}
\renewcommand{\arraystretch}{1.2}
\begin{tabular}{lccccccc}
\toprule
\textbf{Objective} & \textbf{BMRet} & \textbf{ClinBest} & \textbf{S-PubMedB} & \textbf{BGE} & \textbf{PubMedB} & \textbf{TG} & \textbf{\name{}} \\
\midrule
\modeccr & 41.46 & 50.24 & 56.59 & 59.51 & 57.56 & 62.44 & \textbf{93.66} \\
\modeall & 35.98 & 45.73 & 49.39 & 51.22 & 48.17 & 53.66 & \textbf{92.07} \\
\modeallexplore & 50.27 & 56.93 & 67.20 & 68.27 & 70.13 & 70.27 & \textbf{92.53} \\
\bottomrule
\end{tabular}
\caption{SIGIR 2016 results: micro recall against the union of relevant-and-eligible trials found by all methods. Patient-level paired bootstrap CIs in Appendix~\ref{app:ci_tables} show that \name{} significantly outperforms every baseline under all three objectives.\vspace{-1em}}
\label{tab:retrieval_recall_micro}
\end{table}

\subsection{SIGIR 2016}\label{sec:results_sigir}

\textbf{\name{} retrieves more relevant-and-eligible trials per patient.}
Under all three objectives, \name{} retrieves the most useful trials per
patient---e.g., 3.25 vs.\ 2.17 for the strongest baseline (TrialGPT) under
\modeccr---with all TrialGPT-style baselines trailing
(Table~\ref{tab:retrieval_utility_main}; every gain significant by
patient-level paired bootstrap, Appendix~\ref{app:ci_tables}).

\textbf{\name{} recovers the largest fraction of useful trials found by any
method.} Against the union of relevant-and-eligible trials across all methods,
\name{} reaches 92--94\% micro recall under every objective, versus 54--70\%
for the best baseline, TrialGPT (Table~\ref{tab:retrieval_recall_micro}).

\textbf{\name{} improves utility for more individual patients.}
Against TrialGPT, \name{} wins/ties/loses on 20/33/6 patients under
\modeccr, 29/28/2 under \modeall, and 35/17/7 under
\modeallexplore{} (Appendix Table~\ref{tab:retrieval_patientlevel_main}).
The advantage is significant under every objective
(exact sign test, $p<0.01$), showing that the aggregate gains reflect broad
per-patient improvement rather than a few high-yield patients.
Patient-coverage counts are also reported in
Appendix Table~\ref{tab:retrieval_patientlevel_main}; these comparisons are
descriptive because the matched served-vs.-not-served outcomes are not
statistically separated.


\textbf{Remaining \name{} misses are few and mostly parsing- or evidence-related.}
At 200 trials per patient, TrialGPT with GPT-5 misses 253 pairs retrieved by
\name{}, but retrieves only 20, 18, and 36 pairs missed by \name{} under
\modeccr, \modeall, and \modeallexplore{}, respectively
(0.34, 0.31, and 0.61 per patient; 11,800 retrieved instances). We analyze
these 74 \name{} misses. Excluding 17 judge errors or unresolved cases, the
57 clear failures arise mainly from patient-side parsing (14), trial-side
parsing (13), under-specified evidence (8), missing evidence (6), and other
factors (16); see Appendix~\ref{app:failure_case_analysis}.

\textbf{Retrieval is fast:} 146 ms per patient over 3,621 trials
(Machine 2; Appendix~\ref{app:setting}).

\subsection{TREC 2022}\label{sec:results_trec}

\begin{table}[htbp]
\centering
\footnotesize
\setlength{\tabcolsep}{3.8pt}
\renewcommand{\arraystretch}{1.15}
\begin{tabular}{lcrrrcrrr}
\toprule
& & \multicolumn{3}{c}{\textbf{Objective-specific labels}}
& \multicolumn{3}{c}{\textbf{Official TREC labels ($qrel=2$)}} \\
\cmidrule(lr){3-5} \cmidrule(lr){6-8}
\textbf{Objective} & \textbf{$k$}
& \textbf{BM25} & \textbf{TG} & \textbf{\name{}}
& \textbf{BM25} & \textbf{TG} & \textbf{\name{}} \\
\midrule
\modeccr & 20 & 0.252 & 0.247 & \textbf{0.804} & 0.162 & 0.139 & \textbf{0.445} \\
\modeall & 46 & 0.295 & 0.339 & \textbf{0.850} & 0.245 & 0.260 & \textbf{0.480} \\
\modeallexplore & 50 & 0.239 & 0.276 & \textbf{0.862} & 0.252 & 0.268 & \textbf{0.539} \\
\bottomrule
\end{tabular}
\caption{Recall on \textsc{TREC-2022-RetrievalSubset}. Objective-specific recall evaluates trials retrieved by any of the compared methods using our GPT-5 relevance-and-eligibility judge. Official-label recall instead uses the original TREC judgments and counts only trials labeled relevant and eligible ($qrel=2$). $k$ is \name{}'s average number of retrieved trials under each objective. \name{} significantly outperforms both baselines in every comparison using patient-level paired bootstrap confidence intervals (Appendix~\ref{app:trec_ci_tables}).}
\label{tab:trec2022_recall}
\end{table}

\textbf{Gains generalize to TREC 2022.}
On \textsc{TREC-2022-RetrievalSubset}, we evaluate \name{}, BM25, and TrialGPT using both our objective-specific judgments and the original TREC relevance judgments. Under the objective-specific evaluation, we judge the trials retrieved by any of the three methods for relevance to each retrieval objective and patient eligibility. \name{} achieves substantially higher recall than either baseline under all three objectives (Table~\ref{tab:trec2022_recall}), showing that its gains extend beyond the SIGIR benchmark.

\textbf{The two evaluations measure different notions of a useful trial.}
Our objective-specific labels assess relevance to each objective and patient eligibility. The official TREC labels use TREC's original relevance definition and cover only annotated trials. Thus, absolute recall is not directly comparable across evaluations. Nevertheless, both yield the same conclusion: \name{} substantially outperforms BM25 and TrialGPT under every retrieval objective.

The same trend holds when using the official TREC labels. When the target is restricted to trials labeled relevant and eligible ($qrel=2$), \name{} achieves recalls of 0.445--0.539, compared with 0.162--0.252 for BM25 and 0.139--0.268 for TrialGPT. When the target is broadened to include all relevant trials ($qrel=1/2$), \name{} remains best, achieving recalls of 0.357, 0.383, and 0.429, compared with 0.201, 0.277, and 0.285 for BM25 and 0.172, 0.300, and 0.313 for TrialGPT. Together, these results show that \name{} improves retrieval under both our objective-specific definition and the benchmark's original human annotations.

\vspace{-0.6em}

\section{Related Work}\label{sec:related_work}

Most prior pipelines are text-first: they linearize trials and patient records, retrieve candidates by keyword or embedding similarity, and use LLMs for matching and ranking (e.g., TrialGPT-style systems)~\citep{jin2024matching}. Biomedical retrievers improve candidate generation~\citep{gu2021domain,xu2024bmretriever}, but reasoning remains largely implicit.

Entity linking and ontology grounding help canonicalize surface forms~\citep{liu2021self,snomed2025model}. \name{} uses grounding not only for canonicalization but also to guide constraint augmentation with LLM-based informal reasoning (i.e., salience).

Related work uses LLMs to manage, synthesize, and query heterogeneous corpora, including STORM~\citep{shao2024assisting}, Co-STORM~\citep{jiang2024into}, and LOTUS~\citep{patel2024semantic}. These systems reason largely over free text or weakly structured data at query time. By contrast, \name{} pushes retrieval-critical semantics into executable, ontology-grounded constraints for scalable symbolic execution.

\name{} also relates to formal methods and neurosymbolic LLM+symbolic systems. SMT enables reasoning over structured constraints~\citep{barrett2018satisfiability,demoura2008z3}, while systems such as LINC, Logic-LM, and PAL translate natural language into formal programs executed by symbolic engines~\citep{olausson2023linc,pan2023logiclm,gao2023pal}. Unlike these systems, \name{} targets \textit{real-world large-scale clinical retrieval}, emphasizing \textit{scalable, interpretable} constraint execution under missing data.

\vspace{-0.6em}

\section{Conclusion} \label{sec:conclusion}

This paper addressed \textit{per-profile matching}—retrieving profiles whose distinct constraints a candidate satisfies—using clinical trial matching as a high-stakes case study. \name{} combines formal methods with LLMs to solve this at scale, achieving significantly higher recall and precision than the prior state of the art, TrialGPT, due to four factors.
(1) The SNOMED CT model supplies the ontology and subsumption reasoning our representation depends on. (2) Formal methods typically cannot reason about missing data; we identify \textit{salience} as the key to resolving this, and show LLMs can capture salience judgments in an interpretable, controllable formal representation. (3) Clinical language has a depth no fixed symbolic system fully captures; our representation targets the constraints that matter most, and this fidelity drives both high recall and high precision. (4) Projecting SMT constraints into relational algebra achieves scalability, preserving recall while trading away some precision.

More broadly, our results suggest that complex, real-world tasks demanding high recall are best served by dividing labor between LLMs for language understanding and formal methods for reasoning—yielding interpretable representations that are amenable to scalable and verifiably correct reasoning. Limitations are discussed in Appendix~\ref{app:limitations}.

\section*{Acknowledgments}

We acknowledge support from the Verdant Foundation, the Hasso Plattner Institute, Itaú Unibanco, BMO Financial Group, and the Stanford Human-Centered Artificial Intelligence (HAI) Institute. We acknowledge the National Artificial Intelligence Research Resource (NAIRR) Pilot and Microsoft Azure for contributing to the results in this work. Cyrus Zhou is partially supported by the Stanford School of Engineering Fellowship.

We thank Dr. Bryant Lin for suggesting this research topic. We thank Yucheng Jiang for designing and implementing the clinician annotation interface. We are grateful to Shicheng Liu, Jiacheng Sang, Dongwei Jiang, Harshit Joshi, Jiuding Sun, Yucheng Jiang, Sina Semnani, Tamara Czinczoll, Chris Hoenes, Sally Wang, Jungwoo Kim, and Stanford OVAL Lab members for their support, feedback, and discussions. We also thank Dr. James Ford, Dr. Bryant Lin, Dr. James Dickerson, Dr. Jimmy Lin, Dr. Michael Gensheimer, Professor Gill Bejerano, and Lisa S. Lowy for their expert guidance and helpful discussions. We thank Shengguang Wu, Yicheng Qian, Heng Yu, Juze Zhang, Yue Zhao, Hermann Kumbong, Devon Smith, Youngjoong Kwon, Sa Zhou, Yilong Zhao, Jiaming Tang, Alireza Haqi, Hanchen Li, Qizheng Zhang, Fangrui Huang, Haichuan Wang, Jiahao Lu, Ziqi Shu, Shreyas Agarwal, Jie Zhu, Wenyi Wang, Zhijie Huang, Grace Zhang, Jiashan Zhao, Tianxin Wang, Qinghui Wang, Yunong Zhang, Haochen Pan, Yifei Zhang, and Shiqi Kuang for their support and discussions. We further thank Lefan Zhang, Shan Lu, Blase Ur, Nandish Shah, Ning Tang, Youwen Wu, and Qi Hu for their support and perspectives. 
Finally, we thank our families for their enduring support.

\section*{Ethics Statement}

This work presents \name{}, a clinical trial retrieval system for improving clinical trial retrieval for patients. Because such systems operate in a high-stakes medical domain, ethical considerations are important.

\paragraph{Use of synthetic patient data.} Although we use real clinical trial listings, all patient records used in \name{} are synthetically generated and do not correspond to real individuals. Thus, this work does not involve human subject data or raise patient privacy concerns.

\paragraph{Use of LLMs.} LLMs are used as components of \name{} for semantic parsing, salience assessment, and evaluation, as described in the main paper and appendix. They were also used in an assistive role for manuscript writing and for supporting development of the backend, clinical interface, and project website. All LLM outputs used in the system, evaluation, and manuscript were reviewed or validated by the authors, who take full responsibility for the final paper and system.

\bibliography{forma}
\bibliographystyle{colm2026_conference}

\normalsize

\newpage
\appendix

\section*{Appendix}
This appendix is organized into six parts.

\begin{enumerate}[leftmargin=2em]
    \item \textbf{Experimental results and evaluation} (Appendix~\ref{app:experiment_umbrella}).
    Presents aggregate and per-patient results, retrieval objectives,
    the annotation interface, clinician review details, failure-case analysis,
    and experimental settings.

    \item \textbf{Symbolic representation and subsumption} (Appendix~\ref{app:representation_umbrella}).
    Describes subsumption rules, qualifier handling, and the relation
    vocabularies used in the system.

    \item \textbf{Text preprocessing and semantic parsing} (Appendix~\ref{app:parsing_umbrella}).
    Covers requirement preprocessing, canonicalization, SMT programming,
    repair, and patient-side semantic parsing.

    \item \textbf{Database retrieval details} (Appendix~\ref{app:database_umbrella}).
    Explains how symbolic constraints are prepared, projected, and executed
    in the database.

    \item \textbf{Others} (Appendix~\ref{app:others_umbrella}).
    Includes:
    \begin{enumerate}[label=(\alph*), leftmargin=2em]
        \item \textbf{Limitations.}
        Discusses limitations of the current system.

        \item \textbf{Natural-language matching.}
        Describes the auxiliary natural-language matching component.

        \item \textbf{Example SMT program.}
        Provides an example SMT program produced by our semantic parser.

    \end{enumerate}

    \item \textbf{Prompts} (Appendix~\ref{app:prompts}).
    Lists the prompts used throughout the system.
\end{enumerate}

\vspace{1em}
\noindent

\startcontents[sections]
\printcontents[sections]{l}{1}{\setcounter{tocdepth}{2}}
\newpage


\section{Experimental Results and Evaluation}\label{app:experiment_umbrella}
\subsection{SIGIR 2016 Detailed Experimental Results}\label{app:full_results}

This appendix provides a comprehensive view of retrieval performance on the SIGIR 2016 patient--trial test collection across multiple evaluation settings and granularities. We first present the primary SIGIR 2016 evaluation results, including macro recall, pairwise patient-level comparisons, and per-patient visualizations, which together characterize both aggregate performance and the distribution of useful retrieved trials across patients. Throughout this appendix, a retrieved trial is \emph{useful} if it is relevant-and-eligible under the given retrieval objective (Section~\ref{sec:evaluation_methodology}). These analyses highlight how retrieval utility varies across objectives and baselines and where the gains of \name{} arise on SIGIR 2016.

We then report SIGIR 2016 results against the original benchmark labels, including weighted precision and recall under a shared retrieval budget. This setting provides a controlled evaluation that complements the primary results by isolating the effects of retrieval coverage and precision under standardized annotations.

\subsubsection{SIGIR 2016 Aggregate and Per-Patient Results}

\begin{table}[htbp]
\centering
\small
\setlength{\tabcolsep}{4pt}
\renewcommand{\arraystretch}{1.14}
\begin{tabular}{lccccccc}
\toprule
\textbf{Objective} & \textbf{BMRet} & \textbf{ClinBest} & \textbf{S-PubMedB} & \textbf{BGE} & \textbf{PubMedB} & \textbf{TG} & \textbf{\name{}} \\
\midrule
\modeccr & 50.49 & 51.13 & 64.68 & 68.81 & 66.18 & 70.14 & \textbf{93.68} \\
\modeall & 41.66 & 44.42 & 52.40 & 53.51 & 55.26 & 56.50 & \textbf{85.53} \\
\modeallexplore & 49.88 & 48.56 & 65.31 & 66.04 & 68.69 & 69.23 & \textbf{84.52} \\
\bottomrule
\end{tabular}
\caption{SIGIR 2016 results: macro recall against the union of relevant-and-eligible trials found by all compared methods. Per-patient recall is averaged within each objective. Subobjective rows are collapsed by column-wise maximum, and baseline variants are collapsed by row-wise maximum.}
\vspace{2pt}
{\raggedright\footnotesize
\textit{Abbreviations:} BMRet = BMRetriever; ClinBest = best clinician-keyword variant; S-PubMedB = SapBERT-PubMedBERT; BGE = BGE-large-en-v1.5; PubMedB = PubMedBERT; TG = TrialGPT.\par}
\label{tab:retrieval_recall_macro}
\end{table}

\paragraph{SIGIR 2016 macro recall.}
Table~\ref{tab:retrieval_recall_macro} shows macro recall on SIGIR 2016. It demonstrates that \name{} substantially outperforms all baselines across all three retrieval objectives. The improvement is particularly pronounced under the chief-complaint treating objective (\modeccr{}), where \name{} achieves near-complete coverage of the union of relevant-and-eligible trials found by any method. Similar trends hold for the any-condition treating (\modeall{}) and any-condition relevant (\modeallexplore{}) objectives, where \name{} consistently achieves large margins over both dense retrieval models and TrialGPT. Notably, even after selecting the best-performing TrialGPT variant across objective-aware and objective-agnostic prompting strategies and across GPT-4.1 and GPT-5, the gap remains significant. This indicates that the gains of \name{} on SIGIR 2016 are not attributable to baseline prompt configuration choices, but reflect a fundamental improvement in retrieval effectiveness.

\begin{figure}[!htbp]
\centering

\begin{subfigure}{\columnwidth}
\centering
\includegraphics[width=\columnwidth]{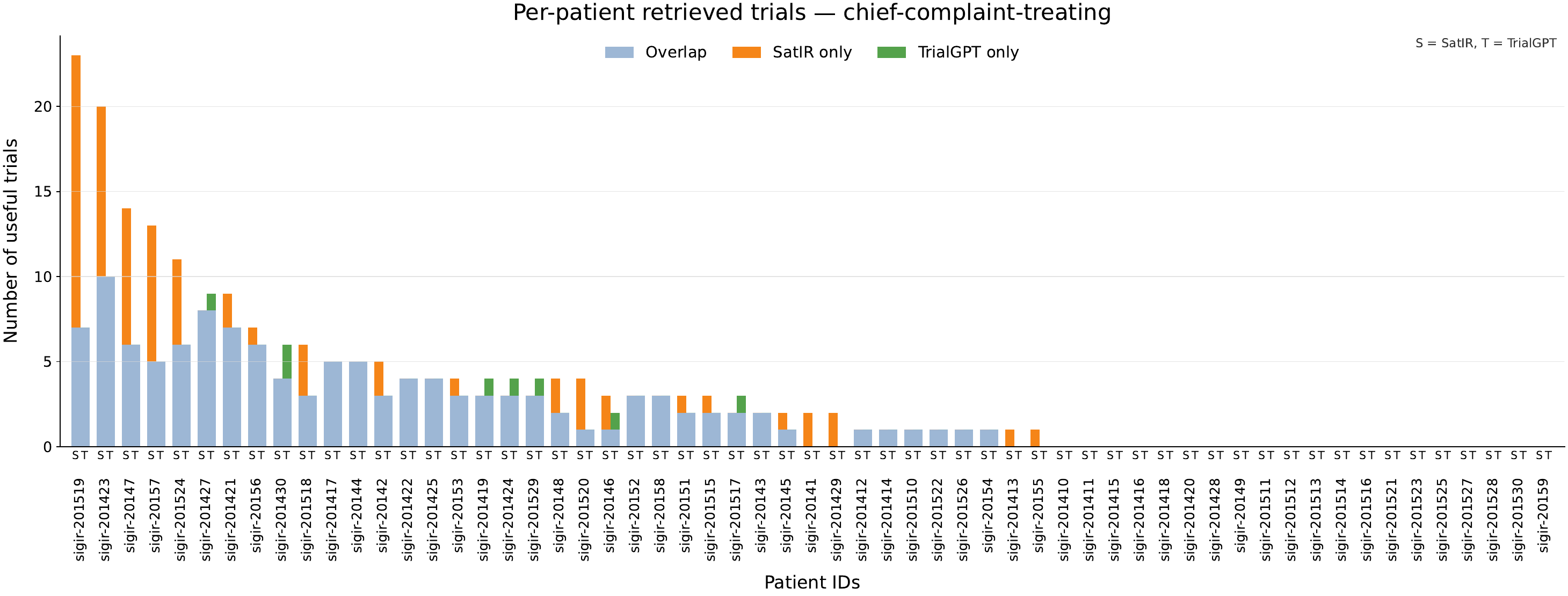}
\caption{\modeccr{}}
\label{fig:per-patient-chief-complaint-treating}
\end{subfigure}

\vspace{0.5em}

\begin{subfigure}{\columnwidth}
\centering
\includegraphics[width=\columnwidth]{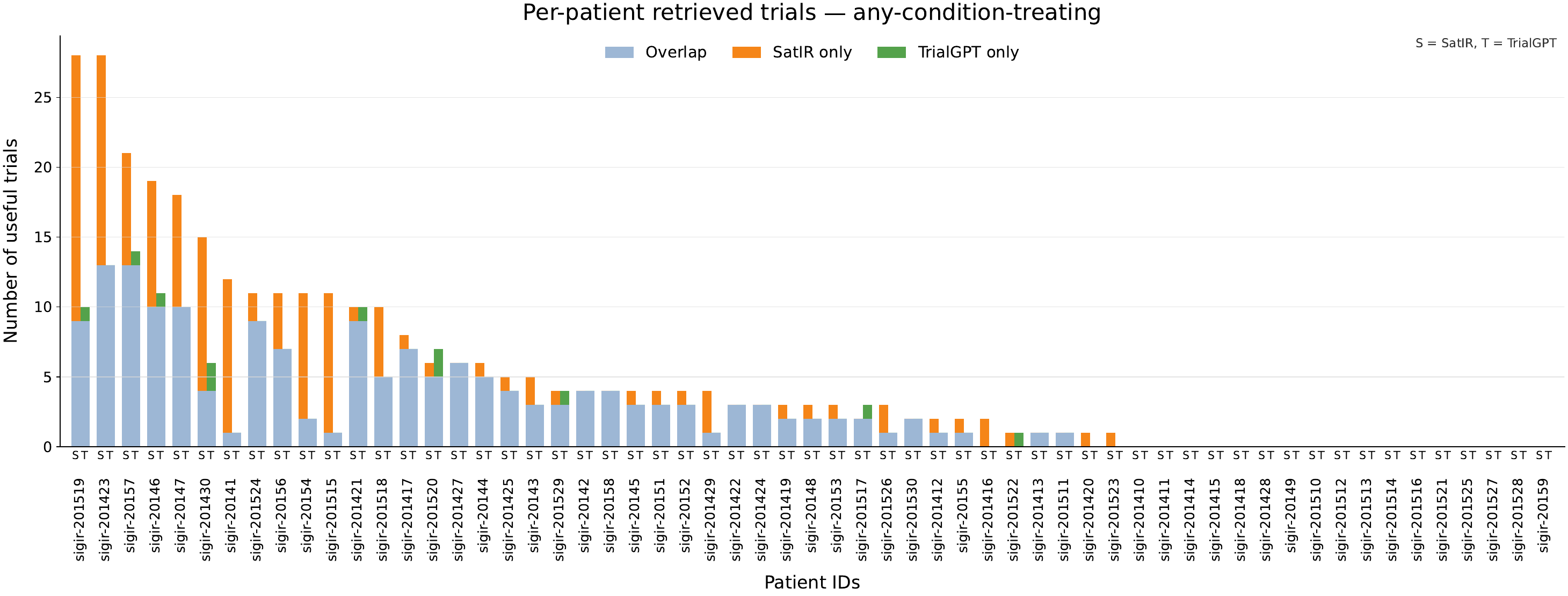}
\caption{\modeall{}}
\label{fig:per-patient-any-condition-treating}
\end{subfigure}

\vspace{0.5em}

\begin{subfigure}{\columnwidth}
\centering
\includegraphics[width=\columnwidth]{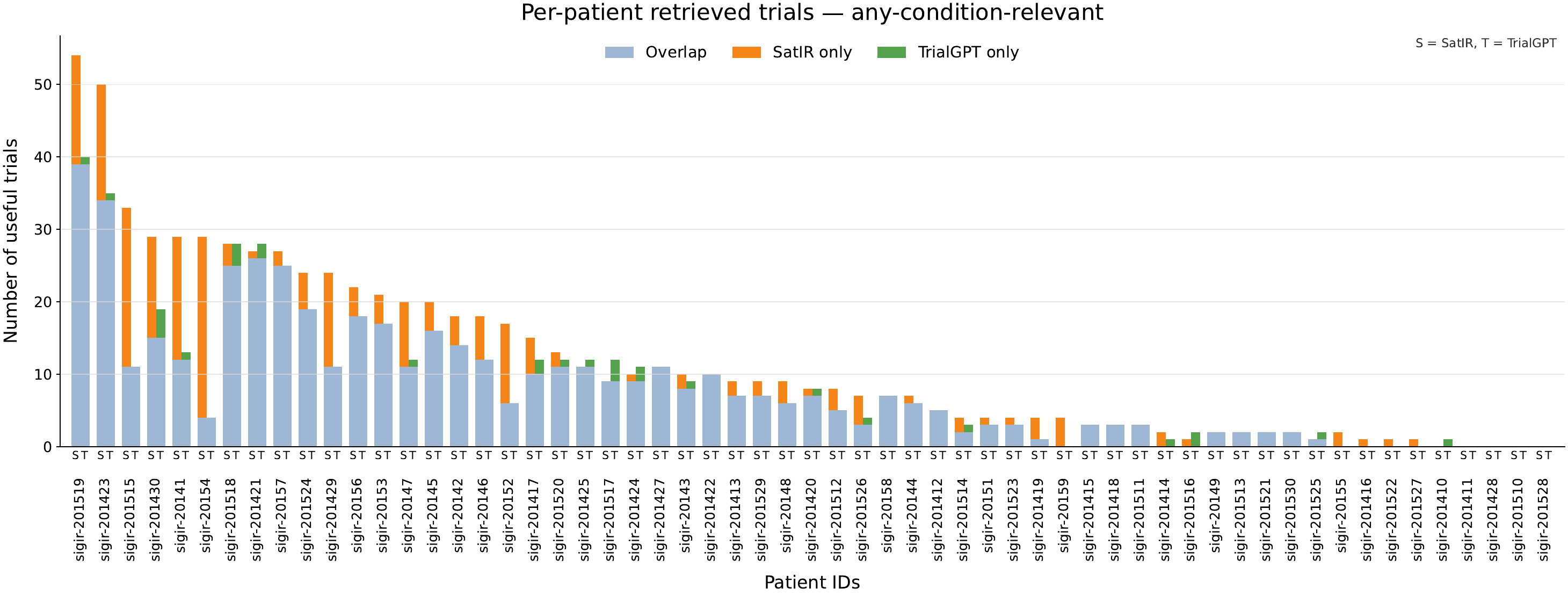}
\caption{\modeallexplore{}}
\label{fig:per-patient-any-condition-relevant}
\end{subfigure}

\caption{SIGIR 2016 per-patient comparison of the numbers of retrieved relevant-and-eligible trials under the three retrieval objectives between \name{} and TrialGPT with GPT-5.}
\label{fig:per-patient-results}
\end{figure}

\begin{figure*}[!htbp]
\centering

\begin{subfigure}[t]{0.32\textwidth}
\centering
\includegraphics[width=\linewidth]{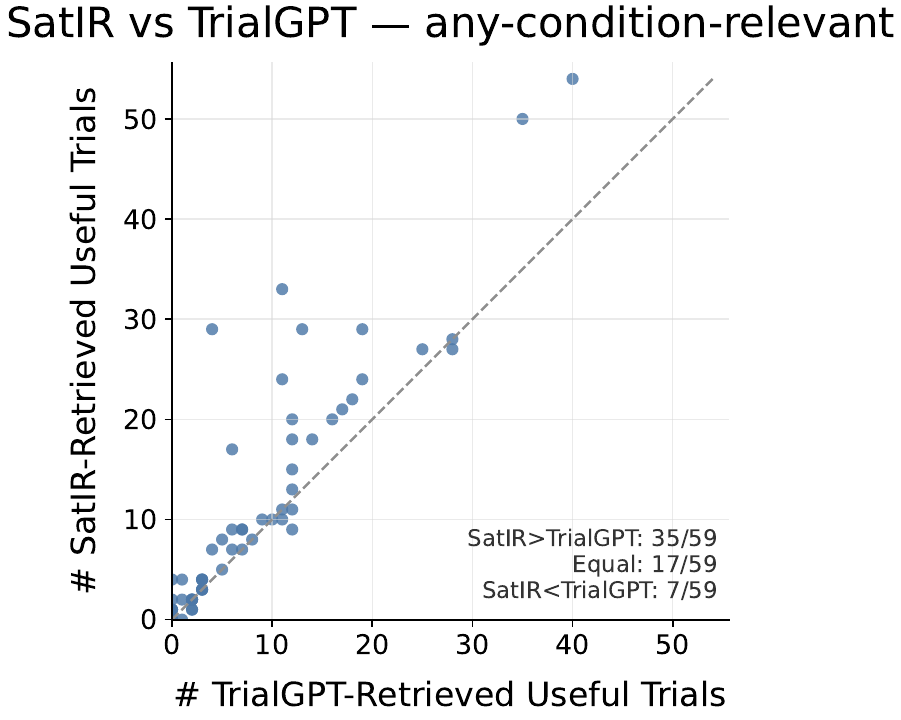}
\caption{\modeallexplore{}}
\label{fig:scatter-any-condition-relevant}
\end{subfigure}
\hfill
\begin{subfigure}[t]{0.32\textwidth}
\centering
\includegraphics[width=\linewidth]{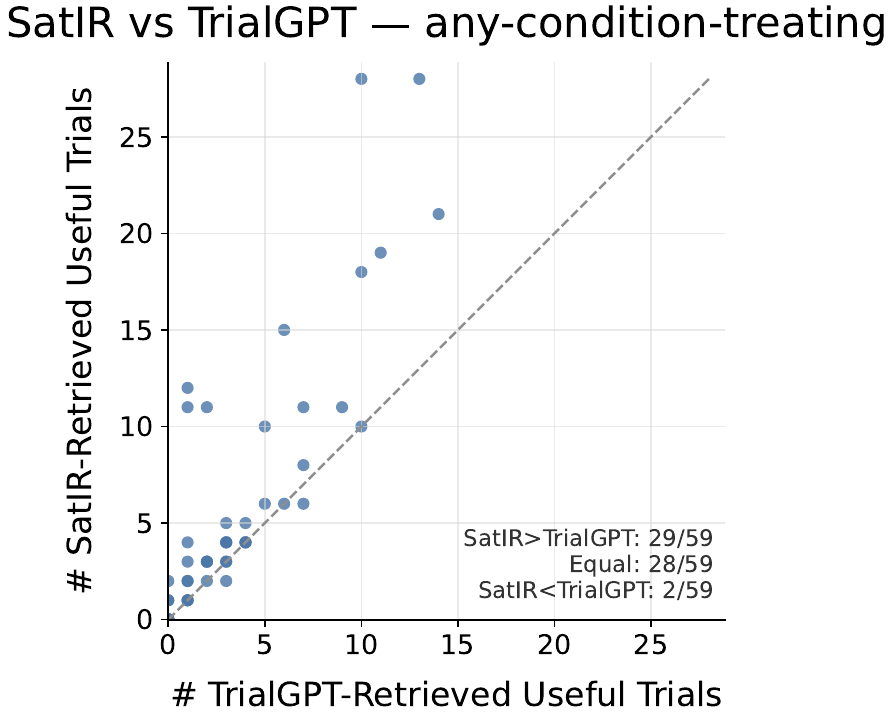}
\caption{\modeall{}}
\label{fig:scatter-any-condition-treating}
\end{subfigure}
\hfill
\begin{subfigure}[t]{0.32\textwidth}
\centering
\includegraphics[width=\linewidth]{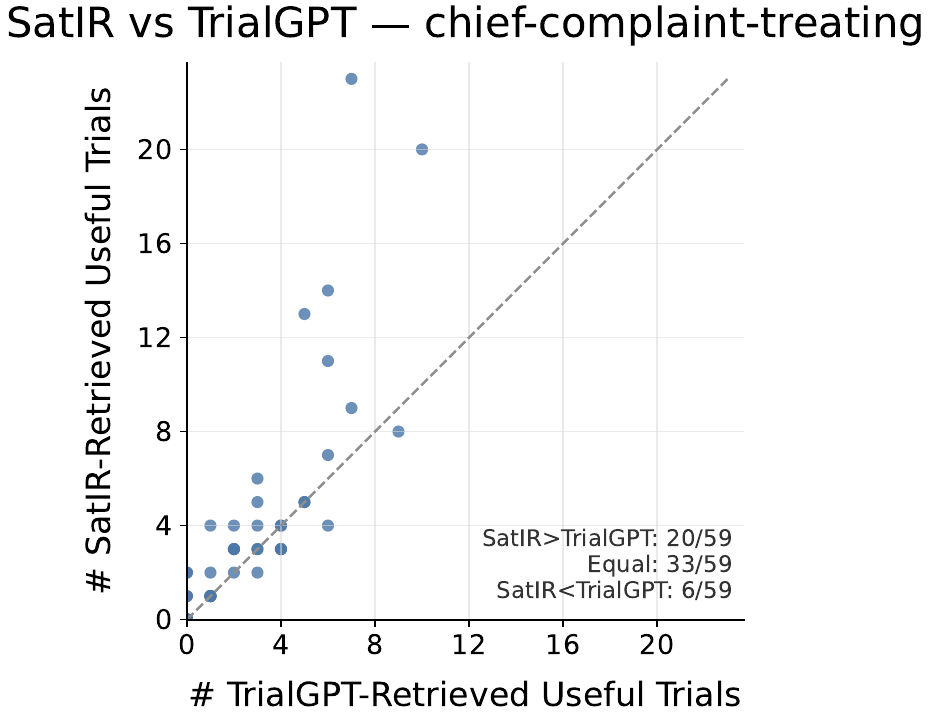}
\caption{\modeccr{}}
\label{fig:scatter-chief-complaint-treating}
\end{subfigure}

\caption{SIGIR 2016 per-patient scatter plots comparing TrialGPT retrieval with GPT-5 and \name{} under three retrieval objectives.}
\label{fig:per-patient-scatter-comparison}
\end{figure*}

\paragraph{SIGIR 2016 per-patient utility decomposition.}
Figure~\ref{fig:per-patient-results} decomposes, for each SIGIR 2016 patient, the useful retrieved trials into three disjoint subsets: trials retrieved by both methods, trials retrieved only by \name{}, and trials retrieved only by TrialGPT. \name{}'s useful retrieval count for a patient is the sum of the overlap and \name{}-only segments, while TrialGPT's count is the sum of the overlap and TrialGPT-only segments.

Across all three retrieval objectives, the \name{}-only segments are consistently more prominent than the TrialGPT-only segments, indicating that \name{} retrieves additional useful trials that TrialGPT does not. At the same time, the overlap is often substantial, showing that \name{} preserves many useful trials that TrialGPT also retrieves. The total bar height represents the union of useful trials retrieved by either method and should not be interpreted as the performance of a single system.

\paragraph{SIGIR 2016 pairwise per-patient comparison.}
Figure~\ref{fig:per-patient-scatter-comparison} compares the per-patient number of useful retrieved trials between TrialGPT and \name{} on SIGIR 2016. Each point corresponds to one patient; points above the diagonal indicate patients for whom \name{} retrieves more useful trials, points on the diagonal indicate ties, and points below the diagonal indicate patients for whom TrialGPT retrieves more useful trials.

Across all three objectives, most points lie above or on the diagonal, indicating that \name{} matches or exceeds TrialGPT for the majority of SIGIR 2016 patients. Points below the diagonal are relatively few and typically close to parity, suggesting that cases where TrialGPT retrieves more useful trials are limited.

\paragraph{SIGIR 2016 patients served.}
Table~\ref{tab:patients_served_appendix} reports the Patients-served outcome (patient coverage) on SIGIR 2016, where a patient is considered served if the method retrieves at least one relevant-and-eligible trial. This binary outcome complements the utility-winner comparison by measuring whether each method retrieves any useful option for a patient, rather than how many useful trials it retrieves.


\paragraph{Summary.}
Together, these SIGIR 2016 figures show that the improvements of \name{} arise from consistently retrieving additional useful trials across patients while retaining much of the shared set identified by TrialGPT. The gains are not driven by a small number of outlier patients, but reflect a broad shift in per-patient useful retrieval counts under all retrieval objectives.

\subsubsection{SIGIR Results Against the Original Benchmark Labels}
\label{app:dataset_faithful_results}

\paragraph{Why these results are limited.}
For completeness, we also report results against the original SIGIR 2016 benchmark labels. In that dataset, judged patient--trial pairs receive label 0 (not relevant), 1 (possible referral), or 2 (highly possible referral). However, the labels were obtained through pooled, budget-limited judging, so many pairs were never assessed; unjudged pairs therefore cannot be treated as negatives. In addition, these labels are not tied to an explicit retrieval objective like ours, which requires a trial to be both relevant under a stated objective and eligible for the patient. We therefore treat this section as a comparison to the original benchmark labels, not as our primary evaluation.

\paragraph{Precision against the original labels.}
Table~\ref{tab:annotated_precision_agg} reports weighted precision over retrieved patient--trial pairs that received original benchmark labels. Across all three objectives, \name{} achieves the highest precision. That is, among retrieved pairs that were judged in the benchmark, a larger fraction of \name{}'s results receive stronger original labels.

Relative to the baselines, including the strongest TrialGPT configuration, \name{} is therefore more selective: under the same retrieval budget, it concentrates more highly rated benchmark pairs in its returned set.

\begin{table}[htbp]
\centering
\small
\setlength{\tabcolsep}{3pt}
\renewcommand{\arraystretch}{1.16}
\begin{tabular}{lcccccccc}
\toprule
\textbf{Objective} & \textbf{Ret./pt.} & \textbf{ClinBest} & \textbf{TG} & \textbf{BGE} & \textbf{S-PubMedB} & \textbf{PubMedB} & \textbf{BMRet} & \textbf{\name{}} \\
\midrule
\modeccr & 36 & 0.289 & 0.328 & 0.332 & 0.330 & 0.337 & 0.351 & \textbf{0.400} \\
\modeall & 64 & 0.284 & 0.310 & 0.312 & 0.313 & 0.313 & 0.334 & \textbf{0.376} \\
\modeallexplore & 121 & 0.271 & 0.292 & 0.293 & 0.298 & 0.293 & 0.320 & \textbf{0.385} \\
\bottomrule
\end{tabular}
\caption{Weighted precision on retrieved patient--trial pairs with original benchmark labels (micro, normalized to $[0,1]$). Baselines are aggregated by family using row-wise maximum within each grouped family, and subcohort rows are collapsed by column-wise maximum. The highest value in each row is bolded.}
\label{tab:annotated_precision_agg}
\end{table}

\paragraph{Recall against the original labels.}
Table~\ref{tab:qrels_recall_agg} reports micro recall against the original benchmark labels, with label 1 and label 2 treated as positive judgments. Under this metric, \name{} does not achieve the highest recall; some dense-retrieval and TrialGPT-based baselines score higher.

This result should be interpreted with caution. Because the benchmark labels are pooled and incomplete, this recall measures recovery of the subset of pairs that happened to be judged, not of all truly relevant patient--trial pairs. It therefore does not directly measure our main task: retrieving trials that are both relevant under an explicit objective and eligible for the patient. Higher recall on the original 0/1/2 labels thus does not necessarily imply better performance for our setting.

Taken together, these results suggest that \name{} trades benchmark-label recall for higher selectivity, returning fewer judged pairs but concentrating more strongly labeled ones.

\begin{table}[htbp]
\centering
\small
\setlength{\tabcolsep}{3pt}
\renewcommand{\arraystretch}{1.16}
\begin{tabular}{lcccccccc}
\toprule
\textbf{Objective} & \textbf{Ret./pt.} & \textbf{BMRet} & \textbf{ClinBest} & \textbf{S-PubMedB} & \textbf{BGE} & \textbf{TG} & \textbf{PubMedB} & \textbf{\name{}} \\
\midrule
\modeccr & 36 & 0.296 & 0.463 & 0.475 & 0.500 & 0.490 & \textbf{0.502} & 0.420 \\
\modeall & 64 & 0.418 & 0.622 & 0.654 & 0.673 & \textbf{0.674} & 0.669 & 0.457 \\
\modeallexplore & 121 & 0.599 & 0.747 & 0.810 & 0.832 & \textbf{0.846} & 0.840 & 0.763 \\
\bottomrule
\end{tabular}
\caption{Micro recall against the original benchmark labels at the shared retrieval budget, with positive judgments weighted by label strength (1 and 2). Baselines are aggregated by family using row-wise maximum within each grouped family, and subcohort rows are collapsed by column-wise maximum. The highest value in each row is bolded.}
\label{tab:qrels_recall_agg}
\end{table}

\subsection{Patient-Level Confidence Intervals (SIGIR 2016)}\label{app:ci_tables}

Because all methods are evaluated on the same 59 SIGIR 2016 patients, the
appropriate inferential statistic for any \name{}-versus-baseline comparison
on this benchmark is the per-patient paired difference, which controls for
patient-level variation in disease, age, and trial-pool size. We report 95\%
patient-level cluster bootstrap CIs on paired differences (5{,}000 replicates)
for three SIGIR 2016 retrieval metrics---the mean useful trials retrieved per
patient and micro recall defined in Section~\ref{sec:evaluation_methodology},
and the macro recall reported in Table~\ref{tab:retrieval_recall_macro}---together with
patient-coverage counts and Gwet's AC1 for the clinician validation. The
external TREC 2022 results in Table~\ref{tab:trec2022_recall} are evaluated over a
separate set of 50 patient topics and 2{,}658 trials; the corresponding
paired per-patient CIs are reported in Appendix~\ref{app:trec_ci_tables}.

\paragraph{Methodology.}
For each (baseline family, retrieval objective, metric) cell, we select a
single baseline variant---the variant with the highest mean across
patients---and use it consistently for both the point estimate and the
bootstrap, matching the convention that populates the corresponding cells
of Tables~\ref{tab:retrieval_utility_main},
\ref{tab:retrieval_recall_micro}, and~\ref{tab:retrieval_recall_macro}.
Subobjective rows ($a$/$b$/$c$/$d$) are collapsed by column-wise maximum
on each patient. For each (method, baseline, retrieval objective) we then
form the vector of per-patient differences $\Delta_i = \mathrm{\name{}}_i -
\mathrm{baseline}_i$ over $N=59$ patients and bootstrap the patient index
with replacement (5{,}000 replicates). For micro recall, each bootstrap
replicate computes the numerator sum and denominator sum over the
resampled patient set before taking the ratio. For macro recall,
patient--objective pairs with empty ground-truth union are excluded from
the paired comparison as uninformative, reducing the effective $n$ to 39,
45, and 57 across the three objectives. For binomial clinician-validation
rates we report Wilson 95\% intervals; for Gwet's AC1 we report a
percentile bootstrap CI over the 27 dual-rated samples. For the binary
patients-served outcome in Table~\ref{tab:retrieval_patientlevel_main}
we report the exact McNemar test on discordant pairs.

\subsubsection{SIGIR 2016: Paired Per-Patient Yield Differences}

\begin{table}[htbp]
\centering
\scriptsize
\setlength{\tabcolsep}{3pt}
\renewcommand{\arraystretch}{1.14}
\begin{tabular}{lcccccc}
\toprule
\textbf{Objective} & \textbf{TG} & \textbf{BGE} & \textbf{PubMedB} & \textbf{S-PubMedB} & \textbf{ClinBest} & \textbf{BMRet} \\
\midrule
\modeccr        & +1.09 [0.44, 1.88] & +1.19 [0.49, 2.05] & +1.25 [0.46, 2.22] & +1.29 [0.54, 2.17] & +1.51 [0.83, 2.37] & +1.81 [1.02, 2.85] \\
\modeall        & +2.14 [1.22, 3.19] & +2.27 [1.29, 3.41] & +2.44 [1.41, 3.61] & +2.37 [1.37, 3.49] & +2.58 [1.61, 3.66] & +3.12 [1.88, 4.53] \\
\modeallexplore & +2.83 [1.56, 4.32] & +3.09 [1.73, 4.66] & +2.85 [1.58, 4.37] & +3.22 [1.85, 4.75] & +4.53 [3.12, 6.07] & +5.37 [3.58, 7.42] \\
\bottomrule
\end{tabular}
\caption{SIGIR 2016: per-patient paired difference in mean useful trials
retrieved, \name{} minus each baseline, with 95\% patient-level cluster
bootstrap CIs ($N=59$ patients, 5{,}000 replicates). All 18 paired
comparisons are significantly positive at 95\%; the smallest gap is
$+1.09$ trials (\modeccr{} vs.\ TG, CI~$[0.44, 1.88]$).}
\label{tab:yield_paired_ci}
\end{table}

\FloatBarrier

\subsubsection{SIGIR 2016: Paired Per-Patient Macro-Recall Differences}

\begin{table}[htbp]
\centering
\scriptsize
\setlength{\tabcolsep}{3pt}
\renewcommand{\arraystretch}{1.14}
\begin{tabular}{lcccccc}
\toprule
\textbf{Objective} & \textbf{TG} & \textbf{PubMedB} & \textbf{BGE} & \textbf{S-PubMedB} & \textbf{ClinBest} & \textbf{BMRet} \\
\midrule
\modeccr        & +23.5 [12.3, 35.7] & +27.5 [15.3, 40.2] & +24.9 [14.3, 36.3] & +29.0 [16.8, 41.6] & +42.6 [30.4, 54.5] & +43.2 [30.5, 55.5] \\
\modeall        & +29.0 [17.8, 39.8] & +30.3 [18.7, 41.4] & +32.0 [20.4, 43.4] & +33.1 [21.3, 44.3] & +41.1 [31.5, 51.1] & +43.9 [33.7, 53.9] \\
\modeallexplore & +15.3 [6.5, 24.2]  & +15.8 [6.8, 24.5]  & +18.5 [9.4, 27.4]  & +19.2 [10.0, 29.2] & +36.0 [26.0, 45.8] & +34.6 [25.1, 43.8] \\
\bottomrule
\end{tabular}
\caption{SIGIR 2016: per-patient paired difference in macro recall
(percentage points), \name{} minus each baseline, with 95\% patient-level
cluster bootstrap CIs. Per-comparison $n$ excludes patient--objective
pairs with empty ground-truth union (uninformative pairs), reducing $n$
to 39, 45, and 57 across the three objectives. All 18 paired
comparisons are significantly positive at 95\%; the smallest gap is
$+15.3$ pp (\modeallexplore{} vs.\ TG, CI~$[6.5, 24.2]$).}
\label{tab:recall_paired_ci}
\end{table}

\FloatBarrier

\subsubsection{SIGIR 2016: Paired Per-Patient Micro-Recall Differences}

\begin{table}[htbp]
\centering
\scriptsize
\setlength{\tabcolsep}{3pt}
\renewcommand{\arraystretch}{1.14}
\begin{tabular}{lcccccc}
\toprule
\textbf{Objective} & \textbf{TG} & \textbf{PubMedB} & \textbf{BGE} & \textbf{S-PubMedB} & \textbf{ClinBest} & \textbf{BMRet} \\
\midrule
\modeccr        & +31.2 [16.0, 43.1] & +36.1 [17.8, 49.7] & +34.2 [18.8, 46.7] & +37.1 [21.3, 49.4] & +43.4 [30.9, 56.0] & +52.2 [37.2, 63.6] \\
\modeall        & +38.4 [27.8, 47.3] & +43.9 [32.7, 53.1] & +40.9 [29.6, 50.5] & +42.7 [32.1, 51.6] & +46.3 [36.4, 55.9] & +56.1 [45.7, 64.5] \\
\modeallexplore & +22.3 [14.0, 30.7] & +22.4 [14.3, 30.4] & +24.3 [15.7, 33.2] & +25.3 [16.8, 33.7] & +35.6 [26.5, 45.6] & +42.3 [34.0, 49.8] \\
\bottomrule
\end{tabular}
\caption{SIGIR 2016: per-patient paired difference in micro recall
(percentage points), \name{} minus each baseline, with 95\% patient-level
cluster bootstrap CIs. Each bootstrap replicate recomputes both methods'
micro recall on the resampled patient set before differencing. All 18
paired comparisons are significantly positive at 95\%; the smallest gap
is $+22.3$ pp (\modeallexplore{} vs.\ TG, CI~$[14.0, 30.7]$).}
\label{tab:micro_recall_paired_ci}
\end{table}
\FloatBarrier

\subsubsection{SIGIR 2016: Per-Patient Head-to-Head and Coverage}

Table~\ref{tab:retrieval_patientlevel_main} reports two complementary
patient-level comparisons. The Utility-winner columns compare the number of
relevant-and-eligible trials retrieved for each patient: \name{} wins when it
retrieves more such trials than TrialGPT, TrialGPT wins under the reverse
comparison, and ties indicate equal yield. The Patients-served columns instead
measure binary coverage, indicating whether each method retrieves at least one
relevant-and-eligible trial for the patient. \name{} wins substantially more
patients on utility under all three objectives, while the coverage counts are
much closer because both systems serve most of the same patients.

\begin{table}[htbp]
\centering
\footnotesize
\setlength{\tabcolsep}{3pt}
\renewcommand{\arraystretch}{1.15}
\begin{tabular}{lccc|cccc}
\toprule
& \multicolumn{3}{c|}{\textbf{Utility winner}}
& \multicolumn{4}{c}{\textbf{Patients served}} \\
\cmidrule(lr){2-4}
\cmidrule(lr){5-8}
\textbf{Objective}
& \textbf{\name{}}
& \textbf{Tie}
& \textbf{TG}
& \textbf{\name{} only}
& \textbf{Both}
& \textbf{TG only}
& \textbf{Neither} \\
\midrule
\modeccr        & 20 & 33 & 6 & 4 & 35 & 0 & 20 \\
\modeall        & 29 & 28 & 2 & 3 & 39 & 0 & 17 \\
\modeallexplore & 35 & 17 & 7 & 5 & 49 & 1 & 4 \\
\bottomrule
\end{tabular}
\caption{SIGIR 2016 patient-level retrieval outcomes over 59 patients,
comparing \name{} with TrialGPT using GPT-5. Utility winner indicates which
method retrieves more relevant-and-eligible trials for a patient. Patients
served indicates whether each method retrieves at least one such trial.
The utility-winner advantage is significant under all three objectives by
exact sign test ($p<0.01$); patient-coverage outcomes are descriptive.}
\label{tab:retrieval_patientlevel_main}
\end{table}

\FloatBarrier

For the per-patient head-to-head ``Utility winner'' counts in
Table~\ref{tab:retrieval_patientlevel_main} we use the exact sign test on
discordant patient pairs. \name{} wins significantly more patients than
TrialGPT under all three objectives ($20{:}6$, $29{:}2$, $35{:}7$; exact
sign test $p<0.01$ in all cases). For the binary served-vs-not-served
outcome we report the corresponding discordant counts and the exact
McNemar test:

\begin{table}[htbp]
\centering
\small
\setlength{\tabcolsep}{4pt}
\renewcommand{\arraystretch}{1.18}
\begin{tabular}{lccccc}
\toprule
\textbf{Objective} & \textbf{\name{} only} & \textbf{Both} & \textbf{TG only} & \textbf{Neither} & \textbf{Exact McNemar $p$} \\
\midrule
\modeccr        & 4 & 35 & 0 & 20 & 0.125 \\
\modeall        & 3 & 39 & 0 & 17 & 0.250 \\
\modeallexplore & 5 & 49 & 1 & 4  & 0.219 \\
\bottomrule
\end{tabular}
\caption{SIGIR 2016: patient-coverage breakdown and exact McNemar tests on
discordant pairs ($N=59$ patients per objective). \name{} matches TrialGPT
on patient-coverage outcomes on this benchmark---the differential is not
statistically separated under the matched-patient analysis. The
inferentially significant retrieval gains on SIGIR 2016 are on yield
(Table~\ref{tab:yield_paired_ci}), macro recall
(Table~\ref{tab:recall_paired_ci}), micro recall
(Table~\ref{tab:micro_recall_paired_ci}), and the per-patient head-to-head
Utility-winner counts in Table~\ref{tab:retrieval_patientlevel_main}.}
\label{tab:patients_served_appendix}
\end{table}

\FloatBarrier

\FloatBarrier

\subsubsection{Clinician Validation of the GPT-5 Judge}

The clinician validation in Section~\ref{sec:evaluation_methodology} audits
the GPT-5 judge that produces the labels underlying every SIGIR 2016 metric
in Tables~\ref{tab:yield_paired_ci}--\ref{tab:micro_recall_paired_ci} and
the objective-specific-judge TREC 2022 metric in Table~\ref{tab:trec2022_recall}.
We report Gwet's AC1 with a bootstrap CI alongside Wilson intervals for
the acceptance rates; the entire AC1 interval lies at or above the
substantial-agreement threshold under the Landis--Koch convention,
consistent with the strong agreement reported in
Section~\ref{sec:evaluation_methodology}.

\begin{table}[htbp]
\centering
\footnotesize
\setlength{\tabcolsep}{8pt}
\renewcommand{\arraystretch}{1.18}
\begin{tabular}{lccl}
\toprule
\textbf{Statistic} & \textbf{$n$} & \textbf{Point} & \textbf{95\% CI} \\
\midrule
Clinician A acceptance rate & 27 & 92.6\% & [76.6, 97.9] (Wilson) \\
Clinician B acceptance rate & 27 & 88.9\% & [71.9, 96.1] (Wilson) \\
Both-accept rate            & 27 & 85.2\% & [67.5, 94.1] (Wilson) \\
Either-accept rate          & 27 & 96.3\% & [81.7, 99.3] (Wilson) \\
Gwet's AC1                  & 27 & 0.82   & [0.65, 0.96] (bootstrap, 5{,}000 reps) \\
\bottomrule
\end{tabular}
\caption{Clinician validation 95\% confidence intervals over the 27
dual-rated samples (SIGIR 2016 patient--trial pairs). The Gwet's AC1
interval lies entirely at or above the substantial-agreement threshold
under the Landis--Koch convention.}
\label{tab:clinician_ci}
\end{table}

\FloatBarrier

\paragraph{Summary.}
The CI analysis supports four classes of conclusion.
\textit{(i) Statistically confirmed on SIGIR 2016 at the per-patient level.}
All 18 paired differences in mean useful trials retrieved
(Table~\ref{tab:yield_paired_ci}), all 18 paired differences in macro
recall (Table~\ref{tab:recall_paired_ci}), and all 18 paired differences
in micro recall (Table~\ref{tab:micro_recall_paired_ci}) are significantly
positive at 95\%, giving 54 of 54 paired comparisons in favor of \name{}
on SIGIR 2016. The smallest gap across all 54 cells is $+1.09$ useful
trials per patient (\modeccr{} vs.\ TG, CI~$[0.44, 1.88]$); the smallest
recall gap is $+15.3$ pp macro recall (\modeallexplore{} vs.\ TG,
CI~$[6.5, 24.2]$). The per-patient head-to-head Utility-winner counts in
Table~\ref{tab:retrieval_patientlevel_main} are likewise significant under
the exact sign test ($p<0.01$ in all three objectives), confirming that the
average gains in Table~\ref{tab:retrieval_utility_main} reflect a broad
per-patient improvement rather than a few high-yield outliers.
\textit{(ii) Matched under the SIGIR 2016 paired analysis.} The binary
patient-coverage outcome ties: discordant counts of $4{:}0$, $3{:}0$, and
$5{:}1$ yield exact McNemar $p>0.1$ in all three objectives
(Table~\ref{tab:patients_served_appendix}). We accordingly do not claim a
patient-coverage advantage in the main paper.
\textit{(iii) External generalization on TREC 2022.} The qualitative
direction and magnitude of the SIGIR 2016 result are reproduced on the
independent TREC 2022 corpus, including against official eligibility-aware
human qrels (Section~\ref{sec:results},
Table~\ref{tab:trec2022_recall}), with paired per-patient CIs in
Appendix~\ref{app:trec_ci_tables}.
\textit{(iv) Clinician--judge agreement.}
Gwet's AC1 is $0.82$ with bootstrap CI~$[0.65, 0.96]$, entirely at or
above the substantial-agreement threshold.

\subsection{TREC 2022 Detailed Experimental Results}\label{app:trec_full_results}

This appendix provides a detailed view of retrieval performance on \textsc{TREC-2022-RetrievalSubset}. We first report aggregate recall and average retrieved useful trials per patient under the objective-specific GPT-5 judge protocol, paralleling the SIGIR 2016 analyses in Appendix~\ref{app:full_results}. We then report recall stratified by the official TREC qrel levels, separating eligibility-aware relevance from topical relevance alone. Finally, we report component ablations isolating the contribution of SNOMED-grounded patient-side ontology entailment and salience-based missingness handling.

\subsubsection{TREC 2022 Aggregate Results}

\paragraph{TREC 2022 macro recall and average retrieved per patient.}
Table~\ref{tab:trec_recall_macro_avg} reports aggregate retrieval results on \textsc{TREC-2022-RetrievalSubset} under the objective-specific GPT-5 judge protocol. We report both macro recall against the union of relevant-and-eligible trials found by any method and the average number of relevant-and-eligible trials retrieved per patient. \name{} substantially outperforms BM25 and TG under all three retrieval objectives. Macro recall gaps of roughly 50--62 percentage points are observed across objectives, indicating that the gains are not driven only by micro-averaging over a small number of high-yield patients.

\begin{table}[htbp]
\centering
\footnotesize
\setlength{\tabcolsep}{4pt}
\renewcommand{\arraystretch}{1.2}
\begin{tabular}{lccccccc}
\toprule
& & \multicolumn{3}{c}{\textbf{Macro recall (\%)}} & \multicolumn{3}{c}{\textbf{Avg retrieved / patient}} \\
\cmidrule(lr){3-5} \cmidrule(l){6-8}
\textbf{Objective} & \textbf{$k$} & \textbf{BM25} & \textbf{TG} & \textbf{\name{}} & \textbf{BM25} & \textbf{TG} & \textbf{\name{}} \\
\midrule
\modeccr & 20 & 25.17 & 24.72 & \textbf{80.37} & 0.50 & 0.70 & \textbf{1.86} \\
\modeall & 46 & 29.48 & 33.93 & \textbf{85.01} & 1.00 & 1.14 & \textbf{2.44} \\
\modeallexplore & 50 & 23.92 & 27.59 & \textbf{86.17} & 1.68 & 2.00 & \textbf{5.40} \\
\bottomrule
\end{tabular}
\caption{Aggregate retrieval results on \textsc{TREC-2022-RetrievalSubset} under the objective-specific GPT-5 judge protocol. Macro recall is averaged per patient against the union of relevant-and-eligible trials found by any method. Average retrieved per patient counts judge-confirmed relevant-and-eligible trials in each method's top-$k$. The union size averages 2.28, 3.00, and 6.38 trials per patient under \modeccr{}, \modeall{}, and \modeallexplore{}, respectively. For each objective, $k$ is set to \name{}'s average number of retrieved trials.}
\vspace{2pt}
{\raggedright\footnotesize}
\label{tab:trec_recall_macro_avg}
\end{table}

\paragraph{TREC 2022 comparison to BM25 and TG.}
The aggregate results show the same pattern as the SIGIR 2016 evaluation: \name{} retrieves substantially more relevant-and-eligible trials than the neural or keyword-based baselines under all three objectives. The largest absolute gain in average retrieved useful trials appears under \modeallexplore{}, where \name{} retrieves 5.40 relevant-and-eligible trials per patient compared with 2.00 for TG and 1.68 for BM25. The macro-recall results show that these gains are broadly distributed across patients rather than being driven only by a few patients with many eligible trials.

\subsubsection{TREC 2022 Recall Stratified by Official Qrel Level}

To provide a broader comparison against the official TREC human judgments,
we additionally report recall under two qrel targets. The primary
eligibility-aware target counts only qrel$=2$ trials, which are judged both
relevant and eligible. For completeness, we also report recall over all
trials with qrel$=1/2$, which measures topical relevance regardless of
eligibility.

\begin{table}[htbp]
\centering
\footnotesize
\setlength{\tabcolsep}{4pt}
\renewcommand{\arraystretch}{1.15}
\begin{tabular}{llcrrr}
\toprule
\textbf{Target} & \textbf{Objective} & \textbf{$k$}
& \textbf{BM25} & \textbf{TG} & \textbf{\name{}} \\
\midrule
Relevant-and-eligible
& \modeccr & 20 & 0.162 & 0.139 & \textbf{0.445} \\
Relevant-and-eligible
& \modeall & 46 & 0.245 & 0.260 & \textbf{0.480} \\
Relevant-and-eligible
& \modeallexplore & 50 & 0.252 & 0.268 & \textbf{0.539} \\
\midrule
All relevant
& \modeccr & 20 & 0.201 & 0.172 & \textbf{0.357} \\
All relevant
& \modeall & 46 & 0.277 & 0.300 & \textbf{0.383} \\
All relevant
& \modeallexplore & 50 & 0.285 & 0.313 & \textbf{0.429} \\
\bottomrule
\end{tabular}
\caption{Recall on \textsc{TREC-2022-RetrievalSubset} against official TREC
human qrels, stratified by qrel target. The primary relevant-and-eligible
target counts qrel$=2$ trials. For
completeness, the broader target counts all qrel$=1$ or qrel$=2$ trials and
therefore measures relevance regardless of eligibility. For each objective,
$k$ is set to the average number of trials retrieved by \name{} on this
dataset.}
\label{tab:trec_qrel_strat}
\end{table}

\paragraph{Eligibility-aware retrieval.}
Under the primary qrel$=2$ target, \name{} improves over the strongest
baseline by 0.283 for \modeccr{}, 0.220 for \modeall{}, and 0.271 for
\modeallexplore{}. These eligibility-aware gains are larger than the
corresponding gains under the broader qrel$=1/2$ target. The supplementary
qrel$=1/2$ results show that \name{} also improves recovery of topically
relevant trials, while the larger qrel$=2$ gains indicate that its advantage
is particularly pronounced for trials judged both relevant and eligible.

\subsubsection{TREC 2022 Component Ablations}\label{app:trec_ablations}

To isolate the contribution of individual design choices, we ran two component ablations on the same TREC 2022 setting: 50 patients, the same retrieval pipeline, the same GPT-5 judge protocol, and a shared patient--trial pair cache. Both ablations use isolated database copies; the canonical pipeline is unchanged. For each ablation, we set the per-patient rank cutoff $K_p$ to the size of the unablated system's \texttt{all\_satisfied} set for that patient. Each ablated condition then counts true positives in the top-$K_p$ of its own \texttt{all\_satisfied} set. Thus, the comparison asks whether an ablated system can recover the same number of true positives under the same per-patient output budget. Average $K_p$ is 17.7 for \modeccr{}, 39.9 for \modeall{}, and 44.6 for \modeallexplore{}.

\paragraph{Ablation 1: SNOMED ontology grounding.}
\name{}'s enriched patient facts are produced by SNOMED ISA, finding--observable-entity, and finding--procedure expansions applied to raw coded patient facts. To ablate ontology grounding, we replace the enriched facts with the raw root coded facts and re-ingest patient inclusion constraints. This changes the patient-side fact database from 48{,}845 enriched fact lines across 49 patients to 2{,}578 raw coded fact lines, an 18.9$\times$ reduction. Trial-side components and the salience layer are unchanged.

\begin{table}[htbp]
\centering
\small
\setlength{\tabcolsep}{4pt}
\renewcommand{\arraystretch}{1.08}
\begin{tabular}{@{}lrrrrrr@{}}
\toprule
& \multicolumn{3}{c}{GPT-5 judge} & \multicolumn{3}{c}{TREC qrel$=2$} \\
\cmidrule(lr){2-4} \cmidrule(l){5-7}
Objective & \name{} & $-$Ont. & $\Delta$ & \name{} & $-$Ont. & $\Delta$ \\
\midrule
\modeccr & 94 & 44 & $-$50 ($-$53.2\%) & 173 & 85 & $-$88 ($-$50.9\%) \\
\modeall & 119 & 78 & $-$41 ($-$34.5\%) & 192 & 124 & $-$68 ($-$35.4\%) \\
\modeallexplore & 270 & 157 & $-$113 ($-$41.9\%) & 215 & 128 & $-$87 ($-$40.5\%) \\
\midrule
\textbf{Total} & 483 & 279 & $-$204 ($-$42.2\%) & 580 & 337 & $-$243 ($-$41.9\%) \\
\bottomrule
\end{tabular}
\caption{TREC 2022 patient-side SNOMED ontology ablation. Disabling ontology entailment reduces true positives by 34.5--53.2\% under the objective-specific GPT-5 judge and by 35.4--50.9\% under official qrel$=2$ labels. Each condition is evaluated at the same per-patient output budget $K_p$, defined as the size of the unablated system's \texttt{all\_satisfied} set for that patient.}
\label{tab:trec_ablation_ontology}
\end{table}

\paragraph{Ablation 2: Salience-based missingness handling.}
\name{}'s salience layer assigns default values to selected variables when the patient note is silent. These defaults allow the constraint solver to reject trials whose inclusion clauses require evidence that is absent from the patient summary. To ablate salience, we empty \name{}'s \texttt{default\_predicate\_values} table, changing it from 1{,}190 rows to 0, and re-run retrieval. Without salience, trials that previously failed inclusion clauses through salience defaults are no longer rejected, so the \texttt{all\_satisfied} set grows by 8--21\% per objective. At fixed $K_p$, these additional satisfied trials enter the ranking and displace some originally top-ranked eligible trials beyond $K_p$.

\begin{table}[htbp]
\centering
\small
\setlength{\tabcolsep}{4pt}
\renewcommand{\arraystretch}{1.08}
\begin{tabular}{@{}lrrrrrr@{}}
\toprule
& \multicolumn{3}{c}{GPT-5 judge} & \multicolumn{3}{c}{TREC qrel$=2$} \\
\cmidrule(lr){2-4} \cmidrule(l){5-7}
Objective & \name{} & $-$Sal. & $\Delta$ & \name{} & $-$Sal. & $\Delta$ \\
\midrule
\modeccr & 94 & 87 & $-$7 ($-$7.4\%) & 173 & 162 & $-$11 ($-$6.4\%) \\
\modeall & 119 & 108 & $-$11 ($-$9.2\%) & 192 & 178 & $-$14 ($-$7.3\%) \\
\modeallexplore & 270 & 251 & $-$19 ($-$7.0\%) & 215 & 199 & $-$16 ($-$7.4\%) \\
\midrule
\textbf{Total} & 483 & 446 & $-$37 ($-$7.7\%) & 580 & 539 & $-$41 ($-$7.1\%) \\
\bottomrule
\end{tabular}
\caption{TREC 2022 salience ablation. Disabling \name{}'s salience-based default values reduces true positives by 7.0--9.2\% under the objective-specific GPT-5 judge and by 6.4--7.4\% under official qrel$=2$ labels. Each condition is evaluated at the same per-patient output budget $K_p$, defined as the size of the unablated system's \texttt{all\_satisfied} set for that patient.}
\label{tab:trec_ablation_salience}
\end{table}

\paragraph{Summary of ablations.}
The ablations show that both ontology grounding and salience-based missingness handling contribute to \name{}'s TREC 2022 recall gains. Ontology grounding is the larger contributor: removing patient-side ontology entailment reduces true positives by 34.5--53.2\% across objectives under the objective-specific GPT-5 judge. Salience-based missingness handling provides a smaller but consistent contribution: removing salience defaults reduces true positives by 7.0--9.2\% across objectives under the objective-specific GPT-5 judge. The same qualitative pattern holds when true positives are measured using official qrel$=2$ labels.

The remaining design factors discussed in Section~\ref{sec:results} are not isolated in these ablations. Representation fidelity is instead supported by examples and failure analysis, and SMT-to-relational projection is supported by the formal projection argument and the measured query time reported in Section~\ref{sec:results}.

\subsection{Patient-Level Confidence Intervals (TREC 2022)}
\label{app:trec_ci_tables}

For methodological consistency with the SIGIR 2016 analysis in
Appendix~\ref{app:ci_tables}, we additionally report 95\% patient-level
cluster bootstrap CIs on paired per-patient differences for the TREC 2022
results in Table~\ref{tab:trec2022_recall}. Because TREC 2022 has two
baselines, BM25 and TrialGPT, we report $2\times 3=6$ comparisons per
metric.

\paragraph{Methodology.}
For each baseline, retrieval objective, and metric on TREC 2022, we form the
vector of per-patient differences
$\Delta_i=\mathrm{\name{}}_i-\mathrm{baseline}_i$
over $N=50$ patient topics and bootstrap the patient index with replacement
for 5{,}000 replicates. For the objective-specific GPT-5 judge protocol,
patient--objective pairs with an empty ground-truth union are excluded as
uninformative, reducing the effective $n$ to 31, 38, and 47 across the three
objectives. For official TREC qrels, we evaluate on patients with non-empty
ground truth under the selected qrel target, giving $n=48$ for qrel$=2$ and
$n=50$ for qrel$=1/2$. The retrieval cutoff matches the main-paper
evaluation: $k=20$, $46$, and $50$ for
\modeccr{}, \modeall{}, and \modeallexplore{}, respectively.

\subsubsection{TREC 2022: Paired Per-Patient Differences against Official
TREC Qrels}

\begin{table}[htbp]
\centering
\footnotesize
\setlength{\tabcolsep}{4pt}
\renewcommand{\arraystretch}{1.18}
\begin{tabular}{lcccc}
\toprule
& \multicolumn{2}{c}{\textbf{Relevant-and-eligible (qrel$=2$)}}
& \multicolumn{2}{c}{\textbf{All relevant (qrel$=1/2$)}} \\
\cmidrule(lr){2-3}
\cmidrule(l){4-5}
\textbf{Objective}
& \textbf{vs.\ BM25}
& \textbf{vs.\ TG}
& \textbf{vs.\ BM25}
& \textbf{vs.\ TG} \\
\midrule
\modeccr
& +28.7 [16.8, 40.9]
& +30.9 [20.4, 41.6]
& +15.9 [8.7, 23.5]
& +18.8 [11.4, 26.9] \\
\modeall
& +23.8 [11.5, 36.0]
& +22.2 [9.8, 34.7]
& +13.1 [5.0, 21.6]
& +10.8 [3.5, 18.3] \\
\modeallexplore
& +29.2 [17.2, 41.2]
& +27.6 [14.9, 40.3]
& +17.2 [8.2, 26.1]
& +14.4 [6.4, 22.3] \\
\bottomrule
\end{tabular}
\caption{TREC 2022 per-patient paired differences in macro recall against
the official TREC human qrels, in percentage points, with 95\%
patient-level cluster bootstrap confidence intervals. The primary
relevant-and-eligible setting uses qrel$=2$ trials. For completeness, the
broader all-relevant setting uses all qrel$=1$ or
qrel$=2$ trials and therefore measures relevance regardless of eligibility.
Each value is \name{} minus the corresponding baseline. Patients with no
ground-truth trials under the selected target are excluded, giving $n=48$
for qrel$=2$ and $n=50$ for qrel$=1/2$. All 12 paired comparisons are
significantly positive at 95\%.}
\label{tab:trec_qrel_paired_ci}
\end{table}

\FloatBarrier

\subsubsection{TREC 2022: Paired Per-Patient Macro-Recall Differences (Objective-Specific GPT-5 Judge)}

\begin{table}[htbp]
\centering
\footnotesize
\setlength{\tabcolsep}{6pt}
\renewcommand{\arraystretch}{1.18}
\begin{tabular}{lcc}
\toprule
\textbf{Objective} & \textbf{vs.\ BM25} & \textbf{vs.\ TG} \\
\midrule
\modeccr & +55.2 [31.2, 76.3] & +55.7 [35.7, 73.7] \\
\modeall & +55.5 [35.3, 73.8] & +51.1 [32.9, 67.7] \\
\modeallexplore & +62.2 [47.2, 75.4] & +58.6 [44.8, 71.3] \\
\bottomrule
\end{tabular}
\caption{TREC 2022: per-patient paired difference in macro recall under the
objective-specific GPT-5 judge protocol (percentage points), \name{} minus each
baseline, with 95\% patient-level cluster bootstrap CIs.
Per-comparison $n$ excludes patient--objective pairs with empty
ground-truth union (uninformative): $n=31, 38, 47$ for
\modeccr/\modeall/\modeallexplore. All 6 paired comparisons are
significantly positive at 95\%; the smallest gap is $+51.1$ pp
(\modeall{} vs.\ TG, CI~$[32.9, 67.7]$).}
\label{tab:trec_macro_recall_paired_ci}
\end{table}




\FloatBarrier

\paragraph{Summary.}
The TREC 2022 confidence-interval analysis confirms that the
external-generalization claims in Section~\ref{sec:results} are
statistically robust. All six paired yield comparisons under the objective-specific
GPT-5 judge, all six objective-specific macro-recall comparisons, and all six
eligibility-aware qrel$=2$ comparisons are significantly positive at 95\%.
For completeness, all six supplementary comparisons under the broader
qrel$=1/2$ target are also significantly positive.

The primary TREC evidence therefore comprises 18 statistically significant
paired comparisons spanning useful-trial yield, objective-specific-judge
recall, and recall against official eligibility-aware human judgments. The
additional qrel$=1/2$ analysis shows that the advantage also extends to
broader topical relevance, although the larger gains under qrel$=2$
indicate that the improvement is especially pronounced for trials judged
both relevant and eligible.

\newpage

\normalsize

    \subsection{Retrieval Objectives}
\label{app:retrieval_objectives}

This appendix defines the retrieval objectives used in our system and explains how relevance is determined under each one. The objectives differ in how narrowly they connect a trial to the patient's clinical needs, from a tight focus on the chief complaint to a broader notion of clinical relatedness.

\paragraph{Objective 1: Chief-Complaint Treating (\modeccr).}
This is the strictest objective. A trial is relevant only if it is aimed at treating the patient's chief complaint, a meaningful higher-level form of that complaint, or an important direct cause whose treatment would substantially relieve it. The standard is intentionally narrow: the condition must be one a patient would realistically search for in order to address the chief complaint.

This objective also includes trials that improve procedures the patient is expected to undergo for treating the chief complaint, but only when the improvement supports treatment of the complaint itself rather than secondary issues such as side effects. Preventive trials are included only when they target diseases explicitly mentioned in the patient note or near-universal downstream outcomes of an existing condition, and only when those diseases are not already present. 

\paragraph{Objective 2: Any Condition Treating (\modeall).}
This objective broadens the scope from the chief complaint to any condition the patient currently has and might reasonably want treated. A trial is relevant if it aims to treat such a condition and the condition is explicitly supported by the patient note rather than inferred.

As in the first objective, preventive trials are included only for explicitly stated risks or near-universal downstream outcomes of existing conditions. Compared with the chief-complaint objective, this objective retrieves a wider set of treatment opportunities while still requiring clear therapeutic intent.

\paragraph{Objective 3: Any Condition Relevant (\modeallexplore).}
This is the broadest objective. A trial is relevant if it is meaningfully related to any condition the patient currently has or is being evaluated for, even if the trial is not directly treating that condition.

To keep this broad objective clinically meaningful, relevance must be based on the trial's real clinical target, not on superficial overlap such as shared symptoms, care settings, or diagnostic workup. The patient must clearly have, or be under evaluation for, the condition targeted by the trial; speculative upgrades to a more specific diagnosis are not allowed.

\paragraph{Summary.}
The three objectives form a spectrum. The first is narrowly focused on treating the chief complaint. The second expands to treating any condition the patient has. The third further broadens the scope to include clinically related trials, while still enforcing safeguards against weak or spurious matches. Together, these objectives let us study different precision--recall trade-offs in clinical trial retrieval. The prompt implementations of the three retrieval objectives can be found in Appendix \ref{app:patient-parsing/relevance-judge}.
\newpage
    \subsection{Clinician Review}
\subsubsection{User Interface for Clinician Annotation} \label{app:clinician_annotation}

\begin{figure*}[!htbp]
  \centering
  \includegraphics[width=0.8\linewidth]{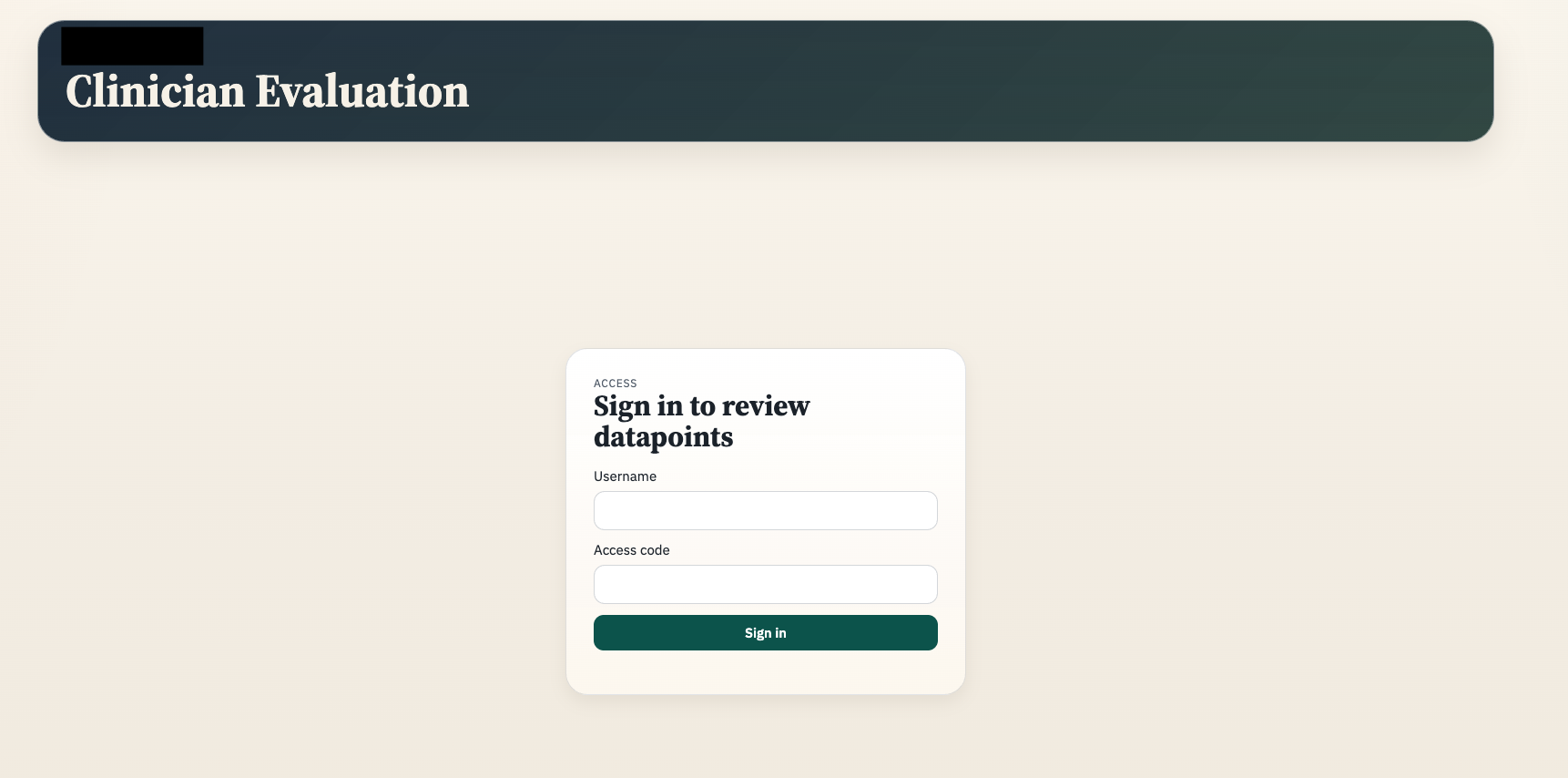}
    \caption{Clinician annotation interface screenshots (1/5): Login Page}
  \label{fig:interface-login}
\end{figure*}

\begin{figure*}[!htbp]
  \centering
  \includegraphics[width=0.8\linewidth]{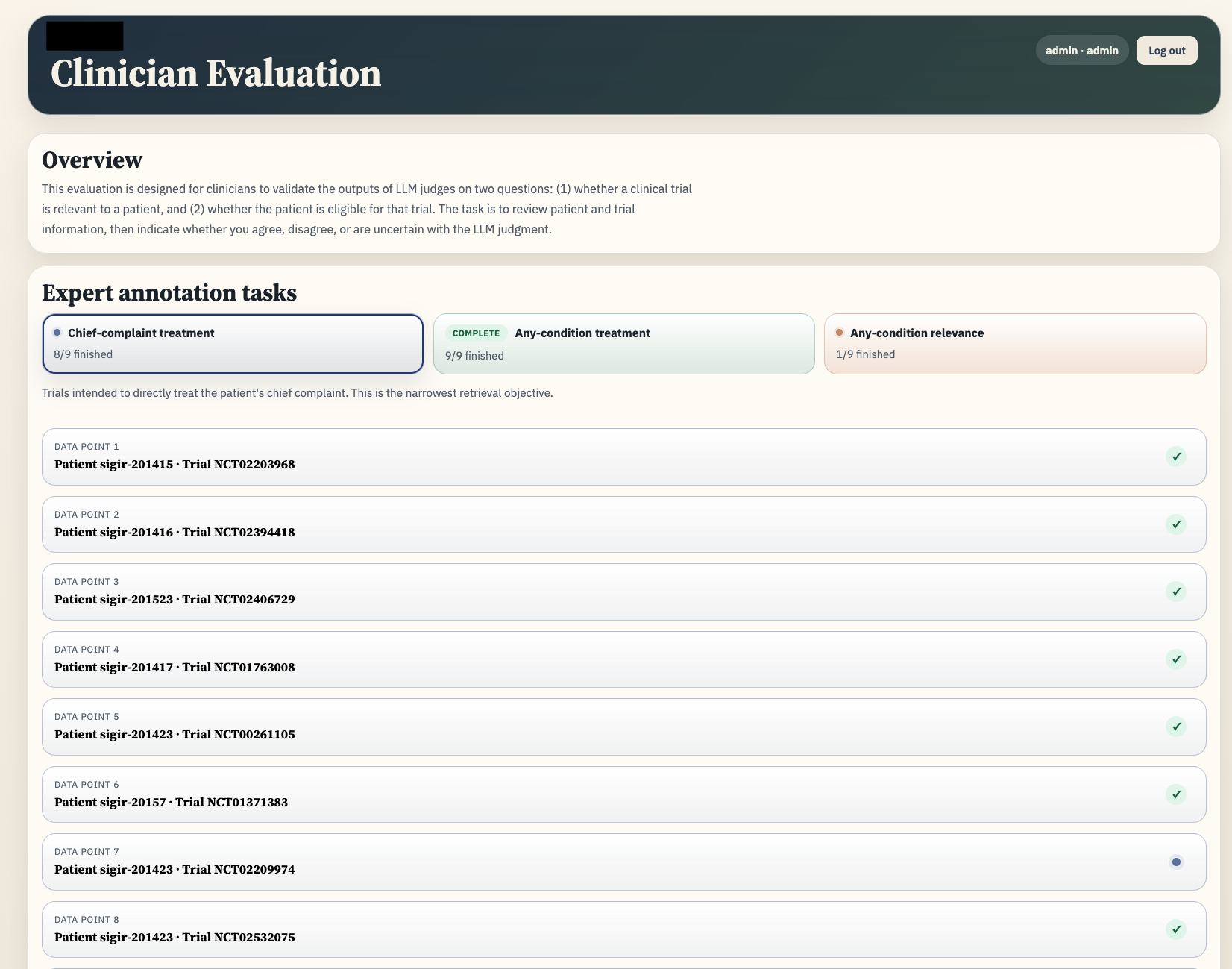}
    \caption{Clinician annotation interface screenshots (2/5): Overview Page}
  \label{fig:interface-overview}
\end{figure*}

\begin{figure*}[!htbp]
  \centering
  \includegraphics[width=0.8\linewidth]{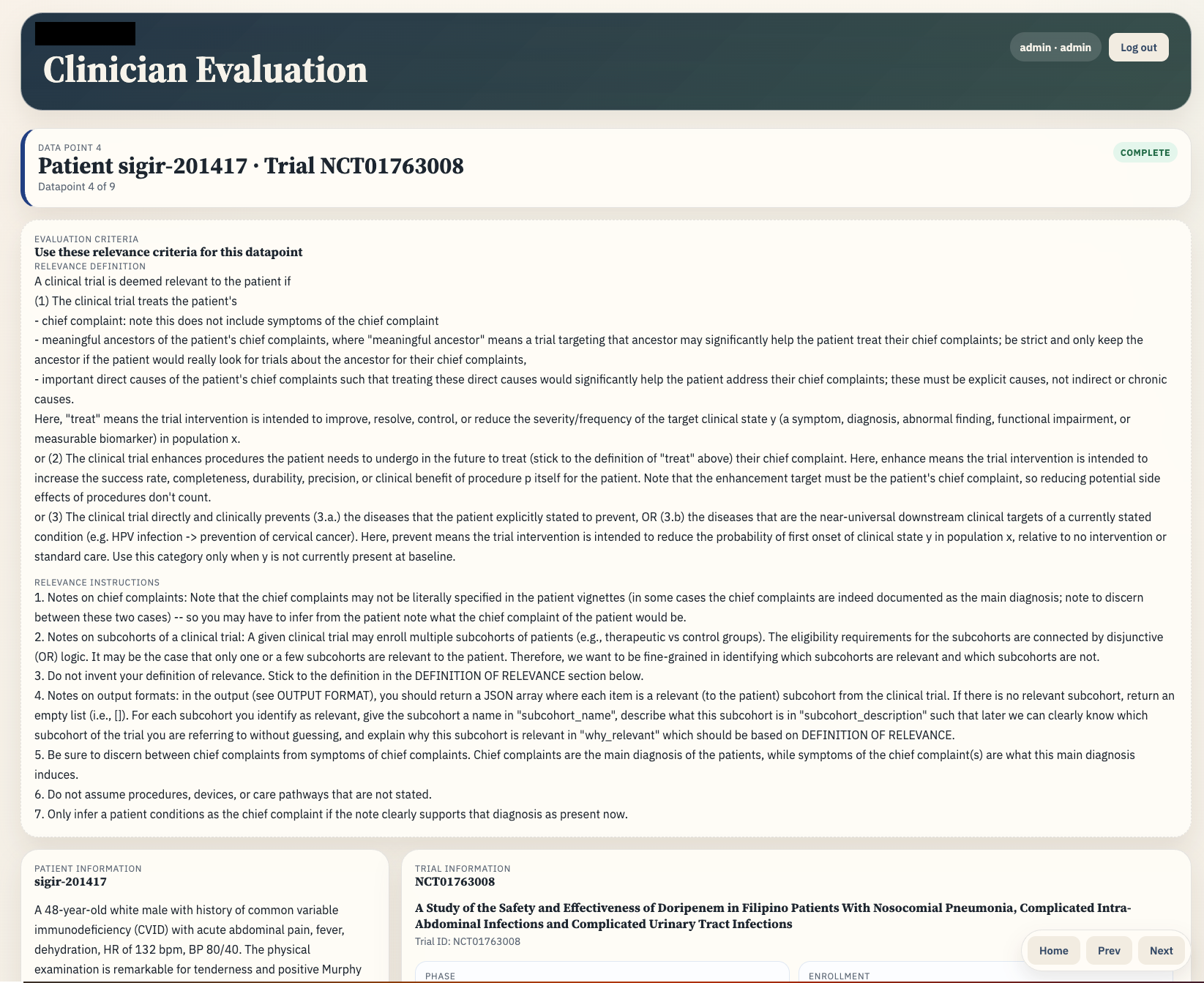}
    \caption{Clinician annotation interface screenshots (3/5): Data Point Page (1/3)}
  \label{fig:interface-datapoint1}
\end{figure*}

\begin{figure*}[!htbp]
  \centering
  \includegraphics[width=0.8\linewidth]{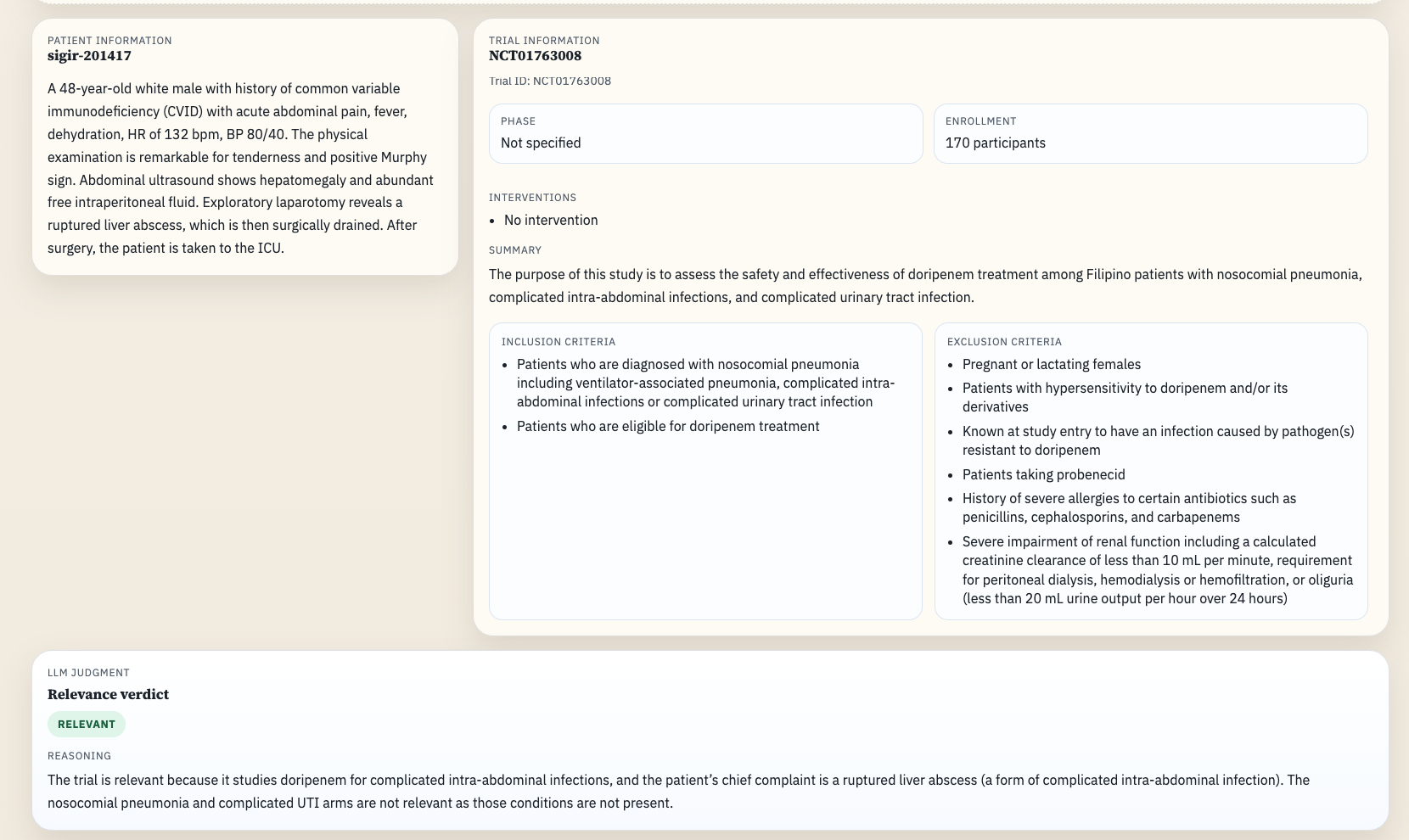}
    \caption{Clinician annotation interface screenshots (4/5): Data Point Page (2/3)}
  \label{fig:interface-datapoint2}
\end{figure*}

\begin{figure*}[!htbp]
  \centering
  \includegraphics[width=0.8\linewidth]{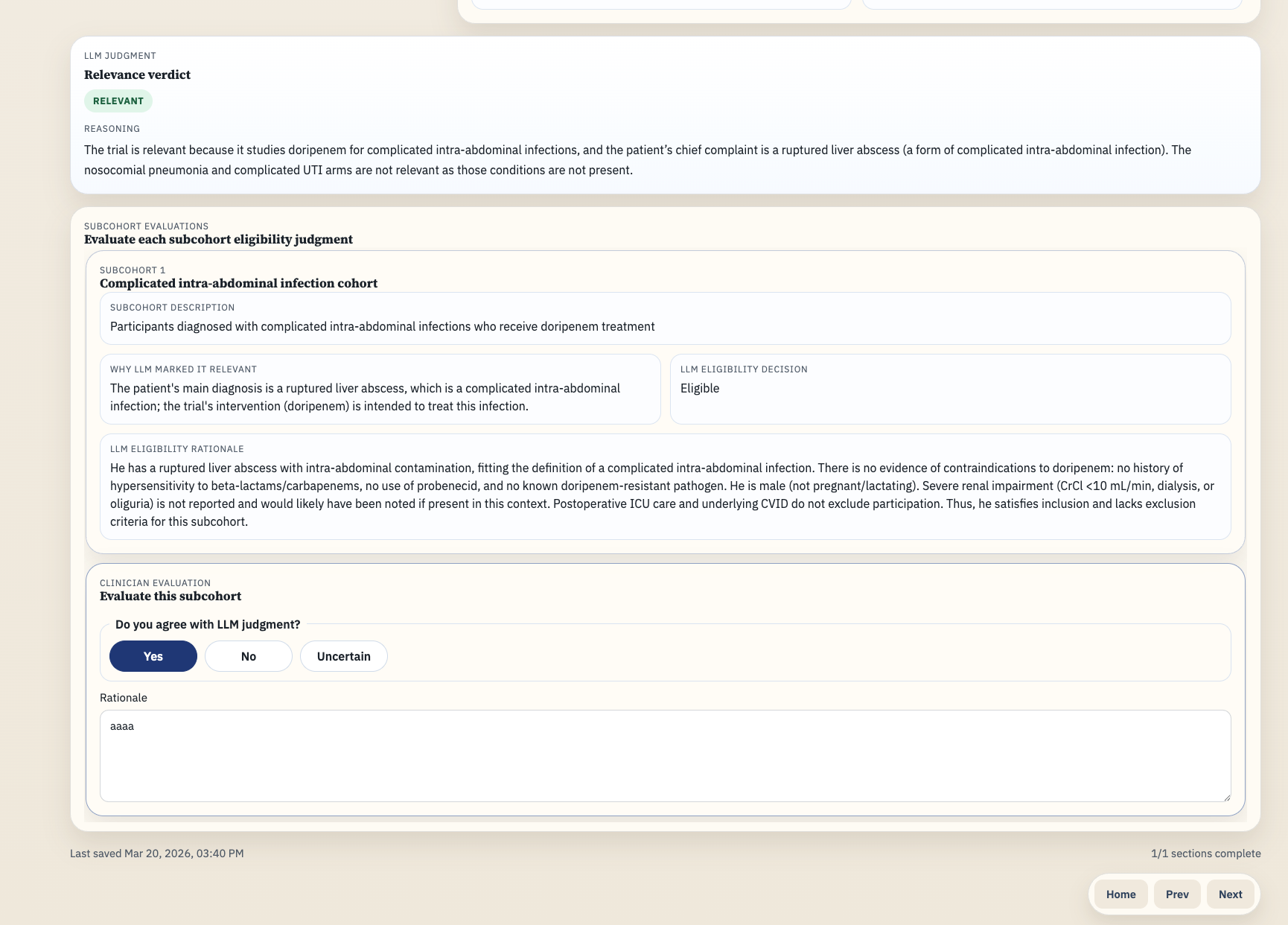}
    \caption{Clinician annotation interface screenshots (5/5): Data Point Page (3/3)}
  \label{fig:interface-datapoint3}
\end{figure*}

In this section, we show screenshots (Figure~\ref{fig:interface-login}, \ref{fig:interface-overview}, \ref{fig:interface-datapoint1}, \ref{fig:interface-datapoint2}, \ref{fig:interface-datapoint3}) of the user interface we provide to clinicians for their annotation. Clinicians are presented with a patient note, a clinical trial description, and the predefined relevance criteria, along with a pre-existing relevance judgment and its associated rationale. Their task is to evaluate this provided judgment. Clinicians are asked to indicate whether they \textit{accept} (agree with), \textit{reject} (disagree with), or are \textit{uncertain} about the given decision, and to provide a brief explanation supporting their assessment. This design allows us to validate both the correctness of the decision and the quality of its justification, while maintaining consistency across annotations. The clinicians' annotation results are presented in Appendix~\ref{app:clinician_review_table}.

\subsubsection{Full Clinician Review Details}
\label{app:clinician_review_table}

\paragraph{Physician-validation sampling.}
To validate the GPT-5 judge, we constructed a physician-review sample by stratified random sampling from the union of judged patient--trial pairs produced by \name{} and TrialGPT. The sampling unit was a tuple \((m,p,t)\), where \(m\) denotes the retrieval objective (\texttt{\modeccr}, \texttt{\modeall}, or \texttt{\modeallexplore}), \(p\) the patient, and \(t\) the trial. We first took the union of all judged tuples appearing in either system's \texttt{patient\_labels} outputs, collapsing duplicates across systems. Each union item was then assigned to one of three judgment buckets according to the GPT-5 judge outcome: \textit{relevant-and-eligible}, \textit{relevant but ineligible}, or \textit{irrelevant}. We then applied balanced allocation: the total sample was distributed as evenly as possible across retrieval objectives and, within each objective, as evenly as possible across the three judgment buckets. If a stratum contained fewer items than its target allocation, it was capped at its population size and the remaining quota was redistributed to other strata with available capacity. Sampling within each stratum was uniform at random with a fixed random seed.

Table~\ref{tab:clinician_review_full} reports the full set of clinician adjudication examples used in our qualitative review. These results are collected through the annotation workflow described in Section~\ref{sec:evaluation_methodology} and the user interface presented in Appendix~\ref{app:clinician_annotation}. Each row corresponds to one patient--trial sample and includes the retrieval objective, the relevant subcohort(s) identified for that trial when applicable, the corresponding eligibility determination, the relevance rationale shown to the reviewer, and the clinicians' judgments of subcohort(s) eligibility.

\begingroup
\tiny
\setlength{\LTleft}{0pt}
\setlength{\LTright}{0pt plus 1fill}
\setlength{\tabcolsep}{1.5pt}
\renewcommand{\arraystretch}{1.08}


\begin{longtable}{
  L{0.25cm}  
  L{0.80cm}  
  L{1.30cm}  
  L{1.00cm}  
  L{2.20cm}  
  L{2.50cm}  
  L{2.50cm}  
  L{1.20cm}  
  L{1.20cm}  
}
\caption{Full clinician review results for all 27 samples} \label{tab:clinician_review_full} \\
\toprule
\textbf{\#} &
\textbf{Patient} &
\textbf{Trial} &
\textbf{Objective} &
\textbf{Relevant Subcohort(s)} &
\textbf{Subcohort(s) Eligibility} &
\textbf{Relevance Rationale} &
\textbf{Clin.\ A} &
\textbf{Clin.\ B} \\
\midrule
\endfirsthead

\multicolumn{9}{c}{\tablename\ \thetable\ -- continued from previous page} \\
\toprule
\textbf{\#} &
\textbf{Patient} &
\textbf{Trial} &
\textbf{Objective} &
\textbf{Relevant Subcohort(s)} &
\textbf{Subcohort(s) Eligibility} &
\textbf{Relevance Rationale} &
\textbf{Clin.\ A} &
\textbf{Clin.\ B} \\
\midrule
\endhead

\midrule
\multicolumn{9}{r}{Continued on next page} \\
\endfoot

\bottomrule
\endlastfoot

1 & sigir-201412 & NCT00287144 & \modeall{} & N/A & N/A & Not relevant. The trial is an observational study assessing long-term risk of hypothyroidism after prior postpartum thyroiditis and does not offer any treatment or prevention. The patient's current complaints suggest hypothyroidism/goiter needing therapy, which this study does not address. & accept \newline \textit{Done.} & accept \\
\addlinespace[3pt]
2 & sigir-201412 & NCT00439127 & \modeall{} & N/A & N/A & Not relevant. The patient's current complaints (fatigue, cold intolerance, hair loss, weight gain) and a uniform midline thyroid enlargement suggest hypothyroidism with diffuse goiter, not confirmed differentiated thyroid carcinoma. The trial focuses on post-thyroidectomy management of histologically proven DTC, evaluating thyroglobulin as a prognostic marker before radioiodine ablation. It does not aim to treat hypothyroidism, goiter, or her symptomatic complaints, nor does it prevent a condition she has stated to avoid. Therefore, no subcohort in this trial treats her present issues. & accept \newline \textit{Done.} & accept \\
\addlinespace[3pt]
3 & sigir-201416 & NCT01689350 & \modeall{} & N/A & N/A & Not relevant. The patient's acute neurologic illness with hydrophobia and dysphagia suggests rabies encephalitis; her complaints are neurologic/spasmodic symptoms. The trial targets systemic lupus erythematosus treatment (genotype-guided cyclophosphamide) and does not treat or prevent rabies or the patient's current complaints. No subcohort addresses her condition. & accept \newline \textit{Done.} & accept \\
\addlinespace[3pt]
4 & sigir-20156 & NCT00677469 & \modeall{} & \textbf{Adjunct cholestyramine for newly diagnosed Graves hyperthyroidism} --- Patients with newly diagnosed Graves-related hyperthyroidism receiving low-dose cholestyramine as adjunctive therapy to reduce thyrotoxicosis --- \textit{Treats the patient's current hyperthyroid/thyrotoxic symptoms by aiming to reduce thyroid hormone levels and control symptoms.} & \textbf{Adjunct cholestyramine for newly diagnosed Graves hyperthyroidism} --- Decision: eligible --- Symptoms and signs (exophthalmos, hyperreflexia, tachycardia, weight loss with increased appetite) strongly suggest Graves' hyperthyroidism. She appears newly diagnosed and there is no indication of prior treatment. No diabetes, kidney, or liver disease is mentioned. She therefore meets inclusion and has no exclusions. & The patient has symptomatic thyrotoxicosis with features suggestive of Graves disease (weight loss despite increased appetite, tachycardia, hyperreflexia, mild exophthalmia). This study evaluates cholestyramine as adjunctive therapy to treat hyperthyroidism by enhancing thyroid hormone clearance, directly targeting her current condition and symptoms. & accept \newline \textit{Done.} & accept \\
\addlinespace[3pt]
5 & sigir-20157 & NCT00963196 & \modeall{} & \textbf{MDD outpatients on antidepressants randomized to adjunctive omega-3 vs placebo} --- Adults (18+) diagnosed with Major Depressive Disorder receiving standard antidepressant therapy, randomized to receive either fish oil capsules containing EPA/DHA (omega-3) or placebo for 12 weeks, with symptom monitoring. --- \textit{The patient currently has depressive symptoms consistent with major depressive disorder (fatigue, anhedonia, increased sleep/appetite, impaired concentration, guilt). This trial is a treatment study targeting MDD symptoms via antidepressant therapy and adjunctive omega-3, aiming to improve symptom control and reduce time to clinical improvement.} & \textbf{MDD outpatients on antidepressants randomized to adjunctive omega-3 vs placebo} --- Decision: eligible --- She is 20 years old and has several months of symptoms consistent with a current Major Depressive Disorder episode (anhedonia, hypersomnia, increased appetite, fatigue, impaired concentration, guilt). There is no evidence of exclusion criteria such as multiple failed antidepressant trials, substance dependence/use, psychosis, bipolar disorder, neurological disorders, recent ECT, unstable medical illness, pregnancy, or imminent suicide risk. Antidepressant therapy can be provided as part of the study, and adjunctive omega-3 vs placebo is appropriate. Contraception requirements for women of reproductive age can be met prior to enrollment. English-speaking and ability to complete study assessments are presumed adequate from context. & The patient currently has depressive symptoms consistent with major depressive disorder (fatigue, anhedonia, increased sleep/appetite, impaired concentration, guilt). This trial is a treatment study targeting MDD symptoms via antidepressant therapy and adjunctive omega-3, aiming to improve symptom control and reduce time to clinical improvement. & accept \newline \textit{Done.} & accept \\
\addlinespace[3pt]
6 & sigir-20157 & NCT02491307 & \modeall{} & \textbf{Ginger.io smartphone intervention arm for mood disorders} --- Adults (18+) with a mood disorder such as depression who receive the Ginger.io smartphone-based platform to enhance communication, monitoring, and proactive outreach within a community behavioral health setting. --- \textit{She has current depressive symptoms (fatigue, anhedonia, hypersomnia, poor concentration, guilt), and this intervention is intended to treat mood disorders like depression by improving clinical and behavioral health outcomes.} & \textbf{Ginger.io smartphone intervention arm for mood disorders} --- Decision: eligible --- Age $\geq$18; clinical features consistent with a current depressive episode (a qualifying mood disorder). No exclusion criteria. Smartphone possession and English literacy are inferred/assumable, and the provider/app operational requirement can be arranged per guidelines. & The patient's presentation is consistent with a current depressive episode. This study's intervention (Ginger.io smartphone-based platform) is designed to treat mood disorders, including depression, by enhancing monitoring and provider outreach to improve clinical outcomes, making the intervention arm relevant to her current complaints. & accept \newline \textit{Done.} & accept \\
\addlinespace[3pt]
7 & sigir-201417 & NCT01147640 & \modeall{} & \textbf{cIAI subcohort---Intraabdominal abscess (liver abscess) requiring surgical intervention} --- Adults with complicated intraabdominal infection due to an intraabdominal abscess, including liver abscess, who require surgical source control (e.g., laparotomy or drainage) in proximity to study drug initiation. --- \textit{The patient has a ruptured liver abscess that was surgically drained, constituting a complicated intraabdominal infection; this study compares IV antibiotic regimens specifically intended to treat cIAI including liver abscesses.} & \textbf{cIAI subcohort---Intraabdominal abscess (liver abscess) requiring surgical intervention} --- Decision: ineligible --- He meets core inclusion criteria (age 48; cIAI due to a liver abscess; source control achieved via laparotomy and drainage), and surgery can be timed relative to drug initiation. However, he has common variable immunodeficiency (CVID), which constitutes an immunocompromising illness and is excluded by the protocol. No clear evidence of other exclusions (e.g., renal failure <50 mL/min CrCl, acute hepatic failure, prior carbapenem use, antibiotic hypersensitivity) is present. Although he was hypotensive and tachycardic, septic shock is not definitively documented; nevertheless, the immunocompromised state alone renders him ineligible. & The trial is directly relevant because it evaluates antibiotic treatment for complicated intraabdominal infections, explicitly including intraabdominal abscesses such as liver abscesses requiring surgical source control---matching the patient's current condition post-drainage. & accept \newline \textit{Done.} & accept \\
\addlinespace[3pt]
8 & sigir-201518 & NCT01647932 & \modeall{} & \textbf{Acute decompensated heart failure treated with diuretics} --- Patients admitted with acute decompensated heart failure who are randomized to combined loop plus thiazide-type diuretics versus loop diuretic alone to improve fluid overload symptoms. --- \textit{The patient has worsening dyspnea on exertion, orthopnea, edema, crackles, and JVD consistent with fluid overload from decompensated heart failure; the trial's interventions are intended to treat and improve these current symptoms.} & \textbf{Acute decompensated heart failure treated with diuretics} --- Decision: ineligible --- Clinically, the presentation is consistent with fluid overload from acute decompensated heart failure (orthopnea, crackles, edema, JVD). However, two key inclusion criteria are not supported: a documented history of chronic heart failure and prior outpatient oral loop diuretic therapy for at least one month at the specified dose range. These are major gate conditions and would likely be mentioned if present; thus, they are presumed unmet at prescreen. No exclusion criterion is clearly triggered from the available data, but absence of required inclusions determines ineligibility. If later records confirm chronic heart failure and qualifying loop diuretic use, the patient could be reconsidered. & This study directly targets treatment of fluid overload in acute decompensated heart failure, aiming to improve dyspnea, orthopnea, and edema---complaints the patient currently has. Therefore the trial is relevant to the patient's present condition. & accept \newline \textit{Done.} & accept \\
\addlinespace[3pt]
9 & sigir-20153 & NCT00711399 & \modeall{} & \textbf{Shortness-of-breath treatment cohort} --- Participants presenting with shortness of breath who receive the study's specific respiratory symptom treatment, with symptom severity measured before and after administration. --- \textit{This arm intends to treat and reduce shortness of breath, which is the patient's current chief complaint.} & \textbf{Shortness-of-breath treatment cohort} --- Decision: ineligible --- The patient meets inclusion by presenting with shortness of breath, and logistical steps (consent and pre/post assessment) are assumed feasible. However, he has signs consistent with respiratory distress (acute dyspnea with a respiratory rate of 35 and tachypnea), which is an explicit exclusion. There is no indication of chest tubes or prohibitive skin lesions, and pregnancy is not applicable. Because the respiratory distress exclusion is met, he is not eligible for this treatment cohort. & Partially relevant. The study includes an interventional component that evaluates a treatment aimed at improving respiratory symptoms such as shortness of breath, directly targeting the patient's current complaint. The database/technology validation aspects are not directly therapeutic. & accept \newline \textit{Done.} & accept \\
\addlinespace[3pt]
10 & sigir-201427 & NCT00059878 & \modeallexplore{} & N/A & N/A & This trial studies therapy for proven or probable invasive aspergillosis or other filamentous fungal infections. The patient's current issue is multiple colonic adenomas (likely familial adenomatous polyposis), with no indication of an invasive fungal infection. Therefore, the trial is not relevant to the patient's present clinical problem. & accept \newline \textit{Done.} & accept \\
\addlinespace[3pt]
11 & sigir-201523 & NCT00740090 & \modeallexplore{} & N/A & N/A & Not relevant. The patient's current problem is an acute dengue-like hemorrhagic febrile illness (fever, leukopenia, hemoconcentration, thrombocytopenia after travel to Asia). The trial targets prevention of Plasmodium falciparum malaria via a vaccine in healthy children aged 1--4 living in Mali, which does not address the patient's present disease or clinical complaint. & accept \newline \textit{Done.} & accept \\
\addlinespace[3pt]
12 & sigir-20157 & NCT00462254 & \modeallexplore{} & N/A & N/A & Not relevant. The trial targets sleep disturbances specifically in patients with Parkinson's disease. The patient is a young adult with fatigue and depressive symptoms and no evidence of Parkinson's disease, so the trial's underlying clinical target does not match the patient's current condition. & accept \newline \textit{Done.} & accept \\
\addlinespace[3pt]
13 & sigir-201427 & NCT00770991 & \modeallexplore{} & \textbf{FAP with rectal polyps} --- Participants diagnosed with familial adenomatous polyposis who have at least 5 rectal polyps $\geq$2 mm and an endoscopically assessable rectal segment. --- \textit{The patient currently has dozens of rectosigmoid adenomas at a young age with a strong family history, consistent with FAP, and the trial targets reduction of rectal polyp burden in FAP.} & \textbf{FAP with rectal polyps} --- Decision: eligible --- Clinical FAP is strongly inferred from dozens of rectosigmoid adenomatous polyps at age 21 and a family history of hundreds of adenomas in siblings; the rectum is endoscopically assessable and there are very likely $\geq$5 rectal polyps $\geq$2 mm. No diabetes or berry allergy is noted, and no need for NSAIDs/COX-2 that cannot be stopped is evident; prior two-month NSAID abstinence requires confirmation but there is no indication it is unmet. & The patient has numerous rectosigmoid adenomas and a family history consistent with familial adenomatous polyposis. This trial specifically targets rectal polyp burden in FAP, making it directly relevant to his current clinical problem. & accept \newline \textit{This patient likely will be eligible for the trial; however, there were missing details required by the inclusion criteria so the system can not judge, which is reasonable.} & uncertain \newline \textit{No enough data provided in the patient information for LLM to make the decision whether or not the patient is eligible.} \\
\addlinespace[3pt]
14 & sigir-20148 & NCT00104663 & \modeallexplore{} & \textbf{Human prion disease (all types)} --- Individuals aged 12+ with a diagnosed human prion disease; trial evaluates quinacrine for prion disorders (e.g., Creutzfeldt-Jakob disease). --- \textit{Patient has clinical, EEG, and cortical biopsy findings consistent with human prion disease (CJD), which this trial targets.} & \textbf{Human prion disease (all types)} --- Decision: eligible --- Age $\geq$12 and diagnostic features consistent with human prion disease (CJD) on clinical presentation, EEG, and cortical biopsy; no evidence of coma/pre-terminal status, quinacrine sensitivity, or recent alternative anti-prion therapy. & The patient's presentation (rapid dementia, myoclonus), EEG with periodic sharp waves, and cortical spongiform changes are diagnostic of Creutzfeldt-Jakob disease, a human prion disease. The trial specifically targets human prion diseases, so it is directly relevant. & accept \newline \textit{Done.} & accept \\
\addlinespace[3pt]
15 & sigir-20156 & NCT00150111 & \modeallexplore{} & \textbf{Graves' disease patients} --- Adults diagnosed with Graves' disease undergoing evaluation of rituximab's effect on biochemical markers (free T4, free T3, TSH, TSH-receptor antibodies, anti-TPO) in a phase II pilot study --- \textit{The patient's current presentation (weight loss with increased appetite, heat intolerance/sweating, tachyarrhythmia, hyperreflexia, and mild exophthalmia) is consistent with active Graves' disease.} & \textbf{Graves' disease patients} --- Decision: eligible --- She is an adult with clinical features strongly consistent with active Graves' disease (including exophthalmos). No exclusion factors are evident: no prior rituximab, no current immunosuppression, no serious comorbidity or active infection, and no indication of pregnancy/breastfeeding; performance status is likely $\leq$2. Contraception can be arranged as needed, and required biochemical assessments can be obtained. & The trial targets Graves' disease, and the patient's current symptoms and exam (hyperthyroid features with exophthalmos) support active Graves' disease, making the study directly relevant. No other trial subgroups are described. & reject \newline \textit{While symptoms were classic, the diagnosis of Grave's diease is not formally established (which was assumed by the system).} & accept \\
\addlinespace[3pt]
16 & sigir-201416 & NCT01646411 & \modeallexplore{} & \textbf{Acute symptomatic infection cohort} --- Adults aged 18--70 with a current acute bacterial or viral infection and active symptoms, used as IgM-positive cases for point-of-care IVD validation --- \textit{The patient is a 28-year-old with acute, symptomatic illness consistent with an acute viral infection (e.g., hydrophobia, dysphagia, agitation), meaning she is currently being managed for an acute infection---the clinical target of this cohort.} & \textbf{Acute symptomatic infection cohort} --- Decision: ineligible --- She is an adult with active symptoms consistent with a current acute viral infection, but there is no confirmed IgM-positive result by an FDA-cleared ELISA, which is required for inclusion in the IgM-positive case cohort. & This trial targets adults with current acute infections to validate an IVD test. The patient presents with an acute encephalitic syndrome (hydrophobia, dysphagia, agitation) consistent with an active viral infection, so the trial's clinical focus aligns with her present condition. & uncertain \newline \textit{The trial summary mentioned that "IgM-positive result by an FDA-cleared ELISA" is reuqired; however this was not specified in the inclusion criteria (missed?).} & reject \newline \textit{The patient is actually eligible for inclusion. The LLM misunderstood the IgM criteria to validate the Point of Care IVD test kit with correlating positive and negative samples with FDA approved ELISA test kit, and misinterpreted as inclusion criteria.} \\
\addlinespace[3pt]
17 & sigir-20153 & NCT00413504 & \modeallexplore{} & \textbf{Adults with DVT/PE requiring anticoagulation} --- Patients aged 18+ with deep vein thrombosis and/or pulmonary embolism who require at least 90 days of anticoagulation, treated with fondaparinux monotherapy instead of warfarin --- \textit{The patient presents with acute dyspnea, pleuritic chest pain, tachypnea, and calf pain two weeks after hip arthroplasty with limited mobility---features concerning for acute DVT/PE for which anticoagulation is indicated. The trial targets long-term anticoagulation management for DVT/PE.} & \textbf{Adults with DVT/PE requiring anticoagulation} --- Decision: ineligible --- He meets age criteria and has a presentation strongly suggestive of acute DVT/PE after recent hip arthroplasty; if confirmed, standard care would be at least 90 days of anticoagulation. There are no clear exclusions such as renal insufficiency >1.5 mg/dL, pregnancy, or known fondaparinux hypersensitivity. However, the trial requires an additional warfarin-unsuitability criterion (e.g., prior warfarin failure, bleeding complications, inability to maintain therapeutic INR, nonbleeding side effects) or being a cancer patient on parenteral monotherapy. He has no history of warfarin use, no prior warfarin adverse events, and no cancer. Therefore, he does not meet the key inclusion gate and is ineligible for this subcohort even if DVT/PE is objectively confirmed. & The clinical picture strongly suggests current evaluation for venous thromboembolism (PE/DVT) after recent hip surgery and immobilization with calf pain and pleuritic chest pain. This trial studies long-term treatment of DVT/PE using fondaparinux monotherapy, which aligns with the patient's current suspected clinical problem. While it emphasizes scenarios avoiding warfarin, the underlying target condition (DVT/PE requiring anticoagulation) matches the patient's presentation. & accept \newline \textit{Done.} & accept \\
\addlinespace[3pt]
18 & sigir-20156 & NCT00383643 & \modeallexplore{} & \textbf{Adults with chronic insomnia} --- Adults aged 18--75 meeting diagnostic criteria for chronic insomnia (ICSD), randomized to sodium oxybate, zolpidem tartrate, or placebo. --- \textit{The patient currently reports insomnia for 9 months, aligning with the clinical problem of chronic insomnia targeted by the study.} & \textbf{Adults with chronic insomnia} --- Decision: ineligible --- Although age and insomnia duration (9 months) fit inclusion, the patient has a clinically significant, uncontrolled medical condition consistent with hyperthyroidism, with abnormal physical exam findings (irregular tachycardia, hyperreflexia, exophthalmos). This violates the requirement to be in good health and meets exclusion criteria for uncontrolled medical conditions and clinically significant exam deviations. Her insomnia is likely secondary to hyperthyroidism and may not meet ICSD chronic insomnia criteria. Sleep diary and logistical requirements could be satisfied later, but current medical instability is a substantive exclusion. & The trial targets chronic insomnia, and the patient has a 9-month history of insomnia. Therefore, the study's clinical focus matches a current complaint and is relevant. & accept \newline \textit{Done.} & accept \\
\addlinespace[3pt]
19 & sigir-201415 & NCT02203968 & \modeccr{} & N/A & N/A & Not relevant. The patient's chief complaint is abnormal uterine bleeding with suspected vesicular mole vs fibroid degeneration. The trial evaluates early fibrinogen concentrate for coagulopathy and hemorrhage in severely injured trauma patients, which does not treat uterine causes of bleeding, does not target molar pregnancy or fibroid degeneration, and does not enhance procedures for managing these gynecologic conditions. & accept \newline \textit{Done.} & accept \\
\addlinespace[3pt]
20 & sigir-201416 & NCT02394418 & \modeccr{} & N/A & N/A & The patient's chief complaint is consistent with acute rabies encephalitis (marked hydrophobia, dysphagia, agitation, spasticity after likely animal exposure). The trial targets management of delirium in mechanically ventilated ICU patients using sedatives; it does not treat rabies itself, a meaningful ancestor of rabies, or a direct cause of the patient's condition, nor does it enhance a procedure to treat rabies. While the patient may exhibit delirium, that is a symptom rather than the chief complaint. Therefore, the trial is not relevant to the patient's primary clinical problem. & accept \newline \textit{Done.} & accept \\
\addlinespace[3pt]
21 & sigir-201523 & NCT02406729 & \modeccr{} & N/A & N/A & The patient's chief complaint is an acute dengue-like illness (fever, hemorrhagic signs, cytopenias). The trial tests a prophylactic tetravalent dengue vaccine to prevent future symptomatic dengue, with efficacy assessed for cases occurring $\geq$28 days after vaccination. It does not treat active dengue, nor enhance a needed therapeutic procedure. Therefore, no subcohort is relevant to this patient's current clinical problem. & accept \newline \textit{Done.} & accept \\
\addlinespace[3pt]
22 & sigir-201417 & NCT01763008 & \modeccr{} & \textbf{Complicated intra-abdominal infection cohort} --- Participants diagnosed with complicated intra-abdominal infections who receive doripenem treatment --- \textit{The patient's main diagnosis is a ruptured liver abscess, which is a complicated intra-abdominal infection; the trial's intervention (doripenem) is intended to treat this infection.} & \textbf{Complicated intra-abdominal infection cohort} --- Decision: eligible --- He has a ruptured liver abscess with intra-abdominal contamination, fitting the definition of a complicated intra-abdominal infection. There is no evidence of contraindications to doripenem: no history of hypersensitivity to beta-lactams/carbapenems, no use of probenecid, and no known doripenem-resistant pathogen. He is male (not pregnant/lactating). Severe renal impairment (CrCl <10 mL/min, dialysis, or oliguria) is not reported and would likely have been noted if present in this context. Postoperative ICU care and underlying CVID do not exclude participation. Thus, he satisfies inclusion and lacks exclusion criteria for this subcohort. & The trial is relevant because it studies doripenem for complicated intra-abdominal infections, and the patient's chief complaint is a ruptured liver abscess (a form of complicated intra-abdominal infection). The nosocomial pneumonia and complicated UTI arms are not relevant as those conditions are not present. & accept \newline \textit{Done.} & accept \\
\addlinespace[3pt]
23 & sigir-201423 & NCT00261105 & \modeccr{} & \textbf{Acute bacterial exacerbation of chronic bronchitis (AECB) outpatient cohort} --- Outpatients with acute bacterial exacerbation of chronic bronchitis treated with telithromycin (Ketek) to assess clinical cure of the primary respiratory infection. --- \textit{The patient's current illness is most consistent with an acute exacerbation of chronic bronchitis/COPD (worsening dyspnea, purulent sputum, hyperinflation on chest X-ray without infiltrate). This subcohort's intervention is aimed at treating AECB itself, directly addressing the patient's chief complaint.} & \textbf{Acute bacterial exacerbation of chronic bronchitis (AECB) outpatient cohort} --- Decision: eligible --- He likely has chronic bronchitic COPD based on heavy smoking, barrel chest, hyperinflation on chest X-ray, and home oxygen, and now meets exacerbation criteria with increased dyspnea and purulent sputum over one week. CAP criteria are not met due to absence of a new infiltrate, and there are no sinus symptoms to suggest acute sinusitis. No clear exclusion criteria are present (male, no known long-QT history, no severe renal disease or interacting drugs documented, no recent prolonged antibiotics noted). The chronic bronchitis duration criterion (>2 years of cough/sputum) is not explicitly stated but is plausible given COPD history and can be confirmed at screening. Therefore, he is an appropriate fit for the AECB cohort pending routine confirmation. & The trial is relevant insofar as it evaluates antibiotic treatment for acute bacterial exacerbation of chronic bronchitis, which matches the patient's likely current diagnosis driving his cough and shortness of breath. The CAP cohort is not relevant (no new infiltrate on chest X-ray), and the acute sinusitis cohort is not relevant (no sinus symptoms). The AECB cohort directly targets the condition responsible for the patient's chief complaint. & accept \newline \textit{Done.} & accept \\
\addlinespace[3pt]
24 & sigir-20157 & NCT01371383 & \modeccr{} & subcohort1: \textbf{Part 1: Omega-3 PUFA monotherapy vs placebo for major depressive disorder} --- Double-blind, placebo-controlled trial testing omega-3 polyunsaturated fatty acids as monotherapy in adults with DSM-IV major depressive disorder. --- \textit{The patient's fatigue is strongly suggestive of being driven by major depressive disorder; treating depression (a direct cause of her chief complaint) is intended to reduce depressive symptom burden and thereby improve fatigue.} \newline subcohort2: \textbf{Part 2: EPA vs DHA effects on depressive symptom clusters} --- Double-blind trial comparing eicosapentaenoic acid (EPA) and docosahexaenoic acid (DHA) to assess their effects on different symptom clusters in patients with DSM-IV major depressive disorder. --- \textit{This subcohort targets depressive symptom clusters, which include fatigue; addressing depression as the direct cause is expected to improve the patient's chief complaint of fatigue.} & subcohort1: \textbf{Part 1: Omega-3 PUFA monotherapy vs placebo for major depressive disorder} --- Decision: eligible --- She exhibits DSM-IV-consistent major depressive disorder (anhedonia, hypersomnia, hyperphagia, impaired concentration, guilt for several months), is 20 years old, and has no major medical illnesses. No history suggestive of exclusionary Axis I/II comorbidities (psychosis, bipolar, primary anxiety disorders, recent substance use disorder, personality disorder) is present. Normal TSH and CBC reduce concern for medical mimics. Informed consent ability is assumed per logistical allowance. \newline subcohort2: \textbf{Part 2: EPA vs DHA effects on depressive symptom clusters} --- Decision: eligible --- The patient meets DSM-IV major depressive disorder criteria with prominent fatigue-related symptom clusters (anhedonia, hypersomnia, hyperphagia, concentration difficulties, guilt), is within the 18--65 age range, and lacks major medical illness or exclusionary psychiatric comorbidities. Nothing in the vignette suggests recent substance use disorder or personality disorders. Consent feasibility is assumed as permitted. & The trial targets major depressive disorder, which is a plausible direct cause of this patient's chief complaint of fatigue given her anhedonia, hypersomnia, hyperphagia, guilt, and concentration difficulties. By treating depression, the interventions aim to reduce depressive symptoms and thereby alleviate fatigue. The study does not focus on procedures or prevention unrelated to her complaint; its relevance stems from addressing the underlying depressive disorder driving the fatigue. & subcohort1: accept -- \textit{Done.} \newline subcohort2: accept -- \textit{Done.} & subcohort1: accept \newline subcohort2: accept \\
\addlinespace[3pt]
25 & sigir-201423 & NCT02209974 & \modeccr{} & \textbf{Ex-smokers with moderate-to-severe COPD receiving inhaled corticosteroids} --- Patients who are ex-smokers with a history >10 pack-years and have moderate to severe chronic obstructive pulmonary disease, enrolled to assess the effect of inhaled corticosteroids on COPD-related inflammation --- \textit{The patient's chief complaints are cough and dyspnea, with clinical features consistent with COPD as the underlying condition. COPD is the meaningful ancestor driving these symptoms. A trial testing inhaled corticosteroids in COPD aims to treat the underlying disease process and could reduce symptom burden and exacerbations, directly addressing the patient's chief complaints.} & \textbf{Ex-smokers with moderate-to-severe COPD receiving inhaled corticosteroids} --- Decision: ineligible --- He likely has moderate-to-severe COPD (barrel chest, hyperinflation) and heavy smoking implies >10 pack-years, satisfying those inclusion elements. However, the trial requires ex-smokers, and the vignette does not indicate he has quit smoking; under prescreen rules this major criterion cannot be inferred and should be considered unmet. No exclusion criteria are provided. Being on inhaled corticosteroids can be addressed as part of the study intervention. Therefore, he is ineligible unless his smoking status can be confirmed as ex-smoker. & The patient's cough and shortness of breath are consistent with COPD exacerbation on a background of COPD (barrel chest, hyperinflation, heavy smoking). This trial targets COPD with inhaled corticosteroids, addressing the underlying disease that drives his chief complaints, and is therefore relevant. & accept \newline \textit{Done.} & reject \newline \textit{The LLM made assumption that the patient is still active smoking and exclude the patient. Smoking hx not provided by the information.} \\
\addlinespace[3pt]
26 & sigir-201423 & NCT02532075 & \modeccr{} & \textbf{COPD patients receiving breathing muscle warm-up prior to exercise} --- Participants with a clinical diagnosis of COPD who perform a breathing muscle warm-up before a 6-minute walk test to assess its effects on distance walked and perceived breathlessness. --- \textit{This intervention is intended to reduce breathlessness and improve functional capacity in COPD, directly addressing the patient's main diagnosis (COPD) rather than unrelated symptoms.} & \textbf{COPD patients receiving breathing muscle warm-up prior to exercise} --- Decision: ineligible --- Although COPD and age criteria are met, he is currently unstable with an acute exacerbation and has required home supplemental oxygen in the past 24 hours; both instability of COPD and requirement for supplemental oxygen are explicit exclusions. Additionally, spinal stenosis likely impairs his ability to perform the walk test. & The patient's presentation (heavy smoking, barrel chest, hyperinflation on CXR) is consistent with COPD. This study tests a COPD-focused intervention aimed at reducing dyspnea and improving walking capacity, which treats the underlying condition's symptom burden and functional impairment. It does not address an acute infectious exacerbation or cough specifically, but is relevant to COPD management. & accept \newline \textit{Done.} & accept \\
\addlinespace[3pt]
27 & sigir-201427 & NCT01422577 & \modeccr{} & \textbf{Bright-NBI vs white-light in screening colonoscopy} --- Asymptomatic, average-risk adults undergoing screening colonoscopy randomized to bright narrow-band imaging versus standard white-light endoscopy to compare colorectal adenoma detection --- \textit{The patient has numerous colonic adenomas; colonoscopic management (with polypectomy during colonoscopy) is used to treat this chief complaint. An imaging modality that increases adenoma detection during colonoscopy can enhance the procedure's completeness and clinical benefit for treating adenomas.} & \textbf{Bright-NBI vs white-light in screening colonoscopy} --- Decision: ineligible --- Ineligible due to age (21, fails inclusion >40 and meets exclusion <50) and not average risk (personal history of multiple adenomas and first-degree family history suggestive of FAP). Colonoscopy is for high-risk surveillance/diagnostic purposes rather than average-risk screening. & Partially relevant: The trial's intervention (bright-NBI) is designed to improve detection of colorectal adenomas during colonoscopy, which can enhance the effectiveness of colonoscopic treatment (polypectomy) for the patient's chief complaint of multiple adenomas. However, the study focuses on average-risk screening rather than patients with known polyposis, so the clinical context does not fully match the patient's high-risk situation. & accept \newline \textit{Done.} & accept \\
\addlinespace[3pt]

\end{longtable}
\endgroup

\subsubsection{Gwet's AC1 Calculation}

We compute inter-rater agreement using Gwet’s AC1 with $N=27$ items and three categories. The observed agreement is $P_o = \frac{23}{27} \approx 0.852$. The average marginal category probabilities are $p_{\text{accept}} = \frac{25/27 + 24/27}{2} = \frac{49}{54}$, $p_{\text{uncertain}} = \frac{1}{27}$, and $p_{\text{reject}} = \frac{1/27 + 2/27}{2} = \frac{1}{18}$. The chance agreement is $P_e = 1 - \sum_k p_k^2 = 1 - \left(\left(\frac{49}{54}\right)^2 + \left(\frac{1}{27}\right)^2 + \left(\frac{1}{18}\right)^2 \right) \approx 0.172$. Thus, $\text{AC1} = \frac{P_o - P_e}{1 - P_e} = \frac{0.852 - 0.172}{1 - 0.172} \approx 0.82$.
    \subsection{Failure Case Analysis}
\label{app:failure_case_analysis}

In this appendix, we present a structured analysis of failure cases described in Section~\ref{sec:results}. These cases were manually reviewed and categorized to better understand the sources of error across different stages of the system.

\begin{figure}[!htbp]
  \centering
  \includegraphics[width=\columnwidth]{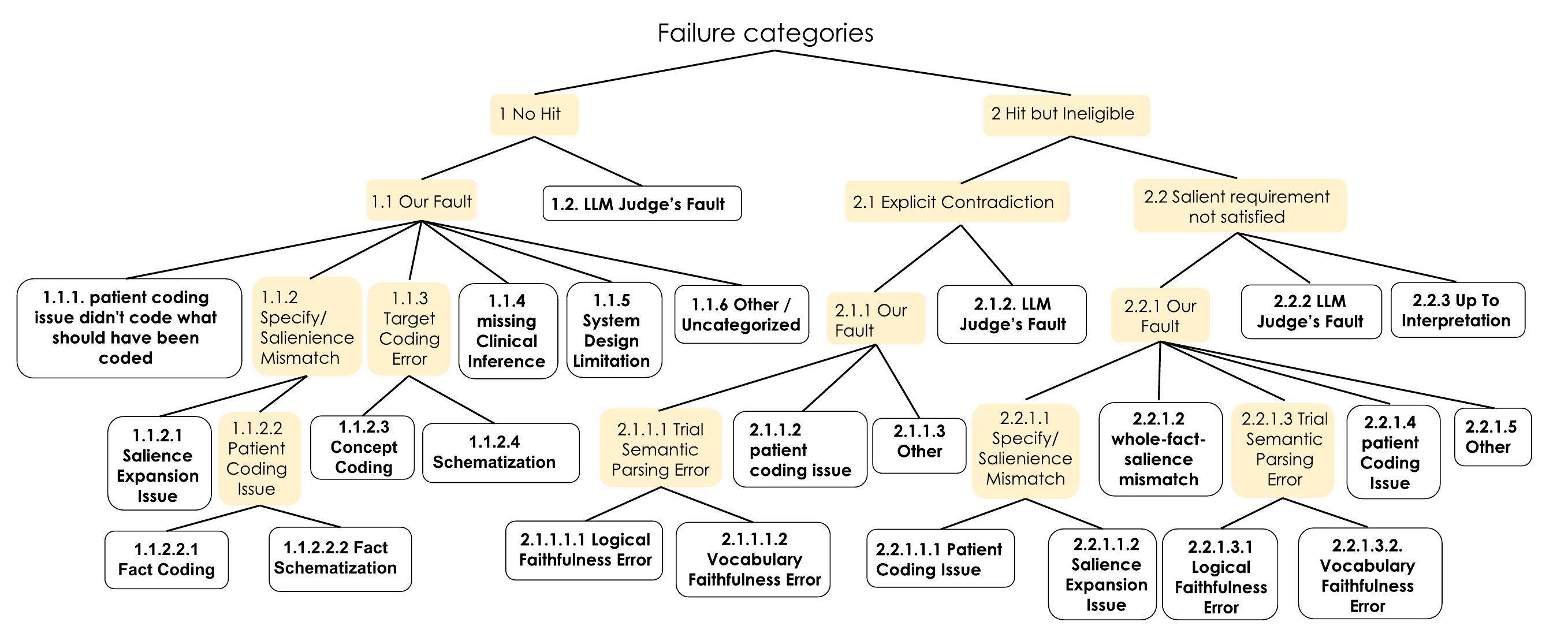}
  \caption{\textbf{Failure case categories.} Observed failure cases are organized into a hierarchical taxonomy. This taxonomy is used to label and analyze all failure cases reported in Table~\ref{tab:failure_case_full}.}
  \label{fig:failure-case-categories-1}
\end{figure}

\begin{figure}[!htbp]
  \centering
  \includegraphics[width=\columnwidth]{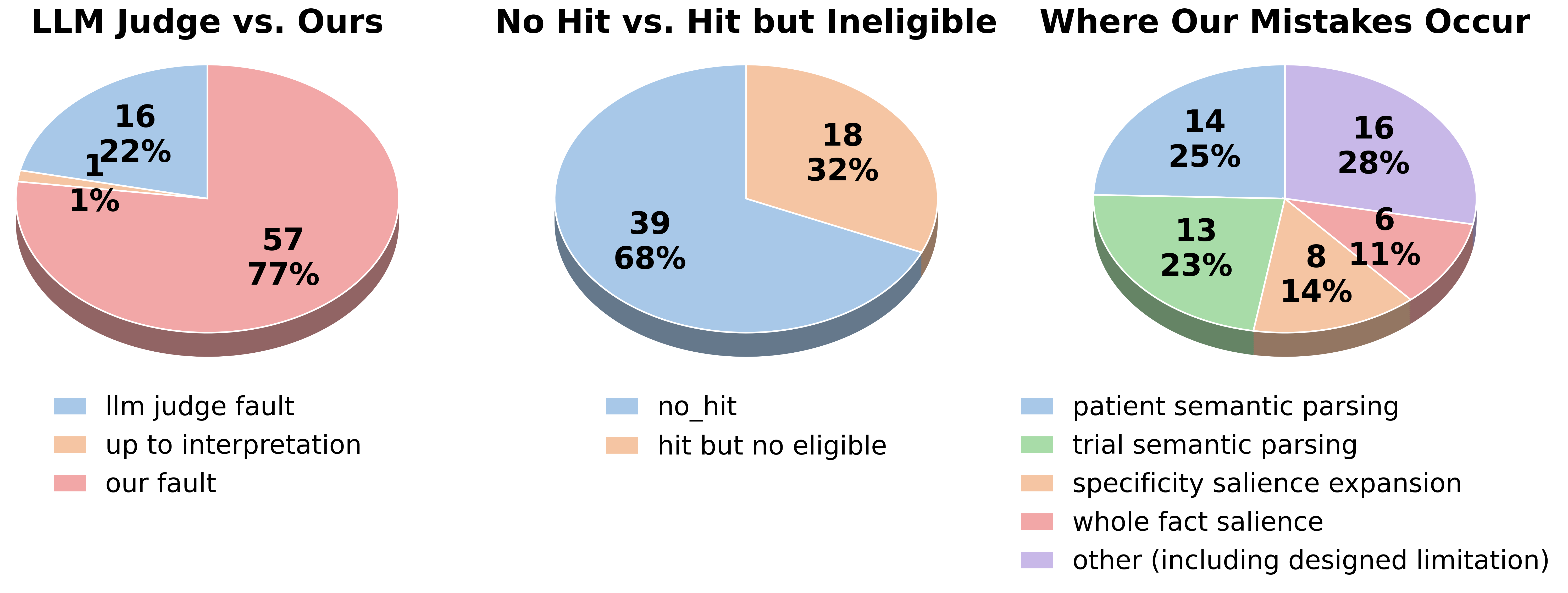}
  \caption{\textbf{Failure case pie charts.} The pie charts summarize the distribution of failure cases reported in Table~\ref{tab:failure_case_full}.}
  \label{fig:failure-case-categories-2}
\end{figure}

\paragraph{Failure case analysis.} Figure~\ref{fig:failure-case-categories-1} presents our failure taxonomy, while Figure~\ref{fig:failure-case-categories-2} summarizes the observed failures from three perspectives. The left pie chart compares disagreement sources between the GPT-5 judge and \name{}, showing that the majority of discrepancies are attributed to our system, with a smaller portion due to judge errors or ambiguous interpretation. The middle pie chart separates failure modes into missed retrievals (no hit) versus retrieved but ineligible trials, indicating that most failures arise from not retrieving relevant candidates rather than eligibility misclassification. The right pie chart breaks down the sources of errors within \name{}, matching the counts in Section~\ref{sec:results}: patient-side semantic parsing (14), trial-side semantic parsing (13), specificity-salience expansion, i.e., under-specified evidence (8), whole-fact salience, i.e., missing evidence (6), and other factors including designed limitations (16), with no single dominant failure mode.

Table~\ref{tab:failure_case_full} reports the full set of analyzed failure cases. Each row corresponds to a patient--trial example and includes the retrieval objective, patient identifier, trial identifier, a concise description of the failure, and its assigned category based on the taxonomy in Figure~\ref{fig:failure-case-categories-1}.
\begingroup
\tiny
\setlength{\LTleft}{0pt}
\setlength{\LTright}{0pt plus 1fill}
\setlength{\tabcolsep}{2pt}
\renewcommand{\arraystretch}{1.08}


\begin{longtable}{
  L{0.35cm}  
  L{1.00cm}  
  L{1.40cm}  
  L{1.90cm}  
  L{7.6cm} 
  L{0.60cm}  
}
\caption{Full failure case table} \label{tab:failure_case_full} \\
\toprule
\textbf{\#} &
\textbf{Objective} &
\textbf{Patient} &
\textbf{Trial} &
\textbf{Description} &
\textbf{Cat.} \\
\midrule
\endfirsthead

\multicolumn{6}{c}{\tablename\ \thetable\ -- continued from previous page} \\
\toprule
\textbf{\#} &
\textbf{Objective} &
\textbf{Patient} &
\textbf{Trial} &
\textbf{Description} &
\textbf{Cat.} \\
\midrule
\endhead

\midrule
\multicolumn{6}{r}{Continued on next page} \\
\endfoot

\bottomrule
\endlastfoot

1 & \modeccr & sigir-20141 & NCT01397175 & not hit. The trial has target Coronary arteriosclerosis (disorder). ACS wasn't salience expanded to Coronary arteriosclerosis (disorder) & 1.1.2.2.2 \\
\addlinespace[3pt]
2 & \modeccr & sigir-20147 & NCT00203450 & The patient should be excluded because the patient is suicidal, and being suicidal is an exclusion criterion & 2.1.2 \\
\addlinespace[3pt]
3 & \modeccr & sigir-20147 & NCT00290914 & Trial target is Major Depressive Disorder, The patient has depressive disorder. But Major Depressive Disorder is not salience-expanded to depressive disorder & 1.1.2.1 \\
\addlinespace[3pt]
4 & \modeccr & sigir-20157 & NCT00946413 & Community Adolescents at Risk for Depression and Suicide should actually be processed as suicidal or suicidal thought instead of this ill-fitting finding concept & 1.1.2.3 \\
\addlinespace[3pt]
5 & \modeccr & sigir-201417 & NCT01188993 & observational/assessment study on septic shock (which patient is inferred to have), not treating & 1.2 \\
\addlinespace[3pt]
6 & \modeccr & sigir-201419 & NCT00826813 & patient\_has\_diagnosis\_of\_primary\_malignant\_neoplasm\_of\_esophagus is not satisfied by the patient. patient has patient\_has\_diagnosis\_of\_neoplasm\_of\_esophagus & 1.1.2.1 \\
\addlinespace[3pt]
7 & \modeccr & sigir-201421 & NCT00430677 & patient\_has\_undergone\_kidney\_biopsy, which is a logistical procedure requirement, is explicitly required & 2.2.1.2 \\
\addlinespace[3pt]
8 & \modeccr & sigir-201422 & NCT00908804 & Age over 15 years is coded as \textgreater{} 15 years not \textgreater{}= 15 years, resulting in requirement contradiction because the patient is exactly 15 years old & 2.1.1.1.1 \\
\addlinespace[3pt]
9 & \modeccr & sigir-201423 & NCT01561794 & Trial treats secondary infection of the chronic disease, which is not really directly chief-complaint-related. & 1.2 \\
\addlinespace[3pt]
10 & \modeccr & sigir-201424 & NCT02297659 & The patient has trauma, and is inferred to likely to undergo damage control laparotomy, this trial tests hypertonic saline to improve primary fascial closure after DCL. This is not directly treating chief complaint & 1.2 \\
\addlinespace[3pt]
11 & \modeccr & sigir-201427 & NCT01584817 & The trial is about telephone education for patients undergoing colonoscopy, which is indirectly related to treating the patient's chief complaint (multiple colonic adenomas) & 1.2 \\
\addlinespace[3pt]
12 & \modeccr & sigir-201427 & NCT02538406 & The trial improves the effectiveness of screening colonoscopy, but we didn't include the part where we infer what procedure the patient needs to undergo & 1.1.4 \\
\addlinespace[3pt]
13 & \modeccr & sigir-201430 & NCT02273232 & The trial asks for intermittent claudication but we failed to code / infer that from the patient side & 2.2.1.4 \\
\addlinespace[3pt]
14 & \modeccr & sigir-201430 & NCT02436200 & The trial asks for intermittent claudication but we failed to code / infer that from the patient side & 2.2.1.4 \\
\addlinespace[3pt]
15 & \modeccr & sigir-201517 & NCT00461760 & Disease list has Cervical Cancer Screening coded, not Cervical Cancer & 1.1.2.3 \\
\addlinespace[3pt]
16 & \modeccr & sigir-201518 & NCT00012818 & The trial is about treating chronic heart failure, the patient most likely has congestive heart failure & 1.2 \\
\addlinespace[3pt]
17 & \modeccr & sigir-201518 & NCT00971386 & The trial is about treating chronic heart failure, the patient most likely has congestive heart failure & 1.2 \\
\addlinespace[3pt]
18 & \modeccr & sigir-201519 & NCT02182856 & patient\_has\_finding\_of\_acute\_exacerbation\_of\_chronic\_obstructive\_airways\_disease is an exclusion requirement & 2.1.2 \\
\addlinespace[3pt]
19 & \modeccr & sigir-201520 & NCT00001776 & Behavioral therapy improving frontal lobe dementia, but we didn't count it as directly treating. May be our target schematization issue or LLM judge issue, depending on what we mean by "directly treating". & 1.1.2.2.2 \\
\addlinespace[3pt]
20 & \modeccr & sigir-201529 & NCT00000520 & Bug. Actually categorized to clinically address Kawasaki, but there is a conflict from the positive literal side, so joined on the positive literal hit instead of disease. & 1.1.6 \\
\addlinespace[3pt]
21 & \modeall & sigir-20143 & NCT00124761 & patient\_has\_diagnosis\_of\_widespread\_metastatic\_malignant\_neoplastic\_disease is required, but this is unlikely to be coded in any patient note, so it is a whole-fact salience defaulting issue & 2.2.1.2 \\
\addlinespace[3pt]
22 & \modeall & sigir-20146 & NCT01307644 & Obesity didn't get kept as a target, it was original in the trial listing. & 1.1.2.3 \\
\addlinespace[3pt]
23 & \modeall & sigir-20157 & NCT00946413 & Community Adolescents at Risk for Depression and Suicide should actually be processed as suicidal or suicidal thought instead of this ill-fitting finding concept & 1.1.2.3 \\
\addlinespace[3pt]
24 & \modeall & sigir-201414 & NCT00311753 & Trial is about prevention of venous thromboembolism, we didn't code VTE as a prevention target, so lack of clinical inference & 1.1.4 \\
\addlinespace[3pt]
25 & \modeall & sigir-201417 & NCT01188993 & observational/assessment study on septic shock (which patient is inferred to have), not treating & 1.2 \\
\addlinespace[3pt]
26 & \modeall & sigir-201419 & NCT00826813 & patient\_has\_diagnosis\_of\_primary\_malignant\_neoplasm\_of\_esophagus is not satisfied by the patient. patient has patient\_has\_diagnosis\_of\_neoplasm\_of\_esophagus & 1.1.2.1 \\
\addlinespace[3pt]
27 & \modeall & sigir-201420 & NCT01729559 & Trial is about prevention of venous thromboembolism, we didn't code VTE as a prevention target, so lack of clinical inference & 1.1.4 \\
\addlinespace[3pt]
28 & \modeall & sigir-201420 & NCT01787773 & Trial is about prevention of venous thromboembolism, we didn't code VTE as a prevention target, so lack of clinical inference & 1.1.4 \\
\addlinespace[3pt]
29 & \modeall & sigir-201421 & NCT00001676 & excludes for patient\_has\_diagnosis\_of\_disorder\_of\_cellular\_component\_of\_blood, the patient does have that, but the trial actually excludes with more specific numerical criteria. Arguably, this patient should still be excluded & 2.1.2 \\
\addlinespace[3pt]
30 & \modeall & sigir-201424 & NCT01276561 & trial enhancing laparoscopic splenectomy which we don't infer for this trauma patient but we can (we have the module but didn't integrate) & 1.1.4 \\
\addlinespace[3pt]
31 & \modeall & sigir-201430 & NCT02273232 & The trial asks for intermittent claudication but we failed to code / infer that from the patient side & 2.2.1.4 \\
\addlinespace[3pt]
32 & \modeall & sigir-201430 & NCT02436200 & The trial asks for intermittent claudication but we failed to code / infer that from the patient side & 2.2.1.4 \\
\addlinespace[3pt]
33 & \modeall & sigir-201517 & NCT00461760 & Disease list has Cervical Cancer Screening coded, not Cervical Cancer & 1.1.2.3 \\
\addlinespace[3pt]
34 & \modeall & sigir-201520 & NCT00001776 & Behavioral therapy improving frontal lobe dementia, but we didn't count it as directly treating. May be our target schematization issue or LLM judge issue, depending on what we mean by "directly treating". & 1.1.2.2.2 \\
\addlinespace[3pt]
35 & \modeall & sigir-201520 & NCT00024908 & Trial is designed to improve planning or social behavior of CNS patients. Doesn't treat any of the patient's complaint directly & 1.2 \\
\addlinespace[3pt]
36 & \modeall & sigir-201520 & NCT02212119 & This trial directly targets a current complaint: paratonic rigidity in a fully dependent patient with severe cognitive impairment. But neither the trial target codes this nor the patient codes this. Both have dementia coded though. So it is more like a clinical inference problem & 1.1.4 \\
\addlinespace[3pt]
37 & \modeall & sigir-201522 & NCT01887457 & Bug.  & 1.1.6 \\
\addlinespace[3pt]
38 & \modeall & sigir-201529 & NCT00000520 & Bug. Actually categorized to clinically address Kawasaki, but there is a conflict from the positive literal side, so joined on the positive literal hit instead of disease. & 1.1.6 \\
\addlinespace[3pt]
39 & \modeallexplore & sigir-20141 & NCT01171911 & patient\_has\_diagnosis\_of\_non\_q\_wave\_myocardial\_infarction not specificity-expanded to myocardial infarction & 1.1.2.1 \\
\addlinespace[3pt]
40 & \modeallexplore & sigir-20143 & NCT00001776 & frontal lobe lesion didn't get coded as a target & 1.1.2.3 \\
\addlinespace[3pt]
41 & \modeallexplore & sigir-20143 & NCT00124761 & patient\_has\_diagnosis\_of\_widespread\_metastatic\_malignant\_neoplastic\_disease is required, but this is unlikely to be coded in any patient note, so it is a whole-fact salience defaulting issue & 2.2.1.2 \\
\addlinespace[3pt]
42 & \modeallexplore & sigir-20143 & NCT00303901 & patient\_has\_finding\_of\_serious\_physical\_health\_problem assumed but should not be defaulted as this is unlikely to be coded & 2.2.1.2 \\
\addlinespace[3pt]
43 & \modeallexplore & sigir-20143 & NCT01566682 & Trial asks for single/multiple lung nodules, didn't get coded from the patient side though should be strongly inferred. We can only entail nearby concepts from current patient facts & 1.1.1 \\
\addlinespace[3pt]
44 & \modeallexplore & sigir-20147 & NCT00563992 & patient has suicidal thoughts but trial asks for suicidal behavior & 2.2.2 \\
\addlinespace[3pt]
45 & \modeallexplore & sigir-20147 & NCT01784666 & bipolar I disorder, current episode depressed and bipolar I disorder, most recent episode depressed are ontology siblings and they show up on both patient and trial sides. Ideally on patient side we should also have most recent episode depressed as alternatives but we didn't consider siblings & 1.1.1 \\
\addlinespace[3pt]
46 & \modeallexplore & sigir-20154 & NCT01076738 & Acute coronary syndrome (disorder) didn't get coded on the patient side, what we have is Acute myocardial infarction which is an ontological sibling & 2.2.1.4 \\
\addlinespace[3pt]
47 & \modeallexplore & sigir-201410 & NCT01599195 & Trial side has carotid bruit, but carotid bruit is not salience alternative lifted to just bruit, which is coded on the patient side & 1.1.2.1 \\
\addlinespace[3pt]
48 & \modeallexplore & sigir-201414 & NCT01075035 & Trial-side traumatic brain injury not lifted to patient-side traumatic injury by specificity-salience expansion & 1.1.2.1 \\
\addlinespace[3pt]
49 & \modeallexplore & sigir-201414 & NCT01196299 & Trial-side traumatic brain injury not lifted to patient-side traumatic injury by specificity-salience expansion & 1.1.2.1 \\
\addlinespace[3pt]
50 & \modeallexplore & sigir-201417 & NCT01188993 & Trial-side septic shock not specificity-salience expanded to patient-side shock & 1.1.2.1 \\
\addlinespace[3pt]
51 & \modeallexplore & sigir-201417 & NCT01231672 & patient\_has\_finding\_of\_disturbance\_of\_consciousness whole-fact salient, but should not be & 2.2.1.2 \\
\addlinespace[3pt]
52 & \modeallexplore & sigir-201420 & NCT01729559 & Trial is about prevention of venous thromboembolism, we didn't code VTE as a prevention target, so lack of clinical inference & 1.1.4 \\
\addlinespace[3pt]
53 & \modeallexplore & sigir-201421 & NCT01031797 & patient\_has\_finding\_of\_ecg\_normal assumed whole fact salient & 2.2.1.2 \\
\addlinespace[3pt]
54 & \modeallexplore & sigir-201423 & NCT01785706 & polarity flip of patient age requirement & 2.1.1.1.1 \\
\addlinespace[3pt]
55 & \modeallexplore & sigir-201424 & NCT01103999 & patient\_has\_diagnosis\_of\_contusion\_of\_spleen, whole fact salient but actually cannot be inferred (too specific to be salient) & 2.2.2 \\
\addlinespace[3pt]
56 & \modeallexplore & sigir-201425 & NCT00452036 & patient\_has\_finding\_of\_nausea whole fact salient but not coded. He clearly has head injury and vomiting, but the text does not explicitly say nausea. also this case does not look like minor head injury. & 2.2.2 \\
\addlinespace[3pt]
57 & \modeallexplore & sigir-201430 & NCT00001368 & Bug. exclusion should not check for overlap but strict containment & 2.1.1.3 \\
\addlinespace[3pt]
58 & \modeallexplore & sigir-201430 & NCT01970332 & patient\_has\_finding\_of\_intermittent\_claudication  salient but didn't get coded & 2.2.1.4 \\
\addlinespace[3pt]
59 & \modeallexplore & sigir-201430 & NCT02273232 & The trial asks for intermittent claudication but we failed to code / infer that from the patient side & 2.2.1.4 \\
\addlinespace[3pt]
60 & \modeallexplore & sigir-201430 & NCT02436200 & patient\_has\_finding\_of\_intermittent\_claudication  salient but didn't get coded & 1.1.1 \\
\addlinespace[3pt]
61 & \modeallexplore & sigir-201512 & NCT01646411 & trial side has viral and bacterial acute infection but patient side entail acute\_infection from meningitis related facts, designed to not hit & 1.1.5 \\
\addlinespace[3pt]
62 & \modeallexplore & sigir-201514 & NCT00842621 & coded \textless{} 19 years old for adult subcohort & 2.1.1.1.1 \\
\addlinespace[3pt]
63 & \modeallexplore & sigir-201516 & NCT00747981 & Experimental logging issue: an entry was accidentally overwritten & 1.1.6 \\
\addlinespace[3pt]
64 & \modeallexplore & sigir-201517 & NCT00461760 & Disease list has Cervical Cancer Screening coded, not Cervical Cancer & 1.1.2.3 \\
\addlinespace[3pt]
65 & \modeallexplore & sigir-201517 & NCT00520117 & abnormal pap results got coded but not hpv status & 1.1.2.3 \\
\addlinespace[3pt]
66 & \modeallexplore & sigir-201518 & NCT00133328 & ``and/or'' criterion admits different interpretations; ambiguous and up to interpretation & 2.2.3 \\
\addlinespace[3pt]
67 & \modeallexplore & sigir-201519 & NCT01628497 & bronchospasm didn't get coded on patient side & 1.1.1 \\
\addlinespace[3pt]
68 & \modeallexplore & sigir-201519 & NCT01785706 & polarity flip of patient age requirement & 2.1.1.1.1 \\
\addlinespace[3pt]
69 & \modeallexplore & sigir-201519 & NCT02182856 & patient\_has\_finding\_of\_acute\_exacerbation\_of\_chronic\_obstructive\_airways\_disease is an exclusion requirement & 2.1.2 \\
\addlinespace[3pt]
70 & \modeallexplore & sigir-201520 & NCT02212119 & This trial directly targets a current complaint: paratonic rigidity in a fully dependent patient with severe cognitive impairment. But neither the trial target codes this nor the patient codes this. Both have dementia coded though. So it is more like a clinical inference problem & 1.1.4 \\
\addlinespace[3pt]
71 & \modeallexplore & sigir-201524 & NCT00428831 & respiratory illness should get coded but didn't & 1.1.2.3 \\
\addlinespace[3pt]
72 & \modeallexplore & sigir-201525 & NCT01396798 & Acutely ill children with suspected serious infection is not a common ontology medical term & 1.1.5 \\
\addlinespace[3pt]
73 & \modeallexplore & sigir-201526 & NCT01261026 & not clear eligible / uncertain. most likely an llm judge issue because not sure if she has mass\_of\_uterine\_adnexa & 2.2.2 \\
\addlinespace[3pt]
74 & \modeallexplore & sigir-201528 & NCT01396798 & Acutely ill children with suspected serious infection is not a common ontology medical term & 1.1.5 \\
\addlinespace[3pt]
\end{longtable}
\endgroup

    \subsection{Experimental Settings}\label{app:setting}

Our experiments were conducted across three machines. Patient-side semantic parsing ran on Machine~1 (Table~\ref{tab:machine_1}). Trial-side semantic parsing, database construction and querying, and the speed and scalability experiments ran on Machine~2 (Table~\ref{tab:machine_2}). Detailed evaluations of individual subpipelines, including preprocessing (Section~\ref{app:preproc_evaluation}) and entity canonicalization (Section~\ref{app:ner-eval}), were performed on Machine~3 (Table~\ref{tab:machine_3}). Pipeline development was carried out across all three machines. Table~\ref{tab:snomed_config} summarizes the SNOMED and terminology-retrieval configuration used for ontology grounding, and Table~\ref{tab:llm_config} summarizes the inference configurations used for GPT-4.1 and GPT-5.

\begin{table}[htbp]
\centering
\small
\begin{tabular}{p{0.30\linewidth} p{0.64\linewidth}}
\toprule
\textbf{Item} & \textbf{Value} \\
\midrule
Machine & MacBook Air / Mac14,15 \\
Chip & Apple M2 \\
CPU Cores & 8 (4 performance and 4 efficiency) \\
GPU & Apple M2, 10 GPU cores, Metal 3 \\
Memory & 16 GB \\
Operating System & Darwin 14.0 \\
Kernel & 23.0.0 \\
Python & 3.12.4 (CPython) \\
Java & OpenJDK 17 \\
Elasticsearch & 7.17.15 \\
Snowstorm & 10.7.0 \\
SNOMED CT & SNOMED CT International RF2 2025-05-01 \\
Key Python Packages & torch=not installed, transformers=not installed, openai=not installed, numpy=1.26.4, pandas=2.2.2 \\
\\
\bottomrule
\end{tabular}
\caption{Machine 1: Experimental environment.}
\label{tab:machine_1}
\end{table}

\begin{table}[htbp]
\centering
\small
\begin{tabular}{p{0.30\linewidth} p{0.64\linewidth}}
\toprule
\textbf{Item} & \textbf{Value} \\
\midrule
Machine & MacBook Pro / Mac14,6 \\
Chip & Apple M2 Max \\
CPU Cores & 12 (8 performance and 4 efficiency) \\
GPU & Apple M2 Max, 38 GPU cores, Metal 3 \\
Memory & 32 GB \\
Operating System & Darwin 13.0 \\
Kernel & 22.1.0 \\
Python & 3.12.2 (CPython) \\
Java & OpenJDK 17 \\
Elasticsearch & 7.17.15 \\
Snowstorm & 10.7.0 \\
SNOMED CT & SNOMED CT International RF2 2025-05-01 \\
Key Python Packages & torch=2.7.1, transformers=4.52.4, openai=1.82.1, numpy=1.26.4, pandas=2.2.3 \\
\\
\bottomrule
\end{tabular}
\caption{Machine 2: Trial-side Parsing Environment, also used for speed and scalability results}
\label{tab:machine_2}
\end{table}

\begin{table}[htbp]
\centering
\small
\begin{tabular}{p{0.30\linewidth} p{0.64\linewidth}}
\toprule
\textbf{Item} & \textbf{Value} \\
\midrule
Machine & ASUS \\
Chip & AMD Ryzen 7 9800X3D \\
CPU Cores & 16 \\
GPU & NVIDIA GeForce RTX 5090 \\
Memory & 60 GB \\
Operating System & Ubuntu 24.04.2 LTS \\
Kernel & 6.14.0-1020-oem \\
Python & 3.12.7 (CPython) \\
Java & OpenJDK 17 \\
Elasticsearch & 7.17.15 \\
Snowstorm & 10.7.0 \\
SNOMED CT & SNOMED CT International RF2 2025-05-01 \\
Key Python Packages & torch=2.9.1, transformers=4.52.4, openai=1.82.1, numpy=1.26.4, pandas=2.2.3 \\
\\
\bottomrule
\end{tabular}
\caption{Machine 3: Experimental environment}
\label{tab:machine_3}
\end{table}

\begin{table}[htbp]
\centering
\small
\setlength{\tabcolsep}{5pt}
\renewcommand{\arraystretch}{1.05}
\begin{tabular}{lcc}
\toprule
\textbf{Setting} & \textbf{GPT-4.1} & \textbf{GPT-5} \\
\midrule
Model & \texttt{gpt-4.1} & \texttt{gpt-5} \\
Temperature & 0 & omitted \\
Top-$p$ & 0.001 & omitted \\
Output-token control & \texttt{max\_tokens} & \texttt{max\_completion\_tokens} \\
Seed & 42 & 42 \\
Connect timeout & 10s & 10s \\
Initial read timeout & 55s & 55s \\
Read-timeout cap & 300s & 300s \\
Max retries & 5 & 5 \\
Backoff & exponential full-jitter & exponential full-jitter \\
Fallback endpoint & optional & optional \\
\bottomrule
\end{tabular}
\caption{Inference configurations used for the GPT-4.1- and GPT-5-based components in our experiments. For GPT-5, sampling parameters such as temperature and top-$p$ are omitted.}
\label{tab:llm_config}
\end{table}

\begin{table}[htbp]
\centering
\small
\begin{tabular}{p{0.30\linewidth} p{0.64\linewidth}}
\toprule
\textbf{Item} & \textbf{Value} \\
\midrule
Terminology server & Snowstorm 10.7.0 \\
Ontology release & SNOMED CT International RF2 2025-05-01 \\
Vector index & \texttt{snomed\_vectors} \\
Embedding model & \texttt{cambridgeltl/sapbert-from-pubmedbert-fulltext} \\
Fuzzy threshold & 0.86 \\
Vector top-$k$ & 5 \\
Vector score cutoff & 0.30 \\
Vector overshoot & 4 \\
Allowed SNOMED top-level roots & 363787002, 71388002, 404684003, 373873005, 105590001 \\
\bottomrule
\end{tabular}
\caption{SNOMED and terminology-retrieval configuration used by the ontology grounding pipeline.}
\label{tab:snomed_config}
\end{table}

\newpage

\newpage
\section{Symbolic Representation and Subsumption}\label{app:representation_umbrella}

\subsection{Patient-Side Implementation of Subsumption Relationships}\label{app:subsumption}

This appendix provides implementation details for the ontology-aware
patient-side expansion described in Section~\ref{sec:formalization}. In the
main text, the ontology is written as
\(O=(R,K,Q,\sqsubseteq)\), with
\[
\sqsubseteq
=
\sqsubseteq_{\mathrm{R}}
\cup
\sqsubseteq_{\mathrm{K}}
\cup
\sqsubseteq_{\mathrm{causal}}
\cup
\sqsubseteq_{\mathrm{Q}}.
\]
The materialization described here concerns patient-side subsumptions under
\(\sqsubseteq_{\mathrm{K}}\), \(\sqsubseteq_{\mathrm{R}}\), and selected
SNOMED-derived \(\sqsubseteq_{\mathrm{causal}}\) enrichments. Before
retrieval, we materialize the resulting supported patient facts, yielding an
ontology-enriched patient representation used for matching against trial
intents and trial-side constraints.

We distinguish two complementary mechanisms. First, we materialize
\emph{entailed patient facts}, which become part of the patient-side symbolic
representation used at retrieval time. Second, we maintain a separate
\emph{relation overlay}, which augments SNOMED with task-specific semantic
links used as admissible alternatives during matching and debugging, but not
as unconditional entailments.

\subsubsection{Patient-side entailment materialization}
\label{app:ontology_entailment_materialization}

Facts are represented as canonical predicates with explicit time windows and are
deduplicated by
\[
\begin{aligned}
(&\texttt{name},\ \texttt{start\_time\_in\_hours},\ \\
 &\texttt{end\_time\_in\_hours},\ \texttt{start\_time\_inclusive},\ \\
 &\texttt{end\_time\_inclusive}).
\end{aligned}
\]
Each derived fact also carries provenance metadata, including its source fact,
the triggering ontology or rule entry, derivation type, and hop depth where
applicable. Materialization is implemented in a streaming, memory-safe manner
with on-disk deduplication.

\paragraph{Concept-level closure (\(\sqsubseteq_{\mathrm{K}}\)).}
For a positive concept-grounded fact, we materialize supporting ancestor facts
under SNOMED \texttt{is-a} relations while preserving the original time window
and qualifiers. For explicit negation, we optionally materialize a bounded set
of negative descendants, using caps and maximum hop limits to control
combinatorial growth. This stage implements the patient-side concept closure
corresponding to \(\sqsubseteq_{\mathrm{K}}\).

\paragraph{Medical-relation closure (\(\sqsubseteq_{\mathrm{R}}\)).}
We also materialize supporting facts induced by subsumption over medical
relations. Concretely, if \(r_p \sqsubseteq_{\mathrm{R}} r_c\), then a fact
with predicate \((r_p,k,q)\) supports one with predicate \((r_c,k,q)\). These
derivations preserve the grounded concept, qualifier bundle, and time window,
while recording the applied relation transformation in provenance. This stage
captures support within \(O_R=(R,\sqsubseteq_{\mathrm{R}})\), such as
\(\textsc{HasDiagnosisOf} \sqsubseteq_{\mathrm{R}} \textsc{HasFindingOf}\).

\paragraph{Causal/dependency closure (\(\sqsubseteq_{\mathrm{causal}}\)).}
Some medically valid supports depend jointly on the relation and grounded
concept and therefore cannot be reduced to concept subsumption or medical-
relation subsumption alone. To capture such cases, we materialize derived facts
when
\[
(r_p,k_p) \sqsubseteq_{\mathrm{causal}} (r_c,k_c).
\]
These derivations preserve source time windows and qualifiers unless a specific
enrichment explicitly normalizes them. 
This stage corresponds to causal or dependency subsumption relations across relations and
concepts derived from SNOMED definitional structure.

\subsubsection{Medical-relation subsumption rules (\(\sqsubseteq_{\mathrm{R}}\))}

Medical-relation subsumption is implemented through a curated, recall-oriented
rule layer over canonical predicate templates. The full rule inventory is
presented in Appendix~\ref{app:medical_relation_subsumption_rules}. Each rule
captures a pattern under which one grounded patient fact should support one or
more additional facts in our retrieval vocabulary. When a rule applies, the
grounded concept \(k\), qualifier bundle \(q\), and temporal context are
preserved, while the medical relation is generalized in a controlled way.

\subsubsection{SNOMED-derived causal/dependency subsumption relations (\(\sqsubseteq_{\mathrm{causal}}\))}
\label{app:causal_subsumption}

The relation
\[
\sqsubseteq_{\mathrm{causal}} \subseteq (R,K)\times(R,K)
\]
captures cases where subsumption depends on the \emph{combination} of a relation
and a concept, rather than on the relation or concept alone. For this reason,
it cannot be reduced to \(\sqsubseteq_{\mathrm{R}}\) or
\(\sqsubseteq_{\mathrm{K}}\). We derive these links from SNOMED CT’s formal
concept definitions.

In SNOMED CT, a concept is defined using typed attributes that connect it to
other concepts. Some attributes may be placed in the same relationship group,
which means they should be interpreted together as part of one coherent
definition. The SNOMED CT concept model, together with the machine-readable
concept model (MRCM), specifies which attributes are allowed, what kinds of
values they can take, and whether they are grouped. This structure lets us
derive ontology-backed causal or dependency-style subsumption relations.

In our implementation, \(\sqsubseteq_{\mathrm{causal}}\) is used in two closely
related ways. For finding \(\rightarrow\) procedure, we use two families of
attributes. The first is the grouped pair \texttt{Interprets}
(\texttt{363714003}) and \texttt{Has interpretation} (\texttt{363713009}). When
a finding concept has an \texttt{Interprets} link to a procedure concept, we
derive a base fact that the patient underwent that procedure. If the same group
also contains a \texttt{Has interpretation} value, we further refine the
entailed procedure fact into a normalized status such as \texttt{positive},
\texttt{negative}, \texttt{normal}, or \texttt{abnormal} using curated
mappings.

The second family includes more indirect procedure-linking attributes:
\texttt{During} (\texttt{371881003}), \texttt{After} (\texttt{255234002}),
\texttt{Due to} (\texttt{42752001}), \texttt{Associated with}
(\texttt{47429007}), and \texttt{Temporally related to}
(\texttt{726633004}). These support more conservative finding
\(\rightarrow\) procedure links and yield only the base procedure fact, without
a normalized status.

The same grouped \texttt{Interprets}/\texttt{Has interpretation} pattern also
supports finding \(\rightarrow\) observable-entity entailments. For example,
\textit{Acute kidney injury} is defined using \texttt{Interprets}
\(\rightarrow\) \textit{Renal function} and \texttt{Has interpretation}
\(\rightarrow\) \textit{Impaired}. From this, we derive an observable-entity
status fact. Similarly, \textit{Postoperative hemorrhage} is defined with
\texttt{After} some \textit{Surgical procedure}, which illustrates the more
conservative attribute family that supports finding \(\rightarrow\) procedure
entailment without status refinement.

We also use the same ontology structure in the reverse direction, to map status
facts back to finding concepts. For observable-entity-status
\(\rightarrow\) finding and procedure-status \(\rightarrow\) finding, we
retrieve candidate findings using SNOMED CT’s Expression Constraint Language
(ECL)~\citep{snomed_ecl}, which is SNOMED’s formal query language for searching concept sets and
attribute-based definitions. Here, we use ECL to find finding concepts whose
definitions include both an \texttt{Interprets} link to the relevant observable
entity or procedure and a \texttt{Has interpretation} value. We then keep only
candidates whose definitions are fully defined and whose relevant attributes
remain in the same relationship group. For example, \textit{Normal renal
function} is defined with \texttt{Interprets} = \textit{Renal function} and
\texttt{Has interpretation} = \textit{Normal}, showing how an observable-entity
status fact can be normalized back into a finding concept.

The curated mappings that convert \texttt{Has interpretation} values into the
normalized status labels \texttt{positive}, \texttt{negative},
\texttt{normal}, and \texttt{abnormal} are shown in
Tables~\ref{tab:f2oe_hasinterpretation_mapping}
and~\ref{tab:f2p_hasinterpretation_mapping}. These correspond to the
finding\(\rightarrow\)observable-entity and finding\(\rightarrow\)procedure
entailment types, respectively.

\subsubsection{Canonicalization of Qualifiers and Subsumption Relations Among Qualifiers (\(\sqsubseteq_{\mathrm{Q}}\))} \label{app:patient_qualifier_subsumption}

In our implementation, we canonicalize temporal qualifiers, together with several recall-oriented, schema-dependent qualifier variables (Appendix~\ref{app:naming_schema}). Temporal qualifiers are normalized into interval representations of the form \((\text{lower bound}, \text{upper bound}, \text{lower inclusiveness}, \text{upper inclusiveness})\). 

For inclusion criteria, subsumption between temporal qualifiers is defined by overlap between the criterion window and the patient’s possible fact window. For exclusion criteria, subsumption is defined more strictly: the exclusion-window requirement must strictly contain the patient fact window. The construction of patient fact windows is described in Appendix~\ref{app:patient_intervalization}. 

More generally, we treat the absence of a qualifier as the weakest form, so any qualifier subsumes no qualifier.

\subsubsection{Multi-pass closure computation}

We compute patient-side expansion in multiple passes to approach a fixpoint without allowing uncontrolled growth. Each pass executes a fixed sequence of stages corresponding to concept-level, relation-level, and fact-level materialization. At the end of each pass, only newly derived facts are retained as seeds for the next pass, after subtracting any fact whose deduplication key has appeared previously. This guarantees monotone growth in unique facts and ensures termination under fixed caps. After the final pass, we export the aggregated and deduplicated ontology-enriched fact set for retrieval and downstream usage.

\subsubsection{Relation-overlay curation}

In addition to the patient-side entailment structures \(\sqsubseteq_{\mathrm{Q}}\), \(\sqsubseteq_{\mathrm{K}}\), \(\sqsubseteq_{\mathrm{R}}\), \(\sqsubseteq_{\mathrm{causal}}\), we curate a separate relation overlay that augments SNOMED with task-specific semantic links. Unlike the entailed facts materialized above, these overlay edges are \emph{not} treated as unconditional logical consequences. Instead, they are used as admissible alternatives during matching and as explanatory support during auditing and debugging.

The relation overlay is intended for cases where strict ontology entailment is too narrow for practical patient--trial matching, but repeated review suggests that certain concept pairs should still be considered connected for retrieval or downstream analysis. These links therefore improve robustness without being folded into the core patient-side logical closure.

We store multiple overlay relation types, each with explicit provenance:
\begin{itemize}
  \item \textbf{Overlap links:} clinically meaningful partial-overlap relationships between two concepts.
  \item \textbf{Synonym and near-synonym links:} semantically close alternatives surfaced during concept discovery and verified by an LLM auditor.
  \item \textbf{Task bridges:} curated mappings that are neither \texttt{is-a} nor strict synonymy, but repeatedly help connect patient phrasing or inferred patient states to trial-side requirements.
\end{itemize}

The overlay is designed to capture reusable task-specific relations that are missing from the native ontology but repeatedly arise in patient--trial matching. Rather than expanding concepts through local ontology-neighborhood traversal, we focus on discovering clinically meaningful missing links that can be verified once and then reused across future patients and trials.

To identify such links, we use an embedding-based discovery pipeline grounded in patient records. Starting from an existing seed set of extracted patient facts, we retrieve vector-nearest SNOMED concepts using dense retrieval over concept descriptions. These retrieved concepts serve as candidates for missing concepts or missing bridges to existing seeds. An LLM-based medical auditor then determines whether each candidate is clearly supported by the patient note and whether it should be linked to an existing concept through one of the overlay relation types. Accepted links are persisted as reusable overlay edges with note-level provenance, allowing them to support future matching beyond the originating example. We run this process for a fixed number of passes and cap both the number of candidates considered per seed and the number of newly accepted links per pass to ensure predictable runtime and conservative growth.

All overlay candidates are passed through a unified verification interface and then persisted to a local SQLite store. Each stored edge includes:
\begin{itemize}
  \item \textbf{edge key:} \((\texttt{base\_conceptId}, \texttt{candidate\_conceptId}, \texttt{relation\_type})\)
  \item \textbf{verifier metadata:} model version, prompt hash, and timestamp
  \item \textbf{decision:} accept/reject with a brief justification
  \item \textbf{provenance:} associated trial identifiers and/or patient-note identifiers, together with the discovery path
\end{itemize}
We cache verifier outputs so repeated runs can reuse prior judgments, and we write to the database only after successful verifier execution. Downstream components query this store to retrieve admissible alternative concepts during matching and to surface supporting explanations during debugging.

\subsubsection{Curated inventory of medical-relation subsumption rules (\(\sqsubseteq_{\mathrm{R}}\))}
\label{app:medical_relation_subsumption_rules}

This section lists the manually curated ruleset used to
operationalize medical-relation subsumption. 

\begin{MyVerbatim}
{
  "rules": [
    {
      "id": "suspicion_implies_finding_same_time",
      "match_template": "patient_has_suspicion_of_{e}_{t}",
      "require_bool": true,
      "produce": [
        { "template": "patient_has_finding_of_{e}_{t}", "type": "Bool", "value": true, "preserve_qualifiers": true }
      ]
    },
    {
      "id": "suspicion_implies_diagnosis_same_time",
      "match_template": "patient_has_suspicion_of_{e}_{t}",
      "require_bool": true,
      "produce": [
        { "template": "patient_has_diagnosis_of_{e}_{t}", "type": "Bool", "value": true, "preserve_qualifiers": true }
      ]
    },
    {
      "id": "suspicion_implies_symptoms_same_time",
      "match_template": "patient_has_suspicion_of_{e}_{t}",
      "require_bool": true,
      "produce": [
        { "template": "patient_has_symptoms_of_{e}_{t}", "type": "Bool", "value": true, "preserve_qualifiers": true }
      ]
    },
    {
      "id": "diagnosis_implies_finding_same_time",
      "match_template": "patient_has_diagnosis_of_{e}_{t}",
      "require_bool": true,
      "produce": [
        { "template": "patient_has_finding_of_{e}_{t}", "type": "Bool", "value": true, "preserve_qualifiers": true }
      ]
    },
    {
      "id": "finding_implies_diagnosis_same_time",
      "match_template": "patient_has_finding_of_{e}_{t}",
      "require_bool": true,
      "produce": [
        { "template": "patient_has_diagnosis_of_{e}_{t}", "type": "Bool", "value": true, "preserve_qualifiers": true }
      ]
    },
    {
      "id": "finding_implies_symptoms_same_time",
      "match_template": "patient_has_finding_of_{e}_{t}",
      "require_bool": true,
      "produce": [
        { "template": "patient_has_symptoms_of_{e}_{t}", "type": "Bool", "value": true, "preserve_qualifiers": true }
      ]
    },
    {
      "id": "finding_implies_clinical_signs_same_time",
      "match_template": "patient_has_finding_of_{e}_{t}",
      "require_bool": true,
      "produce": [
        { "template": "patient_has_clinical_signs_of_{e}_{t}", "type": "Bool", "value": true, "preserve_qualifiers": true }
      ]
    },
    {
      "id": "is_undergoing_implies_has_undergone",
      "match_template": "patient_is_undergoing_{e}_{t}",
      "require_bool": true,
      "produce": [
        { "template": "patient_has_undergone_{e}_{t}", "type": "Bool", "value": true, "preserve_qualifiers": true }
      ]
    },

    {
      "id": "status_positive_implies_not_negative_same_time",
      "match_template": "patients_{e}_is_positive_{t}",
      "require_bool": true,
      "produce": [
        { "template": "patients_{e}_is_negative_{t}", "type": "Bool", "value": false, "preserve_qualifiers": true }
      ]
    },
    {
      "id": "status_negative_implies_not_positive_same_time",
      "match_template": "patients_{e}_is_negative_{t}",
      "require_bool": true,
      "produce": [
        { "template": "patients_{e}_is_positive_{t}", "type": "Bool", "value": false, "preserve_qualifiers": true }
      ]
    },
    {
      "id": "status_abnormal_implies_not_normal_same_time",
      "match_template": "patients_{e}_is_abnormal_{t}",
      "require_bool": true,
      "produce": [
        { "template": "patients_{e}_is_normal_{t}", "type": "Bool", "value": false, "preserve_qualifiers": true }
      ]
    },
    {
      "id": "status_normal_implies_not_abnormal_same_time",
      "match_template": "patients_{e}_is_normal_{t}",
      "require_bool": true,
      "produce": [
        { "template": "patients_{e}_is_abnormal_{t}", "type": "Bool", "value": false, "preserve_qualifiers": true }
      ]
    },

    {
      "id": "allergy_implies_hypersensitivity_same_time",
      "match_template": "patient_has_allergy_to_{e}_{t}",
      "require_bool": true,
      "produce": [
        { "template": "patient_has_hypersensitivity_to_{e}_{t}", "type": "Bool", "value": true, "preserve_qualifiers": true }
      ]
    },
    {
      "id": "nonimmune_hypersensitivity_implies_hypersensitivity_same_time",
      "match_template": "patient_has_nonimmune_hypersensitivity_to_{e}_{t}",
      "require_bool": true,
      "produce": [
        { "template": "patient_has_hypersensitivity_to_{e}_{t}", "type": "Bool", "value": true, "preserve_qualifiers": true }
      ]
    },

    {
      "id": "diagnosis_implies_suspicion_same_time",
      "match_template": "patient_has_diagnosis_of_{e}_{t}",
      "require_bool": true,
      "produce": [
        { "template": "patient_has_suspicion_of_{e}_{t}", "type": "Bool", "value": true, "preserve_qualifiers": true }
      ]
    },
    {
      "id": "finding_implies_suspicion_same_time",
      "match_template": "patient_has_finding_of_{e}_{t}",
      "require_bool": true,
      "produce": [
        { "template": "patient_has_suspicion_of_{e}_{t}", "type": "Bool", "value": true, "preserve_qualifiers": true }
      ]
    },
    {
      "id": "symptoms_implies_suspicion_same_time",
      "match_template": "patient_has_symptoms_of_{e}_{t}",
      "require_bool": true,
      "produce": [
        { "template": "patient_has_suspicion_of_{e}_{t}", "type": "Bool", "value": true, "preserve_qualifiers": true }
      ]
    },
    {
      "id": "clinical_signs_implies_suspicion_same_time",
      "match_template": "patient_has_clinical_signs_of_{e}_{t}",
      "require_bool": true,
      "produce": [
        { "template": "patient_has_suspicion_of_{e}_{t}", "type": "Bool", "value": true, "preserve_qualifiers": true }
      ]
    },

    {
      "id": "is_undergoing_implies_can_undergo_same_time",
      "match_template": "patient_is_undergoing_{e}_{t}",
      "require_bool": true,
      "produce": [
        { "template": "patient_can_undergo_{e}_{t}", "type": "Bool", "value": true, "preserve_qualifiers": true }
      ]
    },
    {
      "id": "will_undergo_implies_can_undergo_same_time",
      "match_template": "patient_will_undergo_{e}_{t}",
      "require_bool": true,
      "produce": [
        { "template": "patient_can_undergo_{e}_{t}", "type": "Bool", "value": true, "preserve_qualifiers": true }
      ]
    },

    {
      "id": "procedure_positive_outcome_implies_has_undergone_same_time",
      "match_template": "patient_has_undergone_{e}_{t}_outcome_is_positive",
      "require_bool": true,
      "produce": [
        { "template": "patient_has_undergone_{e}_{t}", "type": "Bool", "value": true, "preserve_qualifiers": true }
      ]
    },
    {
      "id": "procedure_negative_outcome_implies_has_undergone_same_time",
      "match_template": "patient_has_undergone_{e}_{t}_outcome_is_negative",
      "require_bool": true,
      "produce": [
        { "template": "patient_has_undergone_{e}_{t}", "type": "Bool", "value": true, "preserve_qualifiers": true }
      ]
    },
    {
      "id": "procedure_abnormal_outcome_implies_has_undergone_same_time",
      "match_template": "patient_has_undergone_{e}_{t}_outcome_is_abnormal",
      "require_bool": true,
      "produce": [
        { "template": "patient_has_undergone_{e}_{t}", "type": "Bool", "value": true, "preserve_qualifiers": true }
      ]
    },
    {
      "id": "procedure_normal_outcome_implies_has_undergone_same_time",
      "match_template": "patient_has_undergone_{e}_{t}_outcome_is_normal",
      "require_bool": true,
      "produce": [
        { "template": "patient_has_undergone_{e}_{t}", "type": "Bool", "value": true, "preserve_qualifiers": true }
      ]
    },

    {
      "id": "symptoms_implies_finding_same_time",
      "match_template": "patient_has_symptoms_of_{e}_{t}",
      "require_bool": true,
      "produce": [
        { "template": "patient_has_finding_of_{e}_{t}", "type": "Bool", "value": true, "preserve_qualifiers": true }
      ]
    },
    {
      "id": "clinical_signs_implies_finding_same_time",
      "match_template": "patient_has_clinical_signs_of_{e}_{t}",
      "require_bool": true,
      "produce": [
        { "template": "patient_has_finding_of_{e}_{t}", "type": "Bool", "value": true, "preserve_qualifiers": true }
      ]
    },

    {
      "id": "diagnosis_false_implies_finding_false_same_time",
      "match_template": "patient_has_diagnosis_of_{e}_{t}",
      "require_bool": false,
      "produce": [
        { "template": "patient_has_finding_of_{e}_{t}", "type": "Bool", "value": false, "preserve_qualifiers": true }
      ]
    },
    {
      "id": "finding_false_implies_diagnosis_false_same_time",
      "match_template": "patient_has_finding_of_{e}_{t}",
      "require_bool": false,
      "produce": [
        { "template": "patient_has_diagnosis_of_{e}_{t}", "type": "Bool", "value": false, "preserve_qualifiers": true }
      ]
    },

    {
      "id": "symptoms_false_implies_finding_false_same_time",
      "match_template": "patient_has_symptoms_of_{e}_{t}",
      "require_bool": false,
      "produce": [
        { "template": "patient_has_finding_of_{e}_{t}", "type": "Bool", "value": false, "preserve_qualifiers": true }
      ]
    },
    {
      "id": "clinical_signs_false_implies_finding_false_same_time",
      "match_template": "patient_has_clinical_signs_of_{e}_{t}",
      "require_bool": false,
      "produce": [
        { "template": "patient_has_finding_of_{e}_{t}", "type": "Bool", "value": false, "preserve_qualifiers": true }
      ]
    },

    {
      "id": "not_has_undergone_implies_not_undergoing_same_time",
      "match_template": "patient_has_undergone_{e}_{t}",
      "require_bool": false,
      "produce": [
        { "template": "patient_is_undergoing_{e}_{t}", "type": "Bool", "value": false, "preserve_qualifiers": true }
      ]
    },

    {
      "id": "positive_false_implies_negative_true_same_time",
      "match_template": "patients_{e}_is_positive_{t}",
      "require_bool": false,
      "produce": [
        { "template": "patients_{e}_is_negative_{t}", "type": "Bool", "value": true, "preserve_qualifiers": true }
      ]
    },
    {
      "id": "negative_false_implies_positive_true_same_time",
      "match_template": "patients_{e}_is_negative_{t}",
      "require_bool": false,
      "produce": [
        { "template": "patients_{e}_is_positive_{t}", "type": "Bool", "value": true, "preserve_qualifiers": true }
      ]
    },
    {
      "id": "abnormal_false_implies_normal_true_same_time",
      "match_template": "patients_{e}_is_abnormal_{t}",
      "require_bool": false,
      "produce": [
        { "template": "patients_{e}_is_normal_{t}", "type": "Bool", "value": true, "preserve_qualifiers": true }
      ]
    },
    {
      "id": "normal_false_implies_abnormal_true_same_time",
      "match_template": "patients_{e}_is_normal_{t}",
      "require_bool": false,
      "produce": [
        { "template": "patients_{e}_is_abnormal_{t}", "type": "Bool", "value": true, "preserve_qualifiers": true }
      ]
    },


    {
      "id": "no_hypersensitivity_implies_no_allergy_same_time",
      "match_template": "patient_has_hypersensitivity_to_{e}_{t}",
      "require_bool": false,
      "produce": [
        { "template": "patient_has_allergy_to_{e}_{t}", "type": "Bool", "value": false, "preserve_qualifiers": true }
      ]
    },
    {
      "id": "no_hypersensitivity_implies_no_nonimmune_hypersensitivity_same_time",
      "match_template": "patient_has_hypersensitivity_to_{e}_{t}",
      "require_bool": false,
      "produce": [
        { "template": "patient_has_nonimmune_hypersensitivity_to_{e}_{t}", "type": "Bool", "value": false, "preserve_qualifiers": true }
      ]
    },

    {
      "id": "cannot_undergo_implies_not_undergoing_same_time",
      "match_template": "patient_can_undergo_{e}_{t}",
      "require_bool": false,
      "produce": [
        { "template": "patient_is_undergoing_{e}_{t}", "type": "Bool", "value": false, "preserve_qualifiers": true }
      ]
    },
    {
      "id": "cannot_undergo_implies_not_will_undergo_same_time",
      "match_template": "patient_can_undergo_{e}_{t}",
      "require_bool": false,
      "produce": [
        { "template": "patient_will_undergo_{e}_{t}", "type": "Bool", "value": false, "preserve_qualifiers": true }
      ]
    },

    {
      "id": "not_has_undergone_implies_no_positive_outcome_same_time",
      "match_template": "patient_has_undergone_{e}_{t}",
      "require_bool": false,
      "produce": [
        { "template": "patient_has_undergone_{e}_{t}_outcome_is_positive", "type": "Bool", "value": false, "preserve_qualifiers": true },
        { "template": "patient_has_undergone_{e}_{t}_outcome_is_negative", "type": "Bool", "value": false, "preserve_qualifiers": true },
        { "template": "patient_has_undergone_{e}_{t}_outcome_is_abnormal", "type": "Bool", "value": false, "preserve_qualifiers": true },
        { "template": "patient_has_undergone_{e}_{t}_outcome_is_normal",   "type": "Bool", "value": false, "preserve_qualifiers": true }
      ]
    },

    {
      "id": "finding_false_implies_symptoms_false_same_time",
      "match_template": "patient_has_finding_of_{e}_{t}",
      "require_bool": false,
      "produce": [
        { "template": "patient_has_symptoms_of_{e}_{t}", "type": "Bool", "value": false, "preserve_qualifiers": true }
      ]
    },
    {
      "id": "finding_false_implies_clinical_signs_false_same_time",
      "match_template": "patient_has_finding_of_{e}_{t}",
      "require_bool": false,
      "produce": [
        { "template": "patient_has_clinical_signs_of_{e}_{t}", "type": "Bool", "value": false, "preserve_qualifiers": true }
      ]
    },

    {
      "id": "sex_female_implies_not_male",
      "match_template": "patient_sex_is_female_{t}",
      "require_bool": true,
      "produce": [
        { "template": "patient_sex_is_male_{t}", "type": "Bool", "value": false, "preserve_qualifiers": true },
        { "template": "patient_has_finding_of_female_sex_{t}", "type": "Bool", "value": true, "preserve_qualifiers": true }
      ]
    },
    {
      "id": "sex_male_implies_not_female",
      "match_template": "patient_sex_is_male_{t}",
      "require_bool": true,
      "produce": [
        { "template": "patient_sex_is_female_{t}", "type": "Bool", "value": false, "preserve_qualifiers": true }
      ]
    },
    {
      "id": "need_to_undergo_implies_is_undergoing",
      "match_template": "patient_needs_to_undergo_{e}_{t}",
      "require_bool": true,
      "produce": [
        { "template": "patient_is_undergoing_{e}_{t}", "type": "Bool", "value": true, "preserve_qualifiers": true }
      ]
    },
    {
      "id": "need_to_undergo_implies_will_undergo",
      "match_template": "patient_needs_to_undergo_{e}_{t}",
      "require_bool": true,
      "produce": [
        { "template": "patient_will_undergo_{e}_{t}", "type": "Bool", "value": true, "preserve_qualifiers": true }
      ]
    },
    {
      "id": "need_to_undergo_implies_can_undergo",
      "match_template": "patient_needs_to_undergo_{e}_{t}",
      "require_bool": true,
      "produce": [
        { "template": "patient_can_undergo_{e}_{t}", "type": "Bool", "value": true, "preserve_qualifiers": true }
      ]
    },
    {
      "id": "need_to_undergo_implies_has_undergone",
      "match_template": "patient_needs_to_undergo_{e}_{t}",
      "require_bool": true,
      "produce": [
        { "template": "patient_has_undergone_{e}_{t}", "type": "Bool", "value": true, "preserve_qualifiers": true }
      ]
    }
  ],
  "timeframe_implication": {
    "collapse_timeframes": false
  }
}
\end{MyVerbatim}

\subsubsection{\texttt{Has interpretation} mappings for causal/dependency subsumption}

Tables~\ref{tab:f2oe_hasinterpretation_mapping} and
\ref{tab:f2p_hasinterpretation_mapping} summarize the curated mappings from
observed \texttt{Has interpretation} values in SNOMED CT to the canonical vocabulary \texttt{normal}, \texttt{abnormal}, \texttt{positive}, and
\texttt{negative} for finding$\rightarrow$observable entity and
finding$\rightarrow$procedure entailments, respectively. We maintain separate
mappings for inclusion and exclusion because the two settings have different
error trade-offs. On the inclusion side, the mapping is intentionally
recall-oriented: broader or more ambiguous interpretation values may be allowed
to support canonicalized statuses so that potentially relevant trials are
not missed. On the exclusion side, the mapping is more conservative, since an
overly aggressive canonicalization could incorrectly disqualify a patient by
triggering an exclusion condition from weak or ambiguous evidence.


\begingroup
\tiny
\setlength{\LTleft}{0pt}
\setlength{\LTright}{0pt plus 1fill}
\setlength{\tabcolsep}{2pt}
\renewcommand{\arraystretch}{1.08}

\begin{longtable}{
  L{2.80cm}
  C{0.82cm}
  C{0.82cm}
  C{0.96cm}
  C{0.92cm}
  C{0.92cm}
  C{0.82cm}
  C{0.96cm}
  C{0.92cm}
  C{0.92cm}
}

\caption{Curated mapping from SNOMED CT \texttt{Has interpretation} values to canonicalized vocabulary for finding$\rightarrow$observable-entity entailment. \textbf{Count} reports the number of observed occurrences of each source interpretation term. A value of \texttt{1} in a status column indicates that the source term is mapped to that normalized status in our vocabulary. Columns marked \texttt{(inc)} are used for inclusion matching; columns marked \texttt{(exc)} are used for exclusion matching.}
\label{tab:f2oe_hasinterpretation_mapping}\\
\toprule
\textbf{Term} &
\textbf{Count} &
\textbf{\makecell{normal\\(inc)}} &
\textbf{\makecell{abnormal\\(inc)}} &
\textbf{\makecell{positive\\(inc)}} &
\textbf{\makecell{negative\\(inc)}} &
\textbf{\makecell{normal\\(exc)}} &
\textbf{\makecell{abnormal\\(exc)}} &
\textbf{\makecell{positive\\(exc)}} &
\textbf{\makecell{negative\\(exc)}} \\
\midrule
\endfirsthead

\multicolumn{10}{c}{\tablename\ \thetable\ -- continued from previous page} \\
\toprule
\textbf{Term} &
\textbf{Count} &
\textbf{\makecell{normal\\(inc)}} &
\textbf{\makecell{abnormal\\(inc)}} &
\textbf{\makecell{positive\\(inc)}} &
\textbf{\makecell{negative\\(inc)}} &
\textbf{\makecell{normal\\(exc)}} &
\textbf{\makecell{abnormal\\(exc)}} &
\textbf{\makecell{positive\\(exc)}} &
\textbf{\makecell{negative\\(exc)}} \\
\midrule
\endhead

\midrule
\multicolumn{10}{r}{Continued on next page} \\
\endfoot

\bottomrule
\endlastfoot

Impaired & 2280 &  & 1 &  &  &  & 1 &  &  \\
Abnormal & 2244 &  & 1 &  &  &  & 1 &  &  \\
Below reference range & 1030 &  & 1 &  &  &  &  &  &  \\
Increased & 967 &  & 1 &  &  &  &  &  &  \\
Absent & 962 &  & 1 &  &  &  &  &  &  \\
Decreased & 925 &  & 1 &  &  &  &  &  &  \\
Able & 767 & 1 &  &  &  &  &  &  &  \\
Difficulty & 697 &  & 1 &  &  &  &  &  &  \\
Unable & 678 &  & 1 &  &  &  &  &  &  \\
Does & 642 &  &  &  &  &  &  &  &  \\
Does not & 608 &  &  &  &  &  &  &  &  \\
Above reference range & 584 &  & 1 &  &  &  &  &  &  \\
Present & 527 &  &  & 1 &  &  &  &  &  \\
Normal & 470 & 1 &  &  &  & 1 &  &  &  \\
Altered & 204 &  & 1 &  &  &  &  &  &  \\
Non-patent & 186 &  & 1 &  &  &  &  &  &  \\
Dependent & 164 &  & 1 &  &  &  &  &  &  \\
Patent & 157 & 1 &  &  &  &  &  &  &  \\
Excessive & 135 &  & 1 &  &  &  &  &  &  \\
Slow & 101 &  &  &  &  &  &  &  &  \\
Deficient & 93 &  & 1 &  &  &  &  &  &  \\
High & 84 &  &  &  &  &  &  &  &  \\
Inadequate & 77 &  & 1 &  &  &  &  &  &  \\
Lacking & 55 &  & 1 &  &  &  &  &  &  \\
Positive & 41 &  &  & 1 &  &  &  & 1 &  \\
Within reference range & 37 & 1 &  &  &  & 1 &  &  &  \\
Low & 34 &  &  &  &  &  &  &  &  \\
Hypokinetic & 33 &  &  &  &  &  &  &  &  \\
Akinetic & 28 &  & 1 &  &  &  &  &  &  \\
Independent & 27 & 1 &  &  &  &  &  &  &  \\
Blue color & 26 &  & 1 &  &  &  &  &  &  \\
Detected & 25 &  &  & 1 &  &  &  &  &  \\
Reduced & 23 &  &  &  &  &  &  &  &  \\
Equivocal & 23 &  &  &  &  &  &  &  &  \\
Assisted & 22 &  &  &  &  &  &  &  &  \\
Improved & 21 &  &  &  &  &  &  &  &  \\
Abnormally high & 20 &  & 1 &  &  &  & 1 &  &  \\
Outside reference range & 20 &  & 1 &  &  &  & 1 &  &  \\
Narrow & 19 &  & 1 &  &  &  &  &  &  \\
Thin & 19 &  &  &  &  &  &  &  &  \\
Abnormally low & 18 &  & 1 &  &  &  & 1 &  &  \\
Negative & 16 &  &  &  & 1 &  &  &  & 1 \\
Not detected & 15 &  &  &  & 1 &  &  &  & 1 \\
Small & 15 &  &  &  &  &  &  &  &  \\
Red color & 13 &  &  &  &  &  &  &  &  \\
Decreased relative to previous & 11 &  &  &  &  &  &  &  &  \\
Obstructed & 10 &  & 1 &  &  &  &  &  &  \\
Pale color saturation & 10 &  & 1 &  &  &  &  &  &  \\
Moderate & 9 &  &  &  &  &  &  &  &  \\
Exaggerated & 9 &  &  &  &  &  &  &  &  \\
Inefficient & 8 &  &  &  &  &  &  &  &  \\
Broken & 8 &  &  &  &  &  &  &  &  \\
Asymmetry & 8 &  &  &  &  &  &  &  &  \\
Increased relative to previous & 8 &  &  &  &  &  &  &  &  \\
Bent & 7 &  &  &  &  &  &  &  &  \\
Yellow color & 7 &  &  &  &  &  &  &  &  \\
White color & 6 &  &  &  &  &  &  &  &  \\
Enlarged & 6 &  & 1 &  &  &  &  &  &  \\
Large & 5 &  &  &  &  &  &  &  &  \\
Green colour & 5 &  &  &  &  &  &  &  &  \\
Inconsistent & 5 &  &  &  &  &  &  &  &  \\
Incomplete & 5 &  & 1 &  &  &  &  &  &  \\
Able to and does & 4 & 1 &  &  &  &  &  &  &  \\
Short & 4 &  &  &  &  &  &  &  &  \\
Adequate & 4 & 1 &  &  &  &  &  &  &  \\
Wide & 4 &  &  &  &  &  &  &  &  \\
Decrease & 4 &  &  &  &  &  &  &  &  \\
Pink color & 3 &  & 1 &  &  &  &  &  &  \\
Insufficient & 3 &  &  &  &  &  &  &  &  \\
Complete & 3 & 1 &  &  &  &  &  &  &  \\
Raised & 3 &  &  &  &  &  &  &  &  \\
Normal range & 2 & 1 &  &  &  & 1 &  &  &  \\
Strong & 2 &  &  &  &  &  &  &  &  \\
Weak & 2 &  &  &  &  &  &  &  &  \\
Blue-red color & 2 &  & 1 &  &  &  &  &  &  \\
Hyperkinetic & 2 &  & 1 &  &  &  &  &  &  \\
Decreased relative to baseline & 2 &  & 1 &  &  &  &  &  &  \\
Gray color & 2 &  & 1 &  &  &  &  &  &  \\
Preparation stage & 2 &  &  &  &  &  &  &  &  \\
Good & 2 &  &  &  &  &  &  &  &  \\
Inaccurate & 2 &  &  &  &  &  &  &  &  \\
Black color & 2 &  &  &  &  &  &  &  &  \\
Brown color & 2 &  &  &  &  &  &  &  &  \\
Very low & 2 &  & 1 &  &  &  & 1 &  &  \\
Maintenance stage & 2 &  &  &  &  &  &  &  &  \\
Action stage & 2 &  &  &  &  &  &  &  &  \\
Scanty & 2 &  & 1 &  &  &  &  &  &  \\
Contemplation stage & 2 &  &  &  &  &  &  &  &  \\
Precontemplation stage & 2 &  &  &  &  &  &  &  &  \\
Long duration & 2 &  &  &  &  &  &  &  &  \\
No status change & 2 & 1 &  &  &  &  &  &  &  \\
Borderline high & 1 &  &  &  &  &  &  &  &  \\
Late & 1 &  &  &  &  &  &  &  &  \\
Inconclusive & 1 &  &  &  &  &  &  &  &  \\
Trace & 1 &  &  & 1 &  &  &  &  &  \\
Efficient & 1 &  & 1 &  &  &  &  &  &  \\
Present one plus out of three plus & 1 &  &  & 1 &  &  &  & 1 &  \\
Silver gray color & 1 &  &  &  &  &  &  &  &  \\
Extremely high & 1 &  & 1 &  &  &  &  &  &  \\
Borderline low & 1 &  &  &  &  &  &  &  &  \\
Dark yellow color & 1 &  &  &  &  &  &  &  &  \\
Very slow & 1 &  & 1 &  &  &  &  &  &  \\
Irregular pattern & 1 &  & 1 &  &  &  &  &  &  \\
Present two plus out of three plus & 1 &  &  & 1 &  &  &  & 1 &  \\
Termination stage & 1 &  &  &  &  &  &  &  &  \\
Very high & 1 &  & 1 &  &  &  &  &  &  \\
Inefficient emptying & 1 &  & 1 &  &  &  &  &  &  \\
Indeterminate & 1 &  &  &  &  &  &  &  &  \\
None & 1 &  &  &  & 1 &  &  &  & 1 \\
Present three plus out of three plus & 1 &  &  & 1 &  &  &  & 1 &  \\
Decreased relative to expected & 1 &  &  &  &  &  &  &  &  \\

\end{longtable}
\endgroup

\begingroup
\tiny
\setlength{\LTleft}{0pt}
\setlength{\LTright}{0pt plus 1fill}
\setlength{\tabcolsep}{2pt}
\renewcommand{\arraystretch}{1.08}

\begin{longtable}{
  L{2.80cm}
  C{0.82cm}
  C{0.82cm}
  C{0.96cm}
  C{0.92cm}
  C{0.92cm}
  C{0.82cm}
  C{0.96cm}
  C{0.92cm}
  C{0.92cm}
}

\caption{Curated mapping from SNOMED CT \texttt{Has interpretation} values to canonicalized vocabulary for finding$\rightarrow$procedure entailment. \textbf{Count} reports the number of observed occurrences of each source interpretation term. A value of \texttt{1} in a status column indicates that the source term is mapped to that normalized status in our vocabulary. Columns marked \texttt{(inc)} are used for inclusion matching; columns marked \texttt{(exc)} are used for exclusion matching.}
\label{tab:f2p_hasinterpretation_mapping}\\
\toprule
\textbf{Term} &
\textbf{Count} &
\textbf{\makecell{normal\\(inc)}} &
\textbf{\makecell{abnormal\\(inc)}} &
\textbf{\makecell{positive\\(inc)}} &
\textbf{\makecell{negative\\(inc)}} &
\textbf{\makecell{normal\\(exc)}} &
\textbf{\makecell{abnormal\\(exc)}} &
\textbf{\makecell{positive\\(exc)}} &
\textbf{\makecell{negative\\(exc)}} \\
\midrule
\endfirsthead

\multicolumn{10}{c}{\tablename\ \thetable\ -- continued from previous page} \\
\toprule
\textbf{Term} &
\textbf{Count} &
\textbf{\makecell{normal\\(inc)}} &
\textbf{\makecell{abnormal\\(inc)}} &
\textbf{\makecell{positive\\(inc)}} &
\textbf{\makecell{negative\\(inc)}} &
\textbf{\makecell{normal\\(exc)}} &
\textbf{\makecell{abnormal\\(exc)}} &
\textbf{\makecell{positive\\(exc)}} &
\textbf{\makecell{negative\\(exc)}} \\
\midrule
\endhead

\midrule
\multicolumn{10}{r}{Continued on next page} \\
\endfoot

\bottomrule
\endlastfoot

$\varnothing$ (missing) & 3382 &  &  &  &  &  &  &  &  \\
Below reference range & 1324 &  & 1 &  &  &  & 1 &  &  \\
Above reference range & 971 &  & 1 &  &  &  & 1 &  &  \\
Abnormal & 746 &  & 1 &  &  &  & 1 &  &  \\
Detected & 324 &  &  & 1 &  &  &  & 1 &  \\
Within reference range & 282 & 1 &  &  &  & 1 &  &  &  \\
Normal & 225 & 1 &  &  &  & 1 &  &  &  \\
Present & 158 &  &  & 1 &  &  &  & 1 &  \\
Not detected & 144 &  &  &  & 1 &  &  &  & 1 \\
Outside reference range & 138 &  & 1 &  &  &  & 1 &  &  \\
Isolated & 29 &  &  & 1 &  &  &  & 1 &  \\
Not isolated & 21 &  &  &  & 1 &  &  &  & 1 \\
Positive & 19 &  &  & 1 &  &  &  & 1 &  \\
Equivocal & 17 &  &  &  &  &  &  &  &  \\
Negative & 17 &  &  &  & 1 &  &  &  & 1 \\
Above therapeutic range & 15 &  &  &  &  &  &  &  &  \\
Within therapeutic range & 15 &  &  &  &  &  &  &  &  \\
Below therapeutic range & 14 &  &  &  &  &  &  &  &  \\
Absent & 13 &  &  &  & 1 &  &  &  & 1 \\
Abnormal presence of & 7 &  & 1 & 1 &  &  & 1 & 1 &  \\
Increased & 6 &  &  &  &  &  &  &  &  \\
Not seen & 6 &  &  &  & 1 &  &  &  & 1 \\
Trace & 6 &  &  & 1 &  &  &  & 1 &  \\
+ & 4 &  &  & 1 &  &  &  & 1 &  \\
++ & 4 &  &  & 1 &  &  &  & 1 &  \\
+++ & 4 &  &  & 1 &  &  &  & 1 &  \\
++++ & 4 &  &  & 1 &  &  &  & 1 &  \\
Abnormally low & 4 &  & 1 &  &  &  & 1 &  &  \\
Inconclusive & 3 &  &  &  &  &  &  &  &  \\
Indeterminate & 3 &  &  &  &  &  &  &  &  \\
High & 2 &  &  &  &  &  &  &  &  \\
Present one plus out of three plus & 2 &  &  & 1 &  &  &  & 1 &  \\
Present three plus out of three plus & 2 &  &  & 1 &  &  &  & 1 &  \\
Present two plus out of three plus & 2 &  &  & 1 &  &  &  & 1 &  \\
Very low & 2 &  & 1 &  &  &  & 1 &  &  \\
Abnormally high & 1 &  & 1 &  &  &  & 1 &  &  \\
Altered & 1 &  & 1 &  &  &  &  &  &  \\
Borderline high & 1 &  &  &  &  &  &  &  &  \\
Deficient & 1 &  & 1 &  &  &  &  &  &  \\
Impaired & 1 &  & 1 &  &  &  &  &  &  \\
Improved & 1 &  &  &  &  &  &  &  &  \\
Low & 1 &  &  &  &  &  &  &  &  \\
Nil & 1 &  &  &  & 1 &  &  &  & 1 \\
No status change & 1 &  &  &  &  &  &  &  &  \\
Stable & 1 &  &  &  &  &  &  &  &  \\

\end{longtable}
\endgroup
\subsection{Enriching Constraints with Qualifier Subsumption Relations} \label{app:qualifier_subsumption}

How qualifiers are handled on the patient side is described in Appendix~\ref{app:patient_qualifier_subsumption}. On the trial side, qualifiers follow the \(@@\) notation, as discussed in Appendix~\ref{app:naming_schema}. In the SMT programming part of the semantic parsing pipeline (Appendix \ref{app:smt_programming}), each qualifier is instructed to entail the stem it modifies, so that the qualified form also implies the underlying base constraint. If this entailment is omitted by the parser, we inject it automatically through pattern matching and static checks on the trial-side representation, specifically by detecting uses of \(@@\).

\subsection{Relation Vocabularies}
\label{app:naming_schema}

In our representation, a canonical predicate has the form
\(\pi=(r,k,q)\), where \(r\) is the relation, \(k\) is the grounded
medical concept, and \(q\) is an optional set of qualifiers. This
appendix lists the allowed relation and qualifier vocabularies.

We keep these vocabularies small and fixed so LLMs can turn free-text
clinical statements into symbolic atoms more consistently. Instead of
inventing predicate names freely, the model builds atoms from a standard
relation vocabulary, grounded concepts, and normalized qualifiers. This
makes the representation easier to generate, check, and use later.

The qualifier part \(q\) includes only modifiers that can be normalized
into our fixed qualifier set, such as timeframe, procedure outcome, and
measurement unit. Some useful clinical details do not fit cleanly into
that set.

To preserve details not captured by the standard qualifier vocabulary, we allow SMT-style variable names to carry extra \texttt{@@}-attached fields. These fields are not part of the canonical core; they are used only to retain modifier information that would otherwise be lost.

Thus, a symbolic variable may contain both a canonical core and additional \texttt{@@} fields. The core still has the form \(\pi=(r,k,q)\), while the \texttt{@@} fields store leftover detail outside the canonical representation.

We also use different template families for findings, procedures, numeric
observables, products, substances or exposures, and demographic or
reproductive-status facts. Time is normalized through a small timeframe
grammar, and numeric observations are represented in a numeric-first form
so comparisons can be handled later during reasoning.

The templates below are therefore more than naming conventions: they form
a controlled symbolic vocabulary for building SMT programs from clinical
text.

\subsubsection{Vocabulary of Medical Relations $\Rm$}

We first define the set of canonical medical relations,
\(\Rm \subseteq R\). Each relation describes one standard way a patient
can be connected to a grounded medical concept. To make these relations
easy to read, we present them as stem templates. These templates show the
basic relation together with placeholders for the grounded entity and any
canonical qualifiers.

\small

\begin{tcolorbox}[colback=gray!3, colframe=black!15,
                  title=\textbf{A.1 \quad Canonical Medical Relations}]
\begin{tabularx}{\linewidth}{@{}>{\raggedright\arraybackslash}p{2.2cm} >{\ttfamily\arraybackslash}X@{}}
\textbf{Findings / Conditions} &
patient\_has\_diagnosis\_of\_\{entity\}\_\{timeframe\} \;
patient\_has\_finding\_of\_\{entity\}\_\{timeframe\} \;
patient\_has\_symptoms\_of\_\{entity\}\_\{timeframe\} \;
patient\_has\_clinical\_signs\_of\_\{entity\}\_\{timeframe\} \;
patient\_has\_suspicion\_of\_\{entity\}\_\{timeframe\} \\[0.9em]

\textbf{Procedures} &
patient\_has\_undergone\_\{entity\}\_\{timeframe\} \;
patient\_has\_undergone\_\{entity\}\_\{timeframe\}\_outcome\_is\_positive \;
patient\_has\_undergone\_\{entity\}\_\{timeframe\}\_outcome\_is\_negative \;
patient\_has\_undergone\_\{entity\}\_\{timeframe\}\_outcome\_is\_normal \;
patient\_has\_undergone\_\{entity\}\_\{timeframe\}\_outcome\_is\_abnormal \;
patient\_is\_undergoing\_\{entity\}\_\{timeframe\} \;
patient\_will\_undergo\_\{entity\}\_\{timeframe\} \;
patient\_can\_undergo\_\{entity\}\_\{timeframe\} \\[0.9em]

\textbf{Numeric Observables} &
patient\_\{entity\}\_value\_recorded\_\{timeframe\}\_withunit\_\{unit\} \\[0.9em]

\textbf{Products} &
patient\_is\_taking\_\{entity\}\_\{timeframe\} \;
patient\_has\_taken\_\{entity\}\_\{timeframe\} \;
patient\_has\_hypersensitivity\_to\_\{entity\}\_\{timeframe\} \;
patient\_has\_intolerance\_to\_\{entity\}\_\{timeframe\} \;
patient\_has\_allergy\_to\_\{entity\}\_\{timeframe\} \;
patient\_has\_nonimmune\_hypersensitivity\_to\_\{entity\}\_\{timeframe\} \\[0.9em]

\textbf{Substances / Exposures} &
patient\_is\_exposed\_to\_\{entity\}\_\{timeframe\} \;
patient\_has\_been\_exposed\_to\_\{entity\}\_\{timeframe\} \;
patient\_has\_hypersensitivity\_to\_\{entity\}\_\{timeframe\} \;
patient\_has\_intolerance\_to\_\{entity\}\_\{timeframe\} \;
patient\_has\_allergy\_to\_\{entity\}\_\{timeframe\} \;
patient\_has\_nonimmune\_hypersensitivity\_to\_\{entity\}\_\{timeframe\} \\[0.9em]

\textbf{Demographics \& Reproductive Status} &
patient\_age\_value\_recorded\_\{timeframe\}\_in\_years \;
patient\_age\_value\_recorded\_\{timeframe\}\_in\_months \;
patient\_age\_value\_recorded\_\{timeframe\}\_in\_days \;
patient\_sex\_is\_\{male|female|other\}\_\{timeframe\} \;
patient\_has\_childbearing\_potential\_\{timeframe\} \;
patient\_is\_breastfeeding\_\{timeframe\} \;
patient\_is\_pregnant\_\{timeframe\} \;
patient\_is\_lactating\_\{timeframe\} \\[0.6em]
\end{tabularx}
\end{tcolorbox}

\begin{tcolorbox}[colback=gray!3, colframe=black!15,
                  title=\textbf{A.2 \quad Canonical Qualifier Grammar}]
\ttfamily
timeframe ::= now
 \; | \; inthehistory
 \; | \; inthepast(n,unit)
 \; | \; inthefuture(n,unit) \\[0.25em]
outcome ::= positive \; | \; negative \; | \; normal \; | \; abnormal \\[0.25em]
unit ::= minutes \; | \; hours \; | \; days \; | \; weeks \; | \; months \; | \; years \; | \; \{measurement\_unit\}
\end{tcolorbox}

\normalsize

The qualifier grammar above is only an example, not a complete list. The
main idea is that qualifier values are mapped into a small fixed set
before symbolic execution. This helps reduce variation in wording and
makes both LLM generation and downstream SMT reasoning more reliable.

In use, the templates in Box A.1 correspond to typed atoms
\(\pi=(r,k,q)\): \(r\) gives the base medical relation, \(k\) gives the
grounded concept, and \(q\) gives modifiers such as timeframe, outcome,
and unit.

\subsubsection{Vocabulary of Trial Intents $\Rt$}

In addition to medical relations \(\Rm\), our representation uses a small
controlled vocabulary of \emph{trial intents}, denoted \(\Rt\), to capture
how a trial is meant to act upon a target clinical state, procedure, or
substance. Intuitively, while \(\Rm\) describes \emph{what} fact is under
discussion, \(\Rt\) describes \emph{why} that fact matters in the trial:
for example, because the intervention aims to treat it, prevent it,
monitor it, improve a procedure, or reduce substance-related harm.

As with \(\Rm\), we keep this vocabulary deliberately small and regular so
that LLMs can reliably annotate trial-side predicates with coarse but
useful target semantics. These labels are not intended as a complete
ontology of all real-world trial purposes. Rather, they form a compact
canonical vocabulary sufficient for the retrieval objectives studied in
this work, and they can be extended in future work.

Because different kinds of targets support different clinically meaningful
actions, \(\Rt\) is typed by target category. We distinguish three cases:
(i) findings and conditions, (ii) procedures, and
(iii) products or substances. For each case, an LLM is asked to classify a
canonicalized trial-side variable into one or more intent labels based on
the trial description. These labels are then attached to the corresponding
symbolic atoms as a trial-side semantic annotation layer.

\paragraph{Intent labels for findings and conditions.}
For variables denoting diagnoses, symptoms, findings, signs, or other
clinical states, we use:
\begin{itemize}[leftmargin=*]
    \item \textbf{Treats}: the intervention is intended to improve,
    control, relieve, or otherwise clinically address the target
    condition.
    \item \textbf{Prevents}: the intervention is intended to reduce the
    probability of future onset of the target condition.
    \item \textbf{Other}: the target is clinically relevant, but the
    trial's role is neither primarily treatment nor prevention; examples
    include diagnosis, monitoring, stratification, prediction, or
    characterization.
    \item \textbf{Not Clinically Relevant}: the variable is incidental,
    overly broad, or unrelated background context rather than a meaningful
    clinical focus of the trial.
\end{itemize}

\paragraph{Intent labels for procedures.}
For variables denoting procedures already undergone, ongoing, planned, or
feasible, we use:
\begin{itemize}[leftmargin=*]
    \item \textbf{Improves Effectiveness}: the trial aims to improve the
    success, precision, completeness, durability, or clinical benefit of
    the procedure.
    \item \textbf{Reduces Procedure-Related Harms}: the trial aims to
    reduce adverse effects, complications, or harmful recovery
    consequences associated with the procedure.
    \item \textbf{Other}: the trial modifies how the procedure is
    delivered or experienced in another clinically relevant way, such as
    improving tolerability, shortening recovery, optimizing timing, or
    improving workflow and safety monitoring.
    \item \textbf{Not Clinically Relevant}: the procedure is incidental or
    not a meaningful target of the trial.
\end{itemize}

\paragraph{Intent labels for products and substances.}
For variables denoting drugs, biologics, devices treated as products, or
substances/exposures, we use:
\begin{itemize}[leftmargin=*]
    \item \textbf{Reduces Exposure/Use}: the trial aims to reduce use of
    or exposure to the substance.
    \item \textbf{Mitigates Harms}: given use or exposure, the trial aims
    to reduce its negative effects.
    \item \textbf{Enhances Benefits}: given use or exposure, the trial
    aims to improve its desired effects or benefit--risk profile.
    \item \textbf{Other}: the trial changes the relationship between the
    patient and the substance in another clinically relevant way, such as
    monitoring, detection, dose personalization, relapse prevention, or
    response prediction.
    \item \textbf{Not Clinically Relevant}: the substance is incidental or
    not a meaningful target of the trial.
\end{itemize}

Operationally, then, the surface templates in this appendix should be read
as lexicalizations of typed atoms \(\pi=(r,k,q)\): \(r\) contributes the
base medical relation, \(k\) contributes the grounded concept, and
\(q\) contributes canonical modifiers such as timeframe, outcome status,
and unit. When additional modifier information cannot be canonicalized, we
retain it through \texttt{@@}-attached fields in the concrete symbolic
syntax so that the program remains semantically faithful to the source
text.

\subsubsection{Vocabulary of Patient-Fact Relations $\Rp$}

While \(\Rt\) captures the \emph{trial-side} role of a target, we also use
a small controlled vocabulary of \emph{patient-fact relations}, denoted
\(\Rp\), to capture the \emph{patient-side} role of a fact in
clinical-trial search. Intuitively, these labels describe which patient
facts are central enough to define what the patient is actually looking
for trials about.

This vocabulary supports progressively broader retrieval objectives. The
key idea is that not every fact in the patient record should contribute
equally: some facts correspond to the patient's main problem, some are
closely related to it, some are important enough to justify broader trial
search, and some correspond to diseases the patient is trying to
\emph{avoid} rather than diseases the patient currently has.

As with \(\Rm\) and \(\Rt\), we keep this vocabulary small and regular so
that LLMs can assign patient-side roles in a consistent way. The labels
are not intended to exhaust the full structure of patient intent. Rather,
they provide a compact canonical vocabulary sufficient to define the
retrieval objectives used in this paper.

\paragraph{Chief Complaint.}
A fact is labeled \textbf{Chief Complaint} if it corresponds to the
patient's main presenting problem or primary condition of concern at the
time of trial search. This is the condition, symptom complex, or diagnosis
that most directly answers the question: \emph{what trials is this patient
looking for right now?} In practice, chief complaints are often not stated
literally in the vignette and may need to be inferred from the clinical
presentation. They are usually disease-level conditions rather than
isolated low-level symptoms.

\paragraph{Chief-Complaint Related.}
A fact is labeled \textbf{Chief-Complaint Related} if it is not itself the
chief complaint, but is still closely and clinically meaningfully related
to it. This includes facts that are near the chief complaint in the
diagnostic, etiologic, syndromic, or therapeutic space, such that a trial
targeting the related fact may still plausibly help address the patient's
main concern. This category is stricter than general semantic relatedness:
the connection must be close enough that a patient could realistically
search for trials using that fact in pursuit of help for the chief
complaint.

\paragraph{Any Important Complaint.}
A fact is labeled \textbf{Any Important Complaint} if it represents an
important clinical problem for the patient that may independently justify
a trial search, even if it is not the main presenting complaint and not
especially close to it. This is the broadest complaint-oriented category
used in our objectives. It includes additional meaningful diagnoses,
findings, major symptoms, or significant non-routine interventions that a
patient might deliberately search trials for. However, it excludes overly
general concepts, trivial routine facts, and vague buckets such as
\textit{disease}, \textit{pain}, or broad anatomical findings.

\paragraph{Prevention Target.}
A fact is labeled \textbf{Prevention Target} if it denotes a disease that
the patient is explicitly seeking to prevent, rather than a disease the
patient currently has. This category is handled separately from the
complaint-based categories above. Prevention targets must not be inferred
current diagnoses. They are included only when the note explicitly
indicates preventive intent, or when prevention of a specific downstream
disease is the near-universal clinical goal of the patient's stated
condition. Thus, this category represents what the patient is trying to
\emph{avoid}, not what the patient is presently experiencing.

\section{Text Preprocessing and Semantic Parsing}\label{app:parsing_umbrella}
    \subsection{Requirement Preprocessing: Extended Description (Trial-Side Semantic Parsing)}
\label{app:preprocessing}

\begin{figure*}[!htb]
  \centering
  \includegraphics[width=0.68\linewidth]{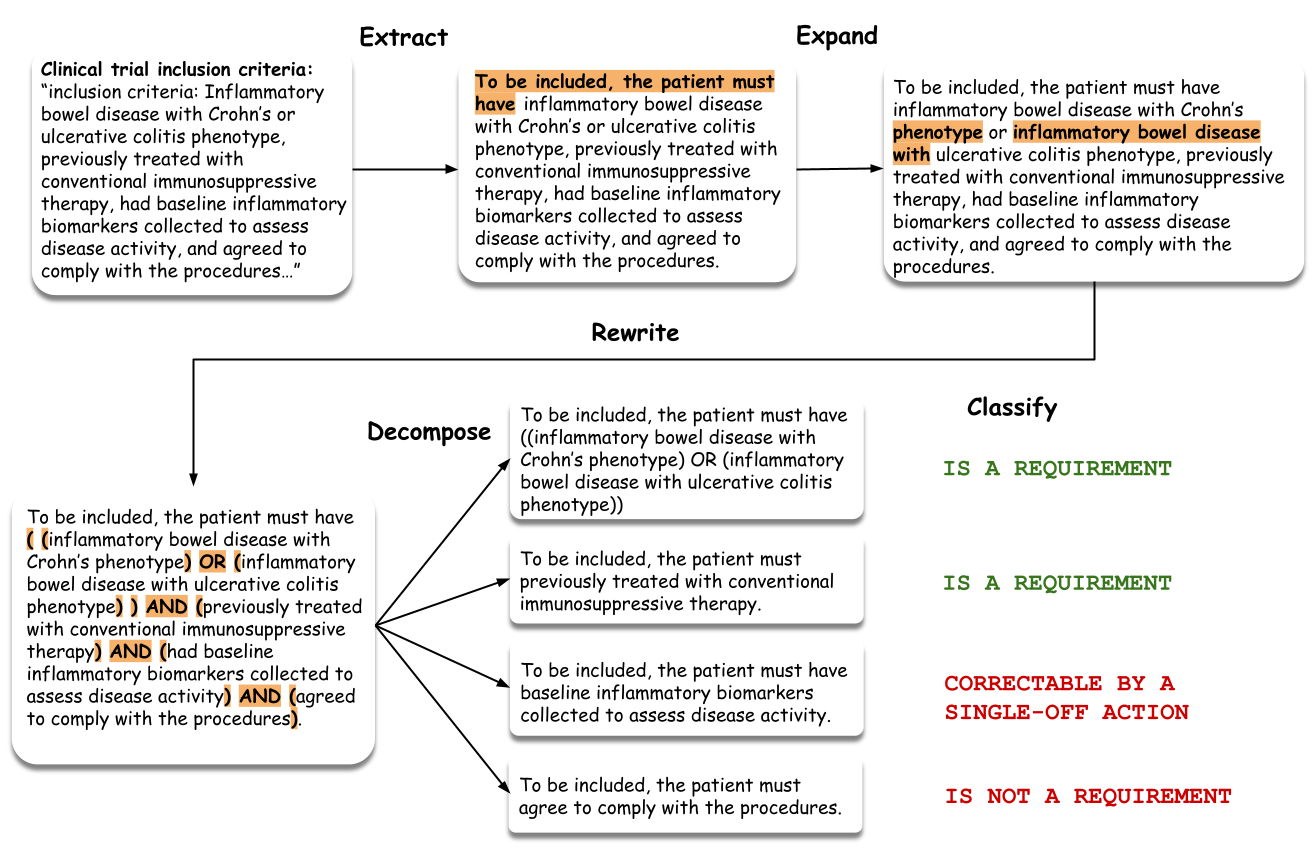}
  \caption{Requirement preprocessing workflow in \name{}. Starting from raw trial eligibility text, the system transforms free-text criteria into parse-ready requirement units through four stages: \textbf{Extract Requirements}, \textbf{Expand Entities}, \textbf{Rewrite Logic}, and \textbf{Decompose Requirements}.}
  \label{fig:preprocess_workflow}
\end{figure*}

This appendix expands the preprocessing stage summarized in Section~\ref{sec:preprocessing_requirements}. The role of preprocessing is to convert raw eligibility text into a representation that is substantially easier for downstream semantic parsing. Rather than asking a model to map long, heterogeneous, and often structurally ambiguous trial text directly into formal logic, \name{} first rewrites the text into \emph{parse-ready} requirement units whose cohort structure, polarity, entity boundaries, and logical scope are made explicit.

This stage can be viewed as a controlled representation transformation: it reduces ambiguity before formalization. The output is not yet canonicalized concepts or SMT assertions, but a cleaner intermediate language that exposes the semantic decisions the later stages must make.

\subsubsection{Input and Output}
\label{app:preproc_io}

The input is free-text trial eligibility text, including inclusion criteria, exclusion criteria, and any cohort-specific enrollment conditions. The output is a cohort-indexed intermediate representation consisting of self-contained requirement statements with preserved provenance and explicit structure.

Each resulting requirement unit is designed to support the next two stages of the pipeline. First, medically meaningful mentions become easier to ground to ontology concepts during entity canonicalization. Second, logical structure becomes easier to translate into formal assertions during incremental SMT programming. In this sense, preprocessing acts as the interface between raw language and symbolic semantic parsing.

\subsubsection{Why Preprocessing Is Needed}
\label{app:preproc_why}

Raw eligibility criteria are difficult for end-to-end semantic parsing for several reasons. First, a single trial often mixes multiple enrollment pathways in one document, so the parser must recover cohort structure before it can interpret a criterion correctly. Second, medically meaningful entities are often expressed in compressed or non-contiguous surface forms, which makes faithful grounding difficult. Third, logical structure is frequently implicit: conjunction, disjunction, precedence, and scope may be obvious to a clinician but not explicit in the text. Finally, individual sentences often bundle several distinct constraints together, making them hard to translate and validate as a single unit.

The preprocessing stage addresses these issues before formalization. Instead of relying on a single model call to simultaneously recover structure, resolve mentions, and generate logic, \name{} decomposes the problem into a sequence of simpler transformations, each of which removes one source of ambiguity. The fine-grained evaluation results for preprocessing step are provided in Section~\ref{app:preproc_evaluation}.

\subsubsection{Step 1: Extract Requirement}
\label{app:preproc_extract}

Clinical trials frequently encode multiple enrollment pathways, such as different treatment arms, disease subtypes, or severity-based cohorts. Since these pathways may share some conditions while differing on others, \name{} first organizes the trial into cohort-specific requirement lists. Prompts used for requirement extraction can be found in Appendix \ref{app:RequirementExtractor/trial_cohort_extractor}, \ref{app:RequirementExtractor/exclusion_requirements_extractor}, and \ref{app:RequirementExtractor/inclusion_requirements_extractor}

Within each cohort, the system extracts requirement-bearing statements from the eligibility text and rewrites them into a standardized form with explicit polarity. Inclusion statements are rewritten into a form such as ``To be included, the patient must \dots'', while exclusion statements are rewritten so that their exclusion meaning is explicit and does not depend on document layout or section headers.

This step corresponds to the \textbf{Extract Requirements} stage in Figure~\ref{fig:preprocess_workflow}. Its purpose is to isolate the requirement content and make the enrollment structure explicit. For downstream semantic parsing, this is important because the same sentence can mean different things depending on which cohort and which polarity context it belongs to.

\subsubsection{Step 2: Expand Entities}
\label{app:preproc_expand}

Eligibility text often expresses clinically meaningful content in forms that are easy for humans to read but difficult for models to ground reliably. For example, coordinated phrases may share a head noun, abbreviations may omit clinically relevant detail, and multiple distinct entities may be packed into a single local span.

To make entity grounding more reliable, \name{} rewrites each requirement so that medically meaningful mentions appear as explicit contiguous phrases. This may involve duplicating shared heads, unpacking coordinated mentions, or expanding abbreviated forms when supported by the local text. The goal is not to add information, but to expose the entity structure already implicit in the criterion. Prompt for entity expansion can be found in Appendix \ref{app:RequirementExtractor/entity_expander}.

This step corresponds to the \textbf{Expand Entities} stage in Figure~\ref{fig:preprocess_workflow}. For the downstream canonicalization module, the benefit is straightforward: contiguous and semantically explicit mentions are easier to align with ontology concepts than compressed surface forms.

\subsubsection{Step 3: Rewrite Logic}
\label{app:preproc_rewrite}

A major difficulty in semantic parsing of eligibility criteria is that the logical structure is often only partially explicit. A requirement may mix conjunction, disjunction, and modifier scope in a way that is under-specified on the surface. This is especially common in long clinical sentences that combine disease phenotype, prior treatment conditions, biomarker requirements, and procedural obligations.

To address this, \name{} rewrites each requirement so that the internal logical structure is more explicit. Conjunctions and disjunctions are surfaced, grouping is clarified, and ambiguous coordination is rewritten into a form whose scope can be interpreted deterministically. Importantly, this step is meaning-preserving: it does not simplify the criterion, but rather makes its latent logical structure visible.

This step corresponds to the \textbf{Rewrite Logic} stage in Figure~\ref{fig:preprocess_workflow}. It makes implicit compositional structure explicit by introducing intermediate symbolic scaffolding, thereby reducing structural ambiguity and simplifying downstream parsing. The prompts for logic rewrite could be found in Appendix \ref{app:RequirementExtractor/inclusion_requirement_logic_rewriter} and Appendix~\ref{app:RequirementExtractor/exclusion_requirement_logic_rewriter}.

\subsubsection{Step 4: Decompose Requirements}
\label{app:preproc_decompose}

Even after logic rewriting, a single requirement sentence may still contain several conceptually distinct constraints. Parsing such sentences monolithically is brittle: errors in one part of the sentence can affect the entire formalization, and validation becomes difficult because the output is not modular.

\name{} therefore decomposes rewritten requirements into smaller self-contained requirement units. For example, a sentence that jointly constrains disease subtype, treatment history, and biomarker evidence may be split into separate requirement statements, each of which can later be canonicalized and translated independently. This decomposition respects the logical structure introduced in the previous stage and avoids splits that would change the scope of negation, disjunction, or qualification.

This step corresponds to the \textbf{Decompose Requirements} stage in Figure~\ref{fig:preprocess_workflow}. The result is a set of localized semantic parsing problems that are easier to ground, easier to translate, and easier to validate. The prompts for decomposition could be found in Appendix \ref{app:RequirementExtractor/inclusion_requirement_decomposer} and Appendix \ref{app:RequirementExtractor/exclusion_requirement_decomposer}.

\subsubsection{Step 5: Classify}
\label{app:preproc_classification}

After decomposition, each requirement unit is further assigned a functional role that determines how it should be treated during early-stage patient--trial matching. While all components are valid eligibility constraints, they differ in how strictly they must be enforced during prescreening, where patient information is often incomplete.

\name{} therefore classifies each atomic-level requirement into a small set of operational categories based on its role in determining relevance under partial information. In particular, the classification distinguishes between (1) requirements that must be explicitly supported by patient notes during prescreening, (2) requirements that do not meaningfully constrain relevance or can be trivially satisfied through simple actions, and (3) requirements that may be unknown at prescreening time but should not exclude a trial unless explicitly contradicted. This distinction reflects a key asymmetry in retrieval: missing information should not prematurely eliminate potentially relevant trials unless the requirement is deemed critical and expected to be known.

The classification is performed independently for each decomposed component, using both the local requirement text and the broader trial context to assess its importance. For example, core clinical attributes that define the target population of the trial are typically treated as strict prescreening constraints, whereas laboratory thresholds, screening measurements, or protocol logistics are often deferred to later stages and treated as permissive. Additionally, requirements that can be satisfied through administrative steps or short-term behavioral adjustments are explicitly separated from true eligibility constraints to avoid unnecessary filtering.

This step corresponds to the \textbf{Classify} stage in Figure~\ref{fig:preprocess_workflow}. The output is a set of requirement components annotated with their prescreening roles, enabling the retrieval system to balance precision and recall by enforcing only the most informative constraints early while deferring others to downstream eligibility checking. The prompts for classification could be found in Appendix \ref{app:RequirementExtractor/inclusion_requirement_classifier} and Appendix~\ref{app:RequirementExtractor/exclusion_requirement_classifier}.

\subsubsection{Faithfulness Verification and Conservative Fallback}

Because preprocessing rewrites text, \name{} applies lightweight verification checks between stages to ensure that the transformed requirements remain faithful to the source. These checks verify that polarity is preserved, clinically meaningful content is not dropped, logical scope is not altered, and decomposition does not introduce unintended semantic changes.

When a check fails, the system retries the transformation with targeted feedback. If the issue cannot be resolved reliably, \name{} falls back to a less processed representation rather than risk semantic drift. For example, a requirement may remain as a larger unit if aggressive decomposition cannot be validated. This conservative fallback policy is important: the goal of preprocessing is to improve parseability, not to force unsupported structure onto the text.

Verification prompts for individual preprocessing stages are documented in the appendix. Specifically, verification procedures for Stage~1 (requirement extraction) are provided in Appendix~\ref{app:RequirementExtractor/inclusion_extraction_verifier} and Appendix~\ref{app:RequirementExtractor/exclusion_extraction_verifier}, corresponding to inclusion and exclusion criteria, respectively. Verification for Stage~2 (entity expansion) is described in Appendix~\ref{app:RequirementExtractor/entity_expansion_verifier}. Verification for Stage~3 (rewrite logic) is described in Appendix~\ref{app:RequirementExtractor/inclusion_logic_rewrite_verifier} and Appendix~\ref{app:RequirementExtractor/exclusion_logic_rewrite_verifier}, corresponding to inclusion and exclusion criteria, respectively. Verification for Stage~4 (decompose) is described in Appendix~\ref{app:RequirementExtractor/inclusion_decomposition_checker} and Appendix~\ref{app:RequirementExtractor/exclusion_decomposition_checker}, corresponding to inclusion and exclusion criteria, respectively.

\subsubsection{Evaluation of the Preprocessing Pipeline} \label{app:preproc_evaluation}
\begin{figure}[!htb]
  \centering
  \includegraphics[width=0.5\linewidth]{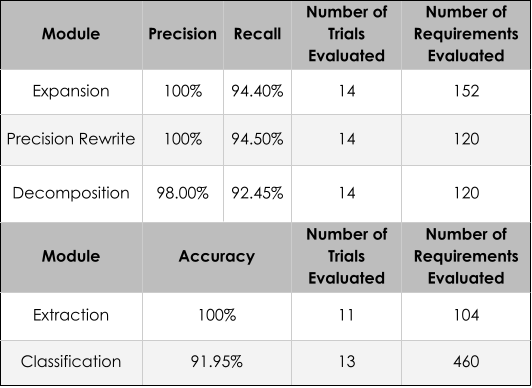}
  \caption{Fine-grained evaluation results for preprocessing modules.}
  \label{fig:preproc_eval}
\end{figure}
We conduct a fine-grained, module-level evaluation of the preprocessing stage, which standardizes free-text eligibility criteria prior to downstream semantic parsing. Three annotators participated in the evaluation. For each module, we compare its input and output and label each requirement against gold annotations to determine whether the module performed the intended operation correctly. Below, we describe the evaluation protocol and results for each module.

\paragraph{Extract Requirement.}
The Extraction module converts bullet points or line-based criteria into a structured list of requirements and prepends relevant preamble context to each requirement. For each trial, we evaluate (i) whether all requirements are successfully extracted and (ii) whether each extracted requirement is semantically consistent with the original text. Extraction performance is measured using accuracy, defined as
\[\frac{\#extracted\ requirements\ that\ are\ semantically\ correct}{\# requirements\ should\ be\ extracted}\]
Across 11 trials (104 requirements), the Extraction module achieves 100\% accuracy.

\paragraph{Expand Entities.}
The Expansion module targets requirements containing medical shorthand or ambiguous terms that should be explicitly expanded for clarity. To evaluate this module, we compare the input and output text and label each requirement as one of four categories: \textbf{true positive} (expansion is required and correctly performed), \textbf{false negative} (expansion is required but missing or incorrect), \textbf{false positive} (unnecessary or extraneous expansion), or \textbf{true negative} (requirement should not be expanded, and no expansion performed). We report precision and recall, defined respectively as
\[
\text{Precision} = \frac{\#\,\text{true positives}}
{\#\,\text{true positives} + \#\,\text{false positives}}, \qquad
\]
\[
\text{Recall} = \frac{\#\,\text{true positives}}
{\#\,\text{true positives} + \#\,\text{false negatives}}.
\]

On 14 trials (152 requirements), Expansion achieves \textbf{100\% precision} and \textbf{94.40\% recall}.

\paragraph{Rewrite Logic.}
Rewrite Logic module rewrites each requirement to make implicit logical structure explicit. Similar to expansion evaluation, we evaluate rewrite logic by comparing the input and output text and label them with: \textbf{true positive} (rewrite required and is performed correctly), \textbf{false negative} (a rewrite was required but missing or incorrect), \textbf{false positive} (a rewrite was not required, yet the system rewrote the text), or \textbf{true negative} (no rewrite was required, and none was performed). We evaluate Rewrite Logic using the same precision and recall definitions as Expansion. On 14 trials (120 requirements), it attains \textbf{100\% precision} and \textbf{94.50\% recall}, suggesting the module captures essentially all rewrite-needed cases, with a small number of over-edits or incorrect rewrites.

\paragraph{Decompose Requirements.}
Decomposition splits each requirement into atomic components that preserve the original meaning. Similar to Expansion and Rewrite Logic evaluation, we compare the input and output text and label each requirement with: \textbf{true positive} (the requirement should be decomposed, and decomposition is performed correctly), \textbf{false negative} (the decomposition did not split something that should have been decomposed), \textbf{false positive} (decomposer splits a requirement that shouldn't be decomposed), or \textbf{true negative} (no decomposition was required, and none was performed). Decomposition is evaluated using the same precision and recall metrics defined for Expansion. On 14 trials (120 requirements), it achieves \textbf{98.00\% precision} and \textbf{92.45\% recall}, showing strong correctness when decomposition is applied, with some missed opportunities to decompose.

\paragraph{Classification.}
The Classification module assigns each atomic requirement to the appropriate preprocessing category. Exclusion requirements are classified into one of two categories: (1)\textbf{not requirement or always satisfiable with one-off action} and (2) \textbf{other requirements}. Inclusion requirements are assigned to one of three categories: (1) \textbf{prescreen notes must completely suffice}, (2) \textbf{not requirement or always satisfiable with action}, and (3) \textbf{other requirements}. For evaluation, we review each requirement in context and count the number of incorrect assignments. Accuracy is computed as \[\frac{\#\ requirements - \#\ incorrectly\ labeled\ requirements}{\#\ requirements}\] On 13 trials (460 atomic requirements), Classification reaches \textbf{91.95\% accuracy}.  We note that this accuracy was evaluated using an earlier version of the classification scheme; subsequent refinements to category definitions may introduce additional distinctions not reflected in this reported score.

\paragraph{Error Interaction Across Preprocessing Modules.}
Although we report per-module precision and recall for diagnostic purposes, errors across preprocessing modules do not compound in a strictly sequential manner. In practice, later modules can compensate for missed operations in earlier stages. For example, if a module fails to apply a required expansion (i.e., incurs a false negative), subsequent Rewrite Logic module may still correct or recover the issue when operating on the same requirement.

This behavior arises from two design choices. First, preprocessing modules are not mutually exclusive in functionality: multiple modules are capable of resolving overlapping forms of ambiguity or incompleteness. Second, each module operates with access to the full contextual information of the requirement, including outputs and preserved context from prior stages, rather than relying solely on the transformed text alone. As a result, missed edits are often recoverable downstream.

Consequently, per-module recall should not be interpreted as a direct indicator of end-to-end preprocessing failure. Instead, recall reflects how often a specific operation is applied at a particular stage, rather than whether the requirement is ultimately normalized correctly. Precision, which captures incorrect or unnecessary edits, is therefore a more critical metric for maintaining semantic fidelity in the overall pipeline.

\subsubsection{Relation to the Downstream Parser}
\label{app:preproc_relation_downstream}

The output of preprocessing is the input to the later semantic parsing stages described in Section~\ref{sec:rq2_faithful_parsing}. The decomposed requirement units produced here are subsequently canonicalized against the medical ontology and then translated into solver-executable assertions through incremental SMT programming.

Preprocessing therefore serves a role analogous to representation shaping in other AI pipelines: it transforms a difficult raw input into an intermediate form whose relevant structure is easier for later components to model correctly. In \name{}, this intermediate form exposes four kinds of information that are particularly important for faithful formalization: enrollment pathway, polarity, entity boundaries, and logical scope.

\paragraph{Summary.}
Requirement preprocessing converts raw trial eligibility text into cohort-aware, polarity-explicit, entity-expanded, logically clarified, and modularized requirement units. These units are not yet formal logic, but they are designed to be \emph{semantic-parsing ready}: they reduce ambiguity in the input and make downstream ontology grounding and symbolic translation substantially more reliable.

    \subsection{Entity Canonicalization Details (Trial-Side Semantic Parsing)}
\label{app:canonicalization}

\begin{figure*}[!htbp]
  \centering
  \includegraphics[width=0.8\linewidth]{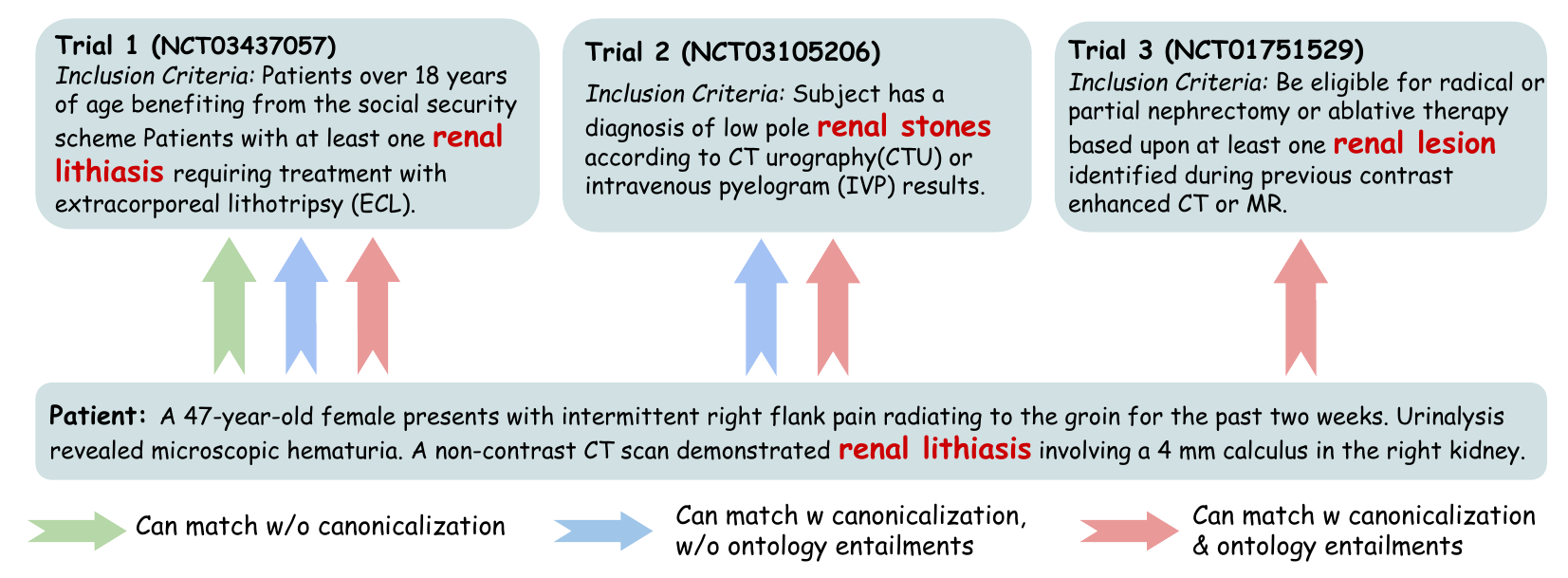}
  \caption{Why canonicalization is necessary for retrieval in \name{}. Pure string-level matching cannot reliably connect synonymous or ontologically related medical concepts. In this example, a patient mention of \emph{renal stones} would fail to retrieve trials written using \emph{renal lithiasis}, and would also miss trials targeting the broader concept \emph{renal lesion} unless ontology-based entailment is applied. Canonicalization collapses lexical variation into a shared concept representation, while ontology reasoning recovers relevant abstraction and subsumption structure. Together, these steps allow all relevant trials to be retrieved.}
  \label{fig:necessitating-canonicalization}
\end{figure*}

\begin{figure*}[!htbp]
  \centering
  \includegraphics[width=\linewidth]{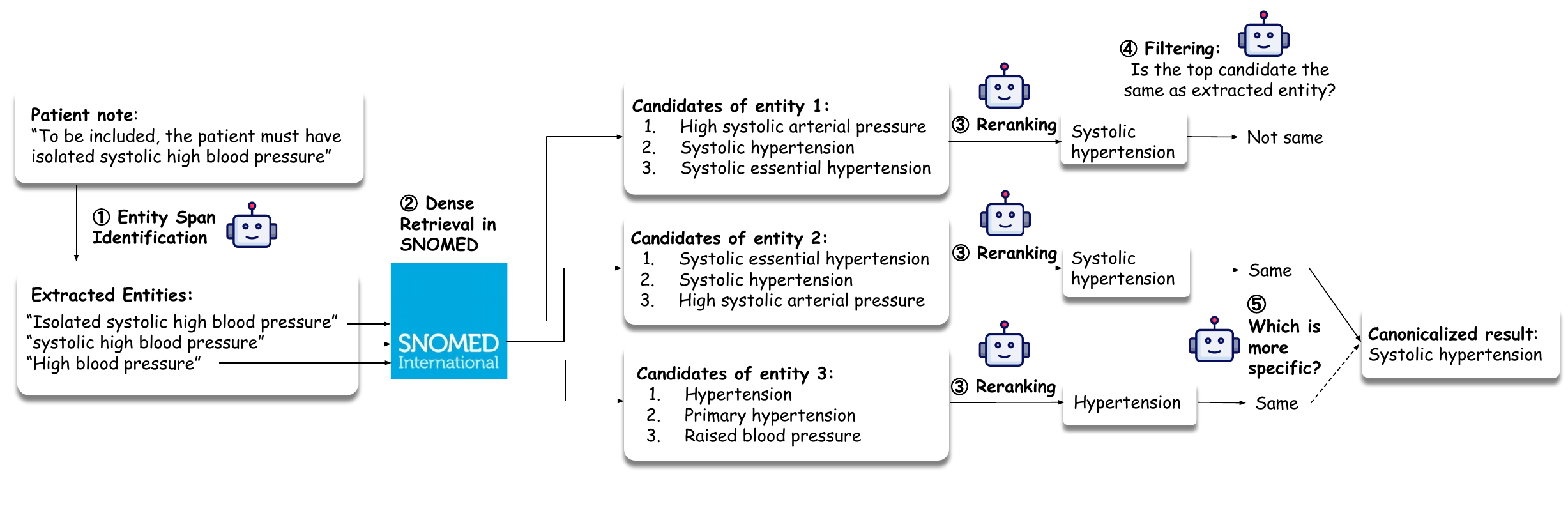}
  \caption{Entity canonicalization workflow in \name{}. Starting from a preprocessed requirement, the system first performs high-recall extraction of medically meaningful spans, including overlapping spans when needed to avoid early information loss (left). For each span, top-$k$ candidate concepts are retrieved from SNOMED CT (center). Candidates are then reranked, audited for exact contextual meaning match, and filtered by policy constraints. When multiple overlapping spans survive, the system retains the most specific faithful concept. The final output is a schema-aware canonical symbol that can be shared across patient parsing, trial parsing, retrieval, and downstream formal reasoning (right).}
  \label{fig:concept_canonicalization}
\end{figure*}

This appendix expands the canonicalization stage summarized in Section~\ref{sec:concept_canonicalization}. The purpose of this stage is to map heterogeneous clinical language into a shared symbolic vocabulary that supports both ontology-aware retrieval and formal reasoning. Rather than treating entity linking as a standalone normalization problem, \name{} uses canonicalization as the interface between free-text clinical language and the structured concept space used throughout the system.

Concretely, entity canonicalization resolves lexical variation, abbreviation, and span ambiguity while maintaining the contextual distinctions needed for correct downstream reasoning.

\subsubsection{Why Canonicalization Is Necessary}
\label{app:canon_why}

Clinical trials and patient records often refer to the same underlying medical concept using different surface forms, abbreviations, levels of specificity, or paraphrases. A retrieval or reasoning system that operates only on raw text spans or surface strings therefore fails in two ways. First, it misses synonymy: semantically equivalent mentions may not match lexically. Second, it misses ontology-mediated generalization: a patient fact expressed at a more specific level should still support a trial target expressed at a more general level.

Figure~\ref{fig:necessitating-canonicalization} illustrates this point. Canonicalization maps diverse surface forms to a shared ontology-grounded concept, while ontology-based entailment recovers medically valid abstraction structure such as subtype--supertype relations. These are both necessary: canonicalization alone does not recover subsumption, and ontology reasoning alone cannot help if semantically equivalent mentions were never grounded to the same concept space.

\subsubsection{Objective and Interface}
\label{app:canon_objective}

The input to canonicalization is a preprocessed requirement unit from the earlier stage of the pipeline. The output is a set of accepted canonical entries, each representing a medically meaningful mention that has been grounded to a standardized concept and converted into a schema-aware symbolic form.

Formally, for each input sentence $t$, the canonicalizer emits a set of entries of the form
\[
\Bigl\{(\textit{span}, [b,e), \textit{conceptId}, \textit{type}, \textit{preferred\_term}, \textit{variable\_name})\Bigr\},
\]
where $[b,e)$ denotes the character span of the mention in the sentence. These outputs form the bridge between raw language and the canonical signature used by later modules.

The goal is not simply to assign ontology nodes. Instead, the goal is to produce canonical symbols that satisfy three constraints simultaneously: they must unify synonymous language across patient and trial text, preserve enough local meaning for correct symbolic use, and remain compatible with the downstream schema used for retrieval and SMT-style formalization. The fine-grained evaluation results for entity canonicalization could be found in Section~\ref{app:ner-eval}

\subsubsection{Stage 1: High-Recall Mention Extraction}
\label{app:canon_extraction}

The first stage identifies as many medically meaningful spans as possible from the preprocessed requirement. We intentionally bias this stage toward recall, since an omitted mention cannot be recovered later and may directly cause false negatives in retrieval or eligibility reasoning.

This high-recall extraction includes overlapping spans when they correspond to distinct plausible semantic units. For example, a longer disease phrase and a nested subtype mention may both be retained at this stage. Administrative or non-medical framing content, such as boilerplate consent language or document structure cues, is suppressed because it does not contribute to the canonical clinical representation.

This design deliberately favors over-generation, allowing later stages to make more informed precision-oriented decisions. This is preferable to aggressive early pruning, which risks permanently discarding clinically important content. Prompts can be found in Appendix \ref{app:EntityCanonicalizer/free_entity_extraction} and \ref{app:EntityCanonicalizer/primary_high_recall_mention_extractor}

\subsubsection{Stage 2: Candidate Retrieval from SNOMED CT}
\label{app:canon_retrieval}

For each extracted span, \name{} retrieves top-$k$ candidate concepts from SNOMED CT. The purpose of this stage is to project the local surface form into a structured concept space that supports shared reasoning across patient and trial text.

Each retrieved candidate includes a concept identifier, preferred term, fully specified name, and semantic type. Candidate retrieval is intentionally permissive, retaining multiple plausible interpretations for each span and deferring disambiguation to later stages where additional context enables more reliable resolution.

\subsubsection{Stage 3: Contextual Link Auditing}
\label{app:canon_link_audit}

Candidate retrieval alone is not sufficient, since nearest-neighbor search over ontology concepts often returns semantically related but not meaning-equivalent candidates. To reduce these false positives, \name{} applies a contextual auditing stage that verifies whether a candidate preserves the meaning of the extracted span at that specific location in the sentence.

The audit accepts a candidate only if it is an exact meaning match in context. This means that the candidate must neither lose clinically relevant semantics nor introduce unsupported semantics. For example, a span referring to a symptom complex should not be collapsed into a disease diagnosis unless the sentence actually licenses that interpretation. Similarly, a candidate that adds severity, temporality, or treatment implications not present in the text is rejected. Prompts can be found at Appendix \ref{app:EntityCanonicalizer/context_aware_concept_linker} and \ref{app:EntityCanonicalizer/semantic_equivalence_verifier}

This stage is important because downstream formal reasoning is highly sensitive to symbol choice. A plausible but slightly incorrect link is often worse than leaving the mention uncanonicalized, since it introduces a false structured commitment into the program.

\subsubsection{Stage 4: Overlap Resolution and Policy Filtering}
\label{app:canon_arbit_filter}

Because the system intentionally allows overlapping spans in the first stage, multiple valid candidate links may survive for partially overlapping text segments. \name{} therefore applies a final arbitration and policy-filtering step. Prompt can be found in Appendix \ref{app:EntityCanonicalizer/final_canonicalization_arbiter}

This stage enforces representation constraints needed by the downstream reasoning system. Mentions whose concept types are not supported by the canonical schema are discarded, as are spans corresponding only to administrative, demographic, or other non-semantic framing content. Among overlapping candidates that remain faithful to the text, the system keeps the most specific acceptable concept. The goal is to preserve the sharpest faithful representation while avoiding redundant or overly broad symbols. In effect, this step consolidates a dense set of local grounding hypotheses into a compact symbolic representation, reducing redundancy while preserving maximal semantic precision.

\subsubsection{Stage 5: Schema-Aware Canonical Symbol Emission}
\label{app:canon_variable_naming}

Once a concept is accepted, \name{} converts it into a schema-aware canonical symbol. This is a crucial step: downstream reasoning does not operate directly over ontology nodes alone, but over variables whose names encode both the grounded medical concept and its semantic role in the structured representation.

The emitted symbol therefore combines the canonical concept with schema information such as whether the concept is being treated as a diagnosis, finding, procedure, medication exposure, or measurable observable. In addition, normalized contextual information such as timeframe may be encoded in a controlled way when it is required by the downstream logic representation.

This design ensures that the same underlying concept is represented consistently across patient parsing, trial parsing, database storage, retrieval, and formal reasoning. It constructs a shared symbolic interface for downstream processing steps.

\subsubsection{Why the Pipeline Is Staged}
\label{app:canon_staged}

The canonicalization workflow is intentionally staged rather than performed in a single end-to-end linking decision. This separation is important because the subproblems have different error profiles and different optimization goals. Mention extraction is recall-sensitive, candidate retrieval is coverage-oriented, contextual auditing is precision-sensitive, and overlap arbitration enforces representation compatibility.

Combining all of these decisions into a single step would make failures harder to diagnose and harder to control. By isolating them, \name{} can maintain high recall early, apply meaning-sensitive precision later, and ensure that the final output satisfies the structural requirements of the downstream symbolic system.

\subsubsection{Evaluation of the Entity Recognition Pipeline} \label{app:ner-eval}

\begin{figure}[!htb]
  \centering
  \includegraphics[width=0.6\linewidth]{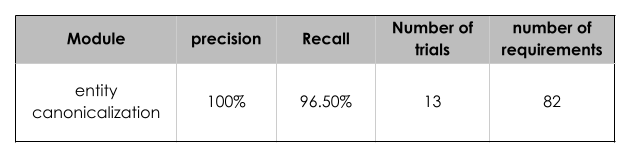}
  \caption{Fine-grained evaluation results for entity canonicalization.}
  \label{fig:canon_eval}
\end{figure}

We evaluate the named entity canonicalization pipeline. For each requirement, we compare the input text with the set of extracted canonical entities and record three quantities: 
(i)~the expected number of gold entities in the requirement \((|TP| + |FN|)\); 
(ii)~the number of correctly extracted entities \((|TP|)\); and 
(iii)~the total number of entities extracted by the system \((|TP| + |FP|)\).

From these counts, the number of false negatives for a requirement is computed as \((|TP| + |FN|) - |TP|\), and the number of false positives as \((|TP| + |FP|) - |TP|\). 
We aggregate \(TP\), \(FP\), and \(FN\) across all evaluated requirements and compute precision and recall using the same definitions as those used for the Expansion module in preprocessing.

Across 13 trials (82 requirements), entity canonicalization achieves \textbf{100\% precision} and \textbf{96.50\% recall}. This result indicates that extracted canonical entities are highly reliable, with no false positives observed in the evaluation set, while a small number of gold entities are missed. Importantly, missed entities do not necessarily result in downstream failure: during SMT program construction, free-form entity recognition can still recover uncanonicalized entities directly from the requirement text. As a result, recall at the canonicalization stage is not a strict bottleneck for end-to-end reasoning correctness, whereas high precision remains essential to avoid introducing spurious constraints.

\subsubsection{End-to-End Summary}
\label{app:canon_summary}

Overall, the entity canonicalization workflow proceeds as follows:
\begin{figure}[!htb]
  \centering
  \includegraphics[width=0.8\columnwidth]{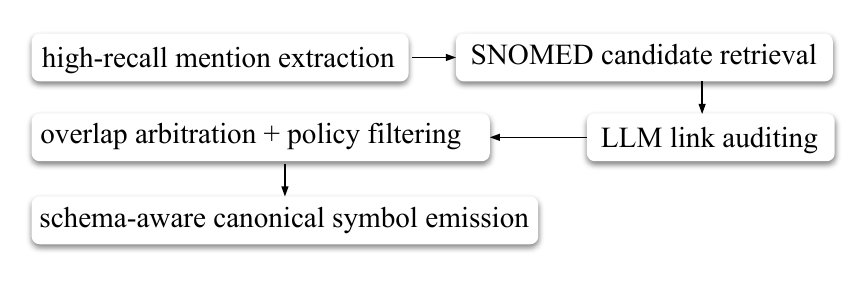}
\end{figure}

Starting from a preprocessed requirement, the system performs high-recall mention extraction, retrieves candidate ontology concepts, audits these candidates for exact contextual meaning match, resolves overlap and policy constraints, and finally emits schema-aware canonical symbols. The result is a unified concept representation shared across patient and trial text, enabling ontology-aware retrieval and formal reasoning in a common symbolic space.

\paragraph{Summary.}
Entity canonicalization transforms heterogeneous clinical language into a shared ontology-grounded symbolic vocabulary. This vocabulary is meaning-preserving, schema-aware, and reusable across patient parsing, trial parsing, retrieval, and formal reasoning. In \name{}, canonicalization is therefore not just an entity-linking component, but a core representation layer that makes symbolic clinical reasoning feasible.
    \subsection{SMT Programming Details: Incremental Orchestration (Trial-Side Semantic Parsing)}
\label{app:smt_programming}

\begin{sidewaysfigure*}[p]
  \centering
  \includegraphics[width=0.8\textheight]{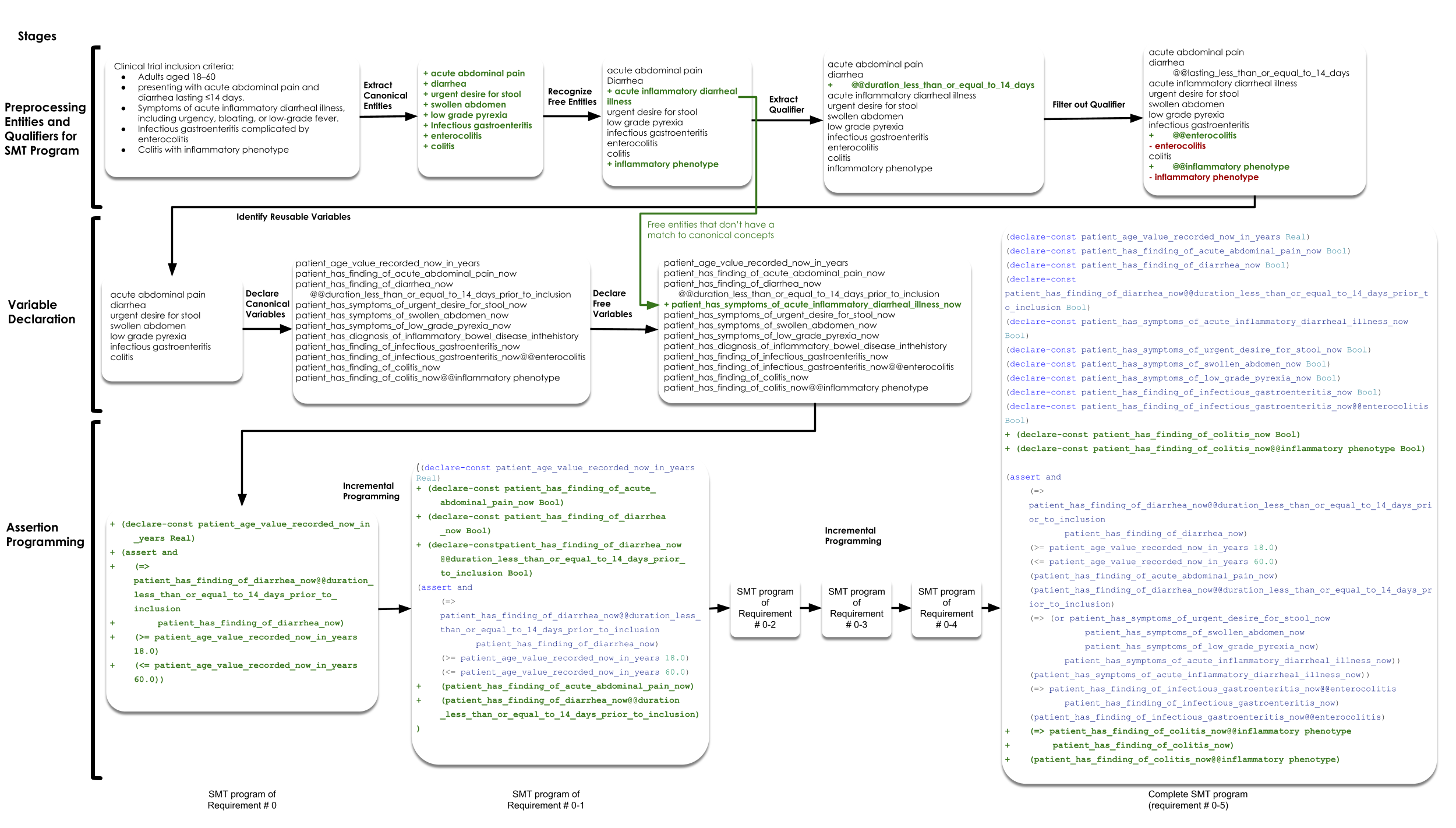}
  \caption{Incremental SMT programming workflow in \name{}. Starting from atomic, preprocessed requirements, the system constructs a solver-executable SMT-LIB program through a staged orchestration pipeline. For each requirement, \name{} first prepares the symbolic ingredients needed for formalization, including canonical entities, free non-canonical entities, and their qualifiers. It then identifies reusable variables from the current partial program and introduces new variables only when necessary, drawing from three sources: demographic variables, canonical medical variables grounded in SNOMED CT and the predefined schema, and free variables for clinically meaningful content that cannot be safely canonicalized. Finally, the system translates the requirement into SMT-LIB, validates the resulting fragment against the evolving program, and commits it only if the fragment passes syntactic, logical, and semantic checks.}
  \label{fig:smt_program_workflow}
\end{sidewaysfigure*}

This appendix expands the incremental SMT programming stage summarized in Section~\ref{sec:programming_smt_assertions}. In this stage, \name{} transforms each preprocessed, canonicalized requirement into a solver-executable SMT fragment and integrates it into a growing formal program. Rather than relying on a single end-to-end text-to-logic generation step, \name{} uses an incremental orchestration procedure in which requirements are translated one at a time, validated immediately, and committed only when they are judged reliable.

SMT programming can be viewed as a constrained program construction loop with structured feedback. Instead of generating a complete eligibility program in a single step, the system decomposes the task into localized decisions---including symbol reuse, variable construction, logical translation, and fragment validation. Each fragment is checked before integration, allowing errors to be detected, localized, and corrected early, and preventing propagation to later parts of the program.

\subsubsection{Input and Output}
\label{app:smt_io}

The input to this stage is a set of atomic, SMT-ready requirements produced by preprocessing, together with the canonicalized entities and qualifiers produced by the earlier grounding stage. Each requirement is already localized enough to support independent formalization, but still requires symbolic variable construction and logical translation.

The output is a single SMT-LIB program consisting of declarations, auxiliary definitions, and a sequence of \texttt{assert} statements that jointly encode the trial's eligibility logic. Each requirement contributes a tagged fragment to this program, and the system records a mapping from requirements to the corresponding SMT lines so that later debugging and attribution remain possible.

\subsubsection{Why Incremental Programming Is Necessary}
\label{app:smt_why}

Direct one-shot translation from free-text eligibility criteria to a full SMT program is brittle. In practice, such generation tends to fail in several recurring ways: variables are redeclared inconsistently, qualifiers are dropped or attached to the wrong concept, polarity is inverted, and the resulting program may be syntactically valid but semantically unsupported by the source text. These failures are particularly damaging in the clinical setting, where a small symbolic mistake can change the meaning of an eligibility constraint.

\name{} therefore adopts an incremental strategy. Each requirement is translated separately against the current partial program, checked immediately, and either committed or rejected. This design offers two advantages. First, it localizes failures to the requirement that caused them, rather than contaminating the full program. Second, it allows the current symbolic state of the program to guide future translation decisions, especially with respect to variable reuse and consistency.

\subsubsection{Overview of the Incremental Driver}
\label{app:smt_driver}

\begin{figure*}[!htbp]
  \centering
  \includegraphics[width=\linewidth]{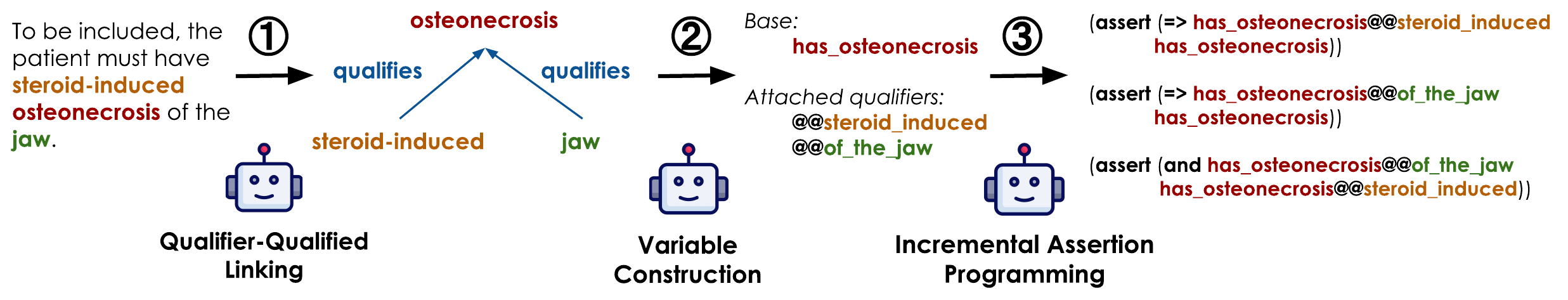}
  \caption{SMT programming workflow example. The phrase ``steroid-induced osteonecrosis of the jaw'' first undergoes qualifier--qualified linking, which identifies \textit{osteonecrosis} as the base concept and attaches \textit{steroid-induced} and \textit{of the jaw} as qualifiers. The system then constructs the corresponding symbolic variable and incrementally commits the resulting SMT assertions only after validation.}
  \label{fig:smt-programming}
\end{figure*}

Even after structured decomposition and entity canonicalization, semantic parsing of clinical requirements into SMT remains challenging. A one-shot LLM translation frequently introduces errors, including dropped qualifiers, polarity reversals, conflation of canonical and non-canonical variables, incorrect qualifier attachment, and logically unsupported program structure. For example, in Figure~\ref{fig:smt-programming}, the requirement ``steroid-induced osteonecrosis of the jaw'' should not be translated by treating \textit{steroid-induced}, \textit{osteonecrosis}, and \textit{jaw} as unrelated symbols. Instead, the system must first determine that \textit{steroid-induced} and \textit{of the jaw} both qualify the base concept \textit{osteonecrosis}, then construct the corresponding symbolic variable, and only then emit the associated SMT assertions. To reduce such errors, \name{} uses an incremental driver that builds the SMT program requirement by requirement, validating each local fragment before it is committed.

For each requirement index $i$, the system runs an attempt loop that follows a simple pattern: build a candidate fragment, check it against the current program prefix, and commit it only if it passes validation. Otherwise, the fragment is rolled back and the system retries with feedback from the failed attempt.

At a high level, the incremental driver separates five distinct subproblems:
\begin{enumerate}
    \item identifying which previously declared symbols should be reused,
    \item resolving which qualifiers modify which entities or predicates,
    \item constructing any new variables needed for the current requirement,
    \item translating the requirement into a local SMT fragment, and
    \item validating that fragment both logically and semantically before committing it.
\end{enumerate}
This decomposition makes the overall synthesis process more stable and more auditable than monolithic generation.

\subsubsection{Stage 0: Free-Entity Prepass}
\label{app:smt_free_prepass}

Before per-requirement translation, \name{} runs a recall-oriented prepass that extracts additional clinically meaningful content not captured by canonical grounding. This includes residual entities or modifiers that remain important for faithful formalization but cannot be safely mapped into the canonical ontology-backed schema.

The purpose of this stage is not to compete with canonicalization but to ensure coverage. When clinically meaningful content falls outside the canonical schema, the system still needs a way to represent it symbolically rather than silently discarding it. The free-entity prepass, therefore, seeds the non-canonical naming pathway used later in translation. Prompts used in this stage could be found in Appendix~\ref{app:SMTProgrammer/SMTPreprocessor/SMTProgrammerFreeEntityExtractor}

\subsubsection{Stage 1: Symbol Reuse Identification}
\label{app:smt_reuse}

Given the current partial SMT program, \name{} first identifies which existing symbols can and should be reused for the current requirement. This step is essential for maintaining global consistency across the evolving program. Many trial criteria refer repeatedly to the same disease, treatment, biomarker, or demographic property, and a correct formalization should represent these repeated mentions with the same underlying symbol whenever their meaning is shared.

This stage acts as a structured memory interface for program synthesis. Rather than regenerating symbols independently for each requirement, the system conditions synthesis on the symbolic state already established by earlier requirements (prompt in Appendix~\ref{app:SMTProgrammer/SMTIncrementalProgrammer/SMTIncrementalReusableVariableIdentifier}).

\subsubsection{Stage 2: Qualifier--Qualified Linking}
\label{app:smt_qualifier_linking}

After identifying reusable symbols, \name{} resolves qualifier--qualified relationships for the current requirement. This step determines which modifiers attach to which underlying entity or predicate before variable construction and logical translation. It is necessary because qualifiers do not stand alone: they refine the meaning of a specific disease, treatment, finding, temporal condition, or comparison statement, and incorrect attachment can change the meaning of the requirement even when the resulting SMT is syntactically valid.

Figure~\ref{fig:smt-programming} illustrates this step with the phrase ``steroid-induced osteonecrosis of the jaw.'' Here, \textit{osteonecrosis} is identified as the base concept, while \textit{steroid-induced} and \textit{of the jaw} are linked as attached qualifiers. The resulting symbolic construction therefore starts from the base meaning and augments it with qualifier structure, rather than introducing disconnected variables. Making this attachment step explicit helps prevent common failures such as dropped qualifiers, wrong modifier scope, and semantically unsupported variable construction. Prompts for stage 2 could be found in Appendix~\ref{app:SMTProgrammer/SMTPreprocessor/SMTProgrammerFreeEntityQualifierIdentifier}, Appendix~\ref{app:SMTProgrammer/SMTPreprocessor/SMTProgrammerFreeEntityQualifierIdentifierVerifier}, and Appendix~\ref{app:SMTProgrammer/SMTIncrementalProgrammer/SMTIncrementalNewVariableFilter}.

\subsubsection{Stage 3: Variable Construction from Three Sources}
\label{app:smt_variable_construction}

When existing symbols are insufficient, \name{} introduces new variables. These variables come from three sources, corresponding to three different kinds of semantic commitments in the input:

\paragraph{Demographic variables.}
Controlled template variables are used for schema-level attributes such as age, sex, pregnancy status, and other population descriptors that are commonly expressed in standardized forms (prompt in Appendix~\ref{app:SMTProgrammer/SMTIncrementalProgrammer/SMTIncrementalDemographicsVariableNamer}).

\paragraph{Canonical medical variables.}
When a medically meaningful mention has been safely grounded to SNOMED CT, linked to its relevant qualifiers, and mapped into the predefined schema, the system emits a canonical variable whose name reflects both the normalized concept and its schema role. These variables provide the ontology-aware symbolic backbone of the program (prompt in Appendix~\ref{app:SMTProgrammer/SMTIncrementalProgrammer/SMTIncrementalCanonicalVariableNamer}).

\paragraph{Free variables.}
Not all clinically meaningful content can be safely canonicalized. When grounding is unsupported or the content falls outside the canonical schema, the system introduces a conservative free variable so that the meaning can still be represented in the formal program. This prevents loss of important information merely because normalization was not possible (prompt in Appendix~\ref{app:SMTProgrammer/SMTIncrementalProgrammer/SMTIncrementalFreeVariableNamer}).

This three-way construction is important because it draws a principled boundary between content that can be represented canonically and content that must remain open-world. The resulting program is therefore both precise where grounding is reliable and conservative where it is not.

\subsubsection{Stage 4: Local Translation to SMT-LIB}
\label{app:smt_translation}

Using the requirement text, its decomposition structure, the resolved qualifier links, and the available variable set, the translator generates a local SMT fragment for the current requirement. This fragment may contain new declarations, auxiliary definitions, and one or more \texttt{assert} statements.

Each assertion is tagged with a structured \texttt{:named} label that records its requirement identity and internal component structure. These tags support later debugging, especially when the solver returns an UNSAT core or when the system needs to trace a contradiction back to a specific requirement fragment.

Translation is also polarity-aware. Inclusion and exclusion requirements must be compiled consistently with the global semantics of the eligibility program, and this stage ensures that the emitted assertions preserve the intended satisfaction behavior (prompts in Appendix~\ref{app:SMTProgrammer/SMTIncrementalProgrammer/SMTIncrementalTranslatorInclusion} and Appendix~\ref{app:SMTProgrammer/SMTIncrementalProgrammer/SMTIncrementalTranslatorExclusion}).

\subsubsection{Stage 5: Solver-Guided Validation and Rollback}
\label{app:smt_solver_validation}

After a candidate fragment is generated, \name{} immediately checks the updated program prefix with the SMT solver. If the fragment introduces a parse error, sort mismatch, or immediate contradiction, the fragment is rejected and the system rolls back to the previous program state. The failure signal is then used to guide the next translation attempt.

This immediate validation is a key design choice. It turns the solver into an active feedback source during program synthesis rather than merely a final execution engine. By checking each fragment as soon as it is generated, the system prevents local synthesis errors from accumulating into opaque downstream failures.

Rollback is strictly local: only the candidate fragment is discarded, while all previously validated requirements remain unchanged. This makes the synthesis process robust and keeps debugging localized.

\subsubsection{Stage 6: Semantic Verification}
\label{app:smt_semantic_verification}

Syntactic validity and solver consistency are necessary but not sufficient for faithful formalization. A fragment can be type-correct and logically satisfiable while still misrepresenting the source requirement. \name{} therefore applies an additional lightweight semantic verifier after solver validation.

The verifier checks whether the fragment preserves the intended meaning of the requirement and whether it uses the expected entities and qualifiers. In practice, this stage is especially useful for catching common generation errors such as polarity inversion, omitted qualifiers, incorrect qualifier attachment, unsupported variable substitutions, or accidental duplication of symbolic content. The verifier ensures that the generated logic remains semantically aligned with the source statement, rather than merely syntactically valid (prompts in Appendix~\ref{app:SMTProgrammer/SMTIncrementalProgrammer/SMTIncrementalVerifierInclusion} and Appendix~\ref{app:SMTProgrammer/SMTIncrementalProgrammer/SMTIncrementalVerifierExclusion}).


\subsubsection{Conservative Fallback and Engineering Robustness}
\label{app:smt_failopen}

In rare cases, a requirement may remain difficult to formalize reliably after multiple attempts. To avoid letting a small number of pathological cases stall the construction of the entire program, \name{} supports a conservative fallback policy. Under this policy, the system may weaken or bypass only the problematic requirement while explicitly recording that this occurred.

This fallback is used only as an engineering safeguard. The broader design principle is to preserve forward progress without silently introducing brittle or unverified logic into the final program.




\subsubsection{SMT Repair Pipeline}
Although the SMT programming workflow is carefully designed, errors can still occur. We therefore introduce a lightweight repair pipeline for generated SMT programs; see Appendix~\ref{app:ir_repair_pipeline}.

\subsubsection{Programming of Query Targets in SMT}

In parallel, we encode the disease list into SMT (Appendix~\ref{app:disease_list_coding}). Once the SMT representation of the disease list and the SMT program of the eligibility criteria are both available, we extract the positive literals from the eligibility-criteria SMT program; these literals define the query targets (Appendix~\ref{app:positive_canonical_literals}). We then schematize both the disease-list SMT and the positive-literal SMT with \(R_T\), following the relation vocabulary described in Appendix~\ref{app:naming_schema}. Using the preparation and execution procedure in Appendix~\ref{app:prep_and_exec_for_retrieval}, these components are converted into retrieval-ready form and added to the full trial-side SMT constraint representation.

\paragraph{Summary.}
Incremental SMT programming in \name{} is a staged program synthesis procedure that translates eligibility requirements into formal logic one fragment at a time. By separating symbol reuse, variable construction, local translation, solver-guided validation, and semantic verification, the system turns a brittle one-shot generation problem into a sequence of smaller, checkable decisions. This design substantially improves reliability, interpretability, and debugging for clinical text-to-logic semantic parsing.
    \subsection{LLM-Orchestrated Repair of Trial-Side SMT Programs (Trial-Side Semantic Parsing)}
\label{app:ir_repair_pipeline}

This appendix describes the post-processing pipeline we use to improve trial-side SMT programs after initial semantic parsing. The goal of this pipeline is not to re-parse eligibility criteria from scratch, but to \emph{repair} and \emph{tighten} existing SMT programs so that they more faithfully encode the intended inclusion and exclusion logic while preserving a conservative retrieval policy.

\subsubsection{Overview}

The repair pipeline operates over a canonical directory of SMT-LIB v2 programs, where each file corresponds to one effective trial--subcohort--side pair. Each program represents either the inclusion or exclusion constraints for a particular subcohort. The pipeline applies a sequence of specialized LLM-based repair stages:
\[
\texttt{repair} \rightarrow \texttt{polarity} \rightarrow \texttt{logic} \rightarrow \texttt{gate} \rightarrow \texttt{underconstraint} \rightarrow \texttt{meaning}.
\]
The canonical directory is treated as the single source of truth: when a stage produces a valid repair, it overwrites the corresponding canonical SMT file in place. Each stage also records a complete before/after snapshot and stores detailed debugging artifacts.

The stages serve different purposes. Early stages correct semantic drift from the initial parser, such as incorrect polarity or malformed logical structure. Later stages identify files that should be removed entirely because the corresponding side has no real natural-language criteria, tighten underconstrained encodings that would otherwise produce false positives, and finally ensure that every declared variable has a self-contained machine-readable inline definition.

\subsubsection{Inputs and Context}

For each SMT file, the repair pipeline assembles a structured context bundle. This bundle includes:

\begin{itemize}[leftmargin=*]
    \item the parent trial identifier and effective subcohort identifier;
    \item the target side (\textit{inclusion} or \textit{exclusion});
    \item shared trial-level context;
    \item subcohort-specific context, when available;
    \item side-specific eligibility text for the current subcohort;
    \item the existing canonical SMT program;
    \item the raw trial record from the clinical-trial corpus, used as a fallback safety source.
\end{itemize}

Subcohort matching is performed conservatively. When the snapshot contains explicit mappings from effective trial identifiers to enrollment cohorts, those mappings are used directly. Otherwise, the pipeline falls back to positional matching only when the corresponding lists are aligned and unambiguous. If no reliable subcohort match is found, a default context is constructed from the available shared trial content. This design allows the repair modules to reason over the correct subcohort whenever possible, while still remaining robust to partially missing upstream metadata.

\subsubsection{Stage 0: General Semantic Repair}

The first stage applies a general-purpose semantic repair prompt to the current SMT program. Its purpose is to correct broad semantic drift introduced during the initial text-to-SMT translation, especially cases where literal parsing would make the program overly strict and exclude patients that should remain eligible.

This stage is conservative and minimal: it prefers the smallest set of edits needed to restore fidelity to the natural-language criteria. It is allowed to introduce qualified variants of existing predicates, such as \texttt{@@previous\_episode} or \texttt{@@clinically\_significant}, when the trial text clearly distinguishes the broad stem from the clinically intended subset. It may also add bridge axioms of the form
\[
\phi_{\mathrm{qualified}} \Rightarrow \phi_{\mathrm{base}}
\]
to preserve the relationship between the refined predicate and the original stem.

A key recurring pattern handled at this stage is the distinction between an \emph{index condition} required for inclusion and a \emph{prior episode} of the same condition used for exclusion. For example, if the inclusion criteria require a current thromboembolic event, then an exclusion such as ``deep vein thrombosis in the preceding three years'' should generally be interpreted as excluding \emph{prior} episodes rather than the current index event itself. The repair stage makes this distinction explicit by introducing a qualified prior-episode predicate and rewriting the exclusion constraint to use it. Prompt for stage 0 can be found in Appendix~\ref{app:repair/relaxer}.

\subsubsection{Stage 1: Polarity Repair}

The second stage operates only on exclusion programs. Its purpose is to correct polarity errors caused by the encoding convention used in the system: a patient is eligible only when both the inclusion program and the exclusion program are satisfiable, so exclusion programs are written in a ``not excluded'' form. Under this convention, a polarity mistake can invert the clinical meaning of an exclusion criterion, making the program satisfiable exactly when the patient should be excluded.

This stage inspects each exclusion SMT file against the natural-language exclusion criteria and repairs only the assertions whose polarity is clearly reversed. It does not attempt to rewrite the entire program unless necessary. If no polarity drift is found and the declaration annotations are already valid, the stage returns \texttt{NO MODIFICATION REQUIRED}. Prompt for stage 1 can be found in Appendix~\ref{app:repair/polarity}.

\subsubsection{Stage 2: Logic Repair}

The third stage also operates only on exclusion programs and focuses on logical structure rather than polarity. It repairs errors such as incorrect qualification scope, dropped conjuncts, or mis-grouped conditions whose interpretation depends on the surrounding trial context. A typical example is a qualifier that has been attached too narrowly to one part of a criterion when the natural language makes clear that it should scope over multiple components.

This stage is intentionally conservative. It is not a re-parser; instead, it repairs local logical drift when the current SMT structure is inconsistent with the natural-language exclusion criteria for the same subcohort. Prompt for stage 2 can be found in Appendix~\ref{app:repair/qualifier}.

\subsubsection{Stage 3: Criteria Gate}

The fourth stage is different from the others: it does not repair SMT content. Instead, it decides whether a side-specific SMT file should exist at all. This stage is needed because some trials or subcohorts lack meaningful inclusion or exclusion criteria on one side, but upstream processing may still produce an empty, placeholder, or otherwise non-substantive SMT file.

The gate prompt is purely LLM-based and is run even when the SMT file is empty. It receives two evidence blocks: a subcohort-scoped authoritative bundle containing the target side's criteria, and the raw trial corpus entry for the entire trial. The decision is conservative: the file is deleted only when the model is very sure that the target side for the current subcohort is genuinely absent or placeholder-only. If there is any real side-specific eligibility content, even a single condition such as an age threshold, the file is kept.

This stage is destructive only in one direction: it may delete a file, but it never rewrites one. Prompts for stage 3 can be found in Appendix~\ref{app:repair/coverage_checker} and Appendix~\ref{app:repair/criteria_gater}.

\subsubsection{Stage 4: Underconstraint Repair}

The fifth stage addresses the opposite failure mode from the initial semantic repair stage. Whereas the first stage corrects overly strict encodings that would create false negatives, the underconstraint stage tightens encodings that are too lax and would admit false positives.

The underconstraint prompt explicitly audits the current program for several classes of failures:

\begin{itemize}[leftmargin=*]
    \item \textbf{Missing gates:} a natural-language requirement is absent from SMT or encoded vacuously.
    \item \textbf{Weakened structure:} the SMT drops conjuncts, weakens thresholds, or otherwise loosens the intended logic.
    \item \textbf{Overly narrow medical meaning:} the SMT uses a predicate that is narrower than the umbrella concept described in the eligibility text.
    \item \textbf{Insufficient evidence modeling at prescreen:} for exclusion criteria, strong clinical support or documented suspicion may need to trigger exclusion when the natural language and prescreen setting justify it.
    \item \textbf{Undercontextualized variable meanings:} the variable definitions themselves are too narrow, too weakly qualified, or attached to the wrong timeframe, causing downstream coding to miss clinically intended matches.
\end{itemize}

A distinctive feature of this stage is that it allows repairs at the level of the \emph{representation} rather than only at the level of constraints. When the natural language expresses a broader umbrella concept than the one captured by the existing variable stem, the repair may introduce an umbrella variable and connect narrower stems to it via bridge axioms. Likewise, when the natural language justifies a qualifier such as \texttt{@@suspected} or \texttt{@@strong\_clinical\_support}, the stage may add the qualified predicate and use it in the constraint while preserving the subset relation to the original stem.

This stage is especially important for reducing false positives in retrieval. Without it, a trial-side program may be formally satisfiable while still omitting clinically essential restrictions that a human prescreener would apply immediately. Prompt for stage 4 can be found in Appendix~\ref{app:repair/constrainer}. 

\subsubsection{Stage 5: Variable-Meaning Enrichment}

The final stage does not modify the program logic. Instead, it repairs only variable declarations. Every \texttt{declare-const} line must contain a valid inline JSON definition on the same line, describing the variable's meaning and the operational conditions under which it should be set to true, false, null, or to a numeric value.

The purpose of this stage is twofold. First, it guarantees that each symbol is self-contained and can be interpreted consistently by downstream coders or patient-fact extraction components. Second, it eliminates ambiguous or detached declaration metadata, such as JSON comment blocks that appear above the declaration instead of inline.

The enrichment prompt derives variable meaning from four sources, in order of priority: the symbol name, its usage in SMT assertions, the side-specific eligibility text, and the broader shared or subcohort context. It is not allowed to change assertions, introduce new declarations, or alter the logical meaning of the program. Its only job is to make declaration annotations complete, valid, and sufficiently contextualized. Prompt for stage 5 can be found in Appendix~\ref{app:repair/meaning_enricher}.

\subsubsection{Strict Validation and In-Place Updates}

All stages except the gate stage are subject to strict validation before their output can overwrite the canonical SMT file. The validator checks:

\begin{itemize}[leftmargin=*]
    \item balanced parentheses after stripping comments;
    \item absence of forbidden constructs such as datatypes, arrays of unsupported shape, or uninterpreted sorts;
    \item allowed declaration forms and sorts;
    \item validity and completeness of the inline JSON declaration comments;
    \item optionally, the presence and format of assertion names.
\end{itemize}

Before validation, the pipeline also normalizes detached JSON declaration blocks into the required inline form when possible. A repaired program is committed only if it passes validation. Otherwise, the repaired candidate and its validation errors are written to inspection artifacts, and the canonical SMT file is left unchanged.

This strict overwrite policy is central to the pipeline's trustworthiness. The LLM is allowed to propose repairs, but it is not allowed to silently degrade the structural integrity of the symbolic artifact.

    \subsection{Recovering Positive Canonical Literals from Trial Logic (Trial-Side Semantic Parsing)}
\label{app:positive_canonical_literals}

This appendix describes how \name{} derives the \emph{positive canonical literals} used in retrieval. Intuitively, these literals identify the clinically meaningful trial-side target conditions that the formal eligibility logic is positively about: diseases, prior procedures, treatment histories, findings, and other canonical clinical states that make a trial relevant to retrieve. These recovered targets are not used merely as raw concepts; they are converted into trial-side target predicates paired with relations from \(R_t\), so that retrieval operates over the same ontology-backed predicate space as the rest of the formalization.

The key point is that these targets are not obtained by simply reading off variable names from the SMT program. Instead, \name{} performs a lightweight semantic recovery procedure over the linked trial logic to identify which canonical clinical propositions are \emph{positively enforced} by the eligibility criteria.

For each trial, the procedure starts from the linked SMT representation together with the canonical variable inventory produced by the normalization pipeline. It then scans the requirement-bearing portions of the SMT program---that is, the assertions corresponding to retained criterion components---rather than treating the full formula as an undifferentiated set of constraints. This keeps the extracted targets aligned with the semantic content of the trial criteria, rather than with auxiliary bookkeeping structure.

Within these retained component formulas, \name{} performs a polarity-aware traversal to recover canonical literals that occur in \emph{positive} positions. Negation reverses polarity, and implication is handled compositionally, so the method distinguishes conditions that the trial is positively seeking from conditions that appear only in exclusions, guards, or logically reversed contexts. In this way, the extracted literals correspond to clinical states whose truth supports satisfying the trial-side component.

To make this recovery robust, \name{} first inlines shallow Boolean helper definitions of the form
\[
(= X \;\phi),
\]
so that positivity can flow through auxiliary symbols into the underlying clinical proposition. This is important because trial logic often introduces intermediate variables for readability or modularity, while retrieval should expose the clinically meaningful concept rather than the local helper name.

\name{} also performs a lightweight implication-based expansion over auxiliary constraints. The motivation is that the most retrieval-useful concept is not always the exact symbol that appears in a component. In many cases, a positively enforced trial-side state implies a more canonical or retrieval-friendly diagnosis, finding, or treatment-related concept. When such implication structure is explicitly encoded in the SMT, \name{} propagates positivity through it and retains the implied canonical target as well. The same idea is used for simple threshold-triggered conditions when a positively enforced numeric predicate implies a clinically meaningful canonical state.

Finally, qualified or context-specific variants are collapsed to representative canonical stems, so that multiple logical variants of the same underlying concept contribute to a single normalized retrieval target. The resulting targets are then passed through the same conservative specificity filtering used elsewhere in retrieval: a target is removed only when a strictly more specific substitute from the same source-specific candidate set is already present. After filtering, each retained target concept is paired with an appropriate trial-intent relation from \(R_t\) (for example, \textsc{Treat}, \textsc{Prevent}, or \textsc{Diagnose}) to form a canonical trial-side target predicate. In this way, positive literal recovery yields not just concepts, but formal retrieval targets that live in the same \((r,k,q)\)-style representation as the rest of \name{}.

This procedure produces a set of retrieval targets that is broader and more semantically grounded than disease-list metadata alone. Disease lists capture coarse trial intent, but positive canonical literals recovered from the trial logic additionally expose the finer-grained clinical structure encoded in eligibility criteria, including prior interventions, treatment context, findings, and other non-disease states. In this sense, positive canonical literal extraction is best viewed as a \emph{semantic target recovery} step: it reads the formal trial logic, identifies the canonical clinical states that the trial is positively seeking, and exports them into the unified trial--patient logic store as reusable retrieval signals.

\paragraph{Example.}
Consider a trial whose linked SMT contains a retained requirement component that positively enforces a variable corresponding to active Graves' hyperthyroidism, together with an auxiliary implication from that variable to a broader canonical diagnosis such as thyrotoxicosis due to Graves' disease. A surface-level extractor might recover only the exact helper variable appearing in the component. In contrast, \name{} first identifies that the component is positively about the active disease state, then propagates positivity through the auxiliary implication, and finally maps the result back to canonical retrieval targets. The final output is therefore a small set of clinically meaningful canonical literals, rather than a brittle set of internal SMT variable names.
    \subsection{Disease List Processing and Canonicalization for Trial Target Conditions}
\label{app:disease_list_coding}

In addition to full eligibility criteria, many clinical trials provide short disease or condition fields that summarize the clinical problems a study aims to treat, mitigate, monitor, prevent, or examine longitudinally. These fields are useful for retrieval because they often expose the trial's target disease space directly. However, they are substantially less structured than eligibility criteria and may contain formatting artifacts, mixed levels of specificity, or contextually related but non-target conditions. \name{} therefore processes disease-list content with a dedicated lightweight pipeline rather than routing it through the full SMT compilation workflow used for eligibility criteria.

The goal of this pipeline is to construct a faithful set of trial-side disease targets that can augment the target set used in retrieval. Concretely, \name{} applies a staged procedure:
\[
\texttt{Extractor} \rightarrow \texttt{Eliminator} \rightarrow \texttt{Preprocessor} \rightarrow \texttt{Canonicalizer}.
\]
This pipeline is lightweight in the sense that it does not attempt to recover full logical structure from disease-list text. Instead, it focuses on high-recall disease recovery, conservative filtering, stable string normalization, and ontology grounding.

\paragraph{Extraction.}
The first stage extracts a comprehensive list of candidate diseases from trial context, using the summary, disease metadata, and inclusion-focused text. Extraction is intentionally recall-oriented, but constrained to explicit evidence: diseases must be explicitly mentioned in the summary or inclusion requirements rather than inferred from loosely related symptoms or background context. Existing disease-list items provided with the trial are preserved rather than dropped, even when their surface form is awkward or metadata-like.

\paragraph{Elimination.}
Because trial summaries and metadata may mention diseases that are adjacent to the study context but not actually part of the target disease set, \name{} next applies a relevance-based elimination step. A disease is retained if it is explicitly relevant to the trial's clinical focus, including treatment, mitigation, screening, prevention, diagnosis, surveillance, or longitudinal study. A disease is not removed merely because patients with that condition may be excluded in some settings, or because only a subtype or stage is ultimately enrolled. This stage improves precision without collapsing away clinically meaningful target conditions.

\paragraph{Preprocessing.}
After elimination, \name{} applies conservative string-level preprocessing to normalize the retained disease list. This stage performs lightweight cleanup such as whitespace normalization, alias handling, optional synonym expansion, and deduplication. A key design choice is that disease entries are treated as atomic by default. In particular, the preprocessor does not split on commas, since many biomedical disease names use inverted modifier order, such as ``Kidney Failure, Chronic'' or ``Pulmonary Disease, Chronic Obstructive.'' Instead, such strings are preserved and normalized only in ways that do not alter their meaning. The purpose of preprocessing is therefore not semantic reinterpretation, but representational stability.

\paragraph{Canonicalization.}
The final stage maps each retained disease mention to SNOMED CT. This step follows the same conservative grounding principle used elsewhere in \name{}: a candidate concept is accepted only if it is either semantically identical to the intended disease string or a valid strict ancestor of it. Exact matches are preferred. If no exact match exists, the system selects the most specific valid ancestor. If no faithful match is available, the disease mention is retained without forcing an incorrect ontology assignment. This prevents false precision and preserves the original disease evidence for downstream use. The canonicalization workflow is roughly the same as the one described in Appendix \ref{app:canonicalization}.

\paragraph{Cohort-specific processing.}
When trials contain multiple enrollment cohorts, \name{} can run this disease pipeline separately for each cohort using cohort-specific contextual text. This is important because different cohorts may correspond to different diseases, subtypes, or treatment settings. Cohort-aware disease processing therefore preserves distinctions that would be lost if the trial were treated as having a single flat disease list.

\paragraph{Role in retrieval.}
The output of this pipeline is a set of normalized disease mentions together with their selected SNOMED concepts when faithful grounding is available. These disease-derived targets are incorporated into the trial-side target representation used during retrieval by pairing them with appropriate trial-intent relations from \(R_t\), alongside targets recovered from the formal eligibility representation.

\paragraph{Summary.}
Overall, \name{} treats trial disease lists as a distinct semantic parsing problem. Rather than compiling them into full eligibility logic, the system extracts target disease mentions from trial context, removes irrelevant items, normalizes them conservatively, and grounds them to SNOMED CT under an exact-or-valid-ancestor policy. This yields a compact but faithful disease-target representation that complements the richer formal structure derived from trial eligibility criteria.
    
\subsection{Patient-Side Semantic Parsing Pipeline}
\label{app:patient_coding_pipeline}

This appendix describes the patient-side semantic parsing pipeline that maps raw clinical notes directly into the symbolic space used for SMT-based trial matching. The central design goal is not merely to extract medical information from text, but to construct \emph{patient-side SMT-ready atomic evidence} over the same ontology-backed symbolic vocabulary used on the trial side. In this sense, the patient-side problem is a semantic parsing problem into SMT from the outset.

The pipeline is intentionally modular. Rather than asking a single model to map a free-text patient note directly into final formal assertions in one step, the system decomposes patient-side semantic parsing into a sequence of narrower transformations, each with its own output contract and verifier. This decomposition improves controllability, makes failure modes easier to localize, and yields intermediate artifacts that remain interpretable to both clinicians and system developers.

At a high level, the pipeline takes as input a free-text patient note and progressively constructs SMT-compatible patient evidence. Intermediate stages include fact extraction, fact normalization, entity grounding, atomic SMT declaration and verification.

The key distinction from the trial side is that the patient-side pipeline does \emph{not} compile a full eligibility formula. Trial text specifies compositional requirements and is therefore compiled into executable logical constraints. Patient notes instead provide evidence about one individual. The patient-side semantic parsing problem is therefore to recover \emph{atomic SMT-grounded assertions} that can later be checked against the trial-side formula. These atomic outputs do not themselves encode eligibility logic; they provide the structured evidence that downstream SMT evaluation consumes. The prompts used for patient-side semantic parsing are listed in Appendix~\ref{app:prompt-patient-side-parsing} (Appendix~\ref{app:patient-parsing/extraction} - \ref{app:patient-parsing/any}).

\subsubsection{Design Goals}

The patient-side pipeline is designed around four requirements.

First, it must preserve \emph{clinical faithfulness}. Patient notes are often incomplete, indirect, and heterogeneous in style. The system therefore separates explicit statements from plausible but uncertain clinical inferences, and it requires every extracted diagnosis, fact, and atomic declaration to be justified by supporting evidence.

Second, it must support \emph{direct symbolic executability}. Downstream trial matching operates over typed variables and logical constraints rather than raw text alone. Patient-side outputs must therefore be normalized into ontology-grounded atomic forms that can be rendered directly as Boolean predicates, numeric assignments and temporalized SMT-ready evidence.

Third, it must provide \emph{fine-grained auditability}. Each stage emits an interpretable artifact with a strict schema, so that failures can be localized to a specific transformation step.

Fourth, it must be \emph{conservative under uncertainty}. On the patient side, unsupported overcommitment can lead to incorrect exclusions or spurious matches. The pipeline therefore prefers explicit support, preserves uncertainty where appropriate, and inserts dedicated verification stages before evidence is promoted into the formal symbolic representation.

\subsubsection{Patient Constraint Formal Representation}
\label{app:patient_pipeline_relation_to_formalism}

In the main text, each patient record \(p \in P\) is represented by a
\emph{patient constraint} \(\PC(p)\), an SMT formula over canonical predicates in
\(\Pi(\Rm \cup \Rp, K, Q)\). The patient-side semantic parsing pipeline is thus
the process of mapping free-text patient descriptions into the ontology-grounded
symbolic space used by the formal representation in the main text.

Specifically, the parser extracts grounded patient evidence in canonical form,
with predicates
\[
\pi = (r,k,q),
\qquad
r \in \Rm \cup \Rp,\;\; k \in K,\;\; q \in Q \cup \{\emptyset\},
\]
and, when needed, atomic constraints
\[
a = (\pi,\mathsf{cmp},t) \in A.
\]
These atoms form the building blocks of \(\PC(p)\).

This differs from the trial side. For each trial \(c \in C\), the main text
defines a \emph{trial constraint} \(\TC(c)\), an SMT formula over predicates in
\(\Pi(\Rm \cup \Rt, K, Q)\), often with richer logical structure such as conjunction, disjunction, counting, and negation. Patient notes,
by contrast, typically provide grounded evidence---diagnoses, findings,
procedures, medications, measurements, histories, and chief complaints---rather
than explicit eligibility logic.

Accordingly, the role of patient-side semantic parsing is not to synthesize an
independent eligibility program, but to construct SMT-consumable patient
evidence in the same symbolic language used by the trial representation. Because
both \(\PC(p)\) and \(\TC(c)\) are grounded in the shared ontology
\[
O = (R, K, Q, \sqsubseteq),
\]
their interaction can be evaluated in a uniform way through the subsumption
family
\[
\sqsubseteq
=
\sqsubseteq_{\mathrm{R}}
\cup
\sqsubseteq_{\mathrm{K}}
\cup
\sqsubseteq_{\mathrm{causal}}
\cup
\sqsubseteq_{\mathrm{Q}}.
\]

In this sense, patient-side semantic parsing is the bridge from free text to the
formal representation: it converts patient descriptions into canonical
predicates and atomic constraints so they can be matched directly against trial
constraints in the shared ontology-backed symbolic space.

\subsubsection{Pipeline Overview}

Before semantic parsing, the system may first infer likely diagnoses and possible prevention targets from the patient note; these stages are described in Appendix~\ref{app:differential_diagnose} and Appendix~\ref{app:disease-to-prevent}. They are not part of the semantic parsing pipeline itself.

The patient-side semantic parsing pipeline then proceeds in four main steps.

\paragraph{Fact extraction.}
The note is broken into a set of self-contained patient facts. Each fact captures one clear piece of patient information together with directly attached details such as severity, location, marker status, or quantity. Each fact is also linked back to its source text for auditing.

\paragraph{Fact normalization.}
The extracted facts are rewritten into a cleaner and more explicit form for downstream processing. This includes expanding shorthand expressions, making shared structure explicit, and preserving the original meaning while making the content easier to encode formally.

\paragraph{Entity grounding.}
Normalized facts are processed by a medical entity recognizer and linked to SNOMED concepts. This step aligns patient evidence with the same canonical concept space used on the trial side.

\paragraph{Atomic SMT declaration.}
The grounded facts are converted into typed atomic declarations for downstream matching. These include demographic variables such as age and sex, as well as canonical clinical facts such as conditions, findings, procedures, substances, and observables. Numeric values are kept in numeric form, and each declaration carries any relevant qualifiers and temporal scope.

\paragraph{Verification.}
Lightweight checks are applied throughout the pipeline to ensure that extracted facts stay faithful to the note, normalization does not change meaning, ontology relationships are valid, and final atomic declarations are supported by the source text.

\subsubsection{Temporal Representation}

Temporal information is attached throughout the patient pipeline. When timing is unclear, the system uses conservative broad intervals rather than inventing precise dates. Final patient declarations therefore record when a fact certainly holds or may hold, so that time-sensitive trial criteria can be evaluated later.

\subsubsection{Patient-Side Semantic Parsing into SMT}

The patient side does not build a full standalone SMT formula like the trial side. Instead, it produces typed, ontology-grounded, time-aware atomic facts that are meant to be consumed directly by downstream SMT evaluation. In this sense, the patient pipeline is still a form of semantic parsing into SMT: its output is not generic structured data, but formal symbolic evidence.

\subsubsection{Resulting Patient Representation}

The final output is a structured patient representation consisting of source-linked facts, normalized facts, grounded medical entities, and atomic SMT-ready evidence with values, qualifiers, and time scope. The trial side provides the logical structure, and the patient side provides the grounded evidence used to evaluate it.
\subsection{Patient-Side Time Intervalization}
\label{app:patient_intervalization}

\definecolor{darkgreen}{rgb}{0.0,0.5,0.0}
\definecolor{darkred}{rgb}{0.5,0.0,0.0} 
\begin{figure*}[!htb]
  \centering
  \includegraphics[width=0.85\linewidth]{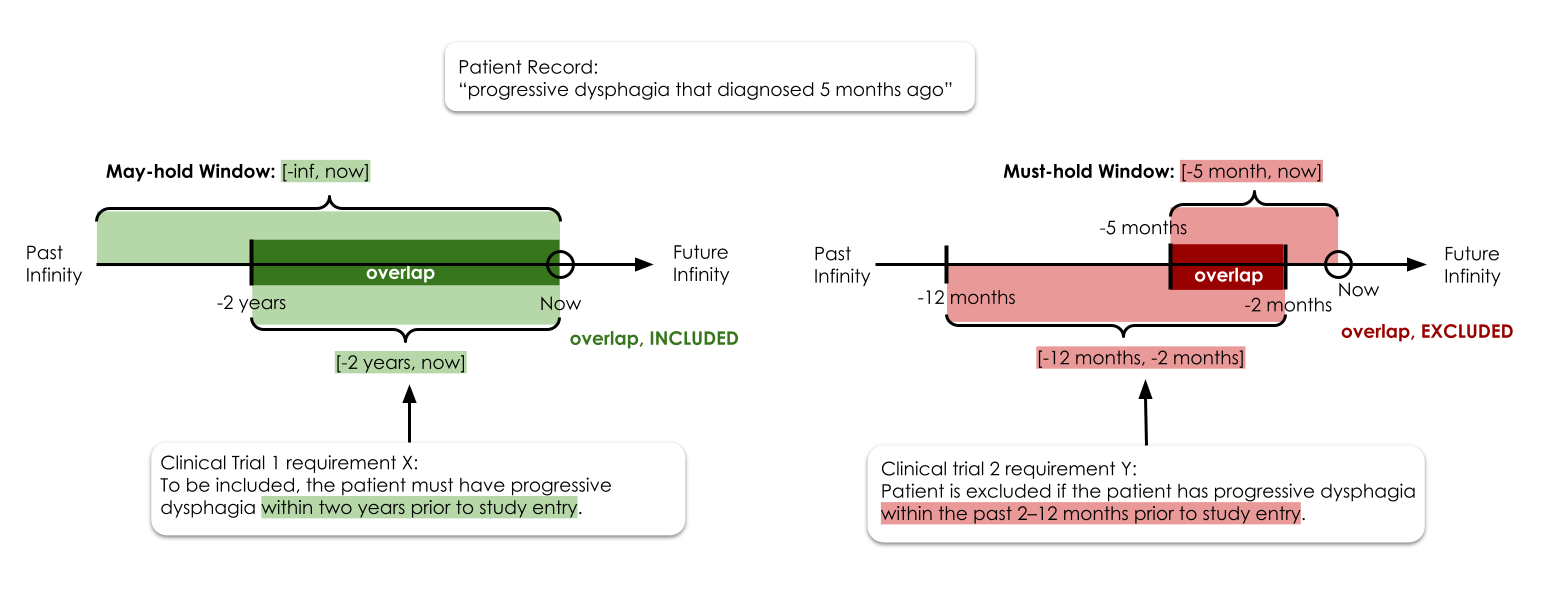}
  \caption{Time intervalization converts implicit temporal information in patient notes into explicit, conservative time windows used for eligibility evaluation. From each patient fact, we extract two temporal intervals: (i) a \textbf{\textcolor{darkgreen}{may-hold window}}, representing the time interval during which the patient \textbf{\textcolor{darkgreen}{may}} have progressive dysphagia, and (ii) a \textbf{\textcolor{darkred}{must-hold window}}, representing the time interval during which the patient \textbf{\textcolor{darkred}{must}} have progressive dysphagia. When an \textbf{inclusion criterion} has any time-interval overlap with the \textbf{\textcolor{darkgreen}{may-hold window}}, the patient is eligible under this criterion. When an \textbf{exclusion criterion} has any time-interval overlap with the \textbf{\textcolor{darkred}{must-hold window}}, the patient is ineligible under this criterion. Prompt implementation can be found at Appendix \ref{app:patient-parsing/canonvar} and \ref{app:patient-parsing/demovar}}
  \label{fig:patient_fact_intervalization}
\end{figure*}

\subsubsection{Two-window representation}
Each extracted patient fact $f$ is represented with two time windows:
(i) a \emph{certain} window $W^{\textsc{cert}}(f)$ during which the fact definitely holds, and
(ii) a \emph{possible} window $W^{\textsc{poss}}(f)$ that upper-bounds where the fact may hold.
Both windows are encoded as start/end bounds with direction ($\{\texttt{past},\texttt{now},\texttt{future}\}$), magnitude, units, and inclusivity. 

\subsubsection{Normalization to numeric bounds}
We normalize bounds to hours relative to the note timestamp (``now''):
\[
(\texttt{direction}, \texttt{magnitude}, \texttt{unit}) \mapsto t \in \mathbb{R}\cup\{\pm\infty\},
\]
with past mapped to negative hours and future to positive hours. We reject invalid windows (future times for diagnoses; start $>$ end) and clip/approximate unbounded windows using a large sentinel when required for engineering convenience.

\subsubsection{SMT encoding}
For a Boolean predicate $p$ representing a fact and a window $[t_s,t_e]$, we compile a time-aware assertion using an explicit time variable or a window-qualified predicate, depending on the chosen SMT signature. Our implementation carries the four window fields
(\texttt{start\_time\_in\_hours}, \texttt{end\_time\_in\_hours}, \texttt{start\_time\_inclusive}, \texttt{end\_time\_inclusive})
with each fact and uses them for (a) rule-based entailment reuse and (b) deduplication.

\subsubsection{Deduplication key}
Derived and extracted facts are deduplicated by:
\[
(\texttt{entity\_variable\_name},\ t_s,\ t_e,\ \mathbb{I}[\text{start inclusive}],\ \mathbb{I}[\text{end inclusive}]),
\]
so the same predicate under different windows is retained.

\subsection{Patient-Side Inference Modules}
\label{app:patient_inference}

\subsubsection{Differential diagnosis from full notes}
\label{app:differential_diagnose}

We run a differential diagnoser over the full note to propose diagnosis candidates with:
\texttt{diagnosis}, \texttt{confidence} $\in[0,1]$, supporting evidence snippets, and time bounds
(\texttt{confirmable\_latest\_start\_time}, \texttt{confirmable\_earliest\_end\_time}).
We generate multiple stochastic samples and merge them by diagnosis name, keeping the maximum confidence and unioning evidence.
We drop candidates whose inferred time bounds are in the future or temporally inconsistent. Prompt implementation can be found in Appendix \ref{app:patient-parsing/ddx}.

\subsubsection{Inferring procedures-to-undergo}
Note that this component is not used in the current experiments, but our failure-case analysis suggests that reintroducing it may be beneficial. We separately infer \emph{significant} procedures the patient likely needs to treat primary conditions (excluding routine diagnostics and logistics). Each inferred procedure is produced with confidence and a forward-looking time window; it is compiled into procedure predicates such as \texttt{patient\_needs\_to\_undergo\_\{proc\}} with guarded activation analogous to diagnoses. Prompt implementation can be found in Appendix \ref{app:patient-parsing/procedure}.

\subsubsection{Diseases-to-prevent}\label{app:disease-to-prevent}
We identify diseases the patient explicitly seeks to prevent (or prevention goals that are near-universal consequences of an explicitly stated condition). These are encoded as prevention-intent predicates and used only in task modes that evaluate preventive matching. Prompt implementation can be found in Appendix \ref{app:patient-parsing/prevention}.

\newpage
\section{Database Retrieval Details}\label{app:database_umbrella}

\subsection{Database Preparation for Retrieval and High-Level Execution Flow of Retrieval}\label{app:prep_and_exec_for_retrieval}

\begin{figure*}[htbp]
  \centering
  \includegraphics[width=\linewidth]{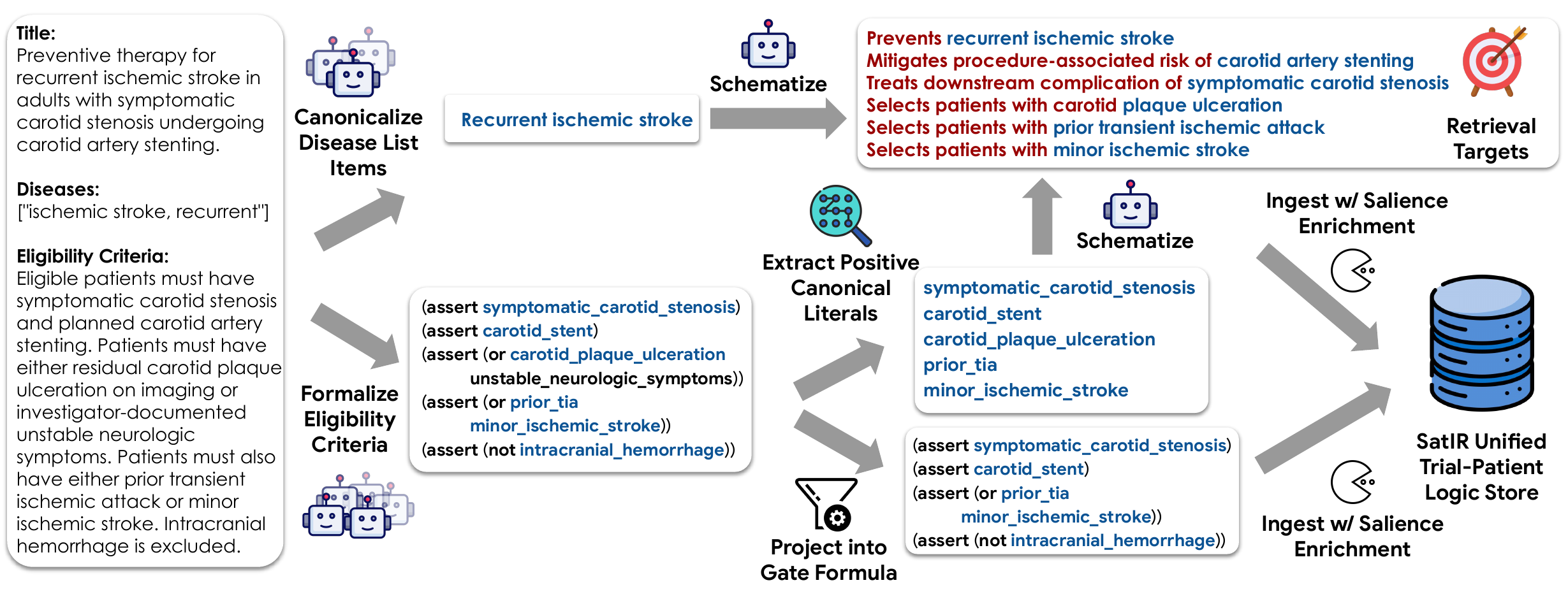}
  \caption{Offline construction of the trial-side retrieval database in \name{}. From each trial, \name{} derives canonical disease-list targets, canonical targets recovered from the formalized eligibility logic, and a recall-preserving eligibility gate formula projected from the full SMT representation. These artifacts are normalized, schematized into retrieval targets , and ingested into the unified trial--patient logic store for query-time retrieval and pruning.}
  \label{fig:offline-db-retrieval}
\end{figure*}

\begin{figure*}[htbp]
  \centering
  \includegraphics[width=\linewidth]{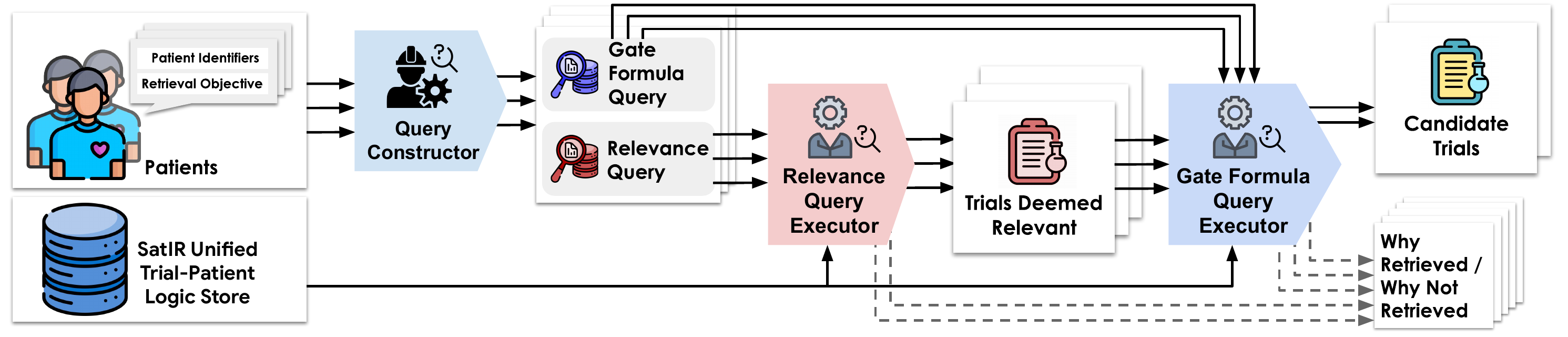}
    \caption{Online query-time retrieval in \name{}. Given a patient record and retrieval objective, \name{} retrieves topically relevant trials by objective-conditioned matching against precomputed trial-side targets, evaluates recall-preserving gate formulas whose disjunctive salience structure is retained on the trial side against the patient's sparse observed facts, and passes the remaining candidates to full SMT-based eligibility checking.}
  \label{fig:online-db-retrieval}
\end{figure*}

As shown in Figure~\ref{fig:offline-db-retrieval}, \name{} builds retrieval around an offline-constructed trial-side index stored in a unified trial--patient logic store. Each trial contributes two kinds of relevance targets, together with an additional gate formula used for high-recall pruning at query time.

The first source of relevance targets is the trial's formalized eligibility logic. From the SMT representation, \name{} extracts \emph{positive canonical literals} that represent clinically meaningful targets explicitly expressed in the criteria, including not only disease conditions but also prior procedures, prior treatments, and other eligibility-defining clinical states. The second source is the trial disease list. These disease-list items are canonicalized through the same concept normalization pipeline described in Section~\ref{sec:concept_canonicalization}, but are generated independently of the SMT-derived literals and mapped into the same retrieval schema. This distinction matters: disease-list targets contribute only disease-oriented trial intent, whereas SMT-derived literals capture a broader set of clinically relevant retrieval signals. In practice, the disease list is especially useful for trials whose central condition is implicit in the eligibility logic, such as prevention or diagnosis-oriented studies.

After canonicalization, both sources are schematized into a shared target representation and filtered conservatively for specificity before ingestion into the store (Figure~\ref{fig:offline-db-retrieval}). A target is removed only when a strictly more specific substitute already exists within the same source-specific candidate set; otherwise it is retained to preserve recall.

\begin{figure*}[!htb]
  \centering
  \includegraphics[width=\linewidth]{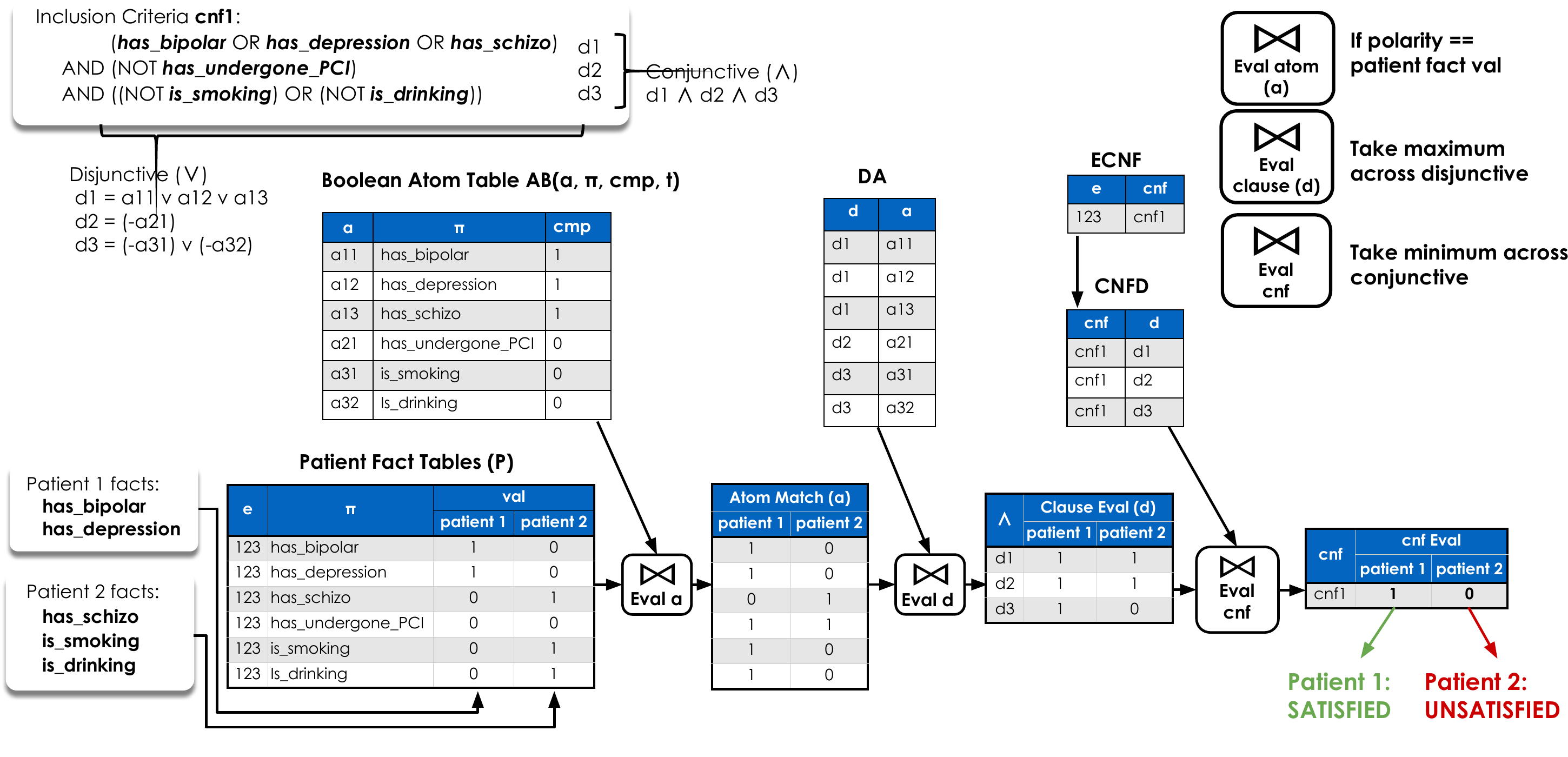}
  \caption{Example execution of the projected-constraint retrieval procedure in the SQL layer. Execution proceeds in three stages: compatible atom matching, clause-level support, and all-clause aggregation over the trial CNF.}
  \label{fig:logical_prefiltering}
\end{figure*}

In addition to relevance targets, each trial $c$ is assigned a gate formula $\Psi{c}$ derived offline from its full eligibility formula $\TrialFormula{c}$, as illustrated in Figure~\ref{fig:offline-db-retrieval}. The gate formula is a recall-preserving projection of the full SMT program onto efficiently testable canonical atoms, satisfying
\[
\TrialFormula{c} \Rightarrow \Psi{c}.
\]
It therefore captures necessary, but generally not sufficient, conditions for full eligibility. Importantly, clinically acceptable salience alternatives are compiled into $\Psi{c}$ on the trial side and retained as disjunctive structure, rather than being exhaustively enumerated on the patient side. At query time, the patient's sparse observed evidence is checked against this retained structure, so whole-fact missingness and specificity-aware uncertainty are handled through which trial-side branches can or cannot be satisfied. Appendix~\ref{app:sql_prefilter} describes how these projected gate formulas are ingested, stored, and executed, and Figure~\ref{fig:logical_prefiltering} shows a concrete example.

At query time, retrieval proceeds in two stages, shown in Figure~\ref{fig:online-db-retrieval}. Given a patient record and retrieval objective, \name{} first performs objective-conditioned matching against the precomputed trial-side targets to identify topically relevant trials. It then evaluates each retained trial's gate formula as a lightweight high-recall filter by checking the patient's sparse observed evidence against the gate's retained trial-side disjunctive structure. The surviving candidates are passed to the full SMT-based eligibility checker, along with lightweight explanations of why each trial was retrieved or filtered.
    \UseRawInputEncoding
\subsection{Projected CNF Storage and SQL-Style Retrieval Execution}
\label{app:sql_prefilter}

This appendix explains how the executable projected constraints from Section~\ref{sec:retrieval} are stored and evaluated in \name{}. The key idea is to compile the retrieval-relevant part of each trial or patient constraint into a relational representation that can be executed efficiently with SQL-style joins and aggregations. This yields a scalable database retrieval layer over the projected constraint space.

As described in Section~\ref{sec:retrieval}, each original constraint formula is conservatively converted into an executable quantifier-free formula over canonical atoms that can be represented and matched in the database while preserving full recall. The executable formula is then written in conjunctive normal form (CNF), so retrieval can be implemented by evaluating atom compatibility, clause support, and all-clause satisfaction directly over relational tables.

A key design choice is that clinically acceptable alternatives---such as ontology-supported generalizations or other retrieval-side substitutions---are materialized on the \emph{trial side} as explicit disjunctions in the executable CNF. We do \emph{not} exhaustively materialize all such alternatives on the patient side. Online retrieval therefore evaluates the patient's sparse projected evidence against the stored trial-side disjunctive structure.

\paragraph{Projected CNF representation.}
For each entity \(e \in C \cup P\), let \(\CNF(e)\) denote its executable projected CNF constraint:
\[
\CNF(e)=\bigwedge_{i=1}^{m_e} d_i,
\qquad
d_i=\bigvee_{j=1}^{n_i} a_{ij},
\]
where each atomic constraint \(a_{ij}\) uses only canonical predicates and retained canonical qualifier fields. Boolean and numerical atoms are stored separately, but both participate in the same clause-level and CNF-level retrieval procedure.

\paragraph{Relational storage.}
As described in Section~\ref{sec:retrieval}, executable CNF constraints are stored as first-class relational objects using five tables:
\[
\ECNF(e,\cnf), \quad
\CNFD(\cnf,d), \quad
\DA(d,a), \quad
\AB(a,\pi,\mathsf{cmp},t), \quad
\AN(a,\pi,\mathsf{cmp},t).
\]
Here, \(\ECNF\) links an entity \(e\) to its executable CNF identifier \(\cnf\); \(\CNFD\) links a CNF to its disjunctive clauses; and \(\DA\) links a clause to its member atoms. Tables \(\AB\) and \(\AN\) store Boolean and numerical atoms, respectively, together with their canonical predicate \(\pi\), comparison field \(\mathsf{cmp}\), and retained qualifier fields \(t\). Entity type is determined by whether \(e\in C\) or \(e\in P\).

This representation allows executable constraints to be stored once and evaluated repeatedly by relational joins, rather than reconstructing logical structure at query time. It is also natural for trial-side alternatives: if a trial clause admits multiple acceptable support paths, these are stored directly as multiple atoms linked to the same clause through \(\DA\).

\paragraph{Objective-conditioned retrieval semantics.}
Retrieval is defined with respect to an objective \(\theta\), which determines which trial-side projected clauses are retrieval-relevant and which patient-side projected atoms may be used to support them. Thus, the SQL layer does not match all stored atoms indiscriminately. Instead, \(\theta\) induces an objective-specific admissibility relation over trial and patient atoms.

For a trial \(c\), let \(\mathcal D_\theta(c)\) denote the set of retrieval-relevant projected clauses in \(\CNF(c)\) under objective \(\theta\). Retrieval proceeds by identifying retrieval-relevant trial atoms and admissible patient atoms, forming compatible atom matches, lifting these matches to supported trial clauses, and retaining only those trial--patient pairs for which every clause in \(\mathcal D_\theta(c)\) is supported.

\paragraph{Patient-side evidence.}
At query time, the patient contributes a sparse set of observed projected atoms. Boolean evidence corresponds to canonical Boolean atoms stored in \(\AB\), while numerical evidence corresponds to canonical numerical atoms stored in \(\AN\). In both cases, the patient-side representation remains sparse: the retrieval layer uses only the projected evidence explicitly retained after patient-side conversion, rather than exhaustively generating all ontology-induced alternatives.

This design is important for both efficiency and control. Trial-side flexibility is represented through disjunction in the stored executable CNF, while the patient side contributes only the projected atoms supported by the patient record.

\paragraph{SQL-style execution.}
Given a retrieval objective \(\theta\), execution follows the projected CNF structure directly.

\textbf{Atom level.}
We first identify retrieval-relevant trial atoms and admissible patient atoms from the stored CNF objects. Concretely, trial and patient entities are linked to CNFs through \(\ECNF\), CNFs to clauses through \(\CNFD\), and clauses to atoms through \(\DA\); Boolean and numerical atom contents are read from \(\AB\) and \(\AN\). Trial and patient atoms are matched when the objective permits the corresponding support relation, their retained qualifier fields are compatible, and their comparison fields are mutually compatible. This yields a relation of compatible atom pairs.

\textbf{Clause level.}
Compatible atom pairs are lifted through \(\DA(d,a)\) to recover the trial clauses containing those atoms. A retrieval-relevant trial clause is treated as supported if at least one of its member atoms has a compatible patient-side match.

\textbf{CNF and entity level.}
Supported trial clauses are aggregated back through \(\CNFD(\cnf,d)\) and \(\ECNF(e,\cnf)\). A trial--patient pair is retained only if every retrieval-relevant trial clause of the trial is supported.

Equivalently, retrieval returns \((c,p)\) only if
\[
\forall d \in \mathcal D_\theta(c), \quad d \text{ is supported by } p.
\]
This semantics maps directly to SQL execution: compute supported trial clauses for each trial--patient pair, then retain only those pairs for which the number of supported retrieval-relevant trial clauses equals the total number of retrieval-relevant trial clauses.

The following pseudo-SQL sketch illustrates the structure:

\begin{lstlisting}[language=SQL]
WITH
-- Retrieval-relevant trial atoms under objective θ
ThetaTrialAtoms AS (
  SELECT
    ec.e   AS trial_id,
    cf.d   AS trial_clause_id,
    da.a   AS trial_atom_id,
    ab.pi  AS pi,
    ab.cmp AS cmp,
    ab.t   AS t
  FROM ECNF ec
  JOIN CNFD cf ON ec.cnf = cf.cnf
  JOIN DA   da ON cf.d = da.d
  JOIN AB   ab ON da.a = ab.a
  WHERE ec.e IN TrialEntities
    AND TrialAtomRelevant(ab.pi, θ)

  UNION ALL

  SELECT
    ec.e   AS trial_id,
    cf.d   AS trial_clause_id,
    da.a   AS trial_atom_id,
    an.pi  AS pi,
    an.cmp AS cmp,
    an.t   AS t
  FROM ECNF ec
  JOIN CNFD cf ON ec.cnf = cf.cnf
  JOIN DA   da ON cf.d = da.d
  JOIN AN   an ON da.a = an.a
  WHERE ec.e IN TrialEntities
    AND TrialAtomRelevant(an.pi, θ)
),

-- Admissible patient atoms under objective θ
ThetaPatientAtoms AS (
  SELECT
    ec.e   AS patient_id,
    da.a   AS patient_atom_id,
    ab.pi  AS pi,
    ab.cmp AS cmp,
    ab.t   AS t
  FROM ECNF ec
  JOIN CNFD cf ON ec.cnf = cf.cnf
  JOIN DA   da ON cf.d = da.d
  JOIN AB   ab ON da.a = ab.a
  WHERE ec.e IN PatientEntities
    AND PatientAtomAdmissible(ab.pi, θ)

  UNION ALL

  SELECT
    ec.e   AS patient_id,
    da.a   AS patient_atom_id,
    an.pi  AS pi,
    an.cmp AS cmp,
    an.t   AS t
  FROM ECNF ec
  JOIN CNFD cf ON ec.cnf = cf.cnf
  JOIN DA   da ON cf.d = da.d
  JOIN AN   an ON da.a = an.a
  WHERE ec.e IN PatientEntities
    AND PatientAtomAdmissible(an.pi, θ)
),

CompatibleAtomPairs AS (
  SELECT
    t.trial_id,
    p.patient_id,
    t.trial_clause_id,
    t.trial_atom_id,
    p.patient_atom_id
  FROM ThetaTrialAtoms t
  JOIN ThetaPatientAtoms p
    ON ObjectiveAllows(t.pi, p.pi, θ)
   AND QualifierCompatible(t.t, p.t, θ)
   AND ComparisonCompatible(t.cmp, p.cmp)
),

SupportedTrialClauses AS (
  SELECT DISTINCT
    trial_id,
    patient_id,
    trial_clause_id
  FROM CompatibleAtomPairs
),

AllThetaTrialClauses AS (
  SELECT DISTINCT
    trial_id,
    trial_clause_id
  FROM ThetaTrialAtoms
),

ReturnedPairs AS (
  SELECT
    a.trial_id,
    s.patient_id
  FROM AllThetaTrialClauses a
  LEFT JOIN SupportedTrialClauses s
    ON a.trial_id = s.trial_id
   AND a.trial_clause_id = s.trial_clause_id
  GROUP BY a.trial_id, s.patient_id
  HAVING COUNT(DISTINCT a.trial_clause_id)
       = COUNT(DISTINCT s.trial_clause_id)
)

SELECT * FROM ReturnedPairs;
\end{lstlisting}

Here, \texttt{TrialEntities} and \texttt{PatientEntities} denote the sets of trial and patient entity identifiers, respectively. The helper predicates \texttt{TrialAtomRelevant}, \texttt{PatientAtomAdmissible}, \texttt{ObjectiveAllows}, \texttt{QualifierCompatible}, and \texttt{ComparisonCompatible} stand in for the objective-specific admissibility test and the atom-level compatibility checks described in Section~\ref{sec:retrieval}.

\paragraph{Equivalent relational-algebra view.}
The same procedure can be expressed more abstractly in relational-algebra terms. First form the retrieval-relevant trial-atom relation and the admissible patient-atom relation by joining
\[
\ECNF \bowtie \CNFD \bowtie \DA \bowtie (\AB \cup \AN)
\]
and filtering by the objective \(\theta\). Then join these two relations using the atom-level compatibility predicate to obtain compatible atom pairs. Next, project to supported trial clauses by retaining \((c,p,d)\) whenever some atom in trial clause \(d\) has a compatible patient-side match. Finally, group by \((c,p)\) and retain only those pairs for which the number of supported trial clauses equals \(|\mathcal D_\theta(c)|\).

\paragraph{Why this preserves full recall.}
The full-recall guarantee comes from the conservative conversion step in Section~\ref{sec:retrieval}. Conversion removes or weakens parts of the original formulas only in a recall-safe direction, so every true match under the full SMT constraints remains satisfiable in the executable projected representation. The SQL retrieval layer then evaluates those executable constraints by checking support for every retrieval-relevant trial clause in \(\mathcal D_\theta(c)\). It may admit extra trial--patient pairs that are later ruled out by deeper reasoning, but it does not eliminate any pair that could satisfy the original constraints. This is exactly the full-recall property proved in Appendix~\ref{app:retriever_full_recall_proof}.

\paragraph{Role in the retrieval pipeline.}
This SQL layer is the executable realization of the projected CNF representation introduced in Section~\ref{sec:retrieval}. Offline, it materializes trial and patient constraints as reusable clause-based relational objects in \(\ECNF,\CNFD,\DA,\AB,\AN\). Online, it evaluates objective-conditioned compatibility between patient-side and trial-side projected atoms, lifts compatible atom matches to supported trial clauses, and returns only those trial--patient pairs whose retrieval-relevant trial clauses are all supported. The surviving candidates are then passed to the downstream reasoning stage for more exact relevance and eligibility evaluation.

Prompts for target schematization on trial side could be found in Appendix~\ref{app:dataset_prep/disease_categorization}, Appendix~\ref{app:dataset_prep/joint_specificity_filter}, Appendix~\ref{app:dataset_prep/positive_literal_finding_schematization}, Appendix~\ref{app:dataset_prep/positive_literal_procedure_schematization}, and Appendix~\ref{app:dataset_prep/positive_literal_substance_schematization}.

    \subsection{Why the Retriever Has Full Recall}
\label{app:retriever_full_recall_proof}

We show that every truly matching trial--patient pair is returned by \name. 
The argument has two parts. First, conversion from the full SMT constraints to the executable CNF constraints stored in the database is conservative, so any true match remains compatible after conversion. Second, the join-based retrieval test is itself a conservative relaxation of exact compatibility over these executable constraints.

\begin{theorem}[Full recall of \name]
Assume the atom-level compatibility test used in the database join is \emph{sound} in the following sense: whenever a trial atom and a patient atom can both hold under the same assignment, the join procedure marks them as compatible.

Then every true match under the full SMT formulation is returned by \name. Equivalently, if
\[
(c,p)\in \CTM(O,\TT,\PT,\OT),
\]
then
\[
(c,p)\in \name(\representation(O,\TT,\PT,\OT)).
\]
\end{theorem}

\begin{proof}
Let \((c,p)\in \CTM(O,\TT,\PT,\OT)\). By definition of the clinical trial matching task, there exists a satisfying interpretation under which the formalized match objective and the full trial and patient constraints all hold.

Let \(\TC(c)\) and \(\PC(p)\) denote the full SMT constraints derived from the free-text trial and patient descriptions, and let \(\theta\) be the formalized objective produced by \(\representation(O,\TT,\PT,\OT)\). Thus the full constraints are jointly satisfiable.

We show that \((c,p)\) is returned by \name in two steps.

\smallskip
\noindent\textbf{Step 1: Conservative conversion to executable CNF constraints.}
Let \(\CNF(c)\) and \(\CNF(p)\) be the executable quantifier-free CNF constraints stored in the database for trial \(c\) and patient \(p\). By construction, the conversion from the full SMT constraints to these executable constraints is conservative: any satisfying assignment of \(\TC(c)\) also satisfies \(\CNF(c)\), and any satisfying assignment of \(\PC(p)\) also satisfies \(\CNF(p)\). Hence the executable constraints for \(c\) and \(p\) are jointly satisfiable. Denote this exact compatibility by
\[
\mathsf{Compat}(c,p).
\]

\smallskip
\noindent\textbf{Step 2: Exact compatibility implies join-based compatibility.}
Assume \(\mathsf{Compat}(c,p)\). Then there exists an assignment under which both \(\CNF(c)\) and \(\CNF(p)\) are satisfied.

Since each CNF is a conjunction of disjunctive clauses, every clause on each side is satisfied under that assignment. Since each clause is a disjunction of atoms, each satisfied clause contains at least one atom that is true under that same assignment.

Consider any satisfied trial clause and any satisfied patient clause. Let \(a_t\) be a true atom from the trial clause and \(a_p\) a true atom from the patient clause under the common satisfying assignment. Because \(a_t\) and \(a_p\) are simultaneously satisfiable, the soundness assumption implies that the database join marks \((a_t,a_p)\) as compatible.

Therefore every satisfied clause is supported by at least one compatible joined atom pair. Since this holds for all required clauses, the pair \((c,p)\) passes the join-based retrieval test. Denote this relational compatibility by
\[
\mathsf{RelCompat}(c,p).
\]
Thus
\[
\mathsf{Compat}(c,p)\Rightarrow \mathsf{RelCompat}(c,p).
\]

\smallskip
Combining the two steps, we obtain
\[
(c,p)\in \CTM(O,\TT,\PT,\OT)
\;\Rightarrow\;
\mathsf{Compat}(c,p)
\;\Rightarrow\;
\mathsf{RelCompat}(c,p).
\]
By definition, \name returns exactly those trial--patient pairs that satisfy the join-based compatibility test over the executable representation derived from \(\representation(O,\TT,\PT,\OT)\). Hence
\[
(c,p)\in \name(\representation(O,\TT,\PT,\OT)).
\]
Therefore every true match is returned by \name.
\end{proof}
\newpage

\section{Others}\label{app:others_umbrella}
    \subsection{Limitations}\label{app:limitations}
\name{} has several limitations. First, our experiments are limited to two benchmark datasets: the SIGIR 2016 benchmark, containing 59 synthetic patient vignettes and 3{,}621 trials, and TREC 2022, containing 50 synthetic patient vignettes and a subset of 26{,}581 trials. We evaluated \name{} on an enriched subset of 2{,}658 TREC trials rather than the full corpus. Although both benchmarks are useful for controlled comparison, they do not fully capture the scale, heterogeneity, and documentation style of real clinical records. Broader validation on real-world EHR data and larger trial corpora remains necessary.

Second, our main evaluation relies on a GPT-5 judge with clinician validation on a sampled subset rather than exhaustive clinician review of all retrieved trial--patient pairs. We adopted this protocol because the original benchmark labels are incomplete and not tied to an explicit retrieval objective, while full clinician adjudication at our evaluation scale would be prohibitively expensive. Nevertheless, this means that our evaluation still depends in part on the quality of the GPT-5 judge.

Third, our formal representation is intentionally conservative. When clinically meaningful content cannot be safely canonicalized into the ontology-backed schema, we often preserve it as non-canonical information but do not fully project it into the symbolic retrieval layer. This choice helps avoid unsupported formal commitments, but it can also leave potentially useful information underutilized. Future work could explore richer handling of such content, for example through ontology extension or retrieval components based on learned representations.

Fourth, \name{} relies heavily on the availability and quality of a medical ontology, especially for canonicalization, subsumption, and retrieval-time matching. It is therefore unclear how well the approach would transfer to domains where such ontology support is weaker. One possible direction is to replace or supplement ontology neighborhoods with neighborhoods induced by learned embeddings, though this would likely trade away some interpretability and formal control.

Fifth, some parts of the system require deployment-specific tuning and policy choices, including the definition of retrieval objectives and the context used for salience judgments. This is both a strength and a limitation: it allows the system to be explicit, controllable, and adaptable to clinical goals, but it also means that deployment requires careful task definition rather than a single universally correct configuration.

Finally, \name{} incurs substantially more upfront semantic-parsing and formalization cost than embedding-based retrieval pipelines. This overhead is the price of constructing an interpretable symbolic representation. In return, the retrieval layer enjoys a formal full-recall guarantee with respect to the formalized constraints and their conservative projection, but the practical benefit of this guarantee still depends on the accuracy of upstream semantic parsing and knowledge grounding.
\subsection{Natural-Language Trial--Patient Matching}
\label{app:nl_match}

Our formal SMT-based \representation is designed to capture strict eligibility and relevance constraints accurately and to support high-recall, high-precision retrieval. However, some aspects of trial--patient matching remain difficult to formalize completely from free text, including nuanced clinical judgments, partially specified cohort definitions, and implicit relationships between patient characteristics and trial intent. To account for these cases, we perform the final matching step directly on the original free-text trial and patient descriptions using an LLM.

\paragraph{Two-stage matching procedure.}
The natural-language matching procedure is decomposed into two stages. First, the LLM is asked to identify all trial cohorts that are relevant to the patient under a predefined relevance definition. This stage considers only whether a cohort is clinically relevant to the patient and does not assess eligibility. The model returns a structured list of relevant cohorts together with a brief clinician-facing explanation.

In the second stage, the LLM evaluates eligibility separately for each relevant cohort returned by the first stage. This cohort-level design is important because many trials contain multiple arms, cohorts, or expansion groups with different disease targets or eligibility requirements. Evaluating eligibility only after relevance identification helps prevent conflating disease or intent mismatch with failure of entry criteria.

\paragraph{Eligibility instructions.}
The eligibility prompt provides detailed prescreening instructions that constrain how the LLM should reason over patient notes and trial text. In particular, it instructs the model to: (1) use clinical inference when appropriate, rather than relying only on explicitly stated facts; (2) project away operational, administrative, and logistics-only requirements that do not reflect the patient's substantive current or past clinical state; (3) reason only over the current and past patient state, rather than assuming future changes in disease status; and (4) apply explicit missingness rules when the patient note does not directly mention a required fact. These rules are intended to make the natural-language matching process more stable, more clinically plausible, and more consistent across trials.

\paragraph{Structured outputs.}
Both stages use structured outputs to support downstream parsing and review. The relevance stage returns the set of relevant cohorts and a short summary of the relevance decision. The eligibility stage returns a brief reasoning block followed by cohort-level eligibility decisions with concise justifications. This structure makes it possible to audit the final matching decision at the level of individual trial cohorts.

See Appendix \ref{app:patient-parsing/relevance-judge} and \ref{app:patient-parsing/eligibilty-judge} for prompt implementation details.

\subsection{An Example of the SMT Semantic Parser Output}\label{app:smt_example}

Here we show the trial input together with the generated SMT programs.  For readability, we omit variable-level metadata comments from the displayed SMT programs and retain only the declarations and logical constraints.

\paragraph{Selected trial: \texttt{NCT00362869}.}

\paragraph{Trial input.}
\begin{MyVerbatim}
Summary: Enteroaggregative E. coli (EAEC) is a bacterium that can cause diarrhea. The purposes of this study are to: determine how much EAEC is needed to cause diarrhea in a healthy person, determine if a genetic factor is important in causing diarrhea, and to see how the body's defenses control EAEC. Participants include 25 healthy adults, ages 18-40. Volunteers will be assigned to 1 of 4 dose levels in groups of 5 volunteers each. One volunteer in each group will receive a sodium bicarbonate placebo solution. Volunteers will be admitted to the University Clinical Research Unit for up to 8 days. Volunteers will receive therapy with levofloxacin to treat the infection either once they develop diarrhea or at Day 5 if they remain asymptomatic. Study procedures will include saliva, blood, and fecal sample collection. An optional study procedure will include an intestinal biopsy. Participants will be involved in study related procedures for up to 223 days.
Inclusion criteria: inclusion criteria: 

 Sign an Institutional Review Board-approved consent prior to any study-related activities. 

 Initiate screening 21 +/- 7 days prior to admission or enrollment. 

 Must accomplish all laboratory and diagnostic examinations at 21 +/- 7 days prior to admission or enrollment. 

 Be at least 18 years of age but not older than 40 years of age at the time of enrollment. 

 Be otherwise healthy with a stable address and telephone where the volunteer can be contacted. 

 Be able to read and write English. 

 Possess a social security number in order to receive compensation. 

 Female participants must have a negative serum pregnancy test at screening and a negative urine pregnancy test on the morning of the challenge and use effective birth control during the entire study period. Methods of effective birth control include: complete abstinence, the use of a licensed hormonal method, intrauterine device, barrier method plus spermicide, or having sexual relations exclusively with a vasectomized partner. Appropriate barrier methods include condoms, cervical sponge, and diaphragm. Females who are not of childbearing potential are defined as those who are physiologically incapable of becoming pregnant, including any female with tubal ligation or who is postmenopausal. For purposes of this study, postmenopausal status will be defined as absence of menses for at least 1 year. 

 Be seronegative for antibodies to dispersin. 

 Have normal laboratory screening values including a white blood cell (WBC) count, hemoglobin, hematocrit, platelets, blood urea nitrogen, glucose, creatinine, alanine aminotransferase (ALT), aspartate aminotransferase (AST), quantitative immunoglobulins, T cell subsets (CD4 and CD8), urinalysis. 

 Have normal chest x-ray and electrocardiogram. 

 Have negative serologies for HIV, hepatitis B virus (HBV), and hepatitis C virus (HCV), and a negative rapid plasma reagin (RPR). 

 Have a negative stool examination for pathogenic ova and pathogenic parasites, and bacterial enteropathogens (EAEC, Salmonella, Shigella, Campylobacter). 

 Have the -251 AA IL-8 genotype. 

 
Exclusion criteria: : 

 Has acute or chronic medical illness (i.e., renal or hepatic disease, hypertension, diabetes mellitus, coronary artery disease, malnutrition, obesity (body mass index >30 kg/m2), HIV, corticosteroid use, cancer or receiving chemotherapy, chronic debilitating illness, syphilis). 

 Has used antibiotics within 7 days of challenge. 

 Has used medications or drugs, including over-the-counter medications such as decongestants, antacids (calcium carbonate or aluminum-based antacids, H2 blockers), anti-diarrheal medications (such as bismuth subsalicylate or loperamide), antihistamines within 7 days of challenge. 

 Has a history of chronic gastrointestinal illness, intra-abdominal surgery, chronic functional dyspepsia, chronic gastroesophageal reflux, documented peptic ulcer disease, gastrointestinal hemorrhage, gallbladder disease, inflammatory bowel disease (Crohn's and ulcerative colitis), diverticulitis, irritable bowel syndrome or frequent diarrhea. 

 Has a history any of the following psychiatric illness(es): 

 Depression not controlled with current drug therapy or involving institutionalization 

 Schizophrenia or psychosis 

 Suicide attempt. 

 Has a history of or current alcohol or illicit drug abuse. 

 Is unable to remain as an inpatient in the University Clinical Research Unit for up to 8 days. 

 Has a known hypersensitivity to latex, heparin, opiates, antiemetics, benzodiazepines, lidocaine, magnesium citrate, or Fleet enema. 

 Has a known hypersensitivity to antibiotics that could be used to treat EAEC infection including fluoroquinolones, amoxicillin, cephalosporins or rifaximin. 

 Has serum antibodies to EAEC dispersin. 

 Recently traveled to a developing country (within 6 months). 

 Has household contacts who are less than 4 years of age or more than 80 years of age. 

 Has household contacts that are infirmed or immunocompromised due to any of the following reasons: 

 Corticosteroid therapy 

 HIV infection 

 Cancer chemotherapy 

 Other chronic debilitating diseases. 

 Works as health care personnel with direct patient care. 

 Works in a day care center for children or the elderly. 

 Is a food handler. 

 Has factors that, in the opinion of the investigator or research personnel, would interfere with the study objectives or increase the risk to the volunteer or his contacts. 

 Is currently participating in a clinical study or had receipt of an investigational drug in the past 30 days. 

 Is pregnant or has a risk of pregnancy or is lactating. 

 Has current excessive use of alcohol or drug dependence. 

 Has evidence of impaired immune function. 

 Has a new positive reaction to purified protein derivative (PPD) (volunteers who are known to be PPD positive that have a negative chest x-ray and have received isoniazid prophylaxis will be eligible). 

 Has a stool culture that demonstrates the presence of pathogenic ova, pathogenic parasites, or bacterial enteropathogens (EAEC, Salmonella, Shigella, and Campylobacter), or that is devoid of normal flora. 

 Has self-reported lactose or soy intolerance or allergy 

 Is a smoker and cannot stop smoking for the duration of the inpatient study. 

 Has abnormal lab results for screening beyond the normal range as defined below: 

 Hematology Hemoglobin: 13-15.0 gm/dL (Females) 14.5-17.0 gm/dL (Males) Hematocrit: 37-46 % (Females) 40-52 % (Males) Platelet count: 140,000-415,000 per mm3 WBC count: 4,000-10,500 per mm3 Neutrophils: 40-74 % or 1,800-7,800 per mm3 Lymphocytes: 14-26 % or 700-4,500 per mm3 Monocytes: 4-13 % or 100-1,000 per mm3 Eosinophils: 0-7 % or 0-400 per mm3 Basophils: 0-3 % or 0-200 per mm3 

 Chemistry BUN 5-25 mg/dL Creatinine 0.5-1.4 mg/dL Glucose (fasting) 69-99 mg/dL ALT 0-40 U/L AST 0-40 U/L 

 Immunology IgG: 596-1584 mg/dL IgA: 71-350 mg/dL IgM: 35-213 mg/dL CD4 T cells: 660-1500 cells/mcl CD8 T cells: 360-850 cells/mcl 

 Urinalysis Urine color: Yellow Turbidity: Clear pH: 5.0-8.0 Protein: Negative Sp. Gravity: 1.003-1.030 Glucose: Negative WBC: 0-2 Cells per HPF RBC: 0 Cells per HPF Bacteria: Rare Ketones: Negative The urinalysis will initially be evaluated for the quality of collection. If urinalysis is found to be poorly collected and demonstrates the presence of squamous epithelial cells and bacteria, results will not be used and a repeat urinalysis will be requested. In the case of menstruating women, the urinalysis collection will be postponed temporarily. A urinalysis may also be repeated once if traces of bile, protein, trace ketones, or Hb are identified. In the case of a properly collected urinalysis, the presence of leukocyte esterase glucose, or nitrates will exclude the participation of the subject. 

 Occult blood (Hemoccult) positive stools on admission to the CRU. 

 Develops gastrointestinal symptoms including nausea, vomiting, anorexia, abdominal pain, cramping, bloating, excessive gas or flatulence, diarrhea, constipation, urgency or tenesmus between the screening period and prior to challenge. 

 Develops a febrile illness during the period of time screening period and prior to challenge.
\end{MyVerbatim}

\paragraph{SMT program outputs.}
\textbf{NCT00362869\_inclusion\_program.smt2} (subcohort: \texttt{main}, side: \texttt{inclusion})
\begin{MyVerbatim}
;; ===================== Declarations for the criterion (REQ 0) =====================
(declare-const patient_has_signed_irb_approved_consent_prior_to_study_activities Bool) ;; "To be included, the patient must sign an Institutional Review Board-approved consent prior to any study-related activities."

;; ===================== Constraint Assertions (REQ 0) =====================
(assert
  (! patient_has_signed_irb_approved_consent_prior_to_study_activities
     :named REQ0_COMPONENT0_NOT_REQUIREMNET_OR_ALWAYS_SATISFIABLE_WITH_ACTION)) ;; "To be included, the patient must sign an Institutional Review Board-approved consent prior to any study-related activities."

;; ===================== Declarations for the criterion (REQ 1) =====================
(declare-const patient_has_undergone_screening_procedure_inthehistory Bool) ;; "Boolean procedure variable indicating whether the patient has undergone a screening procedure at any time in the past (without specifying the exact time window relative to admission/enrollment)." {"when_to_set_to_true":"Set to true if the patient has undergone a screening procedure at any time in the past, regardless of the exact timing relative to admission or enrollment.","when_to_set_to_false":"Set to false if the patient has not undergone a screening procedure at any time in the past.","when_to_set_to_null":"Set to null if it is unknown, not documented, or cannot be determined whether the patient has undergone a screening procedure in the past.","meaning":"Boolean indicating whether the patient has undergone a screening procedure at any time in the past."}
(declare-const patient_has_undergone_screening_procedure_inthehistory@@temporalcontext_within_14_to_28_days_before_admission_or_enrollment Bool) ;; "Boolean procedure variable indicating whether the patient has undergone a screening procedure in the past, specifically within the window of at least 14 days and at most 28 days before admission or enrollment." {"when_to_set_to_true":"Set to true if the patient has undergone a screening procedure at least 14 days and at most 28 days before admission or enrollment.","when_to_set_to_false":"Set to false if the patient has not undergone a screening procedure within this window.","when_to_set_to_null":"Set to null if it is unknown, not documented, or cannot be determined whether the patient has undergone a screening procedure within this window.","meaning":"Boolean indicating whether the patient has undergone a screening procedure at least 14 days and at most 28 days before admission or enrollment."}
(declare-const screening_to_admission_or_enrollment_interval_value_in_days Real) ;; "Numeric variable representing the number of days between the date the patient initiated screening and the date of admission or enrollment." {"when_to_set_to_value":"Set to the number of days between the date the patient initiated screening and the date of admission or enrollment.","when_to_set_to_null":"Set to null if either the date of screening initiation or the date of admission/enrollment is unknown, not documented, or cannot be determined.","meaning":"Numeric value representing the interval in days between screening initiation and admission or enrollment."}

;; ===================== Auxiliary Assertions (REQ 1) =====================
;; The qualifier variable is true iff the interval is at least 14 and at most 28 days
(assert
  (! (= patient_has_undergone_screening_procedure_inthehistory@@temporalcontext_within_14_to_28_days_before_admission_or_enrollment
        (and (>= screening_to_admission_or_enrollment_interval_value_in_days 14.0)
             (<= screening_to_admission_or_enrollment_interval_value_in_days 28.0)
             patient_has_undergone_screening_procedure_inthehistory))
     :named REQ1_AUXILIARY0)) ;; "Boolean procedure variable indicating whether the patient has undergone a screening procedure in the past, specifically within the window of at least 14 days and at most 28 days before admission or enrollment."

;; ===================== Constraint Assertions (REQ 1) =====================
;; Component 0: at least 14 days prior to admission or enrollment
(assert
  (! (>= screening_to_admission_or_enrollment_interval_value_in_days 14.0)
     :named REQ1_COMPONENT0_NOT_REQUIREMNET_OR_ALWAYS_SATISFIABLE_WITH_ACTION)) ;; "To be included, the patient must initiate screening at least 14 days prior to admission or enrollment."

;; Component 1: at most 28 days prior to admission or enrollment
(assert
  (! (<= screening_to_admission_or_enrollment_interval_value_in_days 28.0)
     :named REQ1_COMPONENT1_NOT_REQUIREMNET_OR_ALWAYS_SATISFIABLE_WITH_ACTION)) ;; "To be included, the patient must initiate screening at most 28 days prior to admission or enrollment."

;; ===================== Declarations for the criterion (REQ 2) =====================
(declare-const patient_has_undergone_laboratory_test_inthehistory Bool) ;; "laboratory examinations"
(declare-const patient_has_undergone_laboratory_test_inthehistory@@temporalcontext_within_14_to_28_days_before_admission_or_enrollment Bool) ;; "laboratory examinations at least 14 days and at most 28 days prior to admission or enrollment"
(declare-const patient_has_undergone_diagnostic_assessment_inthehistory Bool) ;; "diagnostic examinations"
(declare-const patient_has_undergone_diagnostic_assessment_inthehistory@@temporalcontext_within_14_to_28_days_before_admission_or_enrollment Bool) ;; "diagnostic examinations at least 14 days and at most 28 days prior to admission or enrollment"

;; ===================== Constraint Assertions (REQ 2) =====================
;; To be included, the patient must accomplish all laboratory examinations at least 14 days prior to admission or enrollment.
(assert
  (! patient_has_undergone_laboratory_test_inthehistory@@temporalcontext_within_14_to_28_days_before_admission_or_enrollment
     :named REQ2_COMPONENT0_NOT_REQUIREMNET_OR_ALWAYS_SATISFIABLE_WITH_ACTION)) ;; "To be included, the patient must accomplish all laboratory examinations at least 14 days prior to admission or enrollment."

;; To be included, the patient must accomplish all laboratory examinations at most 28 days prior to admission or enrollment.
(assert
  (! patient_has_undergone_laboratory_test_inthehistory@@temporalcontext_within_14_to_28_days_before_admission_or_enrollment
     :named REQ2_COMPONENT1_NOT_REQUIREMNET_OR_ALWAYS_SATISFIABLE_WITH_ACTION)) ;; "To be included, the patient must accomplish all laboratory examinations at most 28 days prior to admission or enrollment."

;; To be included, the patient must accomplish all diagnostic examinations at least 14 days prior to admission or enrollment.
(assert
  (! patient_has_undergone_diagnostic_assessment_inthehistory@@temporalcontext_within_14_to_28_days_before_admission_or_enrollment
     :named REQ2_COMPONENT2_NOT_REQUIREMNET_OR_ALWAYS_SATISFIABLE_WITH_ACTION)) ;; "To be included, the patient must accomplish all diagnostic examinations at least 14 days prior to admission or enrollment."

;; To be included, the patient must accomplish all diagnostic examinations at most 28 days prior to admission or enrollment.
(assert
  (! patient_has_undergone_diagnostic_assessment_inthehistory@@temporalcontext_within_14_to_28_days_before_admission_or_enrollment
     :named REQ2_COMPONENT3_NOT_REQUIREMNET_OR_ALWAYS_SATISFIABLE_WITH_ACTION)) ;; "To be included, the patient must accomplish all diagnostic examinations at most 28 days prior to admission or enrollment."

;; ===================== Declarations for the criterion (REQ 3) =====================
(declare-const patient_age_value_recorded_now_in_years Real) ;; "at least 18 years of age AND at most 40 years of age at the time of enrollment"

;; ===================== Constraint Assertions (REQ 3) =====================
;; The patient must be at least 18 years of age at the time of enrollment.
(assert
  (! (>= patient_age_value_recorded_now_in_years 18.0)
     :named REQ3_COMPONENT0_PRESCREEN_NOTES_MUST_COMPLETELY_SUFFICE)) ;; "To be included, the patient must be at least 18 years of age at the time of enrollment."

;; The patient must be at most 40 years of age at the time of enrollment.
(assert
  (! (<= patient_age_value_recorded_now_in_years 40.0)
     :named REQ3_COMPONENT1_PRESCREEN_NOTES_MUST_COMPLETELY_SUFFICE)) ;; "To be included, the patient must be at most 40 years of age at the time of enrollment."

;; ===================== Declarations for the criterion (REQ 4) =====================
(declare-const patient_has_finding_of_fit_and_well_now Bool) ;; "To be included, the patient must be otherwise healthy."
(declare-const patient_has_stable_address_now Bool) ;; "To be included, the patient must have a stable address."
(declare-const patient_has_stable_telephone_now Bool) ;; "To be included, the patient must have a stable telephone where the patient can be contacted."

;; ===================== Constraint Assertions (REQ 4) =====================
;; Component 0: The patient must be otherwise healthy.
(assert
  (! patient_has_finding_of_fit_and_well_now
     :named REQ4_COMPONENT0_OTHER_REQUIREMENTS)) ;; "To be included, the patient must be otherwise healthy."

;; Component 1: The patient must have a stable address.
(assert
  (! patient_has_stable_address_now
     :named REQ4_COMPONENT1_OTHER_REQUIREMENTS)) ;; "To be included, the patient must have a stable address."

;; Component 2: The patient must have a stable telephone where the patient can be contacted.
(assert
  (! patient_has_stable_telephone_now
     :named REQ4_COMPONENT2_OTHER_REQUIREMENTS)) ;; "To be included, the patient must have a stable telephone where the patient can be contacted."

;; ===================== Declarations for the criterion (REQ 5) =====================
(declare-const patient_is_able_to_read_english_now Bool) ;; "To be included, the patient must be able to read English."
(declare-const patient_is_able_to_write_english_now Bool) ;; "To be included, the patient must be able to write English."

;; ===================== Constraint Assertions (REQ 5) =====================
;; Component 0: Patient must be able to read English
(assert
  (! patient_is_able_to_read_english_now
     :named REQ5_COMPONENT0_PRESCREEN_NOTES_MUST_COMPLETELY_SUFFICE)) ;; "To be included, the patient must be able to read English."

;; Component 1: Patient must be able to write English
(assert
  (! patient_is_able_to_write_english_now
     :named REQ5_COMPONENT1_PRESCREEN_NOTES_MUST_COMPLETELY_SUFFICE)) ;; "To be included, the patient must be able to write English."

;; ===================== Declarations for the criterion (REQ 6) =====================
(declare-const patient_has_social_security_number Bool) ;; "To be included, the patient must possess a social security number in order to receive compensation."

;; ===================== Constraint Assertions (REQ 6) =====================
(assert
  (! patient_has_social_security_number
     :named REQ6_COMPONENT0_NOT_REQUIREMNET_OR_ALWAYS_SATISFIABLE_WITH_ACTION)) ;; "To be included, the patient must possess a social security number in order to receive compensation."

;; ===================== Declarations for the criterion (REQ 7, Component 5) =====================
(declare-const patient_is_postmenopausal_now Bool) ;; "postmenopausal status is defined as absence of menses for at least 1 year"
(declare-const patient_has_absence_of_menses_duration_value_recorded_now_in_years Real) ;; "absence of menses for at least 1 year"
(declare-const patient_has_absence_of_menses_duration_value_recorded_now_in_years@@duration_at_least_1_year Bool) ;; "absence of menses for at least 1 year"

;; ===================== Auxiliary Assertions (REQ 7) =====================
;; Definition: duration_at_least_1_year is true iff absence of menses duration >= 1 year
(assert
  (! (= patient_has_absence_of_menses_duration_value_recorded_now_in_years@@duration_at_least_1_year
        (>= patient_has_absence_of_menses_duration_value_recorded_now_in_years 1.0))
     :named REQ7_AUXILIARY0)) ;; "absence of menses for at least 1 year"

;; Definition: postmenopausal status is equivalent to absence of menses for at least 1 year
(assert
  (! (= patient_is_postmenopausal_now
        patient_has_absence_of_menses_duration_value_recorded_now_in_years@@duration_at_least_1_year)
     :named REQ7_AUXILIARY1)) ;; "postmenopausal status is defined as absence of menses for at least 1 year"

;; ===================== Constraint Assertions (REQ 7, Component 5) =====================
(assert
  (! patient_is_postmenopausal_now
     :named REQ7_COMPONENT5_PRESCREEN_NOTES_MUST_COMPLETELY_SUFFICE)) ;; "To be included, for purposes of this study, postmenopausal status is defined as absence of menses for at least 1 year."

;; ===================== Declarations for the criterion (REQ 8) =====================
(declare-const patient_is_exposed_to_antibody_now@@specific_to_dispersin Bool) ;; "To be included, the patient must be seronegative for antibodies to dispersin."
(declare-const patient_is_seronegative_for_antibodies_to_dispersin_now Bool) ;; "To be included, the patient must be seronegative for antibodies to dispersin."

;; ===================== Auxiliary Assertions (REQ 8) =====================
;; Definition: seronegative for antibodies to dispersin means NOT exposed to antibodies to dispersin
(assert
  (! (= patient_is_seronegative_for_antibodies_to_dispersin_now
        (not patient_is_exposed_to_antibody_now@@specific_to_dispersin))
     :named REQ8_AUXILIARY0)) ;; "To be included, the patient must be seronegative for antibodies to dispersin."

;; ===================== Constraint Assertions (REQ 8) =====================
(assert
  (! patient_is_seronegative_for_antibodies_to_dispersin_now
     :named REQ8_COMPONENT0_OTHER_REQUIREMENTS)) ;; "To be included, the patient must be seronegative for antibodies to dispersin."

;; ===================== Declarations for the criterion (REQ 9) =====================

(declare-const patient_has_undergone_white_blood_cell_count_now_outcome_is_normal Bool) ;; "including a white blood cell count"

(declare-const patient_has_finding_of_hemoglobin_finding_now@@is_normal Bool) ;; "including a hemoglobin value"

(declare-const patient_has_undergone_hematocrit_determination_now_outcome_is_normal Bool) ;; "including a hematocrit value"

(declare-const patient_has_undergone_platelet_count_now_outcome_is_normal Bool) ;; "including a platelet count"

(declare-const patient_has_undergone_blood_urea_nitrogen_measurement_now_outcome_is_normal Bool) ;; "including a blood urea nitrogen value"

(declare-const patient_has_undergone_glucose_measurement_now_outcome_is_normal Bool) ;; "including a glucose value"

(declare-const patient_has_finding_of_creatinine_level_finding_now@@is_normal Bool) ;; "including a creatinine value"

(declare-const patient_has_undergone_alanine_aminotransferase_measurement_now_outcome_is_normal Bool) ;; "including an alanine aminotransferase value"

(declare-const patient_has_undergone_aspartate_aminotransferase_measurement_now_outcome_is_normal Bool) ;; "including an aspartate aminotransferase value"

(declare-const patient_has_undergone_immunoglobulin_measurement_now_outcome_is_normal Bool) ;; "including a quantitative immunoglobulin value"

(declare-const patient_has_finding_of_t_cell_subsets_now@@is_normal Bool) ;; "including a T cell subset value (CD4 T cell subset AND CD8 T cell subset)"

(declare-const patient_has_finding_of_cd4_t_cell_subset_now@@is_normal Bool) ;; "CD4 T cell subset"

(declare-const patient_has_finding_of_cd8_t_cell_subset_now@@is_normal Bool) ;; "CD8 T cell subset"

(declare-const patient_has_undergone_urinalysis_now_outcome_is_normal Bool) ;; "including a urinalysis"

;; ===================== Auxiliary Assertions (REQ 9) =====================

;; The umbrella variable for T cell subset is normal if and only if both CD4 and CD8 T cell subset findings are normal.
(assert
  (! (= patient_has_finding_of_t_cell_subsets_now@@is_normal
        (and patient_has_finding_of_cd4_t_cell_subset_now@@is_normal
             patient_has_finding_of_cd8_t_cell_subset_now@@is_normal))
     :named REQ9_AUXILIARY0)) ;; "T cell subset value (CD4 T cell subset AND CD8 T cell subset)"

;; ===================== Constraint Assertions (REQ 9) =====================

(assert
  (! patient_has_undergone_white_blood_cell_count_now_outcome_is_normal
     :named REQ9_COMPONENT0_OTHER_REQUIREMENTS)) ;; "including a white blood cell count"

(assert
  (! patient_has_finding_of_hemoglobin_finding_now@@is_normal
     :named REQ9_COMPONENT1_OTHER_REQUIREMENTS)) ;; "including a hemoglobin value"

(assert
  (! patient_has_undergone_hematocrit_determination_now_outcome_is_normal
     :named REQ9_COMPONENT2_OTHER_REQUIREMENTS)) ;; "including a hematocrit value"

(assert
  (! patient_has_undergone_platelet_count_now_outcome_is_normal
     :named REQ9_COMPONENT3_OTHER_REQUIREMENTS)) ;; "including a platelet count"

(assert
  (! patient_has_undergone_blood_urea_nitrogen_measurement_now_outcome_is_normal
     :named REQ9_COMPONENT4_OTHER_REQUIREMENTS)) ;; "including a blood urea nitrogen value"

(assert
  (! patient_has_undergone_glucose_measurement_now_outcome_is_normal
     :named REQ9_COMPONENT5_OTHER_REQUIREMENTS)) ;; "including a glucose value"

(assert
  (! patient_has_finding_of_creatinine_level_finding_now@@is_normal
     :named REQ9_COMPONENT6_OTHER_REQUIREMENTS)) ;; "including a creatinine value"

(assert
  (! patient_has_undergone_alanine_aminotransferase_measurement_now_outcome_is_normal
     :named REQ9_COMPONENT7_OTHER_REQUIREMENTS)) ;; "including an alanine aminotransferase value"

(assert
  (! patient_has_undergone_aspartate_aminotransferase_measurement_now_outcome_is_normal
     :named REQ9_COMPONENT8_OTHER_REQUIREMENTS)) ;; "including an aspartate aminotransferase value"

(assert
  (! patient_has_undergone_immunoglobulin_measurement_now_outcome_is_normal
     :named REQ9_COMPONENT9_OTHER_REQUIREMENTS)) ;; "including a quantitative immunoglobulin value"

(assert
  (! patient_has_finding_of_t_cell_subsets_now@@is_normal
     :named REQ9_COMPONENT10_OTHER_REQUIREMENTS)) ;; "including a T cell subset value (CD4 T cell subset AND CD8 T cell subset)"

(assert
  (! patient_has_undergone_urinalysis_now_outcome_is_normal
     :named REQ9_COMPONENT11_OTHER_REQUIREMENTS)) ;; "including a urinalysis"

;; ===================== Declarations for the criterion (REQ 10) =====================
(declare-const patient_has_finding_of_tomography_chest_normal_now Bool) ;; "have a normal chest x-ray"
(declare-const patient_has_finding_of_ecg_normal_now Bool) ;; "have a normal electrocardiogram"

;; ===================== Constraint Assertions (REQ 10) =====================
;; Component 0: To be included, the patient must have a normal chest x-ray.
(assert
  (! patient_has_finding_of_tomography_chest_normal_now
     :named REQ10_COMPONENT0_OTHER_REQUIREMENTS)) ;; "To be included, the patient must have a normal chest x-ray."

;; Component 1: To be included, the patient must have a normal electrocardiogram.
(assert
  (! patient_has_finding_of_ecg_normal_now
     :named REQ10_COMPONENT1_OTHER_REQUIREMENTS)) ;; "To be included, the patient must have a normal electrocardiogram."

;; ===================== Declarations for the inclusion criterion (REQ 11) =====================
(declare-const patient_has_finding_of_hiv_negative_now Bool)  ;; "have negative serologies for human immunodeficiency virus infection"
(declare-const patient_has_finding_of_type_b_viral_hepatitis_now@@determined_by_negative_serologies Bool)  ;; "have negative serologies for hepatitis B virus infection"
(declare-const patient_has_finding_of_hepatitis_c_antibody_test_negative_now Bool)  ;; "have negative serologies for hepatitis C virus infection"
(declare-const patient_has_undergone_rapid_plasma_reagin_test_now_outcome_is_negative Bool)  ;; "have a negative rapid plasma reagin test"

;; ===================== Constraint Assertions (REQ 11) =====================
;; Component 0: Negative serologies for HIV infection
(assert
  (! patient_has_finding_of_hiv_negative_now
     :named REQ11_COMPONENT0_OTHER_REQUIREMENTS)) ;; "have negative serologies for human immunodeficiency virus infection"

;; Component 1: Negative serologies for hepatitis B virus infection
(assert
  (! patient_has_finding_of_type_b_viral_hepatitis_now@@determined_by_negative_serologies
     :named REQ11_COMPONENT1_OTHER_REQUIREMENTS)) ;; "have negative serologies for hepatitis B virus infection"

;; Component 2: Negative serologies for hepatitis C virus infection
(assert
  (! patient_has_finding_of_hepatitis_c_antibody_test_negative_now
     :named REQ11_COMPONENT2_OTHER_REQUIREMENTS)) ;; "have negative serologies for hepatitis C virus infection"

;; Component 3: Negative rapid plasma reagin test
(assert
  (! patient_has_undergone_rapid_plasma_reagin_test_now_outcome_is_negative
     :named REQ11_COMPONENT3_OTHER_REQUIREMENTS)) ;; "have a negative rapid plasma reagin test"

;; ===================== Declarations for the inclusion criterion (REQ 12) =====================
(declare-const patient_has_undergone_evaluation_of_stool_specimen_now_outcome_is_negative Bool) ;; "To be included, the patient must have a negative stool examination for pathogenic ova, pathogenic parasites, and bacterial enteropathogens."
(declare-const patient_has_undergone_evaluation_of_stool_specimen_now_outcome_is_negative@@for_pathogenic_ova Bool) ;; "To be included, the patient must have a negative stool examination for pathogenic ova."
(declare-const patient_has_undergone_evaluation_of_stool_specimen_now_outcome_is_negative@@for_pathogenic_parasites Bool) ;; "To be included, the patient must have a negative stool examination for pathogenic parasites."
(declare-const patient_has_undergone_evaluation_of_stool_specimen_now_outcome_is_negative@@for_bacterial_enteropathogens_with_exhaustive_subcategories Bool) ;; "To be included, the patient must have a negative stool examination for bacterial enteropathogens (Enteroaggregative Escherichia coli, Salmonella species, Shigella species, Campylobacter species)."

;; ===================== Auxiliary Assertions (REQ 12) =====================
;; Qualifier variables imply corresponding stem variable
(assert
  (! (=> patient_has_undergone_evaluation_of_stool_specimen_now_outcome_is_negative@@for_pathogenic_ova
         patient_has_undergone_evaluation_of_stool_specimen_now_outcome_is_negative)
     :named REQ12_AUXILIARY0)) ;; "To be included, the patient must have a negative stool examination for pathogenic ova."

(assert
  (! (=> patient_has_undergone_evaluation_of_stool_specimen_now_outcome_is_negative@@for_pathogenic_parasites
         patient_has_undergone_evaluation_of_stool_specimen_now_outcome_is_negative)
     :named REQ12_AUXILIARY1)) ;; "To be included, the patient must have a negative stool examination for pathogenic parasites."

(assert
  (! (=> patient_has_undergone_evaluation_of_stool_specimen_now_outcome_is_negative@@for_bacterial_enteropathogens_with_exhaustive_subcategories
         patient_has_undergone_evaluation_of_stool_specimen_now_outcome_is_negative)
     :named REQ12_AUXILIARY2)) ;; "To be included, the patient must have a negative stool examination for bacterial enteropathogens (Enteroaggregative Escherichia coli, Salmonella species, Shigella species, Campylobacter species)."

;; ===================== Constraint Assertions (REQ 12) =====================
;; Component 0: negative stool exam for pathogenic ova
(assert
  (! patient_has_undergone_evaluation_of_stool_specimen_now_outcome_is_negative@@for_pathogenic_ova
     :named REQ12_COMPONENT0_OTHER_REQUIREMENTS)) ;; "To be included, the patient must have a negative stool examination for pathogenic ova."

;; Component 1: negative stool exam for pathogenic parasites
(assert
  (! patient_has_undergone_evaluation_of_stool_specimen_now_outcome_is_negative@@for_pathogenic_parasites
     :named REQ12_COMPONENT1_OTHER_REQUIREMENTS)) ;; "To be included, the patient must have a negative stool examination for pathogenic parasites."

;; Component 2: negative stool exam for bacterial enteropathogens (exhaustive: Enteroaggregative E. coli, Salmonella, Shigella, Campylobacter)
(assert
  (! patient_has_undergone_evaluation_of_stool_specimen_now_outcome_is_negative@@for_bacterial_enteropathogens_with_exhaustive_subcategories
     :named REQ12_COMPONENT2_OTHER_REQUIREMENTS)) ;; "To be included, the patient must have a negative stool examination for bacterial enteropathogens (Enteroaggregative Escherichia coli, Salmonella species, Shigella species, Campylobacter species)."

;; ===================== Declarations for the criterion (REQ 13) =====================
(declare-const patient_has_interleukin_8_genotype_minus_251_aa Bool) ;; "To be included, the patient must have the -251 AA interleukin-8 genotype."
(declare-const patient_is_exposed_to_interleukin_8_now Bool) ;; variable set includes exposure to interleukin-8
(declare-const patient_is_exposed_to_interleukin_8_now@@genotype_minus_251_aa Bool) ;; variable set includes exposure to interleukin-8 with genotype -251 AA

;; ===================== Auxiliary Assertions (REQ 13) =====================
;; If the patient is exposed to interleukin-8 with genotype -251 AA, then the patient must have the -251 AA genotype of interleukin-8.
(assert
  (! (=> patient_is_exposed_to_interleukin_8_now@@genotype_minus_251_aa
         patient_has_interleukin_8_genotype_minus_251_aa)
     :named REQ13_AUXILIARY0)) ;; "To be included, the patient must have the -251 AA interleukin-8 genotype."

;; ===================== Constraint Assertions (REQ 13) =====================
;; The patient must have the -251 AA interleukin-8 genotype.
(assert
  (! patient_has_interleukin_8_genotype_minus_251_aa
     :named REQ13_COMPONENT0_OTHER_REQUIREMENTS)) ;; "To be included, the patient must have the -251 AA interleukin-8 genotype."
\end{MyVerbatim}

\textbf{NCT00362869\_exclusion\_program.smt2} (subcohort: \texttt{main}, side: \texttt{exclusion})
\begin{MyVerbatim}
;; ===================== Declarations for the exclusion criterion (REQ 0) =====================
(declare-const patient_has_finding_of_acute_disease_now Bool) ;; "the patient has an acute medical illness"

(declare-const patient_has_finding_of_chronic_disease_now Bool) ;; "the patient has a chronic medical illness"

(declare-const patient_has_finding_of_kidney_disease_now Bool) ;; "renal disease"

(declare-const patient_has_finding_of_disease_of_liver_now Bool) ;; "hepatic disease"

(declare-const patient_has_finding_of_hypertensive_disorder_now Bool) ;; "hypertension"

(declare-const patient_has_finding_of_diabetes_mellitus_now Bool) ;; "diabetes mellitus"

(declare-const patient_has_finding_of_coronary_arteriosclerosis_now Bool) ;; "coronary artery disease"

(declare-const patient_has_finding_of_malnutrition_now Bool) ;; "malnutrition"

(declare-const patient_has_finding_of_obesity_now Bool) ;; "obesity (body mass index > 30 kilograms per square meter)"

(declare-const patient_has_finding_of_human_immunodeficiency_virus_infection_now Bool) ;; "human immunodeficiency virus infection"

(declare-const patient_is_taking_corticosteroid_and_corticosteroid_derivative_containing_product_now Bool) ;; "corticosteroid use"

(declare-const patient_has_finding_of_malignant_neoplastic_disease_now Bool) ;; "cancer"

(declare-const patient_is_undergoing_administration_of_antineoplastic_agent_now Bool) ;; "is receiving chemotherapy"

(declare-const patient_has_finding_of_chronic_debilitating_illness_now Bool) ;; "chronic debilitating illness"

(declare-const patient_has_finding_of_syphilis_now Bool) ;; "syphilis"

;; ===================== Auxiliary Assertions (REQ 0) =====================
;; Non-exhaustive examples imply umbrella term for chronic medical illness
(assert
(! (=> (or patient_has_finding_of_kidney_disease_now
           patient_has_finding_of_disease_of_liver_now
           patient_has_finding_of_hypertensive_disorder_now
           patient_has_finding_of_diabetes_mellitus_now
           patient_has_finding_of_coronary_arteriosclerosis_now
           patient_has_finding_of_malnutrition_now
           patient_has_finding_of_obesity_now
           patient_has_finding_of_human_immunodeficiency_virus_infection_now
           patient_is_taking_corticosteroid_and_corticosteroid_derivative_containing_product_now
           patient_has_finding_of_malignant_neoplastic_disease_now
           patient_is_undergoing_administration_of_antineoplastic_agent_now
           patient_has_finding_of_chronic_debilitating_illness_now
           patient_has_finding_of_syphilis_now)
    patient_has_finding_of_chronic_disease_now)
:named REQ0_AUXILIARY0)) ;; "chronic medical illness with non-exhaustive examples ((renal disease) OR (hepatic disease) OR (hypertension) OR (diabetes mellitus) OR (coronary artery disease) OR (malnutrition) OR (obesity (body mass index > 30 kilograms per square meter)) OR (human immunodeficiency virus infection) OR (corticosteroid use) OR (cancer) OR (is receiving chemotherapy) OR (chronic debilitating illness) OR (syphilis))"

;; ===================== Constraint Assertions (REQ 0) =====================
(assert
(! (not patient_has_finding_of_acute_disease_now)
:named REQ0_COMPONENT0_OTHER_REQUIREMENTS)) ;; "The patient is excluded if the patient has an acute medical illness."

(assert
(! (not patient_has_finding_of_chronic_disease_now)
:named REQ0_COMPONENT1_OTHER_REQUIREMENTS)) ;; "The patient is excluded if the patient has a chronic medical illness with non-exhaustive examples ((renal disease) OR (hepatic disease) OR (hypertension) OR (diabetes mellitus) OR (coronary artery disease) OR (malnutrition) OR (obesity (body mass index > 30 kilograms per square meter)) OR (human immunodeficiency virus infection) OR (corticosteroid use) OR (cancer) OR (is receiving chemotherapy) OR (chronic debilitating illness) OR (syphilis))."

;; ===================== Declarations for the exclusion criterion (REQ 1) =====================
(declare-const patient_has_undergone_antibiotic_therapy_inthepast7days Bool) ;; "antibiotics within 7 days prior to challenge"

(declare-const patient_has_undergone_antibiotic_therapy_inthepast7days@@temporalcontext_within7days_before_challenge Bool) ;; "antibiotics within 7 days prior to challenge"

;; ===================== Auxiliary Assertions (REQ 1) =====================
;; Qualifier variable implies corresponding stem variable
(assert
(! (=> patient_has_undergone_antibiotic_therapy_inthepast7days@@temporalcontext_within7days_before_challenge
      patient_has_undergone_antibiotic_therapy_inthepast7days)
:named REQ1_AUXILIARY0)) ;; "antibiotics within 7 days prior to challenge"

;; ===================== Constraint Assertions (REQ 1) =====================
(assert
(! (not patient_has_undergone_antibiotic_therapy_inthepast7days@@temporalcontext_within7days_before_challenge)
:named REQ1_COMPONENT0_OTHER_REQUIREMENTS)) ;; "The patient is excluded if the patient has used antibiotics within 7 days prior to challenge."

;; ===================== Declarations for the exclusion criterion (REQ 2) =====================
(declare-const patient_is_exposed_to_drug_or_medicament_inthepast7days Bool) ;; "medications or drugs within 7 days prior to challenge"

(declare-const patient_is_exposed_to_drug_or_medicament_inthepast7days@@is_over_the_counter_medication Bool) ;; "over-the-counter medications such as decongestants"

(declare-const patient_is_exposed_to_decongestant_inthepast7days Bool) ;; "decongestants"

(declare-const patient_has_taken_antacid_inthepast7days Bool) ;; "antacids"

(declare-const patient_has_taken_calcium_carbonate_containing_product_inthepast7days Bool) ;; "calcium carbonate antacids"

(declare-const patient_has_taken_aluminum_containing_product_inthepast7days Bool) ;; "aluminum-based antacids"

(declare-const patient_is_taking_histamine_h2_receptor_antagonist_containing_product_inthepast7days Bool) ;; "H2 blockers"

(declare-const patient_is_exposed_to_drug_or_medicament_inthepast7days@@is_anti_diarrheal_medication Bool) ;; "anti-diarrheal medications"

(declare-const patient_is_exposed_to_bismuth_subsalicylate_inthepast7days Bool) ;; "bismuth subsalicylate"

(declare-const patient_is_exposed_to_loperamide_inthepast7days Bool) ;; "loperamide"

(declare-const patient_is_exposed_to_histamine_receptor_antagonist_inthepast7days Bool) ;; "antihistamines"

;; ===================== Auxiliary Assertions (REQ 2) =====================
;; Non-exhaustive examples: decongestants imply OTC medication exposure
(assert
(! (=> patient_is_exposed_to_decongestant_inthepast7days
     patient_is_exposed_to_drug_or_medicament_inthepast7days@@is_over_the_counter_medication)
:named REQ2_AUXILIARY0)) ;; "over-the-counter medications such as decongestants"

;; Qualifier variable implies stem variable
(assert
(! (=> patient_is_exposed_to_drug_or_medicament_inthepast7days@@is_over_the_counter_medication
     patient_is_exposed_to_drug_or_medicament_inthepast7days)
:named REQ2_AUXILIARY1)) ;; "over-the-counter medications"

;; Non-exhaustive examples: calcium carbonate, aluminum, H2 blockers imply antacid exposure
(assert
(! (=> (or patient_has_taken_calcium_carbonate_containing_product_inthepast7days
          patient_has_taken_aluminum_containing_product_inthepast7days
          patient_is_taking_histamine_h2_receptor_antagonist_containing_product_inthepast7days)
     patient_has_taken_antacid_inthepast7days)
:named REQ2_AUXILIARY2)) ;; "antacids with non-exhaustive examples (calcium carbonate antacids OR aluminum-based antacids OR H2 blockers)"

;; Non-exhaustive examples: bismuth subsalicylate, loperamide imply anti-diarrheal medication exposure
(assert
(! (=> (or patient_is_exposed_to_bismuth_subsalicylate_inthepast7days
          patient_is_exposed_to_loperamide_inthepast7days)
     patient_is_exposed_to_drug_or_medicament_inthepast7days@@is_anti_diarrheal_medication)
:named REQ2_AUXILIARY3)) ;; "anti-diarrheal medications with non-exhaustive examples (bismuth subsalicylate OR loperamide)"

;; Qualifier variable implies stem variable
(assert
(! (=> patient_is_exposed_to_drug_or_medicament_inthepast7days@@is_anti_diarrheal_medication
     patient_is_exposed_to_drug_or_medicament_inthepast7days)
:named REQ2_AUXILIARY4)) ;; "anti-diarrheal medications"

;; ===================== Constraint Assertions (REQ 2) =====================
;; Exclusion: patient must NOT have used any drug or medicament within 7 days prior to challenge
(assert
(! (not patient_is_exposed_to_drug_or_medicament_inthepast7days)
:named REQ2_COMPONENT0_OTHER_REQUIREMENTS)) ;; "The patient is excluded if the patient has used medications within 7 days prior to challenge ... OR (the patient has used drugs within 7 days prior to challenge)."

;; ===================== Declarations for the exclusion criterion (REQ 3) =====================
(declare-const patient_has_diagnosis_of_chronic_digestive_system_disorder_inthehistory Bool) ;; "the patient has a history of chronic gastrointestinal illness"

(declare-const patient_has_undergone_abdominal_cavity_operation_inthehistory Bool) ;; "the patient has a history of intra-abdominal surgery"

(declare-const patient_has_diagnosis_of_nonulcer_dyspepsia_inthehistory Bool) ;; "the patient has a history of chronic functional dyspepsia"

(declare-const patient_has_diagnosis_of_gastroesophageal_reflux_disease_inthehistory Bool) ;; "the patient has a history of chronic gastroesophageal reflux"

(declare-const patient_has_diagnosis_of_peptic_ulcer_inthehistory Bool) ;; "the patient has a history of documented peptic ulcer disease"

(declare-const patient_has_diagnosis_of_gastrointestinal_hemorrhage_inthehistory Bool) ;; "the patient has a history of gastrointestinal hemorrhage"

(declare-const patient_has_diagnosis_of_disorder_of_gallbladder_inthehistory Bool) ;; "the patient has a history of gallbladder disease"

(declare-const patient_has_diagnosis_of_inflammatory_bowel_disease_inthehistory Bool) ;; "the patient has a history of inflammatory bowel disease"

(declare-const patient_has_diagnosis_of_crohn_s_disease_inthehistory Bool) ;; "Crohn's disease"

(declare-const patient_has_diagnosis_of_ulcerative_colitis_inthehistory Bool) ;; "ulcerative colitis"

(declare-const patient_has_diagnosis_of_diverticulitis_inthehistory Bool) ;; "the patient has a history of diverticulitis"

(declare-const patient_has_diagnosis_of_irritable_bowel_syndrome_inthehistory Bool) ;; "the patient has a history of irritable bowel syndrome"

(declare-const patient_has_history_of_frequent_diarrhea Bool) ;; "the patient has a history of frequent diarrhea"

;; ===================== Auxiliary Assertions (REQ 3) =====================
;; Exhaustive subcategories: inflammatory bowel disease ≡ (Crohn's disease OR ulcerative colitis)
(assert
(! (= patient_has_diagnosis_of_inflammatory_bowel_disease_inthehistory
     (or patient_has_diagnosis_of_crohn_s_disease_inthehistory
         patient_has_diagnosis_of_ulcerative_colitis_inthehistory))
:named REQ3_AUXILIARY0)) ;; "with exhaustive subcategories (Crohn's disease OR ulcerative colitis)"

;; ===================== Constraint Assertions (REQ 3) =====================
(assert
(! (not patient_has_diagnosis_of_chronic_digestive_system_disorder_inthehistory)
:named REQ3_COMPONENT0_OTHER_REQUIREMENTS)) ;; "the patient has a history of chronic gastrointestinal illness"

(assert
(! (not patient_has_undergone_abdominal_cavity_operation_inthehistory)
:named REQ3_COMPONENT1_OTHER_REQUIREMENTS)) ;; "the patient has a history of intra-abdominal surgery"

(assert
(! (not patient_has_diagnosis_of_nonulcer_dyspepsia_inthehistory)
:named REQ3_COMPONENT2_OTHER_REQUIREMENTS)) ;; "the patient has a history of chronic functional dyspepsia"

(assert
(! (not patient_has_diagnosis_of_gastroesophageal_reflux_disease_inthehistory)
:named REQ3_COMPONENT3_OTHER_REQUIREMENTS)) ;; "the patient has a history of chronic gastroesophageal reflux"

(assert
(! (not patient_has_diagnosis_of_peptic_ulcer_inthehistory)
:named REQ3_COMPONENT4_OTHER_REQUIREMENTS)) ;; "the patient has a history of documented peptic ulcer disease"

(assert
(! (not patient_has_diagnosis_of_gastrointestinal_hemorrhage_inthehistory)
:named REQ3_COMPONENT5_OTHER_REQUIREMENTS)) ;; "the patient has a history of gastrointestinal hemorrhage"

(assert
(! (not patient_has_diagnosis_of_disorder_of_gallbladder_inthehistory)
:named REQ3_COMPONENT6_OTHER_REQUIREMENTS)) ;; "the patient has a history of gallbladder disease"

(assert
(! (not patient_has_diagnosis_of_inflammatory_bowel_disease_inthehistory)
:named REQ3_COMPONENT7_OTHER_REQUIREMENTS)) ;; "the patient has a history of inflammatory bowel disease with exhaustive subcategories (Crohn's disease OR ulcerative colitis)"

(assert
(! (not patient_has_diagnosis_of_diverticulitis_inthehistory)
:named REQ3_COMPONENT8_OTHER_REQUIREMENTS)) ;; "the patient has a history of diverticulitis"

(assert
(! (not patient_has_diagnosis_of_irritable_bowel_syndrome_inthehistory)
:named REQ3_COMPONENT9_OTHER_REQUIREMENTS)) ;; "the patient has a history of irritable bowel syndrome"

(assert
(! (not patient_has_history_of_frequent_diarrhea)
:named REQ3_COMPONENT10_OTHER_REQUIREMENTS)) ;; "the patient has a history of frequent diarrhea"

;; ===================== Declarations for the exclusion criterion (REQ 4) =====================
(declare-const patient_has_diagnosis_of_depressive_disorder_inthehistory Bool) ;; "depression"
(declare-const patient_has_diagnosis_of_depressive_disorder_inthehistory@@not_controlled_with_current_drug_therapy Bool) ;; "depression not controlled with current drug therapy"
(declare-const patient_has_diagnosis_of_depressive_disorder_inthehistory@@involving_institutionalization Bool) ;; "depression involving institutionalization"
(declare-const patient_has_diagnosis_of_schizophrenia_inthehistory Bool) ;; "schizophrenia"
(declare-const patient_has_diagnosis_of_psychotic_disorder_inthehistory Bool) ;; "psychosis"
(declare-const patient_has_history_of_suicide_attempt Bool) ;; "suicide attempt"

;; ===================== Auxiliary Assertions (REQ 4) =====================
;; Qualifier variables imply corresponding stem variables
(assert
(! (=> patient_has_diagnosis_of_depressive_disorder_inthehistory@@not_controlled_with_current_drug_therapy
      patient_has_diagnosis_of_depressive_disorder_inthehistory)
   :named REQ4_AUXILIARY0)) ;; "depression not controlled with current drug therapy"

(assert
(! (=> patient_has_diagnosis_of_depressive_disorder_inthehistory@@involving_institutionalization
      patient_has_diagnosis_of_depressive_disorder_inthehistory)
   :named REQ4_AUXILIARY1)) ;; "depression involving institutionalization"

;; ===================== Constraint Assertions (REQ 4) =====================
;; Exhaustive subcategories: history of at least one of the following psychiatric illnesses
(assert
(! (not (or patient_has_diagnosis_of_depressive_disorder_inthehistory@@not_controlled_with_current_drug_therapy
            patient_has_diagnosis_of_depressive_disorder_inthehistory@@involving_institutionalization
            patient_has_diagnosis_of_schizophrenia_inthehistory
            patient_has_diagnosis_of_psychotic_disorder_inthehistory
            patient_has_history_of_suicide_attempt))
   :named REQ4_COMPONENT0_OTHER_REQUIREMENTS)) ;; "has a history of at least one of the following psychiatric illnesses with exhaustive subcategories ((depression not controlled with current drug therapy) OR (depression involving institutionalization) OR (schizophrenia) OR (psychosis) OR (suicide attempt))."

;; ===================== Declarations for the exclusion criterion (REQ 5) =====================
(declare-const patient_has_finding_of_alcohol_abuse_inthehistory Bool) ;; "the patient has a history of alcohol abuse"
(declare-const patient_has_finding_of_illicit_medication_use_inthehistory Bool) ;; "the patient has a history of illicit drug abuse"
(declare-const patient_has_finding_of_alcohol_abuse_now Bool) ;; "the patient has current alcohol abuse"
(declare-const patient_has_finding_of_illicit_medication_use_now Bool) ;; "the patient has current illicit drug abuse"

;; ===================== Constraint Assertions (REQ 5) =====================
(assert
  (! (not patient_has_finding_of_alcohol_abuse_inthehistory)
     :named REQ5_COMPONENT0_OTHER_REQUIREMENTS)) ;; "the patient has a history of alcohol abuse"

(assert
  (! (not patient_has_finding_of_illicit_medication_use_inthehistory)
     :named REQ5_COMPONENT1_OTHER_REQUIREMENTS)) ;; "the patient has a history of illicit drug abuse"

(assert
  (! (not patient_has_finding_of_alcohol_abuse_now)
     :named REQ5_COMPONENT2_OTHER_REQUIREMENTS)) ;; "the patient has current alcohol abuse"

(assert
  (! (not patient_has_finding_of_illicit_medication_use_now)
     :named REQ5_COMPONENT3_OTHER_REQUIREMENTS)) ;; "the patient has current illicit drug abuse"

;; ===================== Declarations for the exclusion criterion (REQ 6) =====================
(declare-const patient_has_finding_of_hospital_patient_now Bool) ;; "inpatient"
(declare-const patient_has_finding_of_hospital_patient_now@@in_university_clinical_research_unit Bool) ;; "inpatient in the University Clinical Research Unit"
(declare-const patient_has_finding_of_hospital_patient_now@@can_remain_for_up_to_8_days Bool) ;; "able to remain as an inpatient for up to 8 days"
(declare-const patient_has_finding_of_hospital_patient_now@@in_university_clinical_research_unit@@can_remain_for_up_to_8_days Bool) ;; "inpatient in the University Clinical Research Unit and able to remain for up to 8 days"

;; ===================== Auxiliary Assertions (REQ 6) =====================
;; Qualifier variable implies stem variable
(assert
(! (=> patient_has_finding_of_hospital_patient_now@@in_university_clinical_research_unit
      patient_has_finding_of_hospital_patient_now)
:named REQ6_AUXILIARY0)) ;; "inpatient in the University Clinical Research Unit"

;; Qualifier variable implies stem variable
(assert
(! (=> patient_has_finding_of_hospital_patient_now@@can_remain_for_up_to_8_days
      patient_has_finding_of_hospital_patient_now)
:named REQ6_AUXILIARY1)) ;; "able to remain as an inpatient for up to 8 days"

;; Double qualifier variable implies both single qualifiers
(assert
(! (=> patient_has_finding_of_hospital_patient_now@@in_university_clinical_research_unit@@can_remain_for_up_to_8_days
      (and patient_has_finding_of_hospital_patient_now@@in_university_clinical_research_unit
           patient_has_finding_of_hospital_patient_now@@can_remain_for_up_to_8_days))
:named REQ6_AUXILIARY2)) ;; "inpatient in the University Clinical Research Unit and able to remain for up to 8 days"

;; ===================== Constraint Assertions (REQ 6) =====================
(assert
(! (not patient_has_finding_of_hospital_patient_now@@in_university_clinical_research_unit@@can_remain_for_up_to_8_days)
:named REQ6_COMPONENT0_OTHER_REQUIREMENTS)) ;; "The patient is excluded if the patient is unable to remain as an inpatient in the University Clinical Research Unit for up to 8 days."

;; ===================== Declarations for the exclusion criterion (REQ 7) =====================
(declare-const patient_has_hypersensitivity_to_latex_now Bool) ;; "The patient has a known hypersensitivity to latex."
(declare-const patient_has_finding_of_allergy_to_heparin_now Bool) ;; "The patient has a known hypersensitivity to heparin."
(declare-const patient_has_hypersensitivity_to_opioid_receptor_agonist_now Bool) ;; "The patient has a known hypersensitivity to opiates."
(declare-const patient_has_hypersensitivity_to_antiemetic_now Bool) ;; "The patient has a known hypersensitivity to antiemetics."
(declare-const patient_has_finding_of_allergy_to_benzodiazepine_now Bool) ;; "The patient has a known hypersensitivity to benzodiazepines."
(declare-const patient_has_finding_of_allergy_to_lidocaine_now Bool) ;; "The patient has a known hypersensitivity to lidocaine."
(declare-const patient_has_hypersensitivity_to_magnesium_citrate_now Bool) ;; "The patient has a known hypersensitivity to magnesium citrate."
(declare-const patient_has_hypersensitivity_to_fleet_enema_now Bool) ;; "The patient has a known hypersensitivity to Fleet enema."

;; ===================== Constraint Assertions (REQ 7) =====================
(assert
(! (not patient_has_hypersensitivity_to_latex_now)
:named REQ7_COMPONENT0_OTHER_REQUIREMENTS)) ;; "The patient is excluded if the patient has a known hypersensitivity to latex."

(assert
(! (not patient_has_finding_of_allergy_to_heparin_now)
:named REQ7_COMPONENT1_OTHER_REQUIREMENTS)) ;; "The patient is excluded if the patient has a known hypersensitivity to heparin."

(assert
(! (not patient_has_hypersensitivity_to_opioid_receptor_agonist_now)
:named REQ7_COMPONENT2_OTHER_REQUIREMENTS)) ;; "The patient is excluded if the patient has a known hypersensitivity to opiates."

(assert
(! (not patient_has_hypersensitivity_to_antiemetic_now)
:named REQ7_COMPONENT3_OTHER_REQUIREMENTS)) ;; "The patient is excluded if the patient has a known hypersensitivity to antiemetics."

(assert
(! (not patient_has_finding_of_allergy_to_benzodiazepine_now)
:named REQ7_COMPONENT4_OTHER_REQUIREMENTS)) ;; "The patient is excluded if the patient has a known hypersensitivity to benzodiazepines."

(assert
(! (not patient_has_finding_of_allergy_to_lidocaine_now)
:named REQ7_COMPONENT5_OTHER_REQUIREMENTS)) ;; "The patient is excluded if the patient has a known hypersensitivity to lidocaine."

(assert
(! (not patient_has_hypersensitivity_to_magnesium_citrate_now)
:named REQ7_COMPONENT6_OTHER_REQUIREMENTS)) ;; "The patient is excluded if the patient has a known hypersensitivity to magnesium citrate."

(assert
(! (not patient_has_hypersensitivity_to_fleet_enema_now)
:named REQ7_COMPONENT7_OTHER_REQUIREMENTS)) ;; "The patient is excluded if the patient has a known hypersensitivity to Fleet enema."

;; ===================== Declarations for the exclusion criterion (REQ 8) =====================
(declare-const patient_has_finding_of_allergic_reaction_caused_by_antibacterial_agent_now Bool) ;; "hypersensitivity to antibiotics"
(declare-const patient_has_finding_of_allergic_reaction_caused_by_antibacterial_agent_now@@known_hypersensitivity Bool) ;; "known hypersensitivity"
(declare-const patient_has_finding_of_allergic_reaction_caused_by_antibacterial_agent_now@@antibiotics_could_be_used_to_treat_enteroadherent_escherichia_coli_gastrointestinal_tract_infection Bool) ;; "antibiotics that could be used to treat enteroaggregative Escherichia coli infection"
(declare-const patient_has_hypersensitivity_to_quinolone_antibacterial_now Bool) ;; "fluoroquinolones"
(declare-const patient_has_hypersensitivity_to_amoxicillin_now Bool) ;; "amoxicillin"
(declare-const patient_has_hypersensitivity_to_cephalosporin_now Bool) ;; "cephalosporins"
(declare-const patient_has_hypersensitivity_to_rifaximin_now Bool) ;; "rifaximin"

;; ===================== Auxiliary Assertions (REQ 8) =====================
;; Non-exhaustive examples imply umbrella term
(assert
(! (=> (or patient_has_hypersensitivity_to_quinolone_antibacterial_now
           patient_has_hypersensitivity_to_amoxicillin_now
           patient_has_hypersensitivity_to_cephalosporin_now
           patient_has_hypersensitivity_to_rifaximin_now)
       patient_has_finding_of_allergic_reaction_caused_by_antibacterial_agent_now@@antibiotics_could_be_used_to_treat_enteroadherent_escherichia_coli_gastrointestinal_tract_infection)
   :named REQ8_AUXILIARY0)) ;; "with non-exhaustive examples (fluoroquinolones OR amoxicillin OR cephalosporins OR rifaximin)."

;; Qualifier variables imply corresponding stem variables
(assert
(! (=> patient_has_finding_of_allergic_reaction_caused_by_antibacterial_agent_now@@antibiotics_could_be_used_to_treat_enteroadherent_escherichia_coli_gastrointestinal_tract_infection
       patient_has_finding_of_allergic_reaction_caused_by_antibacterial_agent_now)
   :named REQ8_AUXILIARY1)) ;; "antibiotics that could be used to treat enteroaggregative Escherichia coli infection"

(assert
(! (=> patient_has_finding_of_allergic_reaction_caused_by_antibacterial_agent_now@@known_hypersensitivity
       patient_has_finding_of_allergic_reaction_caused_by_antibacterial_agent_now)
   :named REQ8_AUXILIARY2)) ;; "known hypersensitivity"

;; ===================== Constraint Assertions (REQ 8) =====================
(assert
(! (not (and patient_has_finding_of_allergic_reaction_caused_by_antibacterial_agent_now@@known_hypersensitivity
             patient_has_finding_of_allergic_reaction_caused_by_antibacterial_agent_now@@antibiotics_could_be_used_to_treat_enteroadherent_escherichia_coli_gastrointestinal_tract_infection))
   :named REQ8_COMPONENT0_OTHER_REQUIREMENTS)) ;; "The patient is excluded if the patient has a known hypersensitivity to antibiotics that could be used to treat enteroaggregative Escherichia coli infection with non-exhaustive examples (fluoroquinolones OR amoxicillin OR cephalosporins OR rifaximin)."

;; ===================== Declarations for the exclusion criterion (REQ 9) =====================
(declare-const patient_is_exposed_to_antibody_now Bool) ;; "antibodies"
(declare-const patient_is_exposed_to_antibody_now@@present_in_serum Bool) ;; "serum antibodies"
(declare-const patient_is_exposed_to_antibody_now@@specific_to_enteroaggregative_escherichia_coli_dispersin Bool) ;; "antibodies to enteroaggregative Escherichia coli dispersin"

;; ===================== Auxiliary Assertions (REQ 9) =====================
;; Qualifier variables imply corresponding stem variable
(assert
(! (=> patient_is_exposed_to_antibody_now@@present_in_serum
      patient_is_exposed_to_antibody_now)
   :named REQ9_AUXILIARY0)) ;; "serum antibodies"

(assert
(! (=> patient_is_exposed_to_antibody_now@@specific_to_enteroaggregative_escherichia_coli_dispersin
      patient_is_exposed_to_antibody_now)
   :named REQ9_AUXILIARY1)) ;; "antibodies to enteroaggregative Escherichia coli dispersin"

;; ===================== Constraint Assertions (REQ 9) =====================
;; Exclusion: patient must NOT have serum antibodies to enteroaggregative Escherichia coli dispersin
(assert
(! (not (and patient_is_exposed_to_antibody_now@@present_in_serum
             patient_is_exposed_to_antibody_now@@specific_to_enteroaggregative_escherichia_coli_dispersin))
   :named REQ9_COMPONENT0_OTHER_REQUIREMENTS)) ;; "The patient is excluded if the patient has serum antibodies to enteroaggregative Escherichia coli dispersin."

;; ===================== Declarations for the exclusion criterion (REQ 10) =====================
(declare-const patient_has_traveled_to_developing_country_within_past_6_months Bool) ;; "has recently traveled to a developing country within the past six months."
(declare-const patient_travel_to_developing_country_date_value_recorded_in_past_6_months_withunit_days Real) ;; "date (in days ago) when the patient traveled to a developing country, recorded if the travel occurred within the past six months."

;; ===================== Constraint Assertions (REQ 10) =====================
(assert
  (! (not patient_has_traveled_to_developing_country_within_past_6_months)
     :named REQ10_COMPONENT0_OTHER_REQUIREMENTS)) ;; "A patient is excluded if the patient has recently traveled to a developing country within the past six months."

;; ===================== Declarations for the exclusion criterion (REQ 11) =====================
(declare-const household_contact_age_value_recorded_now_in_years Real) ;; "age in years of a household contact at the current time"
(declare-const patient_has_household_contact_now Bool) ;; "the patient currently has at least one household contact"

;; ===================== Auxiliary Assertions (REQ 11) =====================
;; None needed: all logic is directly encoded in the constraints.

;; ===================== Constraint Assertions (REQ 11) =====================
;; Exclusion: The patient is excluded if the patient has at least one household contact who is less than four years of age.
(assert
(! (not (and patient_has_household_contact_now
             (< household_contact_age_value_recorded_now_in_years 4)))
   :named REQ11_COMPONENT0_OTHER_REQUIREMENTS)) ;; "The patient is excluded if the patient has at least one household contact who is less than four years of age."

;; Exclusion: The patient is excluded if the patient has at least one household contact who is more than eighty years of age.
(assert
(! (not (and patient_has_household_contact_now
             (> household_contact_age_value_recorded_now_in_years 80)))
   :named REQ11_COMPONENT1_OTHER_REQUIREMENTS)) ;; "The patient is excluded if the patient has at least one household contact who is more than eighty years of age."

;; ===================== Declarations for the exclusion criterion (REQ 12) =====================
(declare-const patient_has_household_contact_infirmed_now Bool) ;; "the patient has at least one household contact who is infirmed"

(declare-const patient_has_household_contact_immunocompromised_now Bool) ;; "the patient has at least one household contact who is immunocompromised"

(declare-const patient_has_household_contact_immunocompromised_now@@due_to_corticosteroid_therapy Bool) ;; "the patient has at least one household contact who is immunocompromised due to corticosteroid therapy"

(declare-const patient_has_household_contact_immunocompromised_now@@due_to_human_immunodeficiency_virus_infection Bool) ;; "the patient has at least one household contact who is immunocompromised due to human immunodeficiency virus infection"

(declare-const patient_has_household_contact_immunocompromised_now@@due_to_cancer_chemotherapy Bool) ;; "the patient has at least one household contact who is immunocompromised due to cancer chemotherapy"

(declare-const patient_has_household_contact_immunocompromised_now@@due_to_other_chronic_debilitating_diseases Bool) ;; "the patient has at least one household contact who is immunocompromised due to other chronic debilitating diseases"

;; ===================== Auxiliary Assertions (REQ 12) =====================
;; Qualifier variables imply corresponding stem variable for household contact immunocompromised
(assert
(! (=> patient_has_household_contact_immunocompromised_now@@due_to_corticosteroid_therapy
      patient_has_household_contact_immunocompromised_now)
    :named REQ12_AUXILIARY0)) ;; "the patient has at least one household contact who is immunocompromised due to corticosteroid therapy"

(assert
(! (=> patient_has_household_contact_immunocompromised_now@@due_to_human_immunodeficiency_virus_infection
      patient_has_household_contact_immunocompromised_now)
    :named REQ12_AUXILIARY1)) ;; "the patient has at least one household contact who is immunocompromised due to human immunodeficiency virus infection"

(assert
(! (=> patient_has_household_contact_immunocompromised_now@@due_to_cancer_chemotherapy
      patient_has_household_contact_immunocompromised_now)
    :named REQ12_AUXILIARY2)) ;; "the patient has at least one household contact who is immunocompromised due to cancer chemotherapy"

(assert
(! (=> patient_has_household_contact_immunocompromised_now@@due_to_other_chronic_debilitating_diseases
      patient_has_household_contact_immunocompromised_now)
    :named REQ12_AUXILIARY3)) ;; "the patient has at least one household contact who is immunocompromised due to other chronic debilitating diseases"

;; ===================== Constraint Assertions (REQ 12) =====================
(assert
(! (not patient_has_household_contact_infirmed_now)
    :named REQ12_COMPONENT0_OTHER_REQUIREMENTS)) ;; "the patient has at least one household contact who is infirmed"

(assert
(! (not patient_has_household_contact_immunocompromised_now@@due_to_corticosteroid_therapy)
    :named REQ12_COMPONENT1_OTHER_REQUIREMENTS)) ;; "the patient has at least one household contact who is immunocompromised due to corticosteroid therapy"

(assert
(! (not patient_has_household_contact_immunocompromised_now@@due_to_human_immunodeficiency_virus_infection)
    :named REQ12_COMPONENT2_OTHER_REQUIREMENTS)) ;; "the patient has at least one household contact who is immunocompromised due to human immunodeficiency virus infection"

(assert
(! (not patient_has_household_contact_immunocompromised_now@@due_to_cancer_chemotherapy)
    :named REQ12_COMPONENT3_OTHER_REQUIREMENTS)) ;; "the patient has at least one household contact who is immunocompromised due to cancer chemotherapy"

(assert
(! (not patient_has_household_contact_immunocompromised_now@@due_to_other_chronic_debilitating_diseases)
    :named REQ12_COMPONENT4_OTHER_REQUIREMENTS)) ;; "the patient has at least one household contact who is immunocompromised due to other chronic debilitating diseases"

;; ===================== Declarations for the exclusion criterion (REQ 13) =====================
(declare-const patient_is_health_care_personnel Bool) ;; "works as health care personnel"
(declare-const patient_is_health_care_personnel@@provides_direct_patient_care Bool) ;; "with direct patient care"

;; ===================== Auxiliary Assertions (REQ 13) =====================
;; Qualifier variable implies corresponding stem variable
(assert
(! (=> patient_is_health_care_personnel@@provides_direct_patient_care
      patient_is_health_care_personnel)
   :named REQ13_AUXILIARY0)) ;; "with direct patient care" implies "works as health care personnel"

;; ===================== Constraint Assertions (REQ 13) =====================
(assert
(! (not patient_is_health_care_personnel@@provides_direct_patient_care)
   :named REQ13_COMPONENT0_OTHER_REQUIREMENTS)) ;; "A patient is excluded if the patient works as health care personnel with direct patient care."

;; ===================== Declarations for the exclusion criterion (REQ 14) =====================
(declare-const patient_works_in_day_care_center_for_children_now Bool) ;; "the patient works in a day care center for children"
(declare-const patient_works_in_day_care_center_for_the_elderly_now Bool) ;; "the patient works in a day care center for the elderly"

;; ===================== Constraint Assertions (REQ 14) =====================
(assert
(! (not patient_works_in_day_care_center_for_children_now)
    :named REQ14_COMPONENT0_OTHER_REQUIREMENTS)) ;; "The patient is excluded if the patient works in a day care center for children."

(assert
(! (not patient_works_in_day_care_center_for_the_elderly_now)
    :named REQ14_COMPONENT1_OTHER_REQUIREMENTS)) ;; "The patient is excluded if the patient works in a day care center for the elderly."

;; ===================== Declarations for the exclusion criterion (REQ 15) =====================
(declare-const patient_is_food_handler_now Bool) ;; "the patient is a food handler"

;; ===================== Constraint Assertions (REQ 15) =====================
(assert
(! (not patient_is_food_handler_now)
:named REQ15_COMPONENT0_OTHER_REQUIREMENTS)) ;; "A patient is excluded if the patient is a food handler."

;; ===================== Declarations for the exclusion criterion (REQ 16) =====================
(declare-const patient_has_factor_that_would_interfere_with_study_objectives_in_opinion_of_investigator_now Bool) ;; "The patient is excluded if the patient has factors that, in the opinion of the investigator, would interfere with the study objectives."

(declare-const patient_has_factor_that_would_increase_risk_to_patient_in_opinion_of_investigator_now Bool) ;; "The patient is excluded if the patient has factors that, in the opinion of the investigator, would increase the risk to the patient."

(declare-const patient_has_factor_that_would_increase_risk_to_patients_contacts_in_opinion_of_investigator_now Bool) ;; "The patient is excluded if the patient has factors that, in the opinion of the investigator, would increase the risk to the patient's contacts."

(declare-const patient_has_factor_that_would_interfere_with_study_objectives_in_opinion_of_research_personnel_now Bool) ;; "The patient is excluded if the patient has factors that, in the opinion of the research personnel, would interfere with the study objectives."

(declare-const patient_has_factor_that_would_increase_risk_to_patient_in_opinion_of_research_personnel_now Bool) ;; "The patient is excluded if the patient has factors that, in the opinion of the research personnel, would increase the risk to the patient."

(declare-const patient_has_factor_that_would_increase_risk_to_patients_contacts_in_opinion_of_research_personnel_now Bool) ;; "The patient is excluded if the patient has factors that, in the opinion of the research personnel, would increase the risk to the patient's contacts."

;; ===================== Constraint Assertions (REQ 16) =====================
(assert
  (! (not patient_has_factor_that_would_interfere_with_study_objectives_in_opinion_of_investigator_now)
     :named REQ16_COMPONENT0_OTHER_REQUIREMENTS)) ;; "The patient is excluded if the patient has factors that, in the opinion of the investigator, would interfere with the study objectives."

(assert
  (! (not patient_has_factor_that_would_increase_risk_to_patient_in_opinion_of_investigator_now)
     :named REQ16_COMPONENT1_OTHER_REQUIREMENTS)) ;; "The patient is excluded if the patient has factors that, in the opinion of the investigator, would increase the risk to the patient."

(assert
  (! (not patient_has_factor_that_would_increase_risk_to_patients_contacts_in_opinion_of_investigator_now)
     :named REQ16_COMPONENT2_OTHER_REQUIREMENTS)) ;; "The patient is excluded if the patient has factors that, in the opinion of the investigator, would increase the risk to the patient's contacts."

(assert
  (! (not patient_has_factor_that_would_interfere_with_study_objectives_in_opinion_of_research_personnel_now)
     :named REQ16_COMPONENT3_OTHER_REQUIREMENTS)) ;; "The patient is excluded if the patient has factors that, in the opinion of the research personnel, would interfere with the study objectives."

(assert
  (! (not patient_has_factor_that_would_increase_risk_to_patient_in_opinion_of_research_personnel_now)
     :named REQ16_COMPONENT4_OTHER_REQUIREMENTS)) ;; "The patient is excluded if the patient has factors that, in the opinion of the research personnel, would increase the risk to the patient."

(assert
  (! (not patient_has_factor_that_would_increase_risk_to_patients_contacts_in_opinion_of_research_personnel_now)
     :named REQ16_COMPONENT5_OTHER_REQUIREMENTS)) ;; "The patient is excluded if the patient has factors that, in the opinion of the research personnel, would increase the risk to the patient's contacts."

;; ===================== Declarations for the exclusion criterion (REQ 17) =====================
(declare-const patient_is_undergoing_clinical_trial_now Bool) ;; "the patient is currently participating in a clinical study"
(declare-const patient_is_exposed_to_drug_or_medicament_inthepast30days Bool) ;; "the patient has received an investigational drug in the past thirty days"

;; ===================== Constraint Assertions (REQ 17) =====================
(assert
(! (not patient_is_undergoing_clinical_trial_now)
:named REQ17_COMPONENT0_OTHER_REQUIREMENTS)) ;; "the patient is currently participating in a clinical study"

(assert
(! (not patient_is_exposed_to_drug_or_medicament_inthepast30days)
:named REQ17_COMPONENT1_OTHER_REQUIREMENTS)) ;; "the patient has received an investigational drug in the past thirty days"

;; ===================== Declarations for the exclusion criterion (REQ 18) =====================
(declare-const patient_is_pregnant_now Bool) ;; "the patient is pregnant"
(declare-const patient_is_lactating_now Bool) ;; "the patient is lactating"
(declare-const patient_is_at_risk_of_pregnancy_now Bool) ;; "the patient is at risk of pregnancy"
(declare-const patient_has_finding_of_contraception_now@@effective Bool) ;; "effective birth control"
(declare-const patient_is_able_to_be_pregnant_now Bool) ;; "physiologically incapable of becoming pregnant"

;; ===================== Constraint Assertions (REQ 18) =====================
;; Exclusion: patient is currently pregnant
(assert
(! (not patient_is_pregnant_now)
:named REQ18_COMPONENT0_OTHER_REQUIREMENTS)) ;; "The patient is excluded if the patient is pregnant."

;; Exclusion: patient is currently lactating
(assert
(! (not patient_is_lactating_now)
:named REQ18_COMPONENT1_OTHER_REQUIREMENTS)) ;; "The patient is excluded if the patient is lactating."

;; Exclusion: patient is at risk of pregnancy AND (does not meet effective birth control OR is not physiologically incapable of becoming pregnant)
(assert
(! (not (and patient_is_at_risk_of_pregnancy_now
             (or (not patient_has_finding_of_contraception_now@@effective)
                 patient_is_able_to_be_pregnant_now)))
:named REQ18_COMPONENT2_OTHER_REQUIREMENTS)) ;; "The patient is excluded if the patient is at risk of pregnancy and does not meet the inclusion criteria for effective birth control or is not physiologically incapable of becoming pregnant as defined in the inclusion criteria."

;; ===================== Declarations for the exclusion criterion (REQ 19) =====================
(declare-const patient_has_finding_of_alcohol_intake_above_recommended_sensible_limits_now Bool) ;; "the patient has current excessive use of alcohol"

(declare-const patient_has_finding_of_psychoactive_substance_dependence_now Bool) ;; "the patient has drug dependence"

;; ===================== Constraint Assertions (REQ 19) =====================
(assert
(! (not patient_has_finding_of_alcohol_intake_above_recommended_sensible_limits_now)
:named REQ19_COMPONENT0_OTHER_REQUIREMENTS)) ;; "the patient has current excessive use of alcohol"

(assert
(! (not patient_has_finding_of_psychoactive_substance_dependence_now)
:named REQ19_COMPONENT1_OTHER_REQUIREMENTS)) ;; "the patient has drug dependence"

;; ===================== Declarations for the exclusion criterion (REQ 20) =====================
(declare-const patient_has_finding_of_disorder_of_immune_function_now Bool) ;; "impaired immune function"

;; ===================== Constraint Assertions (REQ 20) =====================
(assert
  (! (not patient_has_finding_of_disorder_of_immune_function_now)
     :named REQ20_COMPONENT0_OTHER_REQUIREMENTS)) ;; "A patient is excluded if the patient has evidence of impaired immune function."

;; ===================== Declarations for the exclusion criterion (REQ 21) =====================
(declare-const patient_has_finding_of_mantoux_positive_now Bool) ;; "positive reaction to purified protein derivative"
(declare-const patient_has_finding_of_mantoux_positive_now@@new Bool) ;; "new positive reaction to purified protein derivative"
(declare-const patient_has_finding_of_mantoux_positive_now@@known Bool) ;; "the patient is known to be purified protein derivative positive"
(declare-const patient_has_undergone_plain_chest_x_ray_now_outcome_is_negative Bool) ;; "the patient has a negative chest x-ray"
(declare-const patient_has_taken_isoniazid_containing_product_now Bool) ;; "isoniazid"
(declare-const patient_has_taken_isoniazid_containing_product_now@@used_as_prophylaxis Bool) ;; "the patient has received isoniazid prophylaxis"

;; ===================== Auxiliary Assertions (REQ 21) =====================
;; Qualifier variables imply corresponding stem variables
(assert
(! (=> patient_has_finding_of_mantoux_positive_now@@new
       patient_has_finding_of_mantoux_positive_now)
:named REQ21_AUXILIARY0)) ;; "new positive reaction to purified protein derivative"

(assert
(! (=> patient_has_finding_of_mantoux_positive_now@@known
       patient_has_finding_of_mantoux_positive_now)
:named REQ21_AUXILIARY1)) ;; "the patient is known to be purified protein derivative positive"

(assert
(! (=> patient_has_taken_isoniazid_containing_product_now@@used_as_prophylaxis
       patient_has_taken_isoniazid_containing_product_now)
:named REQ21_AUXILIARY2)) ;; "the patient has received isoniazid prophylaxis"

;; ===================== Constraint Assertions (REQ 21) =====================
;; Exclusion: (new positive reaction to PPD) AND NOT (known PPD positive AND negative chest x-ray AND received isoniazid prophylaxis)
(assert
(! (not (and patient_has_finding_of_mantoux_positive_now@@new
             (not (and patient_has_finding_of_mantoux_positive_now@@known
                       patient_has_undergone_plain_chest_x_ray_now_outcome_is_negative
                       patient_has_taken_isoniazid_containing_product_now@@used_as_prophylaxis))))
:named REQ21_COMPONENT0_OTHER_REQUIREMENTS)) ;; "A patient is excluded if (the patient has a new positive reaction to purified protein derivative) AND (NOT ((the patient is known to be purified protein derivative positive) AND (the patient has a negative chest x-ray) AND (the patient has received isoniazid prophylaxis)))."

;; ===================== Declarations for the exclusion criterion (REQ 22) =====================
(declare-const patient_has_undergone_stool_culture_now Bool) ;; "stool culture"
(declare-const patient_has_undergone_stool_culture_now@@demonstrates_presence_of_pathogenic_ova Bool) ;; "demonstrates the presence of pathogenic ova"
(declare-const patient_has_undergone_stool_culture_now@@demonstrates_presence_of_pathogenic_parasites Bool) ;; "demonstrates the presence of pathogenic parasites"
(declare-const patient_has_undergone_stool_culture_now@@demonstrates_presence_of_bacterial_enteropathogens_with_exhaustive_subcategories Bool) ;; "demonstrates the presence of bacterial enteropathogens with exhaustive subcategories (enteroaggregative Escherichia coli, Salmonella, Shigella, Campylobacter)"
(declare-const patient_has_undergone_stool_culture_now@@is_devoid_of_normal_flora Bool) ;; "stool culture that is devoid of normal flora"
(declare-const patient_has_finding_of_enteropathogenic_bacteria_isolated_now Bool) ;; "bacterial enteropathogens"

;; ===================== Auxiliary Assertions (REQ 22) =====================
;; Exhaustive subcategories: bacterial enteropathogens ≡ (enteroaggregative Escherichia coli ∨ Salmonella ∨ Shigella ∨ Campylobacter)
(assert
(! (= patient_has_undergone_stool_culture_now@@demonstrates_presence_of_bacterial_enteropathogens_with_exhaustive_subcategories
     patient_has_finding_of_enteropathogenic_bacteria_isolated_now)
:named REQ22_AUXILIARY0)) ;; "bacterial enteropathogens with exhaustive subcategories (enteroaggregative Escherichia coli, Salmonella, Shigella, Campylobacter)"

;; Qualifier variables imply corresponding stem variables
(assert
(! (=> patient_has_undergone_stool_culture_now@@demonstrates_presence_of_pathogenic_ova
       patient_has_undergone_stool_culture_now)
:named REQ22_AUXILIARY1)) ;; "demonstrates the presence of pathogenic ova"

(assert
(! (=> patient_has_undergone_stool_culture_now@@demonstrates_presence_of_pathogenic_parasites
       patient_has_undergone_stool_culture_now)
:named REQ22_AUXILIARY2)) ;; "demonstrates the presence of pathogenic parasites"

(assert
(! (=> patient_has_undergone_stool_culture_now@@demonstrates_presence_of_bacterial_enteropathogens_with_exhaustive_subcategories
       patient_has_undergone_stool_culture_now)
:named REQ22_AUXILIARY3)) ;; "demonstrates the presence of bacterial enteropathogens with exhaustive subcategories"

(assert
(! (=> patient_has_undergone_stool_culture_now@@is_devoid_of_normal_flora
       patient_has_undergone_stool_culture_now)
:named REQ22_AUXILIARY4)) ;; "stool culture that is devoid of normal flora"

;; ===================== Constraint Assertions (REQ 22) =====================
(assert
(! (not patient_has_undergone_stool_culture_now@@demonstrates_presence_of_pathogenic_ova)
:named REQ22_COMPONENT0_OTHER_REQUIREMENTS)) ;; "The patient is excluded if the patient has a stool culture that demonstrates the presence of pathogenic ova."

(assert
(! (not patient_has_undergone_stool_culture_now@@demonstrates_presence_of_pathogenic_parasites)
:named REQ22_COMPONENT1_OTHER_REQUIREMENTS)) ;; "The patient is excluded if the patient has a stool culture that demonstrates the presence of pathogenic parasites."

(assert
(! (not patient_has_undergone_stool_culture_now@@demonstrates_presence_of_bacterial_enteropathogens_with_exhaustive_subcategories)
:named REQ22_COMPONENT2_OTHER_REQUIREMENTS)) ;; "The patient is excluded if the patient has a stool culture that demonstrates the presence of bacterial enteropathogens with exhaustive subcategories (enteroaggregative Escherichia coli, Salmonella, Shigella, Campylobacter)."

(assert
(! (not patient_has_undergone_stool_culture_now@@is_devoid_of_normal_flora)
:named REQ22_COMPONENT3_OTHER_REQUIREMENTS)) ;; "The patient is excluded if the patient has a stool culture that is devoid of normal flora."

;; ===================== Declarations for the exclusion criterion (REQ 23) =====================
(declare-const patient_has_finding_of_intolerance_to_lactose_now@@self_reported Bool) ;; "the patient has self-reported lactose intolerance"
(declare-const patient_has_intolerance_to_soya_bean_now@@self_reported Bool) ;; "the patient has self-reported soy intolerance"
(declare-const patient_has_allergy_to_lactose_now@@self_reported Bool) ;; "the patient has self-reported lactose allergy"
(declare-const patient_has_finding_of_allergy_to_soy_protein_now@@self_reported Bool) ;; "the patient has self-reported soy allergy"

;; ===================== Constraint Assertions (REQ 23) =====================
(assert
(! (not patient_has_finding_of_intolerance_to_lactose_now@@self_reported)
:named REQ23_COMPONENT0_OTHER_REQUIREMENTS)) ;; "the patient has self-reported lactose intolerance"

(assert
(! (not patient_has_intolerance_to_soya_bean_now@@self_reported)
:named REQ23_COMPONENT1_OTHER_REQUIREMENTS)) ;; "the patient has self-reported soy intolerance"

(assert
(! (not patient_has_allergy_to_lactose_now@@self_reported)
:named REQ23_COMPONENT2_OTHER_REQUIREMENTS)) ;; "the patient has self-reported lactose allergy"

(assert
(! (not patient_has_finding_of_allergy_to_soy_protein_now@@self_reported)
:named REQ23_COMPONENT3_OTHER_REQUIREMENTS)) ;; "the patient has self-reported soy allergy"

;; ===================== Declarations for the exclusion criterion (REQ 24) =====================
(declare-const patient_has_finding_of_smoker_now Bool) ;; "smoker"
(declare-const patient_has_finding_of_tobacco_smoking_behavior_finding_now Bool) ;; "smoking"
(declare-const patient_has_finding_of_tobacco_smoking_behavior_finding_now@@cannot_stop_for_duration_of_inpatient_study Bool) ;; "cannot stop smoking for the duration of the inpatient study"

;; ===================== Auxiliary Assertions (REQ 24) =====================
;; Being a smoker implies engaging in tobacco smoking behavior
(assert
(! (=> patient_has_finding_of_smoker_now
       patient_has_finding_of_tobacco_smoking_behavior_finding_now)
   :named REQ24_AUXILIARY0)) ;; "smoker" ⇒ "engaging in tobacco smoking behavior"

;; Qualifier variable implies corresponding stem variable
(assert
(! (=> patient_has_finding_of_tobacco_smoking_behavior_finding_now@@cannot_stop_for_duration_of_inpatient_study
       patient_has_finding_of_tobacco_smoking_behavior_finding_now)
   :named REQ24_AUXILIARY1)) ;; "cannot stop smoking for the duration of the inpatient study" ⇒ "engaging in tobacco smoking behavior"

;; ===================== Constraint Assertions (REQ 24) =====================
;; Exclusion: patient is a smoker AND cannot stop smoking for the duration of the inpatient study
(assert
(! (not (and patient_has_finding_of_smoker_now
             patient_has_finding_of_tobacco_smoking_behavior_finding_now@@cannot_stop_for_duration_of_inpatient_study))
   :named REQ24_COMPONENT0_OTHER_REQUIREMENTS)) ;; "the patient is a smoker AND the patient cannot stop smoking for the duration of the inpatient study."
\end{MyVerbatim}

\newpage
\section{Prompts}
\label{app:prompts}

This section presents the prompts used in \name{}. To keep this appendix concise, each subsection below summarizes a prompt's purpose and links to its full text hosted on Google Drive. The complete collection is browsable at
  \href{https://drive.google.com/drive/folders/1aJm3vDcnhbvI-Vom6zPfT51y3v7h8QTR}{this Google Drive folder}.

{
\etocsetnexttocdepth{2} 
\etocsettocstyle{\subsection*{Contents of this section}}{}
\localtableofcontents
}

\subsection{Prompts for Trial-Side Semantic Parsing}
\subsubsection{Preprocessing: Trial Cohort Extractor}
\label{app:RequirementExtractor/trial_cohort_extractor}

\paragraph{Purpose.}
This prompt identifies whether a clinical trial defines multiple enrollment groups, where a patient can qualify through any one group. When such structure is present, it separates the trial into explicit cohorts, each with its own inclusion and exclusion criteria, while preserving shared trial-level conditions. This makes the enrollment structure explicit so that each cohort can be processed independently in downstream parsing.

\noindent\textbf{Full prompt:} \href{https://drive.google.com/file/d/1p2acbaXYBC_4ma7KmueULzbOVhT_2bWK/view?usp=sharing}{Google Drive PDF}.

\subsubsection{Preprocessing: Inclusion Requirements Extractor}
\label{app:RequirementExtractor/inclusion_requirements_extractor}

\paragraph{Purpose.}
This prompt extracts inclusion criteria from free-text eligibility text and converts them into a list of self-contained requirement statements. Each requirement is rewritten into a standardized form with explicit inclusion meaning and clear patient-level phrasing, while preserving the original text span for traceability. This ensures that each requirement can be directly translated into a logical condition in downstream processing.

\noindent\textbf{Full prompt:} \href{https://drive.google.com/file/d/19mefTsFyHBVIHa7Jd2Bj0AMXxqeKoAzG/view?usp=sharing}{Google Drive PDF}.

\subsubsection{Preprocessing: Exclusion Requirements Extractor}
\label{app:RequirementExtractor/exclusion_requirements_extractor}

\paragraph{Purpose.}
This prompt extracts exclusion criteria from free-text eligibility text and converts them into a list of self-contained requirement statements. Each requirement is rewritten into a standardized form with explicit exclusion meaning and clear patient-level phrasing, while preserving the original text span for traceability. This ensures that each requirement can be directly translated into a logical condition in downstream processing.

\noindent\textbf{Full prompt:} \href{https://drive.google.com/file/d/1d5VuUb8Yedv1Fz6wZjzfhmOhyMXcAlkM/view?usp=sharing}{Google Drive PDF}.

\subsubsection{Preprocessing: Inclusion Extraction Checker}
\label{app:RequirementExtractor/inclusion_extraction_verifier}

\paragraph{Purpose.}
This prompt verifies that the extracted inclusion requirements fully cover the source text, preserve the intended inclusion meaning, and follow the required patient-level format. In particular, it checks that no criteria are omitted, that the set of requirements correctly reflects the "all conditions must be satisfied" logic of inclusion criteria, and that each requirement is expressed in a consistent, standardized form for downstream processing.

\noindent\textbf{Full prompt:} \href{https://drive.google.com/file/d/1_3c9OEdp2GCQSGUfWk_l1Jgu4uSQNtmN/view?usp=sharing}{Google Drive PDF}.

\subsubsection{Preprocessing: Exclusion Extraction Checker}
\label{app:RequirementExtractor/exclusion_extraction_verifier}

\paragraph{Purpose.}
This prompt verifies that the extracted exclusion requirements fully cover the source text, preserve the intended exclusion meaning, and follow the required patient-level format. In particular, it checks that no criteria are omitted, that the set of requirements correctly reflects the "exclude if any condition is met" logic, and that each requirement is expressed in a consistent, standardized form for downstream processing.

\noindent\textbf{Full prompt:} \href{https://drive.google.com/file/d/1UVrngZ-maqviyhuXKE2-YzTqO17JaMcD/view?usp=sharing}{Google Drive PDF}.

\subsubsection{Preprocessing: Entity Expander}
\label{app:RequirementExtractor/entity_expander}

\paragraph{Purpose.}
This prompt rewrites each requirement so that every medically meaningful entity is expressed as a clear, standalone phrase. It expands compressed or coordinated mentions (e.g., shared heads, abbreviations) while preserving the original wording and logical structure. This makes entity boundaries explicit, allowing each entity to be reliably matched to ontology concepts in downstream processing.

\noindent\textbf{Full prompt:} \href{https://drive.google.com/file/d/1RS_4rDPepxqPaShIOvwe_CSxPVtmGWIB/view?usp=sharing}{Google Drive PDF}.

\subsubsection{Preprocessing: Entity Expansion Checker}
\label{app:RequirementExtractor/entity_expansion_verifier}

\paragraph{Purpose.}
This prompt verifies that entity expansion is correct and meaning-preserving. It checks that all medically relevant entities are expressed as clear, standalone phrases, and that the rewritten requirements remain semantically identical to the originals. In particular, it ensures that no information is lost or added, and that the set of patients described by each requirement remains unchanged.

\noindent\textbf{Full prompt:} \href{https://drive.google.com/file/d/1wI5PwFW3AUos9rffyyyNrfBVIGHC2pxF/view?usp=sharing}{Google Drive PDF}.

\subsubsection{Preprocessing: Inclusion Requirement Logic Rewriter}
\label{app:RequirementExtractor/inclusion_requirement_logic_rewriter}

\paragraph{Purpose.}
This prompt rewrites inclusion requirements to make their logical structure fully explicit and unambiguous. It surfaces logical operators (e.g., AND, OR, NOT), clarifies scope using parentheses, and makes qualifiers and conditions precise, while preserving the original clinical meaning. The result is a standardized form that can be directly translated into formal logical assertions in downstream processing.

\noindent\textbf{Full prompt:} \href{https://drive.google.com/file/d/1RZWK5csjhKH-JH3zHm5RVTd26_rq0GHk/view?usp=sharing}{Google Drive PDF}.

\subsubsection{Preprocessing: Exclusion Requirement Logic Rewriter}
\label{app:RequirementExtractor/exclusion_requirement_logic_rewriter}

\paragraph{Purpose.}
This prompt rewrites exclusion requirements to make their logical structure fully explicit and unambiguous. It surfaces logical operators (e.g., AND, OR, NOT), clarifies scope using parentheses, and makes qualifiers and conditions precise, while preserving the original clinical meaning. The result is a standardized form that can be directly translated into formal logical assertions in downstream processing.

\noindent\textbf{Full prompt:} \href{https://drive.google.com/file/d/1_U2FxtdbKFyG0bVRd041jRnYR_7siICL/view?usp=sharing}{Google Drive PDF}.

\subsubsection{Preprocessing: Inclusion Logic Rewrite Checker}
\label{app:RequirementExtractor/inclusion_logic_rewrite_verifier}

\paragraph{Purpose.}
This prompt verifies that the rewritten inclusion requirements are logically correct, meaning-preserving, and suitable for formal translation. It checks that logical operators and scope are explicitly and correctly represented, that the rewritten requirement corresponds to the same set of patients as the original, and that no information is lost, added, or misinterpreted.

\noindent\textbf{Full prompt:} \href{https://drive.google.com/file/d/1gj4GFr5-eySu6k3gfY-OGu_SeGwMMTz9/view?usp=sharing}{Google Drive PDF}.

\subsubsection{Preprocessing: Exclusion Logic Rewrite Checker}
\label{app:RequirementExtractor/exclusion_logic_rewrite_verifier}

\paragraph{Purpose.}
This prompt verifies that the rewritten exclusion requirements are logically correct, meaning-preserving, and suitable for formal translation. It checks that logical operators and scope are explicitly and correctly represented, that the rewritten requirement corresponds to the same set of patients as the original, and that no information is lost, added, or misinterpreted.

\noindent\textbf{Full prompt:} \href{https://drive.google.com/file/d/16gPO6K7VD1ZiU9jaHz9wKohz9dZq0mZ-/view?usp=sharing}{Google Drive PDF}.

\subsubsection{Preprocessing: Inclusion Requirement Decomposer}
\label{app:RequirementExtractor/inclusion_requirement_decomposer}

\paragraph{Purpose.}
This prompt decomposes each inclusion requirement into a set of independent component conditions. Each component is defined so that if it is satisfied, the original inclusion criterion is satisfied. The decomposition preserves the original meaning while producing smaller, self-contained units that can be evaluated independently in downstream processing.

\noindent\textbf{Full prompt:} \href{https://drive.google.com/file/d/17tVGfqLBbmwSvrgTtsb6nCvIQPLHP6XU/view?usp=sharing}{Google Drive PDF}.

\subsubsection{Preprocessing: Exclusion Requirement Decomposer}
\label{app:RequirementExtractor/exclusion_requirement_decomposer}

\paragraph{Purpose.}
This prompt decomposes each exclusion requirement into a set of independent component conditions. Each component is defined so that if it is satisfied, the original exclusion criterion is satisfied. The decomposition preserves the original meaning while producing smaller, self-contained units that can be evaluated independently in downstream processing.

\noindent\textbf{Full prompt:} \href{https://drive.google.com/file/d/1Bf3fTjMl3NckbTcsdKKrXyfUvxPCikFn/view?usp=sharing}{Google Drive PDF}.

\subsubsection{Preprocessing: Inclusion Decomposition Checker}
\label{app:RequirementExtractor/inclusion_decomposition_checker}

\paragraph{Purpose.}
This prompt checks that the decomposition of each inclusion requirement is logically correct, complete, and meaning-preserving. It checks that components are split according to logical structure (e.g., respecting AND/OR behavior under negation), that no information is lost or added, and that the set of patients defined by the decomposed components is equivalent to the original requirement.

\noindent\textbf{Full prompt:} \href{https://drive.google.com/file/d/1kNnXF7OtDMztPKuAKvwFShSVE_K0EPJ0/view?usp=sharing}{Google Drive PDF}.

\subsubsection{Preprocessing: Exclusion Decomposition Checker}
\label{app:RequirementExtractor/exclusion_decomposition_checker}

\paragraph{Purpose.}
This prompt checks that the decomposition of each exclusion requirement is logically correct, complete, and meaning-preserving. It checks that components are split according to logical structure (e.g., respecting AND/OR behavior under negation), that no information is lost or added, and that the set of patients defined by the decomposed components is equivalent to the original requirement.

\noindent\textbf{Full prompt:} \href{https://drive.google.com/file/d/1QL8u9OKShgLkfd_nwxypDg0J_jhEL_Sq/view?usp=sharing}{Google Drive PDF}.

\subsubsection{Preprocessing: Inclusion Requirement Classifier}
\label{app:RequirementExtractor/inclusion_requirement_classifier}

\paragraph{Purpose.}
This prompt classifies each inclusion component based on how it should be handled during prescreening. It distinguishes criteria that must be explicitly supported by patient notes, criteria that are non-binding or can be satisfied through simple actions, and criteria that may be unknown at prescreening but should not exclude a trial unless contradicted. This allows the system to enforce only the most informative constraints early while maintaining high recall under incomplete patient information.

\noindent\textbf{Full prompt:} \href{https://drive.google.com/file/d/1iiqAi8TFURODD9Awwu7SDDQzBbbYosmu/view?usp=sharing}{Google Drive PDF}.

\subsubsection{Preprocessing: Exclusion Requirement Classifier}
\label{app:RequirementExtractor/exclusion_requirement_classifier}

\paragraph{Purpose.}
This prompt classifies each exclusion component based on whether it should be enforced during early-stage matching. It distinguishes true eligibility constraints that determine patient-trial relevance from requirements that are non-binding or can be satisfied through simple actions. This allows the system to enforce only the most informative constraints during prescreening while deferring others, improving recall when patient information is incomplete.

\noindent\textbf{Full prompt:} \href{https://drive.google.com/file/d/1KSooNO_LU0KhbFYb39SySITWrTheuzjq/view?usp=sharing}{Google Drive PDF}.

\subsubsection{Entity Canonicalization: Free Entity Extraction}
\label{app:EntityCanonicalizer/free_entity_extraction}

\paragraph{Purpose.}
This prompt performs residual high-recall extraction to recover medically meaningful entity spans that were missed during initial canonicalization. Its purpose is to ensure that all constraint-relevant information is captured before downstream linking and filtering, since omissions at this stage cannot be recovered later and may lead to false negatives in retrieval or eligibility reasoning.


\noindent\textbf{Full prompt:} \href{https://drive.google.com/file/d/1jhLyj0-8AWdg-4v9znNGlpM-GCMZaHqJ/view?usp=sharing}{Google Drive PDF}.

\subsubsection{Entity Canonicalization: Primary High-Recall Mention Extractor}
\label{app:EntityCanonicalizer/primary_high_recall_mention_extractor}

\paragraph{Purpose.}
This prompt performs the primary high-recall extraction of medically meaningful entity mentions from clinical trial eligibility criteria. Its goal is to over-generate all plausible constraint-relevant spans, including overlapping candidates, to ensure that no clinically meaningful information is lost prior to ontology linking and downstream reasoning.

\noindent\textbf{Full prompt:} \href{https://drive.google.com/file/d/18MKkSfzdQRG9Bs1fOVldqvb2ulhly0YU/view?usp=sharing}{Google Drive PDF}.

\subsubsection{Entity Canonicalization: Context-Aware Concept Reranker}
\label{app:EntityCanonicalizer/context_aware_concept_linker}

\paragraph{Purpose.}
This prompt performs context-sensitive concept disambiguation by selecting the single best-matching SNOMED concept for each extracted span. Its purpose is to ensure that each link preserves the exact meaning of the span in its local context, rejecting semantically related but incorrect candidates to avoid introducing erroneous structured representations into downstream reasoning.

\noindent\textbf{Full prompt:} \href{https://drive.google.com/file/d/1gIS45lZwUti1_jTBJadwQgZK_HJgKrVE/view?usp=sharing}{Google Drive PDF}.

\subsubsection{Entity Canonicalization: Semantic Equivalence Verifier}
\label{app:EntityCanonicalizer/semantic_equivalence_verifier}

\paragraph{Purpose.}
This prompt verifies candidate SNOMED links by enforcing exact semantic equivalence between each extracted span and its linked concept in the local criterion context. Its purpose is to reject candidates that introduce, omit, or alter clinically relevant meaning, ensuring that only meaning-preserving links are passed to downstream selection and reasoning.

\noindent\textbf{Full prompt:} \href{https://drive.google.com/file/d/15xa4Jr4ECRIgWFK7ucsnQebjL_ajV4rN/view?usp=sharing}{Google Drive PDF}.

\subsubsection{Entity Canonicalization: Final Canonicalization Arbiter}
\label{app:EntityCanonicalizer/final_canonicalization_arbiter}

\paragraph{Purpose.}
This prompt performs final canonicalization by filtering candidate entity mappings, resolving overlapping spans, enforcing schema and policy constraints, and selecting the most specific faithful representations. It also generates schema-aware variable names for each retained entity, producing a compact symbolic representation suitable for downstream logical reasoning.

\noindent\textbf{Full prompt:} \href{https://drive.google.com/file/d/11FidTUS2eWo1yPUbvpF7nzH4T21L1H1M/view?usp=sharing}{Google Drive PDF}.

\subsubsection{SMTProgrammer: Free Entity Extractor}
\label{app:SMTProgrammer/SMTPreprocessor/SMTProgrammerFreeEntityExtractor}

\paragraph{Purpose.}
This prompt identifies medically relevant entities in each eligibility criterion that were not captured by earlier extraction steps. Its goal is to recover meaningful clinical concepts that would otherwise be missed, ensuring that important information is not lost before SMT translation.

\noindent\textbf{Full prompt:} \href{https://drive.google.com/file/d/19zPX2D1NLl2k8IMzSfhRqeh1WXlTzlWf/view?usp=sharing}{Google Drive PDF}.

\subsubsection{SMTProgrammer: Free Entity Qualifier Identifier}
\label{app:SMTProgrammer/SMTPreprocessor/SMTProgrammerFreeEntityQualifierIdentifier}

\paragraph{Purpose.}
This prompt identifies constraint phrases attached to each medical entity in a requirement. These constraints specify how the entity is restricted in context (e.g., time, location, severity, or cause), ensuring that the entity's meaning is fully captured before translation into SMT.

\noindent\textbf{Full prompt:} \href{https://drive.google.com/file/d/1y-cQQzF4EvhUDs649o4wja8N3Oj-6rrv/view?usp=sharing}{Google Drive PDF}.

\subsubsection{SMTProgrammer Free Entity Qualifier Identifier Checker}
\label{app:SMTProgrammer/SMTPreprocessor/SMTProgrammerFreeEntityQualifierIdentifierVerifier}

\paragraph{Purpose.}
This prompt checks whether each extracted constraint correctly and meaningfully restricts the referenced entity. It verifies that the constraint truly narrows the entity's scope, is attached to the correct entity, and does not simply restate inherent properties or unrelated context.

\noindent\textbf{Full prompt:} \href{https://drive.google.com/file/d/1CYLfCTHtBWeJmCJpfqCe8BFZdmxqPJXj/view?usp=sharing}{Google Drive PDF}.

\subsubsection{SMTProgrammer: Top Level Entity Filter}
\label{app:SMTProgrammer/SMTPreprocessor/SMTProgrammerTopLevelEntityFilter}

\paragraph{Purpose.}
This prompt determines which extracted entities represent independent clinical conditions in a requirement and which only serve to refine another entity. It keeps only entities that stand on their own as eligibility conditions, ensuring that only meaningful concepts are carried forward for variable construction.

\noindent\textbf{Full prompt:} \href{https://drive.google.com/file/d/1fpNshd7obsVmaH2RWlWfJ8Ojz3A7LfI_/view?usp=sharing}{Google Drive PDF}.

\subsubsection{SMT Programmer: Canonical Variable Namer}
\label{app:SMTProgrammer/SMTIncrementalProgrammer/SMTIncrementalCanonicalVariableNamer}

\paragraph{Purpose.}
This prompt constructs standardized variables for canonical medical entities in a requirement. It determines which variables should be introduced or reused, assigns consistent names based on predefined templates, and attaches any necessary qualifiers so that each variable accurately represents the intended clinical meaning.

\noindent\textbf{Full prompt:} \href{https://drive.google.com/file/d/1jB0zu3vc1BLhNV-Z84HOWDT8RVPyaY1S/view?usp=sharing}{Google Drive PDF}.

\subsubsection{SMT Programmer: Demographics Variable Namer}
\label{app:SMTProgrammer/SMTIncrementalProgrammer/SMTIncrementalDemographicsVariableNamer}

\paragraph{Purpose. }
This prompt constructs standardized variables for patient demographic attributes, such as age, sex, and reproductive status. It identifies which demographic variables are required for the current requirement and ensures they are represented using consistent templates for downstream SMT encoding.

\noindent\textbf{Full prompt:} \href{https://drive.google.com/file/d/1HBvBMWrCJbxeEQMKVcV5vXq5s5LBDyxy/view?usp=sharing}{Google Drive PDF}.

\subsubsection{SMT Programmer: Free Variable Namer}
\label{app:SMTProgrammer/SMTIncrementalProgrammer/SMTIncrementalFreeVariableNamer}

\paragraph{Purpose. }
This prompt constructs variables for clinically meaningful content that cannot be mapped to canonical representations. It introduces flexible, verbatim-aligned variables to ensure that important information is preserved, even when standardization is not possible.

\noindent\textbf{Full prompt:} \href{https://drive.google.com/file/d/1aVi5mpn_kaCCSIRRUl_Doa_P67hqX2yQ/view?usp=sharing}{Google Drive PDF}.

\subsubsection{SMT Programmer: New Variable Filter}
\label{app:SMTProgrammer/SMTIncrementalProgrammer/SMTIncrementalNewVariableFilter}

\paragraph{Purpose. }
This prompt removes proposed variables that do not represent standalone clinical concepts in the requirement. Its role is to keep only variables that function independently in the eligibility logic, while filtering out variables whose meaning is purely to modify another variable.

\noindent\textbf{Full prompt:} \href{https://drive.google.com/file/d/1Rzf7-tN5ZskrKokINNza6Tap4JOLr1pD/view?usp=sharing}{Google Drive PDF}.

\subsubsection{SMT Programmer: Reusable Variable Identifier}
\label{app:SMTProgrammer/SMTIncrementalProgrammer/SMTIncrementalReusableVariableIdentifier}

\paragraph{Purpose. }
This prompt identifies previously declared variables in the current SMT program that can be reused for the new requirement. Its goal is to preserve consistency across requirements by reusing existing symbols whenever they already capture the needed meaning, rather than introducing unnecessary new variables.

\noindent\textbf{Full prompt:} \href{https://drive.google.com/file/d/1e07Bpjfgp0LhE4ft3Wam05W23TlSmSJ2/view?usp=sharing}{Google Drive PDF}.

\subsubsection{SMT Programmer: Translator Exclusion}
\label{app:SMTProgrammer/SMTIncrementalProgrammer/SMTIncrementalTranslatorExclusion}

\paragraph{Purpose. }
This prompt translates a single exclusion requirement into SMT-LIB code. It generates the corresponding declarations and assertions using previously constructed variables, ensuring that the resulting logic faithfully encodes the requirement's meaning and integrates correctly with the existing SMT program.

\noindent\textbf{Full prompt:} \href{https://drive.google.com/file/d/1zH2GG8FVQe-ekr1RZm718IHd4ScJNR1y/view?usp=sharing}{Google Drive PDF}.

\subsubsection{SMT Programmer: Translator Inclusion}
\label{app:SMTProgrammer/SMTIncrementalProgrammer/SMTIncrementalTranslatorInclusion}

\paragraph{Purpose. }
This prompt translates a single inclusion requirement into SMT-LIB code. It generates the corresponding declarations and assertions using available variables, ensuring that the resulting logic faithfully encodes the requirement's meaning and integrates correctly with the existing SMT program.

\noindent\textbf{Full prompt:} \href{https://drive.google.com/file/d/1MEuL88R8LdHxC7_6ZUwiPxtoIcDI648Q/view?usp=sharing}{Google Drive PDF}.

\subsubsection{SMT Programmer: Verifier Exclusion}
\label{app:SMTProgrammer/SMTIncrementalProgrammer/SMTIncrementalVerifierExclusion}


\paragraph{Purpose. }
This prompt verifies that the generated SMT fragment for an exclusion requirement faithfully captures the original meaning. It checks semantic correctness, polarity, completeness of entities and qualifiers, and consistency with existing variables before the fragment is accepted into the program.

\noindent\textbf{Full prompt:} \href{https://drive.google.com/file/d/1BDovQXPJZBtrhs2Y6tXI7SInVOlw8VDz/view?usp=sharing}{Google Drive PDF}.

\subsubsection{SMT Programmer: Verifier Inclusion}
\label{app:SMTProgrammer/SMTIncrementalProgrammer/SMTIncrementalVerifierInclusion}

\paragraph{Purpose. }
This prompt verifies that the generated SMT fragment for an inclusion requirement faithfully captures the original meaning. It checks semantic correctness, polarity, completeness of entities and qualifiers, and consistency with existing variables before the fragment is accepted into the program.

\noindent\textbf{Full prompt:} \href{https://drive.google.com/file/d/1so6KCGa60vpHoZmA6wZz1HnprGxMnsGQ/view?usp=sharing}{Google Drive PDF}.

\subsubsection{Fix Overly Strict or Misinterpreted Criteria}
\label{app:repair/relaxer}

\paragraph{Purpose.}
This prompt corrects broad semantic drift introduced during the initial translation from natural-language eligibility criteria to SMT. It focuses on cases where literal or overly rigid interpretations make the program too restrictive, causing patients to be incorrectly excluded.

The prompt refines the SMT encoding by aligning it with the clinical intent expressed in the eligibility text. This includes resolving context-dependent meanings, distinguishing current index conditions from prior episodes, and introducing appropriate qualifiers (e.g., clinically significant conditions or prior-event qualifiers) to avoid unintended over-exclusion. It may also adjust how umbrella concepts and example lists are represented to better match their intended scope.

All edits are conservative and minimal, aiming to restore fidelity to the natural-language criteria without introducing new assumptions or weakening valid constraints.

\noindent\textbf{Full prompt:} \href{https://drive.google.com/file/d/18F4JX9H3ZlUZosJy_LuXs8cRVHPW5M75/view?usp=sharing}{Google Drive PDF}.

\subsubsection{Fix Reversed Exclusion Logic}
\label{app:repair/polarity}

\paragraph{Purpose.}
This prompt identifies and corrects polarity errors in exclusion-side SMT programs that arise from the "not excluded" encoding convention. Such errors occur when a condition is encoded in a way that makes the program satisfiable when the patient should actually be excluded, thereby reversing the intended clinical meaning.

The prompt operates at a fine granularity, repairing only those assertions whose logical direction is clearly incorrect while leaving correctly encoded conditions unchanged. All edits are minimal and local, preserving the existing program structure, symbols, and annotations. If no polarity errors are detected, the program is left unchanged.

\noindent\textbf{Full prompt:} \href{https://drive.google.com/file/d/1kYdMIDwTVSu_OPawZFcx2KjN6SVUbJzJ/view?usp=sharing}{Google Drive PDF}.

\subsubsection{Fix Misinterpreted or Mis-scoped Logic}
\label{app:repair/qualifier}

\paragraph{Purpose.}
This prompt corrects logical encoding errors in the SMT program where the structure of the constraints does not faithfully reflect the natural-language eligibility criteria. These errors include incorrect scoping of qualifiers, misplaced conditions, dropped or improperly grouped components, and other context-dependent misinterpretations.

The prompt uses trial and subcohort context to resolve how conditions should be applied, ensuring that constraints are attached to the correct entities and combined with the intended logical structure. All edits are conservative and local, modifying only the parts of the program that clearly deviate from the natural-language meaning, while preserving the overall structure, symbols, and annotations. If no logical inconsistencies are found, the program is left unchanged.

\noindent\textbf{Full prompt:} \href{https://drive.google.com/file/d/1OORXj6BSk_-yQUmhjD-y-woB04WvOCGM/view?usp=sharing}{Google Drive PDF}.

\subsubsection{Check if This Side Has Real Eligibility Criteria}
\label{app:repair/coverage_checker}

\paragraph{Purpose.}
This prompt determines whether a side-specific SMT program meaningfully encodes real eligibility criteria for the current trial subcohort. It acts as a conservative filter to identify cases where the SMT file is empty, placeholder-only, unrelated to the target side, or otherwise fails to represent the natural-language criteria.

The decision is intentionally cautious: the SMT file is marked for removal only when there is strong evidence that no substantive criteria exist for this side. If the program contains any plausible encoding of the target-side eligibility conditions, it is retained. This prevents accidental deletion of valid but incomplete encodings while removing spurious files that would otherwise degrade retrieval and matching quality.

\noindent\textbf{Full prompt:} \href{https://drive.google.com/file/d/1jSTS3I-aFW6KuxkA_ar9u-CfgVpfQi40/view?usp=sharing}{Google Drive PDF}.

\subsubsection{Check if This Subcohort Has Any Eligibility Criteria}
\label{app:repair/criteria_gater}

\paragraph{Purpose.}
This prompt determines whether any real eligibility criteria exist for the target side (inclusion or exclusion) for a specific trial subcohort. It operates directly on natural-language criteria and subcohort context, before considering the SMT representation.

The decision is strictly scoped to the current subcohort. Criteria belonging to other arms or cohorts must not be counted. The prompt uses a conservative policy: it marks the criteria as absent only when there is strong evidence that the target side is missing, placeholder-only, or contains no substantive conditions. If there is any plausible eligibility requirement, even a single condition, the criteria are considered present and retained.

This stage serves as a safeguard against propagating empty or non-substantive side definitions, while avoiding accidental deletion when subcohort attribution is uncertain or upstream preprocessing is incomplete.

\noindent\textbf{Full prompt:} \href{https://drive.google.com/file/d/1JeJmvRPZ82XkXJdxhBRbmW2m4kacmX6G/view?usp=sharing}{Google Drive PDF}.

\subsubsection{Tighten Missing or Weak Requirements}
\label{app:repair/constrainer}

\paragraph{Purpose.}
This prompt identifies and fixes cases where the SMT program is too permissive relative to the natural-language eligibility criteria, leading to false positives. It ensures that all clinically relevant requirements for the current side are properly enforced by (i) adding missing conditions, (ii) restoring weakened logical structure, and (iii) correcting variable definitions that are too narrow, mis-scoped, or insufficiently contextualized. 

In addition to repairing constraints, this stage may refine the underlying representation by introducing broader umbrella concepts, appropriate qualifiers (e.g., suspected or clinically supported conditions), or correct temporal scopes when required by the natural language. All changes are conservative and limited to what is explicitly supported by the eligibility text, with the goal of tightening the program without introducing unsupported restrictions.


\noindent\textbf{Full prompt:} \href{https://drive.google.com/file/d/1Qyi13cUIo9HUqhu5bsrgeFHpfLVRNJ95/view?usp=sharing}{Google Drive PDF}.

\subsubsection{Clarify and Complete Variable Definitions}
\label{app:repair/meaning_enricher}

\paragraph{Purpose.}
This prompt ensures that every variable in the SMT program has a complete, self-contained, and machine-readable definition. It operates only on variable declarations, adding or repairing inline JSON annotations so that each symbol can be consistently interpreted and populated by downstream patient-data extraction components.

In addition to filling missing or invalid definitions, this stage clarifies variable meanings by incorporating relevant context such as time windows, thresholds, units, and evidence sources when supported by the program or eligibility text. It also resolves ambiguity by making implicit assumptions explicit, while avoiding the introduction of unsupported details. The program logic itself is left unchanged; the goal is solely to make each variable definition precise, interpretable, and operational for downstream use.

\noindent\textbf{Full prompt:} \href{https://drive.google.com/file/d/1Z7RJ3P-PNzScKAaBU0GMEtv9zWS29NT9/view?usp=sharing}{Google Drive PDF}.

\subsubsection{Database Preparation: Disease Categorization}
\label{app:dataset_prep/disease_categorization}

\paragraph{Purpose.}
This prompt determines how each disease is relevant to the clinical trial by labeling the trial's intended relationship to that disease (e.g., treating, preventing, monitoring, or not relevant).

The goal is to distinguish which diseases represent true trial targets versus incidental mentions, so that only clinically meaningful disease signals are retained for downstream retrieval. This helps ensure that disease-based matching reflects the trial's actual intent rather than noisy or background conditions.

\noindent\textbf{Full prompt:} \href{https://drive.google.com/file/d/1RaoirvOGeIYYFhpsn_0rzHQU0Ww_VvRd/view?usp=sharing}{Google Drive PDF}.

\subsubsection{Database Preparation: Remove Redundant or Overly General Targets}

\label{app:dataset_prep/joint_specificity_filter}
\paragraph{Purpose.}
This prompt filters the list of candidate diseases and clinical targets by removing items that are too general or redundant given more specific alternatives in the same list.

The goal is to keep only the most informative and specific targets for each trial while preserving recall. General items are retained unless a clearly more precise target already captures the same clinical meaning. This step ensures that the final target set remains concise and meaningful for retrieval without discarding potentially useful signals.

\noindent\textbf{Full prompt:} \href{https://drive.google.com/file/d/1bGwXSw2QdxFgQsRqFbxZvlUWSSwsFZjg/view?usp=sharing}{Google Drive PDF}.

\subsubsection{Database Preparation: Label How the Trial Relates to Clinical Findings}

\label{app:dataset_prep/positive_literal_finding_schematization}
\paragraph{Purpose.}
This prompt determines how each extracted clinical target (from eligibility criteria) is relevant to the trial by labeling the trial's intended relationship to that target (e.g., treating, preventing, monitoring, or not relevant).

The goal is to identify which eligibility-derived signals represent meaningful trial targets versus incidental conditions, so that only clinically relevant targets are retained for retrieval. This complements disease-based labeling by capturing non-disease clinical factors (e.g., treatments, procedures, biomarkers) that reflect the trial's true intent.

\noindent\textbf{Full prompt:} \href{https://drive.google.com/file/d/1BtkwAmimVYV8kr8EFRrM-WEZUP_nt7HK/view?usp=sharing}{Google Drive PDF}.

\subsubsection{Database Preparation: Label How the Trial Relates to Each Procedure}

\label{app:dataset_prep/positive_literal_procedure_schematization}
\paragraph{Purpose.}
This prompt determines how each procedure-related variable is relevant to the clinical trial by labeling the trial's intended relationship to that procedure (e.g., improving effectiveness, reducing complications, modifying delivery, or not relevant).

The goal is to identify which procedure-related signals represent meaningful trial targets versus incidental mentions, so that only clinically relevant procedural aspects are retained for retrieval. This ensures that the system captures trials focused on interventions involving procedures, not just diseases or general clinical conditions.

\noindent\textbf{Full prompt:} \href{https://drive.google.com/file/d/1vq2bb_q2dkLQqc52134kVt3LzbrQZHe3/view?usp=sharing}{Google Drive PDF}.

\subsubsection{Database Preparation: Label How the Trial Relates to Each Substance Use}

\label{app:dataset_prep/positive_literal_substance_schematization}

\paragraph{Purpose.}
This prompt determines how each substance-related variable is relevant to the clinical trial by labeling the trial's intended relationship to that substance (e.g., reducing use, mitigating harms, enhancing benefits, or not relevant).

The goal is to identify which substance-related signals represent meaningful targets of the trial versus incidental mentions, so that only clinically relevant substance-use factors are retained for retrieval. This ensures the system captures trials focused on substance exposure, treatment, or management, rather than unrelated background information.

\noindent\textbf{Full prompt:} \href{https://drive.google.com/file/d/1VJ7xVFVXchdZLui09ORGgQCkGTCKKMEO/view?usp=sharing}{Google Drive PDF}.

\subsection{Prompts for Patient-Side Semantic Parsing}
\label{app:prompt-patient-side-parsing}
\subsubsection{Preprocessing: Patient Fact Extraction}\label{app:patient-parsing/extraction}

\paragraph{Purpose.} This prompt is used to turn a patient note into a clear list of distinct patient facts. Its goal is to make sure each fact is complete on its own, includes the details that belong with it, does not overlap with other facts, and stays faithful to the original note. It also requires each fact to be tied to the exact supporting text so the extraction can be checked later.

\noindent\textbf{Full prompt:} \href{https://drive.google.com/file/d/1vY9UUPfFIFyDnr14cQbB9tmQI1mR1qU1/view?usp=sharing}{Google Drive PDF}.

\subsubsection{Preprocessing: Patient Fact Extraction Checker}\label{app:patient-parsing/extraction_checker}

\paragraph{Purpose.} This prompt is used to check whether the extracted patient facts are a faithful and usable summary of the original note. It verifies that nothing important was missed, nothing new was added, and each fact can stand on its own, then gives a simple pass or fail result with an explanation only when there is a problem.



\noindent\textbf{Full prompt:} \href{https://drive.google.com/file/d/1_Be17a-aCrENXPbJFxUzCvgPVfeVpImN/view?usp=sharing}{Google Drive PDF}.

\subsubsection{Preprocessing: Entity Expander}\label{app:patient-parsing/expander}

\paragraph{Purpose.} This prompt rewrites each patient fact so that every medical condition or finding appears as its own clear, standalone phrase, making later concept matching easier and more reliable. It expands combined entities, spells out shorthand when needed, removes irrelevant always-true wording, and keeps the original meaning and logic unchanged while preserving one rewritten line for each input fact.

\noindent\textbf{Full prompt:} \href{https://drive.google.com/file/d/1225wSqRb3CxlDu9xiB7sMegGxncp4Xv7/view?usp=sharing}{Google Drive PDF}.

\subsubsection{Preprocessing: Logic Rewriter}\label{app:patient-parsing/rewriter}

\paragraph{Purpose.} This prompt rewrites each patient fact into a more precise and logically explicit form so it can be converted into formal rules later. It keeps the medical meaning the same, makes relationships like AND, OR, NOT, and numeric thresholds clear, preserves each medical entity exactly as written, and outputs one cleaned, unambiguous line for each original fact.

\noindent\textbf{Full prompt:} \href{https://drive.google.com/file/d/193TVbOMrIZJYBA3_rWn_AKnqVPp0p4-r/view?usp=sharing}{Google Drive PDF}.

\subsubsection{Preprocessing: Logic Rewrite Checker}\label{app:rewrite_checker}

\paragraph{Purpose.} This prompt checks whether each rewritten patient fact is logically clearer without changing what the original fact meant. It reviews each original-and-rewritten pair for issues like missing parentheses, unclear logic, missed abbreviation expansion, added information, or lost qualifiers, and then returns a pass or fail result for each one, along with a corrected rewrite when needed.

\noindent\textbf{Full prompt:} \href{https://drive.google.com/file/d/1WnsIsH6U_bUF1B7_n2vew67QbsyVrApy/view?usp=sharing}{Google Drive PDF}.

\subsubsection{Entity Canonicalization: Medical Entity Recognizer}\label{app:patient-parsing/entity_recognizer}

\paragraph{Purpose.} This prompt is used to find as many medically meaningful entity mentions as possible in a piece of clinical text. It extracts the exact text span for each entity, allows overlapping entities to improve recall, avoids pulling in non-entity wording like age, sex, consent, or study logistics, and returns each recognized entity with a brief note about what medical concept it may refer to.

\noindent\textbf{Full prompt:} \href{https://drive.google.com/file/d/1WuFscZyGoWCilLHFo6XnpU6UJrf1LFHm/view?usp=sharing}{Google Drive PDF}.

\subsubsection{Entity Canonicalization: Medical Concept Reranker}\label{app:patient-parsing/reranker}

\paragraph{Purpose.} This prompt picks one best SNOMED concept for each extracted medical entity mention. It uses the surrounding criterion text and candidate list to choose either an exact match or, if that is not available, the most specific broader concept, while keeping every input entity in the output even when no suitable concept can be linked.

\noindent\textbf{Full prompt:} \href{https://drive.google.com/file/d/1JLRPx4JM1ttz4gp9A6j8AgeegDrUG3nT/view?usp=sharing}{Google Drive PDF}.

\subsubsection{Entity Canonicalization: Semantic Equivalence Verifier}\label{app:patient-parsing/verifier}

\paragraph{Purpose.} This prompt checks whether a proposed SNOMED concept is acceptable for a given extracted entity mention. It keeps a candidate only if it means the same thing as the extracted span in context or is a broader concept that still correctly covers it, and it records that judgment in a structured output.

\noindent\textbf{Full prompt:} \href{https://drive.google.com/file/d/1G6ebmN9MblUN_gacrOmK7d7oK9VCuWUA/view?usp=sharing}{Google Drive PDF}.

\subsubsection{Entity Canonicalization: Final Canonicalization Arbiter}\label{app:patient-parsing/arbiter}

\paragraph{Purpose.} This prompt is the final filter for entity canonicalization. It decides whether each extracted span and linked concept is precise enough, medically appropriate, and usable in your naming scheme for later formal encoding. It checks that the span really appears in the criterion, maps to an allowed medical concept type, is not just demographic or consent language, does not rely on class or stage labels, matches the linked concept in meaning, can be turned into a valid standardized variable name, and is the most specific valid choice among overlapping candidates. Only entries that pass all checks are kept.

\noindent\textbf{Full prompt:} \href{https://drive.google.com/file/d/1gwy4-UlfMCNdgACTh4onjBpbQQl0LhCO/view?usp=sharing}{Google Drive PDF}.

\subsubsection{Entity Canonicalization: Diagnosis Concept Reranker}\label{app:patient-parsing/diagnosis_reranker}

\paragraph{Purpose.} This prompt links each diagnosis string to the single best matching SNOMED concept from a provided candidate list. It uses the meaning of the diagnosis as written in the note, keeps only one choice per diagnosis, and returns null when none of the candidates correctly represent that diagnosis.

\noindent\textbf{Full prompt:} \href{https://drive.google.com/file/d/1kThY6kcSMpWtmP-g6mnZh29NE38mplPY/view?usp=sharing}{Google Drive PDF}.

\subsubsection{Entity Canonicalization: Diagnosis Concept Semantic Equivalence Verifier}
\label{app:patient-parsing/diagnosis-verifier}

\paragraph{Purpose.} This prompt checks whether a chosen SNOMED concept is a good match for a diagnosis string from the patient note. It determines whether the concept means the same thing as the diagnosis, whether it is a broader parent concept, and whether the two are at least medically related, then returns those judgments with a short explanation.

\noindent\textbf{Full prompt:} \href{https://drive.google.com/file/d/1kbVQt_-gbBSWEWlt77-1YkLbcG3J0T5l/view?usp=sharing}{Google Drive PDF}.

\subsubsection{SMT Programming: Demographic Variable Constructor}\label{app:patient-parsing/demovar}

\paragraph{Purpose.} This prompt identifies demographic variables in a patient fact and converts them into a small set of standardized SMT-ready variables. It only handles age, sex, pregnancy-related status, and ability to be pregnant, and it outputs conservative, template-based variable declarations with values and time windows, without writing any assertions.

\noindent\textbf{Full prompt:} \href{https://drive.google.com/file/d/1NVKucU0NdGmgfRWXpB_mQfkuatdHRLco/view?usp=sharing}{Google Drive PDF}.

\subsubsection{SMT Programming: Canonical Variable Constructor}\label{app:patient-parsing/canonvar}

\paragraph{Purpose.} This prompt converts canonicalized medical entities in a patient fact into standardized SMT variable declarations. It creates Boolean or numeric variables only for entities that already have approved canonical forms, uses the exact canonical strings provided, extracts the corresponding values, and attaches time-window information and any needed qualifiers so the fact can later be encoded and retrieved in a consistent way.

\noindent\textbf{Full prompt:} \href{https://drive.google.com/file/d/1CMtT28ZQ5bfQegKGGrEy5AK2kKn_hefX/view?usp=sharing}{Google Drive PDF}.

\subsubsection{SMT Programming: Diagnosis Clinical Strength Classifier}\label{app:patient-parsing/diagnosis_classifier}

\paragraph{Purpose.} This prompt reviews a full patient note and judges whether each possible diagnosis is strong and clear enough to safely use as an exclusion condition for clinical trial matching. It is designed to be conservative: a diagnosis is marked usable for exclusion only when the note clearly supports that the patient truly has that condition in a clinically meaningful way, and not when the diagnosis is uncertain, negated, historical, or weakly supported.

\noindent\textbf{Full prompt:} \href{https://drive.google.com/file/d/1qX50pSIZctEvUjpr1doneGr8WPd3o8m5/view?usp=sharing}{Google Drive PDF}.

\subsubsection{Patient-Fact Relation Categorizer: Chief Complaints}\label{app:patient-parsing/cc}

\paragraph{Purpose.} This prompt identifies which coded patient facts best represent the patient's main reason for seeking clinical trials. It focuses on the likely chief complaint and, when helpful, a small number of closely related higher-level conditions that a patient would realistically search trials for, while avoiding facts that are too broad, secondary, or about conditions the patient only wants to prevent.

\noindent\textbf{Full prompt:} \href{https://drive.google.com/file/d/1scgQfFp3stvsYiXUTEOtTwdiHoPY3_cu/view?usp=sharing}{Google Drive PDF}.

\subsubsection{Patient-Fact Relation Categorizer: Related to Chief Complaints}\label{app:patient-parsing/ccr}

\paragraph{Purpose.}
This prompt identifies the small set of patient facts that best represent the patient's chief complaint for clinical-trial search. It keeps the complaint itself, meaningful search-relevant ancestors, major procedures directly tied to addressing it, and important direct causes when they are explicitly central. The goal is to focus retrieval on the problems the patient is actually seeking trials for, while excluding background conditions, overly broad concepts, routine workups, and prevention targets.

\noindent\textbf{Full prompt:} \href{https://drive.google.com/file/d/1PrSXN4EFjFMC1t_Cr7TuMuxelhvv1Uqv/view?usp=sharing}{Google Drive PDF}.

\subsubsection{Patient-Fact Relation Categorizer: Any Complaint for Clinical Trials}
\label{app:patient-parsing/any}

\paragraph{Purpose.} This prompt filters a small set of patient facts to those that the patient might realistically search clinical trials for. It is intentionally recall-oriented: it keeps search-worthy findings, disorders, symptoms, and significant non-routine procedures, while excluding overly general concepts and routine clinical events.

\noindent\textbf{Full prompt:} \href{https://drive.google.com/file/d/1CmiriMivVeAIeWEVDHRMog4MfdF2XxUT/view?usp=sharing}{Google Drive PDF}.

\subsection{Prompts for Trial-Side Salience-Based Constraint Augmentation}
\subsubsection{Trial-Constraint Augmentation: Whole-fact Salience Policy Enforcer}\label{app:patient-parsing/whole-fact-salience}

\paragraph{Purpose.}
This prompt enforces a conservative salience policy over patient-fact variables. It keeps only variables whose truth would almost certainly be explicitly stated in a prescreening patient vignette and whose underlying facts are clinically meaningful, canonicalizable, and not routine, vague, demographic, logistical, negative, or easily omitted. The goal is to restrict downstream reasoning to patient facts that are reliably observable from the note itself.

\noindent\textbf{Full prompt:} \href{https://drive.google.com/file/d/1MaZO3BhPNrLgXmUeQuvux-THRT0Qkjj0/view?usp=sharing}{Google Drive PDF}.

\subsubsection{Trial Constraint Augmentation: Specificity-Mismatch Salience Policy Enforcer}

\label{app:patient-parsing/specificity-mismatch-salience}

\paragraph{Purpose.}
This prompt selects a small set of trial-side ancestor concepts that are acceptable documentation-level alternatives to a more specific requirement concept. It keeps only ancestors that plausibly reflect how the condition would be recorded or self-reported when underspecified, while rejecting ancestors that are too generic and would create spurious patient--trial matches. The goal is to preserve recall under realistic specificity mismatch without overly weakening the clinical meaning of the requirement.

\noindent\textbf{Full prompt:} \href{https://drive.google.com/file/d/1OuIr9n7-XRjAPyJUiIpy_w8uCIIg1M8z/view?usp=sharing}{Google Drive PDF}.

\subsection{Prompts for Patient-Side Inferential Constraint Augmentation}

\subsubsection{Patient-Constraint Augmentation: Differential Diagnoser}\label{app:patient-parsing/ddx}

\paragraph{Purpose.}
This prompt augments the patient representation by extracting a ranked differential diagnosis from the full patient note. It proposes concrete diagnoses supported by the note, with calibrated confidence, evidence, rationale, and conservative time bounds, so that downstream retrieval and reasoning can include both explicit and plausibly inferable patient conditions.

\noindent\textbf{Full prompt:} \href{https://drive.google.com/file/d/1QuafCMxCPN6XYVnDuY5p37mvicTORpxb/view?usp=sharing}{Google Drive PDF}.

\subsubsection{Patient-Constraint Augmentation: Prevention Target Inference}\label{app:patient-parsing/prevention}

\paragraph{Purpose.}
This prompt augments the patient representation with diseases the patient is trying to prevent, rather than diseases they already have. It extracts only prevention targets that are explicitly stated or strongly implied by a condition whose near-universal clinical purpose is prevention of a specific downstream disease, while excluding ambiguous risk discussions and diseases already inferable as current diagnoses.

\noindent\textbf{Full prompt:} \href{https://drive.google.com/file/d/1Ua2C5qHUI1KmuNhHkP3VktJFUpoEvYZU/view?usp=sharing}{Google Drive PDF}.

\subsubsection{Patient-Constraint Augmentation: Significant Procedure Inference}\label{app:patient-parsing/procedure}

\paragraph{Purpose.}
This prompt augments the patient representation with significant procedures the patient may reasonably need to treat or mitigate their primary disease. It infers major therapeutic or interventional procedures that could plausibly serve as clinical-trial targets, while excluding routine diagnostic workups, logistical procedures, and interventions aimed only at secondary symptoms.

\noindent\textbf{Full prompt:} \href{https://drive.google.com/file/d/1SfMnTbNPQ21qwWBdqxTY2qgyc2AyXCD-/view?usp=sharing}{Google Drive PDF}.

\subsection{Prompts for Evaluation}
\subsubsection{LLM Relevance Evaluator}\label{app:patient-parsing/relevance-judge}

\paragraph{Purpose.}
This prompt serves as the LLM-based relevance evaluator for patient--trial matching. Given a patient note and a trial description, it determines which trial subcohorts, if any, are clinically relevant to the patient under an explicitly provided relevance definition, while separating relevance from eligibility and producing a brief clinician-facing justification.

\noindent\textbf{Full prompt:} \href{https://drive.google.com/file/d/1d5MJxzJBmRteeAWUp2vwUd3IFASd6_ZB/view?usp=sharing}{Google Drive PDF}.

\subsubsection{LLM Eligibility Evaluator}\label{app:patient-parsing/eligibilty-judge}

\paragraph{Purpose.}
This prompt serves as the LLM-based eligibility evaluator for patient--trial matching. Given a patient note, a trial description, and specific trial subcohorts of interest, it determines subcohort-level eligibility by applying a detailed set of clinical inference, missingness, and projection rules to the patient's current and past clinical state, and returns concise clinician-facing justifications for each decision.

\noindent\textbf{Full prompt:} \href{https://drive.google.com/file/d/1sQCd5c5kL99dx545t4yKQlxIjUW5_RQt/view?usp=sharing}{Google Drive PDF}.

\newpage

\end{document}